\documentclass[11pt]{article}

\usepackage[a4paper,margin=1in]{geometry}
\usepackage{amsmath,amssymb,amsfonts,bm,mathtools}
\usepackage{booktabs,multirow,array,tabularx,longtable,makecell}
\usepackage{placeins}
\usepackage{graphicx}
\usepackage{xcolor}
\usepackage[table]{xcolor}
\usepackage{enumitem}
\usepackage{hyperref}
\usepackage[capitalize,noabbrev]{cleveref}
\usepackage[backend=biber,style=ieee,sorting=none,maxbibnames=99]{biblatex}
\usepackage{microtype}
\usepackage{authblk}
\usepackage{threeparttable}
\usepackage{tabularray}
\usepackage{leftindex}
\usepackage{svg}
\usepackage{multicol}
\usepackage{appendix}
\usepackage{etoc}
\usepackage{float}
\usepackage{wrapfig}
\usepackage{capt-of}
\usepackage{tikz}

\UseTblrLibrary{booktabs}

\usetikzlibrary{arrows.meta, calc, backgrounds, positioning}
\pgfdeclarelayer{foreground}
\pgfsetlayers{background, main, foreground}

\hypersetup{
    colorlinks=true,
    linkcolor=blue!50!black,
    citecolor=blue!50!black,
    urlcolor=blue!50!black
}

\newcommand{\email}[1]{\texttt{#1}}
\definecolor{revisionmagenta}{RGB}{190,70,145}

\newcommand{\R}{\mathbb{R}}

\newcommand{\ten}[1]{\mathcal{#1}}
\newcommand{\mat}[1]{\mathbf{#1}}
\newcommand{\vecb}[1]{\mathbf{#1}}

\newcommand{\set}[1]{\mathrm{#1}}
\newcommand{\llama}{\text{LLaMA-3-8B}}

\title{Tensor Methods for Language Models: From Token\\ Representation to Training, Adaptation, Compression,\\ Inference, and Interpretability}

\date{}

\author[1]{Matvei Tarasov}
 \affil[1]{Innopolis University, Innopolis 420500, Russia\\ \email{m.tarasov@innopolis.university}}

 \author[1]{Salman Ahmadi-Asl}
 \affil[1]{Innopolis University, Innopolis 420500, Russia\\ \email{s.ahmadiasl@innopolis.ru}}

 \author[2]{André L. F. de Almeida}
 \affil[2]{Department of Teleinformatics Engineering, Federal University of Cear\'a, Fortaleza 60455-760, Brazil\\ \email{andre@gtel.ufc.br}}

 \author[3]{Andrzej Cichocki}
 \affil[3]{Systems Research Institute of Polish Academy of Science and Warsaw University of Technology, Poland\\ \email{cichockiand@gmail.com}}

\begin{document}
\maketitle

\begin{abstract}
Large language models (LLMs) are built from structured high-dimensional objects such as token representations, weights, adaptation updates, caches, and activations, whose multilinear structure is underexploited by the conventional matrix-centric view.
Tensor decompositions and tensor networks provide a principled algebraic language for this structure, yet the literature often treats them as isolated compression mechanisms.
This survey organizes tensor methods for LLMs through two complementary views: a seven-stage lifecycle taxonomy covering tokenization, embeddings, pre-training, adaptation, compression, inference, and interpretability, and a component view covering embeddings, attention, and feed-forward networks.
We provide unified notation and theoretical foundations, analyze tensorization strategies for individual Transformer components, and review methods at each lifecycle stage.
We further connect tensor methods to neighboring efficiency techniques and probabilistic tensor networks.
We synthesize open challenges and introduce the compression-realization gap $\rho_{\rm gap}$, a metric that quantifies the discrepancy between theoretical memory reduction and measured system-level speedup.
We compare tensor methods for pre-training and parameter-efficient fine-tuning (PEFT) under matched protocols and use this metric to analyze pre-training efficiency.
More broadly, by treating tensorization as a common structural principle, the survey provides a structured entry point to tensorized language models and clarifies when parameter savings can plausibly translate into memory efficiency, computational efficiency, or interpretability.
The GitHub page dedicated to this paper is accessible at \href{https://github.com/ma-tt-a/awesome-tensor-methods-for-llms}{this https URL}.
\end{abstract}

\section{Introduction}
\label{sec:introduction}

Large language models (LLMs) based on the Transformer architecture \cite{vaswani2017attention} have reshaped deep learning and artificial intelligence by demonstrating broad capabilities and performance gains across successive model generations \cite{brown2020gpt3, openai2024gpt4, touvron2023llama}. Their success stems from three factors: a highly parallelizable architecture that scales efficiently with model size and compute \cite{vaswani2017attention}, self-supervised pre-training on internet-scale text corpora \cite{radford2019language, raffel2023t5}, and the predictable improvements that arise from scaling this combination \cite{kaplan2020scalinglaws, hoffmann2022scalinglaws2}.
Today's production-scale LLMs, such as Qwen \cite{yang2025qwen3}, DeepSeek \cite{deepseekai2026deepseekv4}, and Kimi \cite{kimiteam2026kimik3}, contain tens to hundreds of billions of parameters and are deployed at scale, demanding substantial computational resources for both training and inference.

In parallel, the past several decades have seen active development of tensor decompositions and tensor networks within the numerical linear algebra community.
Emerging independently in psychometrics \cite{tucker1966some, carroll1970analysis, harshman1970foundations} and in quantum many-body physics \cite{orus2014practical, verstraete2008peps, vidal2008mera}, these tools have gradually migrated into general-purpose computational algebra.
Along the way, researchers reformulated the existing formats in the language of numerical analysis and developed stable algorithms for computing them \cite{lathauwer2000hosvd, oseledets2011tensor, zhao2016tensorring}, alongside a broader theoretical effort to systematize and unify the resulting landscape \cite{kolda2009tensor, cichocki2015tensor, cichocki2015tensor2}.
This growing activity has, in turn, fueled interest in applying tensor methods to deep learning \cite{wang2025tnmeetnn}, reflecting a wider recognition that numerical linear algebra sits at the core of modern neural architectures \cite{baggag2025linalg}.

\paragraph{Motivation.}
Despite their impressive capabilities, modern LLMs face three persistent challenges. Training a frontier model requires substantial accelerator time \cite{grattafiori2024llama3}. Inference at scale demands high memory bandwidth, and the KV cache can become the dominant memory bottleneck in long-context applications \cite{kwon2023paggedattn}. Meanwhile, internal representations, including activations and weights, remain largely opaque, and the mechanisms by which these models reason and store knowledge are still poorly understood, motivating a growing body of interpretability research \cite{elhage2021mathematical, elhage2022superposition}.
These challenges are usually treated separately, but a tensor perspective provides a common language for all three. LLMs are conventionally described and implemented in matrix terms: Transformer computations reduce largely to batched matrix multiplication, which is convenient and highly optimized but conceptually incomplete. Many Transformer objects, including stacked multi-head attention projections, layer collections, activation streams, and KV caches, are naturally higher-order, and a purely matrix-centric view leaves their multilinear structure underexploited.

\begin{table}[!htbp]
\centering
\caption{\footnotesize Literature clusters surrounding tensor methods and language models.}
\label{tab:literature-clusters}
\footnotesize
\begin{tblr}{
width = \textwidth,
colspec = {Q[4.9cm,m,c]Q[4.9cm,m,c]X[m,c]},
row{1} = {font=\bfseries},
rowsep = 3pt,
hline{1,6} = {1pt},
hline{2} = {0.6pt},
hline{3,4,5} = {0.6pt, gray7},
}
Cluster & Main objects or methods & Specific gap for this survey \\
Classical tensor decompositions and tensor networks \newline \cite{kolda2009tensor, cichocki2015tensor, cichocki2015tensor2, orus2014practical} & Tensors, tensor-algebraic operations; decompositions, ranks; computation algorithms & Discusses tensors abstractly; does not provide LLM applications of tensors \\
Tensor methods for machine \newline learning and neural networks \newline \cite{sidiropoulos2017tensorsforspandml, ji2019tensorsformlsurvey1, panagakis2021tensorsfornn1, xinwei2024tensorsfornn2, wang2025tnmeetnn, he2026tensorsfornn3}  & Tensorized machine learning; tensorization and decomposition of generic neural modules, layers & Often predates decoder-only LLMs; omits LLM-specific objects such as the KV cache  \\
LLM efficiency techniques \newline \cite{wan2024efficientllmsurvey1, wang2025peftsurvey1, zhu2024compressionsurvey1, tang2024compressionsurvey2, kim2025peftcompressionsurvey1, wang2024compressioninferencesurvey1, zhou2024efficientinferencesurvey1, yuan2024efficientinferencesurvey2, li2025kvcachesurvey1, miao2025efficientservingsurvey1, Laura2026lowranksurvye}  & PEFT; compression: quantization, pruning, knowledge distillation; efficient attention; efficient inference; KV-cache compression & Usually omits tensor counterparts; does not emphasize tensor methods as a separate class \\
LLM mechanistic interpretability \newline \cite{somvanshi2026interpretabilitysurvey1, rai2025interpretabilitysurvey2, bereska2024interpretabilitysurvey3} &  Features, activations, logits, neurons, heads, circuits; activation patching, ablations, SAEs & Overlooks multilinear structure, both as a descriptive language and as an analysis tool \\
\end{tblr}
\end{table}

\paragraph{Current state of the literature.}

The literature on tensors and LLMs is fragmented across several partially overlapping clusters, summarized in \cref{tab:literature-clusters}. We cite representative surveys and reviews, so that the reader can turn to a fuller treatment of each area. Each row identifies a community with its own objects, methods, and terminology; the remainder of this section discusses each cluster in turn, together with the specific gap it leaves for a survey connecting tensor methods to LLMs.

The first cluster is the classical tensor decomposition literature \cite{kolda2009tensor, cichocki2015tensor, cichocki2015tensor2, orus2014practical}. Works from this cluster are mathematically deep and essential for notation, algorithms, and theoretical concepts such as tensor rank. While this literature provides the general algebraic and numerical foundation for tensor computations, it does not map these concepts onto LLM-specific objects.

A second line of work applies tensor decompositions and tensor networks to classical machine learning \cite{sidiropoulos2017tensorsforspandml, ji2019tensorsformlsurvey1} and neural networks (NNs) \cite{panagakis2021tensorsfornn1, xinwei2024tensorsfornn2, wang2025tnmeetnn, he2026tensorsfornn3}. This literature establishes neural-network tensorization as a general design principle. Much of it, however, predates modern decoder-only LLMs or focuses on generic layers, convolutional neural networks (CNNs), and recurrent neural networks (RNNs), leaving LLM-specific objects and constraints untreated.

A third body of work covers the broader landscape of LLM efficiency techniques. It has expanded rapidly over the past few years, spanning many directions \cite{wan2024efficientllmsurvey1}. Some of this work targets parameter-efficient fine-tuning (PEFT) \cite{wang2025peftsurvey1} and post-training compression techniques such as quantization, pruning, and knowledge distillation (KD) \cite{zhu2024compressionsurvey1, tang2024compressionsurvey2, wang2024compressioninferencesurvey1}; some covers PEFT and compression together \cite{kim2025peftcompressionsurvey1}. Other work targets inference-time efficiency \cite{zhou2024efficientinferencesurvey1, yuan2024efficientinferencesurvey2, li2025kvcachesurvey1} and serving \cite{miao2025efficientservingsurvey1}, whereas \cite{Laura2026lowranksurvye} cuts across all these directions by structure rather than application, reviewing low-rank matrix factorizations in weights, adapters, and gradients. Within each of these sub-clusters, tensor decompositions surface only occasionally, mentioned as one technique among many. Yet tensor methods across all these sub-clusters rest on the same underlying algebraic theory, which suggests considering them as a standalone group.

As a fourth cluster, mechanistic interpretability has become an active and rapidly growing research direction, aiming to reverse-engineer language models' internal computations in terms of features, activations, neurons, heads, and circuits, using tools such as activation patching, ablations, and sparse autoencoders (SAEs) \cite{somvanshi2026interpretabilitysurvey1, rai2025interpretabilitysurvey2, bereska2024interpretabilitysurvey3}. Within this broader field, a smaller community studies Transformer circuits specifically through the lens of tensor networks, using graphical tensor notation \cite{taylor2024interpretabilitysurvey4}. Beyond notation, a handful of very recent works make multilinear structure the object of the analysis itself, and no existing interpretability survey covers them. This survey treats both the notation and these methods as a distinct direction (\cref{subsec:interpretability}).

\paragraph{Scope.} 

The same multilinear structure recurs across all four clusters, but the clusters use it differently. The classical tensor literature develops a general theory, whereas tensorized neural modules, LLM efficiency techniques, and mechanistic interpretability typically use that theory locally for a specific task. This survey connects these lines of work through a unified tensor-theoretic lens grounded in LLM-specific objects. Its distinguishing feature is the combination of two complementary organizations: a component view that asks \emph{what} inside a Transformer is tensorized, and a lifecycle view that asks \emph{when and why} tensorization is introduced. To our knowledge, existing surveys do not combine this two-view organization with unified notation, protocol-aware comparisons, and a systems-level analysis of whether theoretical compression is realized in practice.
\paragraph{Contributions}
This survey makes four contributions:
\begin{enumerate}[leftmargin=*]
  \item We frame tensorization as a common structural principle acting on token representations, weights, adaptation updates, caches, and activations. We organize this literature with a seven-stage \emph{lifecycle taxonomy}: tokenization, embeddings, pre-training, adaptation, compression, inference, and interpretability.
  \item We complement the lifecycle taxonomy with a \emph{component view} of embeddings, attention, and feed-forward networks. Together, the two views distinguish the structural compatibility of a decomposition with a model object from the training or deployment objective for which that decomposition is used.
  \item We synthesize open challenges and formulate the compression-realization gap through $\rho_{\rm gap}$, which separates algorithmic overhead from hardware realization and makes explicit why parameter reduction alone does not imply end-to-end speedup.
  \item We compare low-rank, Kronecker, tensor train, and tensor train matrix parameterizations in pre-training and PEFT, evaluating efficiency, model quality, and numerical rank under matched protocols.
\end{enumerate}

\paragraph{Structure of the survey.}
The remainder of this paper is organized as follows.
\Cref{sec:preliminaries} presents the mathematical preliminaries, including notation, tensor-network diagrams, tensor decompositions, and a brief review of the Transformer architecture.
\Cref{sec:lifecycle-overview} introduces the lifecycle taxonomy and illustrates tensorization at each populated stage using LLaMA-3-8B as a running case study.
\Cref{sec:components} develops the component view for individual Transformer modules, whereas \cref{sec:lifecycle-analysis} develops the lifecycle view through formal problem definitions, literature reviews, and comparison tables where coverage permits.
\Cref{sec:neighboring-methods} situates tensor methods within the broader landscape of LLM efficiency techniques, and \cref{sec:prob-tensor-nets} clarifies their connections to probabilistic tensor networks.
\Cref{sec:software-overview} surveys software and presents case studies.
\Cref{sec:future-directions} defines the compression-realization gap, identifies open challenges, and outlines future research directions.
\Cref{sec:experiments} presents the pre-training and PEFT experiments and analyzes the singular value spectra of the learned weights and adapter updates.
Finally, \cref{sec:conclusion} concludes the survey.


\section{Preliminaries}
\label{sec:preliminaries}


\begin{figure}[!htpb]
\centering
\begin{minipage}[t]{0.45\textwidth}
\vspace{0pt}
\small
\begin{tblr}{
  width = \linewidth,
  colspec = {Q[3.15cm,m,c]X[m,c]},
  row{1} = {font=\bfseries},
  rowsep = 2.5pt,
  hline{1,13} = {1pt},
  hline{2} = {0.6pt},
  hline{3,4,5,6,7,8,9,10,11,12} = {0.6pt, gray7},
}
Notation & Meaning \\
$\alpha$ & scalar \\
$\vecb{a}$ & vector \\
$\mat{A}$ & matrix \\
$\ten{A}$ & tensor \\
$\ten{I}$ & identity tensor \\
$\ten{A}_{i_1, \dots, i_N}$ & $(i_1, \dots, i_N)$-th entry \\
$\ten{A}_{i_1, \dots, i_{n-1}, :, i_{n+1}, \dots, i_N}$ & n-th mode slice \\
$\mat{A}\otimes\mat{B}$ & Kronecker product \\
$\vecb{a} \circ \vecb{b}$ & outer product \\
$\ten{A}\times_n\mat{B}$ & mode-$n$ product \\
$\ten{A} \leftindex_{i_1, \dots, i_K}\times^{j_1, \dots, j_K} \ten{B}$ & contraction \\
\end{tblr}
\end{minipage}
\hfill
\begin{minipage}[t]{0.50\textwidth}
\vspace{0pt}
\centering
\resizebox{\linewidth}{!}{%
\begin{tikzpicture}[
  tnode/.style={circle, draw=black!80, line width=1.0pt,
               minimum size=0.5cm, inner sep=0pt, font=\tiny},
  knode/.style={circle, draw=black!80, line width=1.0pt,
               minimum size=0.52cm, inner sep=0pt, font=\tiny},
  leg/.style={line width=1.0pt, black!80},
  bond/.style={line width=1.0pt, black!80},
  lbl/.style={font=\tiny, align=center, text width=1.9cm},
  idx/.style={font=\tiny, text=black},
  pnl/.style={font=\tiny\bfseries},
  idnode/.style={circle, fill=black!80, minimum size=4pt, inner sep=0pt}
]

\definecolor{ColA}{RGB}{102,194,165}
\definecolor{ColB}{RGB}{252,141, 98}

\pgfmathsetmacro{\Rad}{0.25} 
\pgfmathsetmacro{\Lleg}{0.30}  
\pgfmathsetmacro{\Dx}{2.10}
\pgfmathsetmacro{\CapDy}{0.70}
\pgfmathsetmacro{\CapDyTensor}{0.8} 
\pgfmathsetmacro{\CCentre}{2.10} 

\pgfmathsetmacro{\RowOneY}{3.0}
\pgfmathsetmacro{\RowTwoY}{1.0}
\pgfmathsetmacro{\RowThreeY}{-1.0}
\pgfmathsetmacro{\CapRowOneY}{\RowOneY-\CapDyTensor}  


\node[tnode, fill=ColA!35] (v) at (0*\Dx, \RowOneY) {$\vecb{a}$};
\draw[leg] (v.east) -- ++(\Lleg, 0);
\node[idx] at ($(v.east)+(\Lleg+0.08,0.11)$) {$I$};
\node[lbl] at (0*\Dx, \CapRowOneY) {(a) vector};

\node[tnode, fill=ColA!35] (m) at (1*\Dx, \RowOneY) {$\mat{A}$};
\draw[leg] (m.west) -- ++(-\Lleg, 0);
\draw[leg] (m.east) -- ++( \Lleg, 0);
\node[idx] at ($(m.west)+(-\Lleg-0.08,0.11)$) {$I$};
\node[idx] at ($(m.east)+(\Lleg+0.08,0.11)$) {$J$};
\node[lbl] at (1*\Dx, \CapRowOneY) {(b) matrix};

\node[tnode, fill=ColA!35] (t) at (2*\Dx, \RowOneY) {$\ten{A}$};
\draw[leg] (t.225) -- ++(-0.26, -0.26);
\draw[leg] (t.315) -- ++( 0.26, -0.26);
\node[idx] at ($(t.225)+(-0.26-0.2,-0.26-0.02)$) {$I_1$};
\node[idx] at ($(t.315)+( 0.26+0.2,-0.26-0.02)$) {$I_N$};
\node[font=\tiny] at ($(t.center)+(0,-0.50)$) {$\cdots$};
\node[lbl] at (2*\Dx, \CapRowOneY) {(c) tensor};


\pgfmathsetmacro{\KRadCore}{0.25}
\pgfmathsetmacro{\KDyABH}{0.36}
\pgfmathsetmacro{\KFanHalf}{\KDyABH+\KRadCore}
\pgfmathsetmacro{\KFanDepth}{0.26}
\pgfmathsetmacro{\KBondLen}{0.35}
\pgfmathsetmacro{\KExtLleg}{0.28}

\pgfmathsetmacro{\OGap}{0.95}
\pgfmathsetmacro{\KHalfWidth}{\KRadCore+\KBondLen+\KFanDepth+\KExtLleg}
\pgfmathsetmacro{\OHalfWidth}{0.5*\OGap+\Rad}
\pgfmathsetmacro{\RowTwoGap}{0.9} 
\pgfmathsetmacro{\RowTwoHalf}{\KHalfWidth+0.5*\RowTwoGap+\OHalfWidth}
\pgfmathsetmacro{\KCoreX}{\CCentre+\RowTwoHalf-\KHalfWidth}

\pgfmathsetmacro{\KLFanBX}{\KCoreX-\KRadCore-\KBondLen}
\pgfmathsetmacro{\KLFanX}{\KLFanBX-\KFanDepth}
\pgfmathsetmacro{\KRFanBX}{\KCoreX+\KRadCore+\KBondLen}
\pgfmathsetmacro{\KRFanX}{\KRFanBX+\KFanDepth}

\pgfmathsetmacro{\KTopY}{\RowTwoY+\KDyABH}
\pgfmathsetmacro{\KBotY}{\RowTwoY-\KDyABH}
\pgfmathsetmacro{\KFanTopY}{\RowTwoY+\KFanHalf}
\pgfmathsetmacro{\KFanBotY}{\RowTwoY-\KFanHalf}

\draw[leg] (\KLFanX, \RowTwoY) -- ++(-\KExtLleg, 0);
\draw[leg] (\KRFanX, \RowTwoY) -- ++( \KExtLleg, 0);
\node[idx] at ($(\KLFanX-\KExtLleg-0.08, \RowTwoY+0.15)$) {$I_1 I_2$};
\node[idx] at ($(\KRFanX+\KExtLleg+0.08, \RowTwoY+0.15)$) {$J_1 J_2$};

\fill[black!9, draw=black!80, line width=1.0pt]
    (\KLFanBX, \KFanTopY)
    arc[start angle=90, end angle=270, x radius=\KFanDepth, y radius=\KFanHalf]
    -- cycle;
\fill[black!9, draw=black!80, line width=1.0pt]
    (\KRFanBX, \KFanBotY)
    arc[start angle=-90, end angle=90, x radius=\KFanDepth, y radius=\KFanHalf]
    -- cycle;

\node[knode, fill=ColA!35] (kA) at (\KCoreX, \KTopY) {$\mat{A}$};
\node[knode, fill=ColB!35] (kB) at (\KCoreX, \KBotY) {$\mat{B}$};
\draw[bond] (\KLFanBX, \KTopY) -- (kA.west);
\draw[bond] (\KLFanBX, \KBotY) -- (kB.west);
\draw[bond] (kA.east) -- (\KRFanBX, \KTopY);
\draw[bond] (kB.east) -- (\KRFanBX, \KBotY);
\node[idx] at ($(\KLFanBX,\KTopY)!0.5!(kA.west)+(0.025, 0.17)$) {$I_1$};
\node[idx] at ($(\KLFanBX,\KBotY)!0.5!(kB.west)+(0.025,-0.17)$) {$I_2$};
\node[idx] at ($(kA.east)!0.5!(\KRFanBX,\KTopY)+(-0.025, 0.17)$) {$J_1$};
\node[idx] at ($(kB.east)!0.5!(\KRFanBX,\KBotY)+(-0.025,-0.17)$) {$J_2$};

\pgfmathsetmacro{\CapRowTwoY}{\RowTwoY-\KFanHalf-0.48}
\node[lbl] at (\KCoreX, \CapRowTwoY) {(e) Kronecker product};

\pgfmathsetmacro{\OX}{\CCentre-\RowTwoHalf+\Rad}
\pgfmathsetmacro{\OLleg}{0.24}
\node[tnode, fill=ColA!35] (o1) at (\OX,       \RowTwoY) {$\vecb{a}$};
\node[tnode, fill=ColB!35] (o2) at (\OX+\OGap, \RowTwoY) {$\vecb{b}$};
\draw[leg] (o1.south) -- ++(0, -\OLleg);
\draw[leg] (o2.south) -- ++(0, -\OLleg);
\node[idx] at ($(o1.south)+(-0.16,-\OLleg)$) {$I$};
\node[idx] at ($(o2.south)+( 0.16,-\OLleg)$) {$J$};
\node[lbl] at ({\OX+0.5*\OGap}, \CapRowTwoY) {(d) outer product};


\pgfmathsetmacro{\CGap}{1.15}
\pgfmathsetmacro{\Cmid}{\CCentre-1.35}
\pgfmathsetmacro{\CX}{\Cmid-0.5*\CGap}
\node[tnode, fill=ColA!35] (c1) at (\CX,       \RowThreeY) {$\mat{A}$};
\node[tnode, fill=ColB!35] (c2) at (\CX+\CGap, \RowThreeY) {$\mat{B}$};
\draw[leg] (c1.west) -- ++(-\Lleg, 0);
\draw[bond] (c1.east) -- (c2.west);
\draw[leg] (c2.east) -- ++(\Lleg, 0);
\node[idx] at ($(c1.west)+(-\Lleg-0.08,0.15)$) {$I$};
\node[idx] at ($(c1.east)!0.5!(c2.west)+(0,0.15)$) {$K$};
\node[idx] at ($(c2.east)+(\Lleg+0.08,0.15)$) {$J$};
\node[lbl] at (\Cmid, \RowThreeY-\CapDy) {(f) contraction};

\pgfmathsetmacro{\IdX}{\CCentre+1.55}
\node[idnode] (id) at (\IdX, \RowThreeY) {};
\draw[leg] (id.west) -- ++(-0.45, 0) node[above, idx] {$I$};
\draw[leg] (id.east) -- ++( 0.45, 0) node[above, idx] {$I$};
\draw[leg] (id.north) -- ++(0, 0.38) node[right, idx] {$I$};
\node[lbl] at (\IdX, \RowThreeY-\CapDy) {(g) identity\\tensor};

\end{tikzpicture}%
}
\end{minipage}
\caption{\footnotesize Notation reference. 
\textit{Left}: symbols used throughout this survey for scalars, vectors, matrices, tensors, and the main tensor-algebraic operations. \textit{Right}: tensor-network diagram representations of the same objects, with examples of the operations.}
\label{fig:notation-overview}
\end{figure}

\subsection{Notation}

\paragraph{Objects} Calligraphic uppercase letters $\ten{A}\in\R^{I_1\times I_2\times \cdots\times I_N}$ denote an $N$th-order tensor; matrices and vectors are denoted by bold uppercase $\mat{A}$ and lowercase letters $\vecb{a}$. The symbols in \cref{fig:notation-overview} (left) collect the notation used throughout the survey; the tensor-network diagrams \cite{penrose1971tensordiagrams} in \cref{fig:notation-overview} (right) give the corresponding graphical illustration for each object and operation.
The $N$th-order identity tensor $\ten{I}\in\R^{I\times\cdots\times I}$ has entries $\ten{I}_{i_1,\dots,i_N}=\prod_{n=2}^{N}\delta_{i_1 i_n}$, where $\delta$ is the Kronecker delta. We draw it as a filled dot, as in \cref{fig:notation-overview}(g).

\paragraph{Outer product} Given $N$ vectors $\vecb{a}^{(1)}\in\R^{I_1},\ldots,\vecb{a}^{(N)}\in\R^{I_N}$, their outer product is the $N$th-order tensor $\vecb{a}^{(1)}\circ\cdots\circ\vecb{a}^{(N)}\in\R^{I_1\times\cdots\times I_N}$ with entries:
\begin{equation}
\bigl(\vecb{a}^{(1)}\circ\cdots\circ\vecb{a}^{(N)}\bigr)_{i_1\ldots i_N}= \vecb{a}^{(1)}_{i_1}\vecb{a}^{(2)}_{i_2}\cdots \vecb{a}^{(N)}_{i_N}.
\end{equation}
This tensor is called a canonical rank-one tensor. For $N=2$, this reduces to the familiar rank-one matrix $\vecb{a}\circ\vecb{b}\in\R^{I\times J}$ with entries $(\vecb{a}\circ\vecb{b})_{ij}=a_ib_j$. In tensor-network diagrams, the outer product is represented by placing the tensors next to each other, as shown in \cref{fig:notation-overview}(d).

\paragraph{Kronecker product and decomposition} The Kronecker product of two matrices $\mat{A}\in\R^{I_1\times J_1}$ and $\mat{B}\in\R^{I_2\times J_2}$ is the block matrix obtained by replacing each entry of $\mat{A}$ with a copy of $\mat{B}$ scaled by that entry,
\begin{equation}
  \mat{A}\otimes\mat{B}=
  \begin{bmatrix}
  A_{11}\mat{B} & \cdots & A_{1J_1}\mat{B} \\
  \vdots & \ddots & \vdots \\
  A_{I_11}\mat{B} & \cdots & A_{I_1J_1}\mat{B}
  \end{bmatrix}
  \in\R^{I_1I_2\times J_1J_2}.
\end{equation}
The corresponding tensor-network diagram is shown in \cref{fig:notation-overview}(e). This diagram has a strong connection with the outer product; see \cite{yokota2024tensorbasics} for more details.

For a fixed $R \in \mathbb{N}$, one may look for the best approximation of a matrix by a sum of Kronecker products
\begin{equation}
\label{eq:kd}
\min_{\mat{U}_r,\mat{V}_r}\ \| \mat{A}-\sum_{r=1}^{R}\mat{U}_r\otimes\mat{V}_r\|_F,
\end{equation}
where $\mat{A}\in\R^{I_1I_2\times J_1J_2}$, $\mat{U}_r\in\R^{I_1\times J_1}$, and $\mat{V}_r\in\R^{I_2\times J_2}$. This problem admits a closed-form solution; rearranging the $I_2\times J_2$ blocks of $\mat{A}$ into the rows of a matrix $\mathcal{R}(\mat{A})\in\R^{I_1J_1\times I_2J_2}$ turns each Kronecker term into a rank-one matrix, so Eq. \eqref{eq:kd} reduces to a low-rank approximation of $\mathcal{R}(\mat{A})$ and is solved by its rank-$R$ truncated SVD, whose singular vectors are then reshaped back into $\mat{U}_r$ and $\mat{V}_r$; see \cite{golubvanloan} for the precise construction. This is known as the Kronecker product SVD, and we refer to it as the Kronecker decomposition. The smallest $R$ for which the approximation is exact is called the Kronecker rank of $\mat{A}$. The decomposition is not intrinsic to $\mat{A}$ alone, since the factor shapes $I_1I_2=I$ and $J_1J_2=J$ can be chosen in several ways, and each choice defines a different problem \eqref{eq:kd} with its own Kronecker rank. We consider only the matrix case, although the construction generalizes to tensors \cite{batselier2017ktd}.

\paragraph{Contractions} Contraction can be viewed as summing with respect to one or more paired modes of two tensors, while leaving other modes free. Let $\ten{A}\in\R^{I_1\times I_2\times \cdots\times I_N}$ be an $N$th-order tensor and $\ten{B}\in\R^{J_1\times J_2 \times \cdots\times J_M}$ an $M$th-order tensor, and suppose there are $K$ mode pairs $(i_1,j_1),\ldots,(i_K,j_K)$ satisfying
\begin{equation}
\left\{
\begin{array}{ll}
  1\le i_k\le N, \quad 1\le j_k\le M,  &  k\in\{1,2,\ldots,K\} \\
  i_k \neq i_\ell, \quad  j_k\neq j_\ell, &  \forall k \neq \ell \\
  I_{i_k}=J_{j_k}, & k\in \{1,2,\ldots,K\}.
\end{array}
\right.
\end{equation}
Then the contraction of $\ten{A}$ and $\ten{B}$ over these $K$ mode pairs, denoted as
\begin{equation}
    \ten{A}\leftindex_{i_1,\ldots,i_K}\times^{j_1,\ldots,j_K}\ten{B},
\end{equation}
is an $(N+M-2K)$th-order tensor, obtained by summing over the $K$ matched indices and keeping every remaining mode of $\ten{A}$ and $\ten{B}$ free. In tensor-network diagram notation, contractions are represented by connecting the legs of the contracted modes, while the free modes remain unconnected. This abstract operation corresponds to the \texttt{einsum} function in modern array programming libraries such as NumPy \cite{harris2020numpy}, PyTorch \cite{paszke2019pytorch}, and JAX \cite{bradbury2018jax}.

For example, given $\mat{A} \in \R^{I \times K}, \mat{B} \in \R^{K \times J}$, the contraction over the mode pair $(2,1)$ yields the standard matrix product:
\begin{equation}
  \mat{C} = \mat{A} \mat{B} = \mat{A} \leftindex_{2} \times^{1} \mat{B} \in \R^{I \times J}, \quad \mat{C}_{ij} = \sum_{k=1}^{K} \mat{A}_{ik} \mat{B}_{kj}.
\end{equation}
This example is depicted in \cref{fig:notation-overview}(f), where modes of size $K$ are contracted and modes of size $J$ and $I$ remain free. Expressed using  pseudocode \texttt{einsum} notation, the same operation is:
\begin{equation}
  \mat{C} = \texttt{einsum}({"} ik, \ kj \to ij {"}, \mat{A}, \mat{B}).
\end{equation}

\paragraph{Mode-$n$ product} The mode-$n$ product contracts an $N$th-order tensor $\ten{A}\in\R^{I_1\times I_2 \times \cdots\times I_N}$ with a matrix $\mat{B}\in\R^{J\times I_n}$ over the mode pair $(n_1,m_1)=(n,2)$:
\begin{equation}
\begin{gathered}
    \ten{A}\times_n\mat{B} = \ten{A} \ \leftindex_{n} \times^{2} \ \mat{B} \in\R^{I_1\times\cdots\times I_{n-1}\times J\times I_{n+1}\times\cdots\times I_N}, \\
    (\ten{A}\times_n\mat{B})_{i_1\dots i_{n-1}\,j\,i_{n+1}\dots i_N}=\sum_{i_n=1}^{I_n}\ten{A}_{i_1\dots i_N} \mat{B}_{j,i_n}.
\end{gathered}
\end{equation}
In \texttt{einsum} notation, this operation can be expressed as:
\begin{equation}
    \ten{C} = \texttt{einsum}({"} i_1\dots i_n \dots i_N, \ j i_n \to i_1 \dots j \dots i_N {"}, \ten{A}, \mat{B}).
\end{equation}
Intuitively, this operation multiplies every fiber (i.e., every 1D line) running along the $n$-th dimension of the tensor by the matrix $\mat{B}$. If one unfolds the tensor (i.e., reshapes the tensor into a matrix) so that the $n$-th dimension becomes the rows of a 2D matrix, then $\times_n$ reduces to standard matrix multiplication on that unfolded matrix, after which the result is folded back into a tensor of the new shape.

\subsection{Tensor Decompositions}
\label{subsec:tensor-decompositions}
This section reviews the main tensor decomposition formats relevant to this work. We cover the CP decomposition (sum of canonical rank-one terms), the Tucker decomposition (core tensor with mode-specific factors and its Tucker-1/2/3 variants), the TT decomposition and its matrix variant (TTM), and the block term decomposition, which generalizes both CP and Tucker. For each model, we present the mathematical formulation, highlight key properties, and discuss uniqueness and trade-offs. Throughout this section, $\ten{X} \in \R^{I_1 \times I_2 \times \cdots \times I_N}$ stands for an $N$th-order tensor.

\subsubsection{Canonical Polyadic Decomposition}

The canonical polyadic  (CP) decomposition \cite{hitchcock1927cp, carroll1970analysis, harshman1970foundations} approximates a tensor by a sum of rank-one outer products
\begin{equation}
\label{eq:cp-decomposition}
    \ten{X} \approx \sum_{r=1}^{R}\lambda_r \vecb{a}^{(1)}_r\circ\vecb{a}^{(2)}_r\circ\cdots\circ\vecb{a}^{(N)}_r,
\end{equation}
or equivalently in scalar form
\begin{equation}
\label{eq:cp-decomposition-scalar}
    \ten{X}_{i_1, i_2, \dots, i_N} \approx \sum_{r=1}^{R}\lambda_r \mat{A}^{(1)}_{i_1, r}\mat{A}^{(2)}_{i_2, r}\cdots\mat{A}^{(N)}_{i_N, r},
\end{equation}
where $\mat{A}^{(n)} = [\vecb{a}^{(n)}_1, \vecb{a}^{(n)}_2, \dots, \vecb{a}^{(n)}_R] \in \mathbb{R}^{I_n \times R}$ is the factor matrix for mode $n$, and $\lambda_r$ are scaling coefficients.
When equality holds in Eqs. \eqref{eq:cp-decomposition} and \eqref{eq:cp-decomposition-scalar}, the decomposition is exact and we refer to it as a \emph{CP representation} of $\ten{X}$. The minimum admissible value of $R$ in such a representation is the tensor rank, or CP rank, of $\ten{X}$.
An exact representation is unique up to trivial scaling and column permutations under the Kruskal condition $\sum_{n=1}^N k_{\mat{A}^{(n)}} \ge 2R + (N-1)$, which is sufficient but not necessary, where $k_{\mat{A}}$ is the k-rank of the factor matrix \cite{kruskal1977three, sidiropoulos2000cpuniqueness, kolda2009tensor}.
CP decomposition is compact and often interpretable, but it behaves poorly as an approximation problem because computing CP rank is NP-hard \cite{haastad1990tensor, hillar2013most}. Consequently, $R$ must be selected heuristically, and for a fixed $R \ge 2$, a best rank-$R$ approximation may fail to exist on a set of tensors of positive volume \cite{de2008tensor}. Fitting algorithms inherit these difficulties and become numerically unstable at high orders \cite{kolda2009tensor, cichocki2015tensor}.
Additional constraints such as non-negativity, orthogonality, or sparsity are therefore often imposed to improve stability and accuracy, and to relax the uniqueness conditions \cite{cichocki2015tensor}.

\subsubsection{Tucker Decomposition} 

The Tucker decomposition \cite{tucker1966some} admits the following model:
\begin{equation}
\label{eq:tucker-decomposition}
\ten{X} \approx \ten{G}\times_1\mat{A}^{(1)}\times_2\mat{A}^{(2)} \times_3 \cdots \times_N\mat{A}^{(N)},
\end{equation}
where $\ten{G} \in \R^{R_1 \times R_2 \times \cdots \times R_N}$ is an $N$th-order core tensor, and each $\mat{A}^{(n)} \in \R^{I_n \times R_n}$ is the mode-$n$ factor matrix. In scalar form, the decomposition expresses each element of $\ten{X}$ as: 
\begin{equation}
\label{eq:tucker-decomposition-scalar}
\ten{X}_{i_1, i_2, \dots, i_N} \approx \sum_{r_1=1, \dots, r_N=1}^{R_1, \dots, R_N}
\ten{G}_{r_1, r_2, \dots, r_N} \mat{A}^{(1)}_{i_1, r_1} \mat{A}^{(2)}_{i_2, r_2} \cdots \mat{A}^{(N)}_{i_N, r_N}.
\end{equation}
If exact equality in Eqs. \eqref{eq:tucker-decomposition} and \eqref{eq:tucker-decomposition-scalar} holds, we refer to it as a \emph{Tucker representation} of $\ten{X}$, and the element-wise minimal $N$-tuple $(R_1, R_2, \dots, R_N)$ for which such a representation exists is called the Tucker, or multilinear, rank of $\ten{X}$ (such a tuple always exists, with each $R_n$ equal to the rank of the mode-$n$ matricization of $\ten{X}$ \cite{kolda2009tensor}).
The Tucker decomposition is non-unique since, for any invertible $\mat{S}_n \in \R^{R_n \times R_n}$, replacing the factor matrix $\mat{A}^{(n)}$ with $\mat{A}^{(n)}\mat{S}_n$ and compensating in the core with $\mat{S}_n^{-1}$ leaves the reconstructed tensor unchanged. Orthogonality restricts, but does not in general remove, this gauge freedom. Stronger uniqueness statements require additional assumptions, such as non-degenerate mode singular subspaces and a specified canonical convention.
Common algorithms such as HOSVD \cite{lathauwer2000hosvd} therefore fix a particular choice, producing orthogonal factor matrices $(\mat{A}^{(n)})^\top \mat{A}^{(n)} = \mat{I}_{R_n}$ and a core with additional constraints (see \cite{lathauwer2000hosvd} for more details).
Because of this non-uniqueness, unlike in the CP decomposition, the vectors given by the Tucker factors and the values in the core are not directly interpretable.

For the third-order case, additional terminology is used, indexed by the number of compressed modes. Tucker-3 denotes the standard Tucker decomposition applied to a 3rd-order tensor, where all three modes are compressed.
If $\ten{X} \in \R^{I \times J \times K}$ has only two modes of large size (say, $I$ and $J$), and a dimensionality reduction is required only along those modes, the corresponding decomposition is called the Tucker-2 model, $\ten{X} \approx \ten{G} \times_1 \mat{A} \times_2 \mat{B}$,
where $\ten{G} \in \R^{R_1 \times R_2 \times K}$, $\mat{A} \in \R^{I \times R_1}$ and $\mat{B} \in \R^{J \times R_2}$; $R_1 < I$, $R_2 < J$ are the reduced ranks; and the third mode ($K$) remains uncompressed.
Similarly, if the tensor has only one mode of large size (say, $I$), and reduction is needed only in that mode, the decomposition is called the Tucker-1 model, $\ten{X} \approx \ten{G} \times_1 \mat{A}$,
with $\ten{G} \in \R^{R_1 \times J \times K}$, $\mat{A} \in \R^{I \times R_1}$, $R_1 < I$; this reduces to a truncated SVD along the first mode.
In both cases, the core tensor retains the remaining uncompressed modes in full, making these models efficient when only a subset of modes is highly dimensional.

\subsubsection{Tensor Train Decomposition}

The tensor train (TT) decomposition \cite{oseledets2011tensor} expresses a higher-order tensor $\ten{X}\in\R^{I_1\times I_2\times \cdots\times I_N}$ in chain-like form as follows:
\begin{equation}
    \ten{X} \approx \ten{G}^{(1)} \leftindex_{-1} \times^{1} \ \ten{G}^{(2)} \leftindex_{-1} \times^{1} \ \cdots \leftindex_{-1} \times^{1} \ \ten{G}^{(N)},
\end{equation}
where the $-1$ in the left contraction index denotes the last mode of the left tensor, following Python-style indexing. In scalar form, this is equivalent to
\begin{equation}
\ten{X}_{i_1, i_2,\dots, i_N} \approx \sum_{\alpha_0, \alpha_1, \dots, \alpha_N=1}^{R_0, R_1, \dots, R_N} \ten{G}^{(1)}_{\alpha_0, i_1, \alpha_1} \ten{G}^{(2)}_{\alpha_1, i_2, \alpha_2} \cdots \ten{G}^{(N)}_{\alpha_{N-1}, i_N, \alpha_N},
\end{equation}
where each core $\ten{G}^{(n)} \in \R^{R_{n-1} \times I_n \times R_n}$ is 3rd-order, and the $(N+1)$-tuple $(R_0, R_1, \ldots, R_N)$ is called the TT-rank, with $R_0=R_N=1$ always holding. This decomposition can also be expressed in terms of the core slices:
\begin{equation}
\ten{X}_{i_1, i_2,\dots, i_N} \approx \mat{G}^{(1)}_{i_1}\mat{G}^{(2)}_{i_2}\cdots\mat{G}^{(N)}_{i_N},
\end{equation}
where $\mat{G}^{(n)}_{i_n} = \ten{G}^{(n)}_{:, i_n, :} \in\R^{R_{n-1}\times R_n}$. 
This form of the decomposition motivates the alternative name Matrix Product State (MPS).
If equality holds in the above equations, we refer to it as a \emph{TT representation} of $\ten{X}$.
The TT decomposition is unique only up to gauge transformations. Specifically, for any sequence of invertible matrices $\mat{S}_n$ inserted between adjacent cores, the cores can be transformed via $\mat{G}^{(n)}_{i_n} \mapsto \mat{S}_{n-1} \mat{G}^{(n)}_{i_n} \mat{S}_n^{-1}$ without changing the reconstructed tensor. This non-uniqueness is a fundamental property of the TT decomposition, although orthogonality conditions on the cores restrict this gauge freedom and yield a canonical form \cite{cichocki2015tensor}.
TT decomposition is attractive for high-order tensors because it scales favorably with order. Specifically, the chain-like structure ensures that the total number of parameters grows only as $\mathcal{O}(N I R^2)$ (assuming uniform dimensions and ranks), which is linear in the order $N$, in contrast to the exponential growth of the full tensor.

The Tensor Train Matrix (TTM) format \cite{oseledets2010ttm} applies the TT decomposition to matrices. More precisely, a given matrix $\mat{T}\in\R^{I\times J}$ with $I=\prod_{n=1}^N I_n$ and $J=\prod_{n=1}^N J_n$ is reshaped into a higher-order tensor $\ten{T} \in \R^{I_1 \times J_1 \times \dots \times I_N \times J_N}$, which is then approximated via the TT-like structure
\begin{equation}
\label{eq:ttm-decomposition}
    \ten{T} \approx \ten{G}^{(1)} \leftindex_{-1} \times^{1} \ \ten{G}^{(2)} \leftindex_{-1} \times^{1} \ \cdots \leftindex_{-1} \times^{1} \ \ten{G}^{(N)},
\end{equation}
where, unlike the TT decomposition, each core $\ten{G}^{(n)} \in \R^{R_{n-1} \times I_n \times J_n \times R_n}$ is 4th-order. In scalar form, this can be expressed as
\begin{equation}
\label{eq:ttm-decomposition-scalar}
  \ten{T}_{i_1, j_1,\ldots,i_N ,j_N} \approx \sum_{\alpha_0, \alpha_1, \dots, \alpha_N=1}^{R_0, R_1, \ldots, R_N} \ten{G}^{(1)}_{\alpha_0, i_1, j_1, \alpha_1} \ten{G}^{(2)}_{\alpha_1, i_2, j_2, \alpha_2} \cdots\ten{G}^{(N)}_{\alpha_{N-1},i_N,j_N, \alpha_N},
\end{equation}
where $(R_0, R_1, \ldots, R_N)$ is the TTM-rank, with $R_0=R_N=1$. As in the MPS form of TT, TTM can also be expressed in terms of the matrix slices of the cores
\begin{equation}
\label{eq:ttm-decomposition-slice}
  \ten{T}_{i_1, j_1,\dots, i_N, j_N} \approx \mat{G}^{(1)}_{i_1, j_1}\mat{G}^{(2)}_{i_2, j_2}\cdots\mat{G}^{(N)}_{i_N, j_N},
\end{equation}
where $\mat{G}^{(n)}_{i_n, j_n} = \ten{G}^{(n)}_{:, i_n, j_n, :} \in\R^{R_{n-1}\times R_n}$.
In the TTM decomposition, each $(i_1, j_1, \ldots, i_N, j_N)$-th entry of $\ten{T}$ maps one-to-one to the $(\overline{i_1, \ldots, i_N}, \overline{j_1, \ldots, j_N})$-th entry of $\mat{T}$, where $\overline{i_1, \ldots, i_N}$ denotes the \emph{multi-index} and equals
\begin{equation*}
\overline{i_1, \ldots, i_N} = i_N + (i_{N-1} - 1)I_N + \cdots + (i_1 - 1)I_2 \cdots I_N.
\end{equation*}
This mapping corresponds to row-major reshape order, with the last index varying fastest. The column multi-index follows the same convention with mode sizes $J_1,\ldots,J_N$.
Since this mapping is a bijection, if equality holds in Eqs. \eqref{eq:ttm-decomposition}, \eqref{eq:ttm-decomposition-scalar}, and \eqref{eq:ttm-decomposition-slice}, we refer to it as a \emph{TTM representation} of the matrix $\mat{T}$ itself.
The advantage of TTM is that it natively represents matrices as multilinear operators, preserving the operator structure, in contrast to TT. That is why, in quantum physics, TTM is often referred to as a Matrix Product Operator (MPO) \cite{schollwock2011dmrg}. We refer to TTM and MPO interchangeably, preferring TTM, and use MPO when the paper under discussion adopts that terminology.

\subsubsection{Block Term Decomposition}

The block term (BT) decomposition \cite{lathauwer2008btd} writes $\ten{X}$ as a sum of $K$ Tucker terms
\begin{equation}
\label{eq:btd}
\ten{X} \approx \sum_{k=1}^{K} \left[ \ten{G}_k\times_1\mat{A}_k^{(1)}\times_2\mat{A}_k^{(2)}\cdots\times_N\mat{A}_k^{(N)} \right],
\end{equation}
where each $\ten{G}_k\in\R^{R_{k1}\times R_{k2}\times \cdots\times R_{kN}}$ is a small core tensor and each $\mat{A}_k^{(n)}\in\R^{I_{n}\times R_{kn}}$ is a factor matrix associated with term $k$; the ranks $(R_{k1},R_{k2},\ldots,R_{kN})$ need not be shared across terms, so each term can have its own core size, in addition to its own core and factor matrices. In scalar form:
\begin{equation}
\label{eq:btd-scalar}
\ten{X}_{i_1, \dots, i_N} \approx \sum_{k=1}^{K} \left[ \sum_{l_1=1, \dots, l_N=1}^{R_{k1}, \ldots, R_{kN}} (\ten{G}_k)_{l_1, \dots, l_N}\,(\mat{A}_k^{(1)})_{i_1,l_1}\cdots(\mat{A}_k^{(N)})_{i_N,l_N} \right].
\end{equation}
When equality holds in Eqs. \eqref{eq:btd} and \eqref{eq:btd-scalar}, we refer to it as a \emph{BT representation} of $\ten{X}$. BT decomposition interpolates between CP and Tucker. Setting $K=1$ removes the outer sum and recovers the Tucker decomposition with core $\ten{G}_1$ and ranks $(R_{11},R_{12},\ldots,R_{1N})$, whereas setting $R_{kn}=1$ for every term $k$ and mode $n$ instead collapses every core $\ten{G}_k$ to a scalar $\lambda_k$, recovering the CP decomposition as a sum of $K$ rank-one terms.
Its uniqueness is more nuanced than that of the standard Tucker decomposition, as a BT decomposition carries the within-term ambiguity of Tucker in every block, and the $K$ blocks may also be permuted arbitrarily. Uniqueness beyond these indeterminacies requires additional conditions on the factor matrices, which are known for third-order tensors \cite{lathauwer2008btd}.
Its main feature is that, by choosing $K$ and the per-term ranks $(R_{k1},R_{k2},\ldots,R_{kN})$ jointly, BT decomposition can trade off CP's compactness against Tucker's per-mode flexibility, at the cost of more hyperparameters and a more involved fitting procedure than either decomposition alone \cite{lathauwer2008btd}.

\subsection{Attention Mechanism}

Attention mechanism is a parametric neural component that enables a model to selectively concentrate on salient features of an input sequence while processing each element of an output sequence. It overcomes the bottleneck of fixed-dimensional context vectors in RNNs by allowing direct access to all previous hidden states at each decoding step \cite{bahdanau2016attention}. Attention computes a context vector as a convex combination of values, with coefficients derived from a compatibility function between queries and keys.

\subsubsection{Single-Head Attention}

The most widely adopted formulation, introduced in the Transformer architecture \cite{vaswani2017attention}, is \textit{scaled dot-product attention}. Given a query matrix $\mat{Q} \in \R^{N \times d_k}$, a key matrix $\mat{K} \in \R^{M \times d_k}$, and a value matrix $\mat{V} \in \R^{M \times d_v}$, the attention output is computed as
\begin{equation}
\text{Attention}(\mat{Q}, \mat{K}, \mat{V}) = \text{Softmax}\left( \frac{\mat{Q} \mat{K}^\top}{\sqrt{d_k}} \right) \mat{V},
\label{eq:scaled_dot_product}
\end{equation}
where $N$ and $M$ denote the sequence lengths of queries and keys/values, respectively, and $d_k$, $d_v$ are the dimensionalities of keys and values. The scaling factor $\frac{1}{\sqrt{d_k}}$ normalizes the variance of the softmax input, stabilizing training.

Eq. \ref{eq:scaled_dot_product} is the general formulation, in which queries and keys may come from different sequences and differ in length. As our survey focuses on decoder-only LLMs, two restrictions apply throughout. First, all three matrices are derived from the same sequence of $T$ tokens, so $N = M = T$; this kind of attention is called \emph{self-attention}. Second, a token must not attend to its successors, which is enforced by adding a mask matrix $\mat{M} \in \R^{T \times T}$ with zeros on and below the diagonal and $-\infty$ entries above it:
\begin{equation}
\label{eq:masked_attention}
  \text{MaskedAttention}(\mat{Q}, \mat{K}, \mat{V}) = \text{Softmax}\left( \frac{\mat{Q} \mat{K}^\top}{\sqrt{d_k}} + \mat{M} \right) \mat{V},
\end{equation}
where now $\mat{Q}, \mat{K} \in \R^{T \times d_k}$ and $\mat{V} \in \R^{T \times d_v}$. The $-\infty$ entries vanish under the softmax, so the $i$-th output depends only on the tokens $1,2, \ldots, i$. This \emph{masked self-attention} is what makes autoregressive decoding possible, and it is the variant we assume from here on.

\subsubsection{Interpretation of the Attention Weights}

The Softmax operation is applied row-wise to produce a probability distribution over the keys for each query
\begin{equation}
\label{eq:att-score}
\alpha_{ij} = \frac{\exp\left( \vecb{q}_i^\top \vecb{k}_j / \sqrt{d_k} \right)}{\sum_{l=1}^{M} \exp\left( \vecb{q}_i^\top \vecb{k}_l / \sqrt{d_k} \right)},
\end{equation}
where $\alpha_{ij}$ represents the attention score assigned by the $i$-th query to the $j$-th key. The output for query $i$ is then:
\begin{equation}
\vecb{o}_i = \sum_{j=1}^{M} \alpha_{ij} \vecb{v}_j.
\label{eq:context_vector}
\end{equation}
In the masked case, the same expressions hold with the summations restricted to $j \le i$, since the mask sets the remaining weights to zero.
Each term can be understood with the following interpretation:
\begin{itemize}
    \item \textbf{Queries ($\mat{Q}$)}: Represent the current position for which we seek relevant context. In self-attention, queries are derived from the same input sequence as keys and values.
    \item \textbf{Keys ($\mat{K}$)}: Act as labels or identifiers for each element in the input sequence. The dot product $\mat{Q} \mat{K}^\top$ measures pairwise compatibility, or similarity, between each query and all keys.
    \item \textbf{Values ($\mat{V}$)}: Contain the actual information to be aggregated. The final output is a weighted sum of values, where the weighting is determined by the attention scores.
\end{itemize}
This formulation lets the model dynamically weigh the importance of each input element, creating a context-aware representation for each position in the output sequence.

\subsubsection{Multi-Head Attention}

To capture different types of relationships and dependencies simultaneously, the Transformer runs several attention operations in parallel, each with its own learned projection matrices. At this point, the input is no longer $\mat{Q}$, $\mat{K}$, and $\mat{V}$ themselves, but the hidden states from which they are projected: given $\mat{X}_Q \in \R^{N \times d}$ and $\mat{X}_K, \mat{X}_V \in \R^{M \times d}$, where $d$ is the model dimension
\begin{equation}
\text{MultiHeadAttention}(\mat{X}_Q, \mat{X}_K, \mat{X}_V) = \text{Concat}(\text{head}_1, \dots, \text{head}_H) \mat{W}^O,
\label{eq:multi_head}
\end{equation}
where each head is computed as
\begin{equation}
\text{head}_i = \text{Attention}(\mat{X}_Q \mat{W}_i^Q, \mat{X}_K \mat{W}_i^K, \mat{X}_V \mat{W}_i^V).
\label{eq:head}
\end{equation}
Here, $\mat{W}_i^Q \in \R^{d \times d_k}$, $\mat{W}_i^K \in \R^{d \times d_k}$, $\mat{W}_i^V \in \R^{d \times d_v}$ are projection matrices for the $i$-th head, and $\mat{W}^O \in \R^{H d_v \times d}$ projects the concatenated outputs back to the model dimension. Typically, $d_k = d_v = \frac{d}{H}$, where $H$ is the number of heads. We will denote $d_h = \frac{d}{H}$ as the head dimension, so $d_h = d_k = d_v$ in standard implementations.

In the decoder-only setting, the two restrictions introduced above carry over to every head. A single hidden-state matrix $\mat{X} \in \R^{T \times d}$ supplies all three projections, and each head applies masked self-attention,
\begin{equation}
\label{eq:mha-decoder}
\begin{gathered}
\text{MultiHeadAttention}(\mat{X}) = \text{Concat}(\text{head}_1, \dots, \text{head}_H) \mat{W}^O, \\
\text{head}_i = \text{MaskedAttention}(\mat{X}\mat{W}_i^Q, \mat{X} \mat{W}_i^K, \mat{X} \mat{W}_i^V).
\end{gathered}
\end{equation}
This combination, called \emph{multi-head masked self-attention}, is the standard attention mechanism of decoder-only LLMs, and we refer to it as MHA. In standard implementations, $d_k = d_v = d/H$, and we write $d_h = d/H$ for this common head dimension.
Some architectures let several query heads share one key-value projection. From here on, $H_{\rm Q}$ denotes the number of query heads and $H_{\rm KV}$ the number of distinct key-value heads; for MHA the two coincide, $H_{\rm Q} = H_{\rm KV} = H$.

\subsubsection{KV-cache}

Let us consider autoregressive decoding and observe that, within a head, the attention output at time step $t$ is Eq. \eqref{eq:context_vector} restricted to $j \le t$:
\begin{equation}
\label{eq:kv-output}
\vecb{o}_t = \sum_{j=1}^{t} \alpha_{tj} \vecb{v}_j .
\end{equation}
Of the queries, only $\vecb{q}_t$ enters this expression, through $\alpha_{tj}$ in Eq. \eqref{eq:att-score}, whereas the keys and values of all tokens $1, \ldots, t$ are required.
The keys and values of the first $t-1$ tokens have already been computed in previous steps, so they can be stored and reused instead of being recomputed. Before appending the current token's KV pair, an implementation holds exactly these $t-1$ cached entries. This technique is known as \textit{KV-cache} and is widely used in LLMs to speed up inference by avoiding redundant computations, at the cost of an increased memory footprint.
Reducing this footprint is the goal of multi-query attention (MQA) \cite{shazeer2019mqa}, which sets $H_{\rm KV} = 1$, and grouped-query attention (GQA) \cite{ainslie2023gqa}, which takes $1 < H_{\rm KV} < H_{\rm Q}$. Multi-head latent attention (MLA) \cite{deepseekai2024deepseekv2} instead caches one low-rank latent vector per token, from which the per-head keys and values are reconstructed.

\subsubsection{Computational and Memory Complexity}

Three regimes have to be distinguished. During \emph{training} the whole sequence is processed at once. At inference, the model first runs a single forward pass over the input prompt, called \textit{prefill}, and then generates tokens one at a time; each generated token requires one \textit{decode} step.
The cost of an attention layer splits into two contributions that scale differently across these regimes. The projection cost covers the computation of $\mat{Q}$, $\mat{K}$, $\mat{V}$, and the output projection $\mat{W}^O$, while the attention cost covers the formation of $\mat{Q}\mat{K}^\top$ and the aggregation of $\mat{V}$. Which dominates depends on whether the sequence is longer than the model is wide.
For memory, training and prefill must handle the attention matrix, which occupies $\mathcal{O}(T^2)$ memory if materialized fully. During a decode step, thanks to the KV cache, only the last row of this matrix is needed, which requires $\mathcal{O}(T)$ memory.
The cache itself, however, occupies $M_{\rm KV} = 2 L B T H_{\rm KV} d_h b$ bytes,
where $L$ is the number of layers, $B$ the batch size, and $b$ the number of bytes per scalar.
For full MHA one has $H_{\rm KV} d_h = d$, while MQA and GQA take $H_{\rm KV} < H_{\rm Q}$. This expression does not cover MLA, which caches a latent vector of dimension $d_c$ per token and layer instead of $H_{\rm KV}$ key-value heads.
\Cref{tab:attention-complexity} summarizes the computational cost and corresponding memory for one attention layer in the three regimes.
\begin{table}[h]
\centering
\caption{Per-sequence, per-layer complexity of one MHA block. $T$ denotes the sequence length, and in the decode column, the number of already cached tokens; $d$ is the model dimension, $H_{KV}$ is the number of key-value heads, and $d_h$ is the head dimension.}
\label{tab:attention-complexity}
\begin{tabular}{lccc}
\toprule
 & \multirow{2}{*}{Training} & \multicolumn{2}{c}{Inference} \\
\cmidrule(lr){3-4}
 & & Prefill & Decode (per step) \\
\midrule
Projection cost & $\mathcal{O}(T d^2)$ & $\mathcal{O}(T d^2)$ & $\mathcal{O}(d^2)$ \\
Attention cost & $\mathcal{O}(T^2 d)$ & $\mathcal{O}(T^2 d)$ & $\mathcal{O}(T d)$ \\
\midrule
Attention-matrix memory & $\mathcal{O}(T^2)$ & $\mathcal{O}(T^2)$ & $\mathcal{O}(T)$ \\
KV-cache memory & -- & $\mathcal{O}(T H_{\rm KV} d_h)$ & $\mathcal{O}(T H_{\rm KV} d_h)$ \\
\bottomrule
\end{tabular}
\end{table}

Training and prefill share the same asymptotics because both process the whole sequence at once; attention allows efficient parallel training on GPUs \cite{vaswani2017attention}, as all pairwise interactions and heads can be computed simultaneously. A decoding step, in contrast, processes a single token against the cache, which removes one factor of $T$ from both cost rows.
Consequently, training and prefill are commonly compute-bound, while decoding is commonly memory-bound \cite{yuan2024efficientinferencesurvey2}, since a decoding step performs only $\mathcal{O}(Td + d^2)$ operations but must read the whole cache from High Bandwidth Memory (HBM). This is typical, as actual behavior depends on batch size, context length, kernel implementation, numerical precision, and hardware.

\subsection{Feed-Forward Network}

The feed-forward network (FFN) is a critical component of the Transformer architecture and typically accounts for about two-thirds of the model's parameters \cite{geva2021ffnmemory}. In modern decoder-only LLMs, the FFN follows a gated architecture \cite{shazeer2020glu}:

\begin{equation}
\operatorname{FFN}(\vecb{x}) = \left[ \sigma(\vecb{x}^\top \mat{W}^{\text{gate}}) \odot \vecb{x}^\top \mat{W}^{\text{up}} \right] \mat{W}^{\text{down}},
\label{eq:ffn_gated}
\end{equation}
where $\mat{W}^{\text{gate}}, \mat{W}^{\text{up}} \in \R^{d \times d_{\text{ff}}}$ are the gate and up projections, $\mat{W}^{\text{down}} \in \R^{d_{\text{ff}} \times d}$ is the down projection, $\sigma(\cdot)$ denotes a non-linear activation function (commonly SiLU/$\text{Swish}_1$ \cite{ramachandran2017swish} or GELU \cite{hendrycks2023gelu}), and $\odot$ represents element-wise multiplication. The expansion factor $\frac{d_{\text{ff}}}{d}$ typically ranges from $2$ to $4$ in practice. The FFN is applied independently and identically at each sequence position; that is, the same weights are reused for every token, so, unlike attention, it does not mix information across positions.

Two properties make this component distinctive. First, it holds a large share of the weights, yet consists only of dense matrix multiplications. Since general matrix multiplication (GEMM) \cite{goto2008anatomy} is highly optimized for modern accelerator architectures such as GPUs, an FFN runs very fast despite its size, and any structured replacement must compete with that baseline. Second, FFNs have been argued to act as key-value memories \cite{geva2021ffnmemory} and to mediate some factual associations \cite{meng2023ffnfactuallocation}, making their factorization relevant to interpretability, not only efficiency.

\subsection{Overview of Tensor Decompositions for LLMs}

\Cref{tab:decompositions} compares the five formats along the properties discussed above. CP is the most compact and the only one with a uniqueness guarantee, making it the natural choice when the factors themselves are meant to carry meaning. Tucker assigns a separate rank to every mode and therefore suits tensors whose modes have distinct semantics, such as layers and heads, at the price of a core that grows with the order. TT and TTM keep the parameter count linear in the order and are the usual choice for large matrices, including embedding layers, projection matrices, and adapters; TTM additionally preserves the operator structure of the matrix it replaces. BT sits in between: richer than CP, but more structured than a single large Tucker core.

\begin{table}[!htbp]
\centering
\caption{\footnotesize Overview of the tensor decomposition formats used in language models: parameter count, main strengths and weaknesses, and representative use cases. The notation follows \cref{subsec:tensor-decompositions}.}
\label{tab:decompositions}
\footnotesize
\begin{tblr}{
  width = \textwidth,
  colspec = {Q[2.0cm,m,c]Q[2.75cm,m,c]Q[2.9cm,m,c]Q[2.9cm,m,c]X[m,c]},
  row{1} = {font=\bfseries},
  rowsep = 3pt,
  hline{1,7} = {1pt},
  hline{2} = {0.6pt},
  hline{3,4,5,6} = {0.6pt, gray7},
}
Decomposition & Number of parameters & Strengths & Weaknesses & Natural use cases in LLMs \\
CP \newline \cite{hitchcock1927cp, carroll1970analysis, harshman1970foundations} & $R \sum_n I_n$ & Most compact; unique under mild condition, hence interpretable factors & NP-hard CP-rank; a best approximation may not exist; unstable fitting & Compact embedding tables \cite{zhou2026tensorizingengram}; factored KV-cache \cite{zhang2026tpa} \\
Tucker \cite{tucker1966some} & $\prod_n R_n+\sum_n I_nR_n$ & Per-mode ranks; flexible compression & Core grows exponentially in the order $N$; gauge freedom leaves factors not directly interpretable & Cross-head factor sharing \cite{gu2025tensorllm, li2026lestd}; cross-head redundancy in the KV-cache \cite{klein2026tuckerattention} \\
TT \cite{oseledets2011tensor} & $\sum_n R_{n-1}I_nR_n$ & Parameters linear in the order $N$ & Gauge freedom; TT-rank depends on the chosen reshaping & Tensorized embedding layers \cite{xu2023tensorgpt}, projection matrices \cite{yang2024comera}, adapters \cite{anjum2024ttlora} \\
TTM \cite{oseledets2010ttm} & $\sum_n R_{n-1}I_nJ_nR_n$ & Preserves the operator structure & Same gauge freedom as TT & Tensorized embedding layers \cite{hrinchuk2020tensorized}, projection matrices \cite{chekalina2023ttm}, adapters \cite{hu2024dota} \\
BT \cite{lathauwer2008btd} & $\sum_k \Bigl[ \Pi_n R_{kn} + \sum_n I_{n} R_{kn} \Bigr]$ & Interpolates between CP compactness and Tucker flexibility & More hyperparameters ($K$ and per-term ranks) & Attention rewritten as a BT representation \cite{ma2019tensorized} \\
\end{tblr}
\end{table}


\section{Lifecycle Overview}
\label{sec:lifecycle-overview}

This section introduces the lifecycle taxonomy that organizes the survey. \Cref{subsec:taxonomy} defines the stages, their boundaries, and the rule for assigning methods that affect more than one stage. \Cref{subsec:illustrative-examples} then grounds the taxonomy with concrete examples based on LLaMA-3-8B. The lifecycle view is complemented by the component view in \cref{sec:components}, which asks which Transformer objects are tensorized, and by \cref{sec:lifecycle-analysis}, which formalizes the objective and reviews the literature at each stage.

\subsection{The Lifecycle Taxonomy}
\label{subsec:taxonomy}

We organize tensor methods according to the \emph{lifecycle} of a language model: tokenization, embeddings, pre-training, adaptation, compression, inference, and interpretability. Adaptation is represented primarily by parameter-efficient fine-tuning (PEFT), although the stage also includes other procedures that modify a pre-trained model for downstream use. The term \emph{lifecycle} identifies the point at which tensor structure is introduced, exploited, or analyzed; it does not imply that every model follows one irreversible linear sequence. For example, adaptation and compression may be repeated or applied in either order, and interpretability can be performed before, during, or after deployment.

We distinguish the stages using four attributes: (i) the \emph{intervention time} in the model lifecycle, (ii) the primary \emph{tensor object}, (iii) the \emph{objective} optimized by the method, and (iv) the \emph{evaluation criterion}. Tokenization acts on a discrete vocabulary and segmentation rule; embeddings act on continuous token representations; pre-training learns the base model parameters; adaptation learns task- or domain-specific updates to a pre-trained model; compression transforms an already trained model to reduce its resource footprint; inference operates on runtime objects such as attention factors and the KV cache; and interpretability analyzes, reformulates, or deliberately imposes multilinear structure to expose model mechanisms. These boundaries matter, since two methods may use the same decomposition while solving different optimization problems and requiring different evidence of success.


The lifecycle view and the component view are therefore orthogonal taxonomies. A method can be located by both \emph{when} it introduces tensorization and \emph{what} it acts on. For example, a TTM representation of an FFN weight may be learned during pre-training, fitted post hoc for compression, or used only for an adaptation update. For methods that span several stages, we assign the primary lifecycle label based on where the tensor constraint is introduced and which objective is optimized; we record downstream effects as cross-stage consequences. A complete method description should consequently specify at least the lifecycle stage, component or tensor object, tensorization scheme, decomposition family and ranks, training regime, and evaluation metrics. \Cref{fig:lifecycle} presents the taxonomy with representative works, while the corresponding subsections of \cref{sec:lifecycle-analysis} give formal problem definitions and broader literature coverage.
\begin{figure}[H]
\centering
\resizebox{\textwidth}{!}{%
\begin{tikzpicture}[
  stage/.style={
    draw=black!70, line width=1.1pt,
    rounded corners=7pt,
    fill=ColStage!55,
    minimum width=3.0cm, minimum height=0.75cm,
    font=\small\bfseries, align=center,
    inner sep=3pt, outer sep=0pt
  },
  paperbox/.style={
    draw=black!60, line width=0.75pt,
    rounded corners=4pt,
    fill=white,
    text width=2.90cm,
    font={\scriptsize\hyphenpenalty=10000\relax\exhyphenpenalty=10000\relax},
    align=left,
    inner sep=5pt, outer sep=0pt
  },
  nowork/.style={
    draw=black!40, line width=0.75pt,
    rounded corners=4pt,
    fill=ColNone!60,
    text width=2.90cm,
    font=\scriptsize,
    align=left,
    inner sep=5pt, outer sep=0pt
  },
  flow/.style={
    -{Stealth[length=6pt,width=4pt]},
    line width=1.0pt, black!70
  },
  drop/.style={
    -{Stealth[length=4pt,width=3pt]},
    line width=0.75pt, black!55
  },
  wrap/.style={
    -{Stealth[length=6pt,width=4pt]},
    line width=1.0pt, black!70,
    rounded corners=11pt
  },
]
\definecolor{ColStage}{RGB}{141,160,203}
\definecolor{ColNone}{RGB}{220,220,220}
\def\Dg{0.50}

\node[stage] (tok) at (2.25,0) {Tokenization};
\node[stage] (emb) at (6.30,0) {Embeddings};
\node[stage, minimum width=4.8cm] (pre) at (11.40,0) {Pre-training};
\draw[flow] (tok.east) -- (emb.west);
\draw[flow] (emb.east) -- (pre.west);

\node[nowork, below=\Dg cm of tok] (tok_p) {%
  \makebox[\linewidth][c]{\textbf{\S\ref{subsec:tokenization}}}\\[2pt]%
  \rule{\linewidth}{0.4pt}\\[2pt]%
  \makebox[\linewidth][c]{\textit{no published works}}%
};
\draw[drop] (tok.south) -- (tok_p.north);
\node[paperbox, below=\Dg cm of emb] (emb_p) {%
  \makebox[\linewidth][c]{\textbf{\S\ref{subsec:embeddings}}}\\[2pt]%
  \rule{\linewidth}{0.4pt}\\[2pt]%
  TT-embeddings~\cite{hrinchuk2020tensorized}, TensorGPT~\cite{xu2023tensorgpt}, TN-gram~\cite{zhou2026tensorizingengram}, 
  MorphTE~\cite{gan2022morphte}%
};
\draw[drop] (emb.south) -- (emb_p.north);
\node[paperbox, text width=5.0cm, below=\Dg cm of pre] (pre_p) {%
  \makebox[\linewidth][c]{\textbf{\S\ref{subsec:pre-training}}}\\[2pt]%
  \rule{\linewidth}{0.4pt}\\[2pt]%
  Tuformer~\cite{liu2022tuformer}, Multi-linear Attention~\cite{ma2019tensorized},
  GPT-TTM~\cite{chekalina2023ttm}, Hypoformer~\cite{li2022hypoformer}, CoMERA~\cite{yang2024comera},
  Shapeshifter~\cite{pahani2021shapeshifter}, Monarch~\cite{dao2022monarch}, BTT~\cite{qiu2024btt},
  BTT-MoE~\cite{potapczynski2024generaleinsum}%
};
\draw[drop] (pre.south) -- (pre_p.north);

\coordinate (secondrow) at ([yshift=-1.35cm]pre_p.south);
\node[stage, minimum width=4.6cm] (peft) at ({2.15},0 |- secondrow) {Adaptation};
\node[stage] (compr) at ({6.85},0 |- secondrow) {Compression};
\node[stage] (infer) at ({10.81},0 |- secondrow) {Inference};
\node[stage] (interp) at ({14.48},0 |- secondrow) {Interpretability};

\draw[flow] (peft.east) -- (compr.west);
\draw[flow] (compr.east) -- (infer.west);
\draw[flow] (infer.east) -- (interp.west);

\node[paperbox, text width=5.0cm, below=\Dg cm of peft] (peft_p) {%
  \makebox[\linewidth][c]{\textbf{\S\ref{subsec:peft}}}\\[2pt]%
  \rule{\linewidth}{0.4pt}\\[2pt]%
  KronA~\cite{edalati2022krona}, SaMA~\cite{nguyen2026sama}, DoTA / QDoTA~\cite{hu2024dota},
  LoRETTA~\cite{yang2024loretta}, ChainDoRA~\cite{yeafi2026chaindora}, TRAC~\cite{ye2026trac},
  TT-LoRA~\cite{anjum2024ttlora}, TeRA~\cite{gu2026tera}, QuanTA~\cite{chen2024quanta},
  Quantum-PEFT~\cite{koikeakino2025quantumpeft}, LoRTA~\cite{hounie2024lorta},
  LoTR~\cite{bershatsky2024lotr}, LoRA-CRAFT~\cite{dewage2026loracraft}, MetaTT~\cite{lopezpiqueres2025metatt}%
};
\draw[drop] (peft.south) -- (peft_p.north);
\node[paperbox, text width=3.25cm, below=\Dg cm of compr] (compr_p) {%
  \makebox[\linewidth][c]{\textbf{\S\ref{subsec:compression}}}\\[2pt]%
  \rule{\linewidth}{0.4pt}\\[2pt]%
  KnGPT~\cite{edalati2021kroneckergpt}, TQCompressor~\cite{abronin2024tqcompressor},
  TensorLLM~\cite{gu2025tensorllm}, LeSTD~\cite{li2026lestd}, CompactifAI~\cite{compactifai},
  Saten~\cite{solgi2025saten}, LatentLLM~\cite{koikeakino2025latentllm}%
};
\draw[drop] (compr.south) -- (compr_p.north);
\node[paperbox, text width=3.15cm, below=\Dg cm of infer] (infer_p) {%
  \makebox[\linewidth][c]{\textbf{\S\ref{subsec:inference}}}\\[2pt]%
  \rule{\linewidth}{0.4pt}\\[2pt]%
  TPA~\cite{zhang2026tpa}, KPA~\cite{nikpour2026kpa}, Tucker Attention~\cite{klein2026tuckerattention},
  Spectral Analysis~\cite{krishnan2026kvspectra}, JoLT~\cite{krishnan2026jolt},
  DecoQuant~\cite{liu2024decoquant}, EinSort~\cite{koikeakino2026einsort}%
};
\draw[drop] (infer.south) -- (infer_p.north);
\node[paperbox, below=\Dg cm of interp] (interp_p) {%
  \makebox[\linewidth][c]{\textbf{\S\ref{subsec:interpretability}}}\\[2pt]%
  \rule{\linewidth}{0.4pt}\\[2pt]%
  Tensor notation~\cite{taylor2024interpretabilitysurvey4}, Bilinear MLPs~\cite{pearce2025bilinearmlp},
  PolySAE~\cite{koromilas2026polysae}, TensorLens~\cite{atad2026tensorlens}%
};
\draw[drop] (interp.south) -- (interp_p.north);

\coordinate (turnright) at ([xshift=0.25cm]pre_p.east);
\coordinate (turnleft) at ([xshift=-0.25cm]peft_p.west);
\coordinate (rowgap) at ($(pre_p.south)!0.5!(peft.north)$);
\draw[wrap] (pre.east)
  -- (turnright |- pre.east) -- (turnright |- rowgap)
  -- (turnleft |- rowgap) -- (turnleft |- peft.west) -- (peft.west);
\end{tikzpicture}}
\caption{The seven-stage lifecycle taxonomy, with representative tensor methods at each stage. The arrows show a common model-development and deployment flow: stages may be revisited, reordered, or analyzed retrospectively. Each stage corresponds to a subsection of \cref{sec:lifecycle-analysis}.}
\label{fig:lifecycle}
\end{figure}

Each lifecycle stage exposes a different structured object and a different notion of success. \Cref{tab:lifecycle-stages} summarizes the primary tensorization target, expected benefit, and characteristic risk at each stage. The listed risk is stage-specific; cross-cutting issues, including rank selection, tensorization-shape sensitivity, optimization stability, kernel support, and the compression-realization gap, are discussed in \cref{subsec:gaps}.

\begin{table}[!htbp]
\centering
\caption{\footnotesize Lifecycle view of tensor methods: the primary tensorization target, expected benefit, and characteristic risk at each stage.}
\label{tab:lifecycle-stages}
\footnotesize
\begin{tblr}{
  width = \textwidth,
  colspec = {Q[2.2cm,m,c]Q[3.4cm,m,c]Q[4.0cm,m,c]X[m,c]},
  row{1} = {font=\bfseries},
  rowsep = 3pt,
  hline{1,9} = {1pt},
  hline{2} = {0.6pt},
  hline{3-8} = {0.6pt, gray7},
}
Stage & Tensorization target & Expected benefit & Key risk or cost \\
Tokenization & Vocabulary, text segmentation & Vocabulary structured by token relatedness \emph{(no published work)} & A discrete, non-differentiable map offers no object to factorize \\
Embeddings & Token representations & Compact vocabulary-scale embedding table & Less expressive token representations \\
Pre-training & Weights, optimizer states & Lower memory for weights and optimizer states; structural priors & Optimization stability; unknown scaling behavior \\
Adaptation (primarily PEFT) & Adaptation updates & Ultra-parameter-efficient fine-tuning & Unmerged adapters add contractions to every forward pass \\
Compression & Pre-trained weights & Memory reduction without full retraining & Quality loss and increased latency \\
Inference & KV-cache, attention factors & Per-token cache memory savings at long context & Compression that does not translate into decoding speedup \\
Interpretability & Activations, weights & Explicit multilinear structure; circuits as diagrams & Imposing multilinearity requires pre-training from scratch \\
\end{tblr}
\end{table}

The taxonomy also clarifies why decomposition names alone are insufficient for comparison. TT and TTM, for example, occur in embeddings, pre-training, adaptation, and compression, but their roles differ: they may define a trainable base-model parameterization, constrain an update, or approximate a fixed pre-trained weight. Conversely, one lifecycle stage may admit several decomposition families with different rank notions and contraction costs. Cross-stage dependencies further complicate evaluation: ranks selected during pre-training may constrain later adaptation capacity; post-training compression changes the kernels and memory traffic observed during inference; and interpretability can study either the original or tensorized model. Comparisons should therefore be made within a shared lifecycle objective and should report both stage-specific quality and system-level consequences.

\subsection{Illustrative Examples for $\llama$}
\label{subsec:illustrative-examples}

To ground the abstract lifecycle stages in concrete terms, this section walks through one representative method per stage, whereas a formal description and comprehensive comparison of methods at each stage appear in \cref{sec:lifecycle-analysis}.
Tokenization has no example since, although it remains part of the taxonomy, no published work tensorizes this stage (\cref{subsec:tokenization}); consequently, the walkthrough starts with embeddings.
Throughout, $\llama$ \cite{grattafiori2024llama3} serves as a running example, with configuration $|V|{=}128{,}000$, $d{=}4096$, $L{=}32$, $H_{\rm Q} = 32$, $H_{\rm KV} = 8$, $d_{\rm ff}{=}14336$, $d_h{=}128$, ${\approx}8.03$B parameters total. Not every stage has a published method targeting this exact model,so where necessary, we adapt the notationd accordingly.


\paragraph{Embeddings.}
\begin{wraptable}{r}{0.48\textwidth}
\centering
\resizebox{\linewidth}{!}{%
\begin{tikzpicture}[
  vec/.style={circle, draw=black!80, line width=1.0pt,
              minimum size=1.0cm, inner sep=0pt, font=\small},
  core/.style={circle, draw=black!80, line width=1.0pt,
              minimum size=0.75cm, inner sep=0pt, font=\scriptsize},
  bond/.style={line width=1.0pt, black!80},
  leg/.style={line width=1.0pt, black!80},
  lbl/.style={font=\small},
  slbl/.style={font=\scriptsize, text=black},
]
\definecolor{ColI} {RGB}{102,194,165}
\definecolor{ColII}{RGB}{252,141, 98}
\definecolor{ColV} {RGB}{141,160,203}
\pgfmathsetmacro{\RadCirc}{0.50}
\pgfmathsetmacro{\RadCore}{0.375}
\pgfmathsetmacro{\Dy}{1.65}
\pgfmathsetmacro{\Dx}{1.15}
\pgfmathsetmacro{\Lleg}{0.50}
\pgfmathsetmacro{\Llegv}{0.80}
\pgfmathsetmacro{\PadX}{0.3}
\pgfmathsetmacro{\PadY}{0.75}
\pgfmathsetmacro{\Lcx}{0.0}
\pgfmathsetmacro{\Rx}{5.0}
\pgfmathsetmacro{\Ytop}{2.8}
\pgfmathsetmacro{\Ybot}{-2.8}
\pgfmathsetmacro{\Yone}{\Ytop - \PadY}
\pgfmathsetmacro{\Ytwo}{\Yone - \Dy}
\pgfmathsetmacro{\YV}{\Ybot + \PadY}
\pgfmathsetmacro{\Ymid}{0.5*(\Ytop+\Ybot)}
\pgfmathsetmacro{\Yelps}{0.5*(\Ytwo+\YV)}
\pgfmathsetmacro{\Rshift}{-0.5*\RadCore}
\pgfmathsetmacro{\RYone}{\Yone - \Rshift}
\pgfmathsetmacro{\RYtwo}{\Ytwo - \Rshift}
\pgfmathsetmacro{\RYV}{\YV   - \Rshift}
\pgfmathsetmacro{\LboxL}{\Lcx - \RadCirc - \PadX}
\pgfmathsetmacro{\LboxR}{\Lcx + \RadCirc + \Llegv + \PadX}
\pgfmathsetmacro{\RboxL}{\Rx - \RadCore - \PadX}
\pgfmathsetmacro{\RboxR}{\Rx + 3*\Dx + \RadCore + \PadX}
\pgfmathsetmacro{\ArrLen}{1.0}
\pgfmathsetmacro{\ArrMid}{0.5*(\LboxR + \RboxL)}
\pgfmathsetmacro{\ArrL}{\ArrMid - 0.5*\ArrLen}
\pgfmathsetmacro{\ArrR}{\ArrMid + 0.5*\ArrLen}
\pgfmathsetmacro{\ABH}{0.10}
\pgfmathsetmacro{\AHH}{0.22}
\pgfmathsetmacro{\AHD}{0.26}
\node[vec, fill=ColI!45]  (E1) at (\Lcx, \Yone) {$\mathbf{e}_1$};
\node[vec, fill=ColII!45] (E2) at (\Lcx, \Ytwo) {$\mathbf{e}_2$};
\node[lbl]                     at (\Lcx, \Yelps) {$\vdots$};
\node[vec, fill=ColV!45]  (EV) at (\Lcx, \YV)   {$\mathbf{e}_V$};
\foreach \N in {E1, E2, EV}{
  \draw[leg] (\N.east) -- ++(\Llegv, 0) node[above, slbl] {$d$};
}
\fill[white, draw=black!80, line width=0.9pt]
    (\ArrL, {\Ymid+\ABH}) -- ({\ArrR-\AHD}, {\Ymid+\ABH})
    -- ({\ArrR-\AHD}, {\Ymid+\AHH}) -- (\ArrR, \Ymid)
    -- ({\ArrR-\AHD}, {\Ymid-\AHH}) -- ({\ArrR-\AHD}, {\Ymid-\ABH})
    -- (\ArrL, {\Ymid-\ABH}) -- cycle;
\foreach \k in {1,2,3,4}{
  \node[core, fill=ColI!45] (R1\k) at ({\Rx+(\k-1)*\Dx}, \RYone) {$\ten{G}^{(1)}_\k$};
}
\foreach \k/\kk in {1/2, 2/3, 3/4}{
  \draw[bond] (R1\k) -- (R1\kk) node[midway, above=1pt, slbl] {$R_\k$};
}
\foreach \k in {1,2,3,4}{
  \draw[leg] (R1\k.south) -- ++(0, -\Lleg) node[right=1pt, slbl] {$d_\k$};
}
\foreach \k in {1,2,3,4}{
  \node[core, fill=ColII!45] (R2\k) at ({\Rx+(\k-1)*\Dx}, \RYtwo) {$\ten{G}^{(2)}_\k$};
}
\foreach \k/\kk in {1/2, 2/3, 3/4}{
  \draw[bond] (R2\k) -- (R2\kk) node[midway, above=1pt, slbl] {$R_\k$};
}
\foreach \k in {1,2,3,4}{
  \draw[leg] (R2\k.south) -- ++(0, -\Lleg) node[right=1pt, slbl] {$d_\k$};
}
\node[lbl] at ({\Rx + 1.5*\Dx}, \Yelps) {$\vdots$};
\foreach \k in {1,2,3,4}{
  \node[core, fill=ColV!45] (RV\k) at ({\Rx+(\k-1)*\Dx}, \RYV) {$\ten{G}^{(V)}_\k$};
}
\foreach \k/\kk in {1/2, 2/3, 3/4}{
  \draw[bond] (RV\k) -- (RV\kk) node[midway, above=1pt, slbl] {$R_\k$};
}
\foreach \k in {1,2,3,4}{
  \draw[leg] (RV\k.south) -- ++(0, -\Lleg) node[right=1pt, slbl] {$d_\k$};
}
\begin{pgfonlayer}{foreground}
\node[font=\small\bfseries] at ({0.5*(\LboxL+\RboxR)}, \Ytop + 0.5)
    {Per-Row TT Compression of the Embedding Table};
\pgfmathsetmacro{\Sw}{0.30}
\draw[black, line width=1.3pt, line cap=rect]
    ({\LboxL+\Sw}, \Ytop) -- (\LboxL, \Ytop) -- (\LboxL, \Ybot) -- ({\LboxL+\Sw}, \Ybot);
\draw[black, line width=1.3pt, line cap=rect]
    ({\LboxR-\Sw}, \Ytop) -- (\LboxR, \Ytop) -- (\LboxR, \Ybot) -- ({\LboxR-\Sw}, \Ybot);
\node[slbl] at (\ArrMid, \Ymid + 0.75) {Reshape};
\node[slbl] at (\ArrMid, \Ymid + 0.45) {$+$ TT-SVD};
\draw[black, line width=1.3pt, line cap=rect]
    ({\RboxL+\Sw}, \Ytop) -- (\RboxL, \Ytop) -- (\RboxL, \Ybot) -- ({\RboxL+\Sw}, \Ybot);
\draw[black, line width=1.3pt, line cap=rect]
    ({\RboxR-\Sw}, \Ytop) -- (\RboxR, \Ytop) -- (\RboxR, \Ybot) -- ({\RboxR-\Sw}, \Ybot);
\end{pgfonlayer}
\end{tikzpicture}}
\captionof{figure}{\footnotesize Each embedding row $\vecb{e}_i$ is reshaped into a tensor $\ten{E}^{(i)}$ and approximated by TT cores $\ten{G}^{(i)}_n \in \R^{R_{n-1} \times d_n \times R_n}$ via TT-SVD \cite{oseledets2011tensor}.}
\label{fig:example-embeddings}
\end{wraptable}
One way to tensorize an embedding table is per-row factorization: each row is reshaped into a higher-order tensor and decomposed independently. TensorGPT \cite{xu2023tensorgpt} compresses each token embedding individually. This preserves row independence, which suits the lookup table nature of the embedding layer and keeps the scheme valid under a changing vocabulary. In $\llama$, the embedding table $\mat{E}\in\R^{128000\times 4096}$ holds roughly 525M parameters, and since the model does not tie input and output embeddings \cite{wolf2017weighttying1, inan2017weighttying2}, the table alone accounts for about $6.5\%$ of all parameters. The feature axis factors cleanly as $d = 4096 = 8^4$, so each row becomes a fourth-order tensor and admits a TT decomposition with TT-rank $(1, R_1, R_2, R_3, 1)$, as illustrated in \cref{fig:example-embeddings}. The technique is training-free, as TT-SVD \cite{oseledets2011tensor} is applied directly to the pre-trained embeddings. With $R_1 = R_2 = R_3 = R$, the whole table costs $128{,}000 \times 16R(1+R) \approx 2R(1+R) \times 10^6$ parameters. The rank governs the trade-off: $R=1$ compresses the table by $128{\times}$, while $R=4$ gives $12.8{\times}$ and retains more of the original embeddings, since the TT-SVD truncation error decreases as the rank grows. The empirical quality cost of such compression is reported in \cite{xu2023tensorgpt}.

\paragraph{Pre-training.}
Training a language model with tensor-structured weight matrices from scratch is the most direct way to use tensor structure as an inductive bias.
One natural approach is to parameterize each large dense weight matrix with a TTM representation before training \cite{chekalina2023ttm}. The dense matrix is then never materialized. Instead, the input $\vecb{x}$ is reshaped into a tensor and contracted directly with the TTM cores, so the weights and their optimizer states scale with the cores.
As a concrete illustration, consider applying the TTM representation to the down-projection $\mat{W}\in\R^{14336\times 4096}$ of one FFN block in $\llama$, whose input and output dimensions factor as $14336 = 8\times 14\times 16\times 8$ and $4096 = 8^4$, respectively. As shown in \cref{fig:tt-ttm-tensor-diagram} (right), this yields a TTM chain of four cores
\begin{equation*}
\ten{G}^{(1)}\in\R^{1\times 8\times 8\times R_1},\quad
\ten{G}^{(2)}\in\R^{R_1\times 14\times 8\times R_2},\quad
\ten{G}^{(3)}\in\R^{R_2\times 16\times 8\times R_3},\quad
\ten{G}^{(4)}\in\R^{R_3\times 8\times 8\times 1}.
\end{equation*}
If we take $R_1 = R_2 = R_3 = R$, the total parameter count is $128R + 240R^2$, compared with $14336\times 4096\approx 58.7\text{M}$ for the dense matrix. At rank $R=64$, for instance, the TTM stores approximately 1M parameters, corresponding to a compression ratio of roughly $60{\times}$ for this single projection. Chekalina et al. \cite{chekalina2023ttm} demonstrate that training GPT-2 with such TTM-parametrized layers (at a sufficiently high rank) incurs only a small perplexity increase, suggesting that the dense parameterization is not necessary to reach this quality level


\begin{figure}[!htpb]
\centering
\resizebox{\textwidth}{!}{%
\begin{tikzpicture}[
  mat/.style={circle, draw=black!80, line width=1.0pt,
              minimum size=1.60cm, inner sep=0pt, font=\normalsize},
  core/.style={circle, draw=black!80, line width=1.0pt,
              minimum size=1.05cm, inner sep=0pt, font=\small},
  fac/.style={circle, draw=black!80, line width=1.0pt,
              minimum size=1.05cm, inner sep=0pt, font=\tiny},
  bond/.style={line width=1.0pt, black!80},
  leg/.style={line width=1.0pt, black!80},
  lbl/.style={font=\small},
  slbl/.style={font=\scriptsize, text=black},
]

\definecolor{ColW}{RGB}{141,160,203}
\definecolor{ColA}{RGB}{102,194,165}
\definecolor{ColB}{RGB}{252,141, 98}
\pgfmathsetmacro{\RadMat}{0.80}
\pgfmathsetmacro{\RadCore}{0.525}
\pgfmathsetmacro{\Lleg}{0.72}
\pgfmathsetmacro{\PadX}{0.32}
\pgfmathsetmacro{\PadY}{0.45}
\pgfmathsetmacro{\LWx}{0.0}
\pgfmathsetmacro{\LWy}{0.0}
\pgfmathsetmacro{\DyAB}{1.70}
\pgfmathsetmacro{\DyABH}{\DyAB/2}
\pgfmathsetmacro{\NegDyABH}{-\DyAB/2}
\pgfmathsetmacro{\FanHalf}{\DyABH + \RadCore}
\pgfmathsetmacro{\NegFanHalf}{-\DyABH - \RadCore}
\pgfmathsetmacro{\FanDepth}{0.72}
\pgfmathsetmacro{\BondLen}{0.58}
\pgfmathsetmacro{\LFanX}{4.40}
\pgfmathsetmacro{\LFanBX}{\LFanX + \FanDepth}
\pgfmathsetmacro{\CoreX}{\LFanBX + \BondLen + \RadCore}
\pgfmathsetmacro{\RFanBX}{\CoreX + \RadCore + \BondLen}
\pgfmathsetmacro{\RFanX}{\RFanBX + \FanDepth}
\pgfmathsetmacro{\LboxL}{\LWx - \RadMat - \Lleg - \PadX}
\pgfmathsetmacro{\LboxR}{\LWx + \RadMat + \Lleg + \PadX}
\pgfmathsetmacro{\RboxL}{\LFanX - \Lleg - \PadX}
\pgfmathsetmacro{\RboxR}{\RFanX + \Lleg + \PadX}
\pgfmathsetmacro{\PlusX}{0.5*(\LboxR + \RboxL)}
\node[mat, fill=ColW!40] (W) at (\LWx, \LWy) {$\mat{\Delta W}$};
\draw[leg] (W.west) -- ++(-\Lleg, 0) node[above=1pt, slbl] {$I_1I_2$};
\draw[leg] (W.east) -- ++(\Lleg,  0) node[above=1pt, slbl] {$J_1J_2$};
\node[font=\large] at (\PlusX, \LWy) {$=$};
\draw[leg] (\LFanX, 0) -- ++(-\Lleg, 0) node[above=1pt, slbl] {$I_1I_2$};
\draw[leg] (\RFanX, 0) -- ++( \Lleg, 0) node[above=1pt, slbl] {$J_1J_2$};
\fill[black!9, draw=black!80, line width=1.0pt]
    (\LFanBX, \FanHalf)
    arc[start angle=90, end angle=270, x radius=\FanDepth, y radius=\FanHalf]
    -- cycle;
\fill[black!9, draw=black!80, line width=1.0pt]
    (\RFanBX, \NegFanHalf)
    arc[start angle=-90, end angle=90, x radius=\FanDepth, y radius=\FanHalf]
    -- cycle;
\node[core, fill=ColA!40] (A) at (\CoreX, \DyABH)    {$\mat{A}$};
\node[core, fill=ColB!40] (B) at (\CoreX, \NegDyABH) {$\mat{B}$};
\draw[bond] (\LFanBX, \DyABH)    -- (A.west) node[midway, above=1pt, slbl] {$I_1$};
\draw[bond] (\LFanBX, \NegDyABH) -- (B.west) node[midway, above=1pt, slbl] {$I_2$};
\draw[bond] (A.east) -- (\RFanBX, \DyABH) node[midway, above=1pt, slbl] {$J_1$};
\draw[bond] (B.east) -- (\RFanBX, \NegDyABH) node[midway, above=1pt, slbl] {$J_2$};
\begin{pgfonlayer}{foreground}
\node[font=\small\bfseries] at ({0.5*(\LboxL+\RboxR)}, 2.57)
    {Kronecker Parameterization of the Adapter Update};
\path (0, 2.77) -- (0, -3.25);
\end{pgfonlayer}

\draw[black!35, line width=0.6pt] (9.59, -3.25) -- (9.59, 2.77);
\begin{scope}[shift={(11.93, 0)}]
\tikzset{
  core/.style={circle, draw=black!80, line width=1.0pt,
              minimum size=1.80cm, inner sep=0pt, font=\small},
  fac/.style={circle, draw=black!80, line width=1.0pt,
              minimum size=1.05cm, inner sep=0pt, font=\tiny},
}
\definecolor{ColG}{RGB}{252,141, 98}
\definecolor{ColU}{RGB}{102,194,165}
\pgfmathsetmacro{\RadW}{0.80}
\pgfmathsetmacro{\RadG}{0.90}
\pgfmathsetmacro{\RadU}{0.525}
\pgfmathsetmacro{\Lleg}{0.72}
\pgfmathsetmacro{\PadX}{0.32}
\pgfmathsetmacro{\PadY}{0.45}
\pgfmathsetmacro{\LWx}{0.0}
\pgfmathsetmacro{\LWy}{0.0}
\pgfmathsetmacro{\Rx}{6.95}
\pgfmathsetmacro{\Gy}{0.0}
\pgfmathsetmacro{\DxU}{1.80}
\pgfmathsetmacro{\DyU}{1.80}
\pgfmathsetmacro{\DiagHalf}{0.58}
\pgfmathsetmacro{\DashN}{9}
\pgfmathsetmacro{\DashLen}{2*\DiagHalf*sqrt(2)*28.4528/(2*\DashN-1)}
\pgfmathsetmacro{\ArrLen}{1.0}
\pgfmathsetmacro{\ArrMid}{0.5*((\LWx + \RadW + \Lleg + \PadX) + (\Rx - \DxU - \RadU - \Lleg - \PadX))}
\pgfmathsetmacro{\ArrL}{\ArrMid - 0.5*\ArrLen}
\pgfmathsetmacro{\ArrR}{\ArrMid + 0.5*\ArrLen}
\pgfmathsetmacro{\ABH}{0.10}
\pgfmathsetmacro{\AHH}{0.22}
\pgfmathsetmacro{\AHD}{0.26}
\pgfmathsetmacro{\LboxL}{\LWx - \RadW - \Lleg - \PadX}
\pgfmathsetmacro{\LboxR}{\LWx + \RadW + \Lleg + \PadX}
\pgfmathsetmacro{\LboxT}{\LWy + \RadW + \Lleg + \PadY}
\pgfmathsetmacro{\LboxB}{\LWy - \RadW - \Lleg - \PadY}
\pgfmathsetmacro{\RboxL}{\Rx - \DxU - \RadU - \Lleg - \PadX}
\pgfmathsetmacro{\RboxR}{\Rx + \DxU + \RadU + \Lleg + \PadX}
\pgfmathsetmacro{\RboxT}{\Gy + \RadG + \Lleg + \PadY}
\pgfmathsetmacro{\RboxB}{\Gy - \DyU - \RadU - \Lleg - \PadY}
\pgfmathsetmacro{\BoxT}{max(\LboxT, \RboxT)}
\pgfmathsetmacro{\BoxB}{min(\LboxB, \RboxB)}
\node[mat, fill=ColW!40] (W) at (\LWx, \LWy) {$\ten{W}$};
\draw[leg] (W.west)  -- ++(-\Lleg,  0) node[above=1pt, slbl] {$d$};
\draw[leg] (W.east)  -- ++( \Lleg,  0) node[above=1pt, slbl] {$d_h$};
\draw[leg] (W.north) -- ++(0,  \Lleg)  node[right=1pt, slbl] {$H$};
\draw[leg] (W.south) -- ++(0, -\Lleg)  node[right=1pt, slbl] {$4$};
\fill[white, draw=black!80, line width=0.9pt]
    (\ArrL,            {\LWy+\ABH})
    -- ({\ArrR-\AHD},  {\LWy+\ABH})
    -- ({\ArrR-\AHD},  {\LWy+\AHH})
    -- (\ArrR,          \LWy)
    -- ({\ArrR-\AHD},  {\LWy-\AHH})
    -- ({\ArrR-\AHD},  {\LWy-\ABH})
    -- (\ArrL,         {\LWy-\ABH})
    -- cycle;
\node[slbl] at (\ArrMid, \LWy + 0.75) {Tucker};
\node[slbl] at (\ArrMid, \LWy + 0.45) {$+$ sparsity};
\node[core, fill=ColG!35] (G) at (\Rx, \Gy) {};
\node[font=\small] at (\Rx + 0.24, {\Gy + 0.36}) {$\ten{G}$};
\draw[black!60, line width=1.0pt,
      dash pattern=on \DashLen pt off \DashLen pt]
    ({\Rx - \DiagHalf}, {\Gy + \DiagHalf}) -- ({\Rx + \DiagHalf}, {\Gy - \DiagHalf});
\draw[leg] (G.north) -- ++(0, \Lleg) node[right=1pt, slbl] {$H$};
\node[fac, fill=ColU!40] (U1) at ({\Rx - \DxU}, \Gy)         {$\mat{U}^{(1)}$};
\node[fac, fill=ColU!40] (U2) at ({\Rx + \DxU}, \Gy)         {$\mat{U}^{(2)}$};
\node[fac, fill=ColU!40] (U3) at (\Rx,          {\Gy-\DyU})  {$\mat{U}^{(3)}$};
\draw[bond] (G.west)  -- (U1.east)  node[midway, above=1pt, slbl] {$R_1$};
\draw[bond] (G.east)  -- (U2.west)  node[midway, above=1pt, slbl] {$R_2$};
\draw[bond] (G.south) -- (U3.north) node[midway, right=1pt, slbl] {$R_3$};
\draw[leg] (U1.west)  -- ++(-\Lleg,  0) node[above=1pt, slbl] {$d$};
\draw[leg] (U2.east)  -- ++( \Lleg,  0) node[above=1pt, slbl] {$d_h$};
\draw[leg] (U3.south) -- ++(0, -\Lleg)  node[right=1pt, slbl] {$4$};
\begin{pgfonlayer}{foreground}
\node[font=\small\bfseries] at ({0.5*(\LboxL+\RboxR)}, \BoxT + 0.50)
    {Tucker Compression with a Sparse Core};
\end{pgfonlayer}
\end{scope}
\end{tikzpicture}}
\caption{\textit{Left}: the weight update is parameterized as a Kronecker product of two factor matrices, $\mat{A}\in\R^{I_1\times J_1}$ and $\mat{B}\in\R^{I_2\times J_2}$, so that $\Delta\mat{W}\in\R^{I_1I_2\times J_1J_2}$. \textit{Right}: the attention projections of one layer are stacked into a fourth-order tensor $\ten{W}\in\R^{d\times d_h\times 4\times H}$, whose modes index, in order, the model dimension $d$, the head dimension $d_h$, the projection type ($\mat{W}^Q,\mat{W}^K,\mat{W}^V,\mat{W}^O$), and the $H$ attention heads. The tensor is approximated by a Tucker decomposition with core $\ten{G}\in\R^{R_1\times R_2\times R_3\times H}$ and factor matrices $\mat{U}^{(1)}\in\R^{d\times R_1}$, $\mat{U}^{(2)}\in\R^{d_h\times R_2}$, $\mat{U}^{(3)}\in\R^{4\times R_3}$, leaving the head mode uncompressed. The core is then sparsified, as indicated by the dashed diagonal.}\label{fig:ex-peft-compr}
\end{figure}

\paragraph{Adaptation.}
At the adaptation stage, the incremental weight update $\Delta \mat{W}$ is either parameterized directly as a tensor network or structured through tensor-algebraic operations such as the Kronecker product. Instead of adding a low-rank matrix update $\mat{U}\mat{V}^\top \in\R^{I\times J}$ \cite{hu2022lora} with $\mat{U}\in \R^{I \times R}$ and $\mat{V} \in \R^{J \times R}$, KronA \cite{edalati2022krona} applies a Kronecker product increment $\mat{W}_0 \leftarrow \mat{W}_0 + \alpha(\mat{A}\otimes\mat{B})$, where $\mat{A}\in\R^{I_1\times J_1}$, $\mat{B}\in\R^{I_2\times J_2}$, with scaling $\alpha$, and $I_1 I_2 = I$, $J_1 J_2 = J$. \Cref{fig:ex-peft-compr} (left) shows the corresponding tensor network diagram. By the standard identity $\mathrm{rank}(\mat{A}\otimes\mat{B})=\mathrm{rank}(\mat{A})\cdot\mathrm{rank}(\mat{B})$ \cite{golubvanloan}, a single Kronecker term reaches matrix rank up to $\min(I_1,J_1)\cdot\min(I_2,J_2)$ at a cost of $I_1J_1+I_2J_2$ parameters. A LoRA update of rank $R$, by contrast, has matrix rank up to $R$ and costs $R(I+J)$ parameters. Thus, its rank grows linearly with the parameter budget, whereas the Kronecker form multiplies the ranks of its factors while only adding their sizes.
The gap is concrete for $\llama$. For the per-head query projection $\mat{W}_Q\in\R^{4096\times 128}$, the factorization $(I_1,J_1)=(64,16)$, $(I_2,J_2)=(64,8)$ gives a single-term KronA adapter of $1{,}536$ parameters whose rank can reach $128$, the full column rank. LoRA spends $4{,}224$ parameters per unit of rank here, so the same rank costs $540$K, and even $R=1$ costs nearly three times the whole Kronecker adapter. The comparison concerns attainable rank rather than attainable updates, since Kronecker terms span only a structured subset of the matrices of that rank, while the low-rank factorization is unconstrained.

\paragraph{Compression.}
At the compression stage, tensor decompositions are applied post-hoc to pretrained weights to reduce their memory footprint, exploiting the low-rank structure that emerges in trained networks. LeSTD \cite{li2026lestd} targets the MHA block as a whole. For each head, the four projections, namely query, key, value, and output, are stacked into a third-order tensor, and the per-head tensors are stacked along a fourth mode indexing the heads. Compression is data-free and proceeds in two stages. In the first stage, a Tucker decomposition over the first three modes is computed, so all heads share the factor matrices $\mat{U}^{(1)},\mat{U}^{(2)},\mat{U}^{(3)}$ while head-specific information stays in the corresponding slice of the core $\ten{G}\in\R^{R_1\times R_2\times R_3\times H}$; the head mode is left uncompressed. In the second stage, the core is sparsified by an importance score. \Cref{fig:ex-peft-compr} (right) illustrates both transformations jointly. In $\llama$, GQA prevents a single head mode, given that $H_{\rm Q}=32$ whereas $H_{KV}=8$, so the construction could be split into $\ten{W}_{\rm QO}\in\R^{4096\times128\times2\times32}$ and $\ten{W}_{\rm KV}\in\R^{4096\times128\times2\times8}$, holding $33.6$M and $8.4$M parameters. The smaller $K/V$ tensor uses the lower mode-1 rank since its mode-1 factor would otherwise dominate. With $70\%$ core sparsity and Tucker ranks $(1024,64,2,H_{\rm Q})$ and $(256,64,2,H_{\rm KV})$, the two shrink to $5.5$M and $1.1$M, a factor of $6.4\times$ for the block. What remains is dominated by the shared factors $\mat{U}^{(1)}$ ($5.2$M of $6.6$M), so the benefit grows with the number of heads because these factors are amortized across heads, while only the core scales with $H$. GQA works against this, since the $K/V$ tensor offers only eight heads to amortize over.


\paragraph{Inference.}
\begin{wrapfigure}{r}{0.48\textwidth}
\centering
\resizebox{\linewidth}{!}{%
\begin{tikzpicture}[scale=1.4,
  mat/.style={circle, draw=black!80, line width=1.0pt,
              minimum size=1.61cm, inner sep=0pt, font=\small},
  core/.style={circle, draw=black!80, line width=1.0pt,
              minimum size=1.26cm, inner sep=0pt, font=\small},
  bond/.style={line width=1.0pt, black!80},
  leg/.style={line width=1.0pt, black!80},
  slbl/.style={font=\scriptsize, text=black},
]
\definecolor{ColK}{RGB}{141,160,203}
\definecolor{ColA}{RGB}{102,194,165}
\definecolor{ColB}{RGB}{252,141, 98}
\pgfmathsetmacro{\RadMat}{0.575}
\pgfmathsetmacro{\RadCore}{0.45}
\pgfmathsetmacro{\Lleg}{0.65}
\pgfmathsetmacro{\LlegS}{0.40}
\pgfmathsetmacro{\PadX}{0.30}
\pgfmathsetmacro{\PadXR}{0.60}
\pgfmathsetmacro{\PadY}{0.30}
\pgfmathsetmacro{\ABGap}{1.80}
\pgfmathsetmacro{\LQx}{0.0}
\pgfmathsetmacro{\LPanY}{0.0}
\pgfmathsetmacro{\LRx}{3.0}
\pgfmathsetmacro{\BoxRowA}{\LPanY + 1.45}
\pgfmathsetmacro{\BoxRowB}{\LPanY - 1.45}
\pgfmathsetmacro{\BoxL}{\LRx - \RadCore - \PadX}
\pgfmathsetmacro{\BoxR}{\LRx + \ABGap + \RadCore + \Lleg + \PadXR}
\pgfmathsetmacro{\BoxTop}{\BoxRowA + \RadCore + \PadY}
\pgfmathsetmacro{\BoxBot}{\BoxRowB - \RadCore - \LlegS - \PadY}
\pgfmathsetmacro{\LboxR}{\LQx + \RadMat + \Lleg + \PadX}
\pgfmathsetmacro{\EqXL}{0.5*(\LboxR + \BoxL)}
\pgfmathsetmacro{\FigL}{\LQx - \RadMat - \Lleg - \PadX}
\pgfmathsetmacro{\Sw}{0.30}
\begin{pgfonlayer}{background}
\draw[black, line width=1.3pt, line cap=rect]
    ({\BoxL+\Sw}, \BoxTop) -- (\BoxL, \BoxTop) -- (\BoxL, \BoxBot) -- ({\BoxL+\Sw}, \BoxBot);
\draw[black, line width=1.3pt, line cap=rect]
    ({\BoxR-\Sw}, \BoxTop) -- (\BoxR, \BoxTop) -- (\BoxR, \BoxBot) -- ({\BoxR-\Sw}, \BoxBot);
\end{pgfonlayer}
\node[mat, fill=ColK!40] (K) at (\LQx, \LPanY) {$\ten{K}$};
\draw[leg] (K.west)  -- ++(-\Lleg, 0) node[above=1pt, slbl] {$T$};
\draw[leg] (K.south) -- ++(0, -\Lleg) node[right=1pt, slbl] {$H$};
\draw[leg] (K.east)  -- ++(\Lleg, 0)  node[above=1pt, slbl] {$d_h$};
\node[font=\normalsize] at (\EqXL, \LPanY) {$=$};
\node[core, fill=ColA!40] (A1) at (\LRx,          \BoxRowA) {$\mat{A}_1$};
\node[core, fill=ColB!40] (B1) at ({\LRx+\ABGap}, \BoxRowA) {$\mat{B}_1$};
\draw[bond] (A1.east) -- (B1.west) node[midway, above=2pt, slbl] {$R_{\rm K}$};
\draw[leg]  (A1.south) -- ++(0, -\LlegS) node[right=1pt, slbl] {$H$};
\draw[leg]  (B1.east)  -- ++(\Lleg, 0)   node[above=1pt, slbl] {$d_h$};
\node[font=\large] at ({0.5*(2*\LRx + \ABGap)}, \LPanY) {$\vdots$};
\node[core, fill=ColA!40] (AT) at (\LRx,          \BoxRowB) {$\mat{A}_T$};
\node[core, fill=ColB!40] (BT) at ({\LRx+\ABGap}, \BoxRowB) {$\mat{B}_T$};
\draw[bond] (AT.east) -- (BT.west) node[midway, above=2pt, slbl] {$R_{\rm K}$};
\draw[leg]  (AT.south) -- ++(0, -\LlegS) node[right=1pt, slbl] {$H$};
\draw[leg]  (BT.east)  -- ++(\Lleg, 0)   node[above=1pt, slbl] {$d_h$};
\begin{pgfonlayer}{foreground}
\node[font=\small\bfseries]
    at ({0.5*(\FigL + \BoxR)}, {\BoxTop + 0.60}) {Key Tensor in TPA};
\end{pgfonlayer}
\end{tikzpicture}}
\caption{\footnotesize In TPA \cite{zhang2026tpa}, the key tensor $\ten{K}$ stacks the per-token matrices $\mat{K}_t=\mat{A}_t^\top\mat{B}_t/R_{\rm K}$. The diagram omits the factor $1/R_{\rm K}$.}
\label{fig:example-inference}
\end{wrapfigure}
The natural target for tensorization at the inference stage is the KV-cache, whose size grows linearly with sequence length. Tensor Product Attention (TPA) \cite{zhang2026tpa} factorizes the query, key, and value matrices of each token, so that the cached keys and values are never materialized. For token $t$, the key matrix $\mat{K}_t\in\R^{H\times d_h}$ is written as $\frac{1}{R_{\rm K}}\mat{A}_t^\top\mat{B}_t$ with $\mat{A}_t\in\R^{R_K\times H}$ and $\mat{B}_t\in\R^{R_K\times d_h}$, and analogously for $\mat{Q}_t$ and $\mat{V}_t$. \Cref{fig:example-inference} shows this two-factor case; variants with more factors are also possible. Caching the factors costs $(R_K+R_V)(H+d_h)$ values per token instead of $2Hd_h$: for a $\llama$-sized layer with $H=32$ and $d_h=128$, that is $640$ against $8{,}192$, assuming $R_{\rm K} = R_{\rm V} = 2$; a reduction of $12.8\times$. The baseline here is multi-head attention since TPA subsumes both: MHA, MQA, and GQA are its non-contextual special cases. For comparison, the grouped-query attention of $\llama$ shares keys and values across $H_{\rm KV}=8$ groups and therefore stores $2H_{\rm KV}d_h=2{,}048$ values per token; the factored representation is $3.2\times$ smaller still.

\paragraph{Interpretability.} 
Bilinear MLPs \cite{pearce2025bilinearmlp} drop the element-wise nonlinearity from the gated FFN, which turns the layer into an exact contraction with a third-order tensor $\ten{B} \in \R^{d \times d \times d}$ (\cref{sec:component-ffn}, Eq. \eqref{eq:bilinear-forward-pass}). For $\llama$, this means replacing all $32$ gated FFN blocks with bilinear ones. The parameter count stays the same, but SwiGLU disappears from the architecture, so the model has to be pre-trained from scratch.
The resulting tensor admits three modes of analysis, differing in how much is already known about the features. Given dictionaries for both the input $\mat{F}^{\rm in} \in \R^{f\times 4096}$ and the output $\mat{F}^{\rm out} \in \R^{f\times 4096}$ features of a layer, obtained for example from sparse autoencoders, contracting them into all three modes rewrites $\ten{B} \in \R^{4096 \times 4096 \times 4096}$ in the feature basis $\tilde{\ten{B}} = \ten{B} \times_1 \mat{F}^{\rm out} \times_2 \mat{F}^{\rm in} \times_3 \mat{F}^{\rm in} \in \R^{f \times f \times f}$, so that each entry of $\tilde{\ten{B}}$ is the interaction of a pair of input features for one output feature. Given only an output feature $\vecb{u} \in \R^{4096}$, such as an unembedding row, contracting the output mode leaves the symmetric interaction matrix $\mat{Q} = \ten{B} \times_1 \vecb{u}^\top \in \R^{4096 \times 4096}$, whose eigenvectors are the input directions driving $\vecb{u}$ and whose eigenvalues weight their quadratic contributions. With no features, the tensor could be analyzed directly, for example through a HOSVD. The three routes therefore differ in cost and prerequisites, since contracting features away keeps the analysis at matrix scale, whereas the last route touches the full tensor, with $d^3 \approx 69$B entries per layer.


\section{Tensorizing the Transformer}
\label{sec:components}

The current literature tensorizes the Transformer module primarily in a component-wise fashion. Specifically, existing work mostly focuses on a single Transformer submodule, such as the embedding layer \cite{xu2023tensorgpt}, the attention mechanism \cite{ma2019tensorized}, or the FFN \cite{chekalina2023ttm}. This section follows the same component-wise organization, which we call the \emph{component view} of the survey, given the substantial differences in structural constraints across the three submodules. First, \cref{sec:tensor-strategies} introduces a two-way classification of tensorization strategies. Then, \cref{sec:component-emb,sec:component-att,sec:component-ffn} discuss each submodule in turn, covering where the tensor object arises, which tensorization strategies are structurally compatible with it, and what the key challenges are. Finally, \cref{sec:component-sum} gives a structural overview.

\subsection{Tensorization Strategies}
\label{sec:tensor-strategies}

Tensorization strategies can be classified by whether a tensor object's modes carry semantic meaning. If the modes have known semantics, such as position in the sequence or the attention-head index, we call this \emph{mode-specific} tensorization. In this case, knowledge of the semantic meaning allows us to exploit the tensor's structure. For instance, decomposition factors can be shared across structurally similar modes, reducing memory consumption.
\emph{Imposed} tensorization arises when modes are assigned arbitrarily, for example by reshaping a weight matrix into a compatible shape. In this case, nothing in principle restricts the choice of tensor network; instead, it may be motivated by the inductive bias one wishes to impose on the parameterization \cite{hamreras2025tensorization}.
This distinction organizes the component analysis in the following sections.

\subsection{Embedding Layer}
\label{sec:component-emb}

Three structurally distinct lines of work on embedding tensorization can be identified, differing in how the tensor object arises and which decompositions are structurally compatible.

\paragraph{Classical embedding table.}
Let $\mat{E} \in \R^{|\set{V}| \times d}$  be the standard token embedding matrix, where each row is an independent token representation. Since the rows are independent, the matrix carries no higher-order organization. Tensorization is therefore imposed in one of two ways. Either each row is reshaped into a tensor independently, or the full matrix is treated as a tensor. While the row-wise approach preserves row independence, tensorizing the full matrix can hurt it. In both cases, however, decomposed embeddings require a chain of contractions to recover the full vector at inference time, unlike a direct table lookup. This creates a tradeoff between compression ratio and inference latency.

\paragraph{N-gram embedding table.}
A structurally different tensor arises when the classical embedding layer is replaced by an n-gram embedding $\mat{E}_{\rm ngram} \in \R^{|\set{V}|^n\times d}$ \cite{huang2025overtokenized}, where each entry represents a sequence of $n$ tokens. After reshaping the vocabulary axis, each mode of the resulting tensor corresponds to a distinct position within the n-gram. Therefore, tensorization here is mode-specific. For $n\ge 2$ and any realistic vocabulary size, the full table is prohibitively large to store in memory \cite{zhou2026tensorizingengram}. Tensor decomposition thus serves not as a post-hoc compression choice, but as a structural necessity for operating on this object at all. Parameterizing the n-gram table as a tensor decomposition introduces a rank hyperparameter. This rank directly controls the expressiveness of the learned representations, with higher ranks yielding lower final loss \cite{zhou2026tensorizingengram}. This highlights the key tradeoff in selecting the optimal rank under memory constraints.

\paragraph{Morpheme-based embedding table.} 
Instead of storing a representation for each token, one can maintain a smaller table over morphemes, i.e., the sub-token units that compose tokens. This morpheme-based matrix $\mat{E}_{\rm morpheme}$  has size $|\set{M}| \times d$, where $|\set{M}| \ll |\set{V}|$, making it substantially more compact than the standard embedding layer. Token vectors are then constructed from morpheme embeddings via tensor-algebraic operations \cite{gan2022morphte}, making this a mode-specific tensorization, as each mode corresponds to a morpheme position within the word. However, this introduces additional hyperparameters and may yield less expressive word representations than direct lookup.

\subsection{Attention Mechanisms}
\label{sec:component-att}

The attention mechanism exposes several structurally distinct tensor objects, each admitting a different tensorization rationale. We organize them into three cases.

\paragraph{Attention schemes.}
MHA can be represented by a tensor diagram that combines contractions with softmax \cite{wang2023tensor, liu2022tuformer}. In \cref{fig:mha-tensor-diagram}, the identity tensors on the head mode keep the attention computation within each head. This representation suggests how to design other attention schemes. For example, Tuformer replaces the fixed coupling with a learnable core so that queries and keys from one head can contribute to another \cite{liu2022tuformer}.

\begin{figure}[!htbp]
\centering
\resizebox{0.72\linewidth}{!}{%
\begin{tikzpicture}[
  tnode/.style={circle, draw=black!80, line width=1.0pt,
                minimum size=0.90cm, inner sep=0pt, font=\small},
  idnode/.style={circle, fill=black!80, minimum size=4pt, inner sep=0pt},
  op/.style={draw=black!80, line width=1.0pt, rounded corners=2pt,
             minimum size=0.45cm, inner sep=1pt, font=\scriptsize},
  bond/.style={line width=1.0pt, black!80},
  leg/.style={line width=1.0pt, black!80},
  slbl/.style={font=\scriptsize, text=black},
]
\definecolor{ColX}{RGB}{102,194,165}
\definecolor{ColW}{RGB}{252,141, 98}
\definecolor{ColS}{RGB}{141,160,203}

\node[tnode, fill=ColX!35] (Xq) at (-3.0,  1.65) {$\mat{X}$};
\node[tnode, fill=ColX!35] (Xk) at ( 3.0,  1.65) {$\mat{X}$};
\node[tnode, fill=ColX!35] (Xv) at ( 3.0, -1.65) {$\mat{X}$};
\node[tnode, fill=ColW!40] (Wq) at (-1.55,  1.65) {$\ten{W}_Q$};
\node[tnode, fill=ColW!40] (Wk) at ( 1.55,  1.65) {$\ten{W}_K$};
\node[tnode, fill=ColW!40] (Wv) at ( 1.55, -1.65) {$\ten{W}_V$};
\node[tnode, fill=ColW!40] (Wo) at (-1.55, -1.65) {$\ten{W}_O$};
\node[idnode] (IdQK) at (0,  0.76) {};
\node[idnode] (IdVO) at (0, -0.76) {};
\node[op, fill=ColS!40] (S) at (4.05, 1.65) {$\mathsf{S}$};

\begin{pgfonlayer}{background}
\draw[ColS!85!black, line width=0.8pt, dashed, rounded corners=4pt]
    (-3.78, 0.06) rectangle (4.50, 2.40);
\end{pgfonlayer}

\draw[leg] (Xq.west) -- (-4.65,  1.65) node[above right, slbl] {$T$};
\draw[leg] (Wo.west) -- (-4.65, -1.65) node[above right, slbl] {$d$};
\draw[bond] (Xq.east) -- (Wq.west) node[midway, above=1pt, slbl] {$d$};
\draw[bond] (Wk.east) -- (Xk.west) node[midway, above=1pt, slbl] {$d$};
\draw[bond] (Wv.east) -- (Xv.west) node[midway, below=1pt, slbl] {$d$};
\draw[bond] (Wq.east) -- (Wk.west) node[midway, above=1pt, slbl] {$d_h$};
\draw[bond] (Wo.east) -- (Wv.west) node[midway, below=1pt, slbl] {$d_h$};

\draw[bond] (Wq.330) -- (IdQK) -- (Wk.210);
\draw[bond] (Wo.30)  -- (IdVO) -- (Wv.150);
\draw[bond] (IdQK) -- (IdVO) node[pos=0.7, right=2pt, slbl] {$H$};
\draw[bond] (Xk.east) -- (S.west);
\draw[bond] (S.east) -- (4.90, 1.65) -- (4.90,-1.65)
    node[midway, right=1pt, slbl] {$T$} -- (Xv.east);
\end{tikzpicture}}
\caption{\footnotesize Tensor diagram of multi-head self-attention for a single sequence, adapted from \cite{liu2022tuformer}. The three copies of $\mat{X}$ represent the same input. The projection tensors $\ten{W}_Q$, $\ten{W}_K$, $\ten{W}_V$, and $\ten{W}_O$ each have a head mode. The dashed box encloses the score computation, and $\mathsf{S}$ denotes scaling by $1/\sqrt{d_h}$ followed by softmax along the key-position leg. In causal attention, we add the mask in \eqref{eq:masked_attention} after scaling and before softmax.}
\label{fig:mha-tensor-diagram}
\end{figure}

\paragraph{Q, K, V tensors.}
For hidden states $\ten{X}\in\R^{B\times T\times d}$, MHA produces query, key, and value tensors
\begin{equation}
\ten{Q} \in \R^{B \times H_{\rm Q} \times T \times d_h}, \quad \ten{K},\ten{V}\in\R^{B\times H_{\rm KV}\times T\times d_h},
\end{equation}
where the four modes are batch, head, sequence position, and head dimension. These tensors are activations rather than static weights, so we also use the notation $\ten{Q}(x), \ten{K}(x),\ten{V}(x)$. Since each mode carries a distinct semantic role, tensorization here is mode-specific. The primary object of interest is the KV-cache, namely the $\ten{K}, \ten{V}$ tensors. Exploiting the structure of these tensors can significantly reduce the KV-cache memory footprint and the volume of inference-time data transfer.
However, exploiting this structure remains non-trivial for the following reasons. First, tensor methods must either be compatible with efficient Attention implementations such as FlashAttention \cite{dao2022flashattention} or achieve comparable GPU utilization. Second, tensor methods must match or exceed non-tensor counterparts such as GQA and MLA in terms of both compression ratio and model quality. Third, RoPE \cite{su2023rope} compatibility is an additional constraint, given that RoPE applies position-dependent rotations to $\mat{Q}$ and $\mat{K}$ at inference time; consequently, any factored or compressed KV-cache representation must support these rotations. The crucial design challenges for a tensorized KV cache are therefore efficient GPU implementation, competitiveness with non-tensor schemes, and RoPE compatibility.

\paragraph{Projection matrices.}
We distinguish two approaches to tensorizing the projection matrices. The first one treats each matrix $\mat{W}_Q, \mat{W}_K, \mat{W}_V, \mat{W}_O$ independently via imposed reshaping, which is structurally identical to FFN tensorization and is discussed in \cref{sec:component-ffn}. The second one stacks projections across layers, heads, and projection types to form a joint tensor, introducing semantic modes. The projection matrices $\mat{W}_{Q, h}^{(\ell)}, \mat{W}_{K, h}^{(\ell)}, \mat{W}_{V, h}^{(\ell)} \in\R^{d\times d_h}$ (per head $h$ and layer $\ell$) and $\mat{W}_{O}^{(\ell)} \in\R^{Hd_h\times d}$ differ in shape: $\mat{W}_{O}^{(\ell)}$ maps from the concatenated outputs of all $H$ heads and is therefore $H$ times wider along one dimension, which must be accounted for when forming a joint tensor. If $\mat{W}_{O}^{(\ell)}$ is excluded, the query, key, and value projections assemble as:

\begin{equation}
\label{eq:WQKV}
\ten{W}_{QKV} = \operatorname{stack}_{\ell,p,h}\!\left(\mat{W}_{p,h}^{(\ell)}\right) \in\R^{L\times 3\times H\times d\times d_h},
\quad p\in\{Q,K,V\},
\end{equation}
where $\ell$ and $h$ are the layer and head indices, respectively. To include $\mat{W}_{O}^{(\ell)}$, we divide $(\mat{W}_{O}^{(\ell)})^\top$ into $H$ blocks $\mat{W}_{O,h}^{(\ell)} \in\R^{d\times d_h}$, each corresponding to a single head. Then:

\begin{equation}
\label{eq:QKVO}
\ten{W}_{QKVO} = \operatorname{stack}_{\ell,p,h}\!\left(\mat{W}_{p,h}^{(\ell)}\right) \in\R^{L\times 4 \times H \times d \times d_h},
\quad p\in\{Q,K,V,O\}.
\end{equation}
For $\ten{W}_{QKV}$ and $\ten{W}_{QKVO}$, the layer, projection type, and head modes carry distinct semantic meaning. This mode-specific structure enables cross-layer and cross-head factor sharing, which existing methods have widely exploited \cite{li2026lestd}. The key trade-off is between the compression gains from sharing and the increased coupling between modes that sharing introduces.

\subsection{Feed-Forward Network}
\label{sec:component-ffn}

The FFN matrices $\mat{W}_{\rm gate}$, $\mat{W}_{\rm up}$, and $\mat{W}_{\rm down}$ can be treated independently as linear maps or organized into a joint tensor. The following two paragraphs discuss these approaches, which fall on opposite sides of the mode-specific/imposed distinction.

\paragraph{TT and TTM.}
Under imposed tensorization, the projection matrix $\mat{W}$ is reshaped into a tensor $\ten{W}$ with a chosen shape and mode order. TT and TTM are common choices for the tensor network. A two-core TT with one input mode and one output mode gives a low-rank matrix factorization, with the internal rank bounding the matrix rank. While a two-core TTM with internal rank $R$ is the sum of $R$ Kronecker products.

Let $I = \prod_{n=1}^N I_n$ and $J = \prod_{m=1}^M J_m$. Under \emph{TT tensorization}, the projection $\mat{W} \in \R^{I \times J}$ is reshaped into $\ten{W} \in \R^{I_1 \times \cdots \times I_N \times J_1 \times \cdots \times J_M}$ and the input $\vecb{x} \in \R^{I}$ into $\ten{X} \in \R^{I_{1} \times \cdots \times I_{N}}$. Representing $\ten{W}$ in the TT format with third-order cores $\ten{G}^{(1)}, \dots, \ten{G}^{(N+M)}$, the output $\vecb{y}^\top = \vecb{x}^\top \mat{W}$ is computed entrywise as
\begin{equation}
  \ten{Y}_{j_1, \dots, j_M} = \sum_{i_1=1, \dots, i_N=1}^{I_1, \dots, I_N} \ten{X}_{i_1, \dots, i_N} \ten{G}^{(1)}_{:, i_1, :} \cdots \ten{G}^{(N)}_{:, i_N, :} \cdot \ten{G}^{(N+1)}_{:, j_1, :} \cdots \ten{G}^{(N+M)}_{:, j_M, :},
\end{equation}
and $\ten{Y} \in \R^{J_1 \times \cdots \times J_M}$ is reshaped back into $\vecb{y}$.
\emph{TTM tensorization} instead requires the two factorizations to have the same length, so that $I = \prod_{n=1}^N I_n$ and $J = \prod_{n=1}^N J_n$. The projection is reshaped into $\ten{W} \in \R^{I_1 \times J_1 \times \cdots \times I_N \times J_N}$, pairing each input mode with the corresponding output mode, and the contraction becomes
\begin{equation}
  \ten{Y}_{j_1, \dots, j_N} = \sum_{i_1=1, \dots, i_N=1}^{I_1, \dots, I_N} \ten{X}_{i_1, \dots, i_N} \ten{G}^{(1)}_{:, i_1, j_1, :} \cdots \ten{G}^{(N)}_{:, i_N, j_N, :}, 
\end{equation}
where $\ten{G}^{(n)} \in \R^{R_{n-1} \times I_n \times J_n \times R_n}$ are the TTM cores. The two formats differ in how they order the modes: TT chains all $N+M$ modes in sequence, whereas TTM couples the $n$-th input mode with the $n$-th output mode within a single core. \Cref{fig:tt-ttm-tensor-diagram} illustrates both contractions.

\begin{figure}[!htbp]
\centering
\resizebox{0.80\linewidth}{!}{%
\begin{tikzpicture}[
  core/.style={circle, draw=black!80, line width=1.0pt,
               minimum size=1.10cm, inner sep=0pt, font=\scriptsize},
  bond/.style={line width=1.0pt, black!80},
  leg/.style={line width=1.0pt, black!80},
  slbl/.style={font=\scriptsize, text=black},
]
\definecolor{ColX}{RGB}{102,194,165}
\definecolor{ColW}{RGB}{252,141, 98}
\pgfmathsetmacro{\RadCore}{0.55}
\pgfmathsetmacro{\Lleg}{0.65}
\pgfmathsetmacro{\Rx}{0}
\pgfmathsetmacro{\Dx}{1.55}
\pgfmathsetmacro{\RyTop}{1.10}
\pgfmathsetmacro{\RyBot}{-1.10}

\pgfmathsetmacro{\TTLeft}{\Rx+2*\Dx-\RadCore}
\pgfmathsetmacro{\TTRight}{\Rx+3*\Dx+\RadCore}
\draw[draw=black!80, line width=1.0pt, fill=ColX!35, rounded corners=5pt]
    (\TTLeft, {\RyTop-\RadCore}) rectangle (\TTRight, {\RyTop+\RadCore});
\node[font=\small] at ({0.5*(\TTLeft+\TTRight)}, \RyTop) {$\ten{X}$};
\foreach \k in {1,2,3,4}{
  \node[core, fill=ColW!40] (T\k) at ({\Rx+(4-\k)*\Dx}, \RyBot) {$\ten{G}^{(\k)}$};
}
\foreach \a/\b/\lab in {1/2/1, 2/3/2, 3/4/3}{
  \draw[bond] (T\a) -- (T\b) node[midway, above=1pt, slbl] {$R_\lab$};
}
\foreach \k in {1,2}{
  \draw[bond] ({\Rx+(4-\k)*\Dx}, {\RyTop-\RadCore}) -- (T\k.north)
    node[midway, right=2pt, slbl] {$I_\k$};
}
\foreach \k/\lab in {3/1, 4/2}{
  \draw[leg] (T\k.south) -- ++(0, -\Lleg) node[right=1pt, slbl] {$J_\lab$};
}
\node[font=\small\bfseries] at ({\Rx+1.5*\Dx}, 2.08) {TT};

\begin{scope}[xshift=7cm]
\pgfmathsetmacro{\RectL}{\Rx - \RadCore}
\pgfmathsetmacro{\RectR}{\Rx + 3*\Dx + \RadCore}
\pgfmathsetmacro{\RectCx}{0.5*(\RectL + \RectR)}
\pgfmathsetmacro{\RectT}{\RyTop + \RadCore}
\pgfmathsetmacro{\RectB}{\RyTop - \RadCore}
\draw[draw=black!80, line width=1.0pt, fill=ColX!35, rounded corners=5pt]
    (\RectL, \RectB) rectangle (\RectR, \RectT);
\node[font=\small] at (\RectCx, \RyTop) {$\ten{X}$};
\foreach \k in {1,2,3,4}{
  \node[core, fill=ColW!40] (W\k) at ({\Rx+(\k-1)*\Dx}, \RyBot) {$\ten{G}^{(\k)}$};
}
\foreach \a/\b/\lab in {1/2/1, 2/3/2, 3/4/3}{
  \draw[bond] (W\a) -- (W\b) node[midway, above=1pt, slbl] {$R_\lab$};
}
\foreach \k in {1,2,3,4}{
  \draw[leg] (W\k.south) -- ++(0, -\Lleg) node[right=1pt, slbl] {$J_\k$};
}
\foreach \k in {1,2,3,4}{
  \draw[bond] ({\Rx+(\k-1)*\Dx}, \RectB) -- (W\k.north) node[midway, right=2pt, slbl] {$I_\k$};
}
\node[font=\small\bfseries] at (\RectCx, 2.08) {TTM};
\end{scope}
\end{tikzpicture}}
\caption{\footnotesize Contraction of a reshaped input $\ten{X}$ with a tensorized projection. \textit{Left}: the TT chain is drawn from right to left, with the two input modes on the first two cores and the two output modes on the remaining cores. \textit{Right}: each of the four TTM cores pairs an input mode $I_n$ with an output mode $J_n$. The free $J_n$ legs specify the shape of $\ten{Y}$.}
\label{fig:tt-ttm-tensor-diagram}
\end{figure}

In TT tensorization, the cut between cores $N$ and $N+1$ separates all input modes from all output modes, so the rank of the reconstructed matrix $\mat{W}$ is bounded by $R_{N}$ \cite{oseledets2011tensor}. For TTM tensorization, Theorem 1 of \cite{hrinchuk2020tensorized} states that almost every TTM whose ranks are bounded by a fixed set $\set{R}$ reconstructs to a matrix of full rank $\min(I, J)$. An arbitrary learned TTM projection can therefore represent a full-rank matrix in principle. In practice, however, Li et al. \cite{li2022hypoformer} report that low-rank TTM parameterizations struggle to retain expressivity.

The key tradeoff here is between compression and latency. A tensorized layer replaces a single GEMM with a sequential chain of small contractions, whereas modern GPUs are well optimized for large matrix computations but handle such small-size contractions poorly, so a tensorized model can cut parameters and FLOPs while still running slower than its dense counterpart; realizing an actual speedup requires kernel-level work, such as optimizing the contractions themselves and removing backend overhead \cite{yang2024comera}. The general solution to this challenge remains an open direction (\cref{sec:future-directions}).

\paragraph{Bilinear FFN}
A gated FFN becomes \emph{bilinear} when the element-wise nonlinearity is dropped \cite{shazeer2020glu}:
\begin{equation}
    \vecb{z}^\top = (\vecb{x}^\top \mat{W}^{\rm gate}) \odot ( \vecb{x}^\top \mat{W}^{\rm up}), \quad \vecb{y}^\top = \vecb{z}^\top \mat{W}^{\rm down}.
\end{equation}
This case differs from those above: the layer is multilinear in $\vecb{x}$, so it admits an exact tensor representation. Fix a hidden coordinate $\alpha$. Since $(\vecb{x}^\top \mat{W})_\alpha = \vecb{x}^\top \mat{W}_{:, \alpha}$, the two branches combine into a quadratic form:
\begin{equation}
  \vecb{z}_\alpha = (\vecb{x}^\top \mat{W}^{\rm gate})_\alpha \cdot (\vecb{x}^\top \mat{W}^{\rm up})_\alpha = \bigl(\vecb{x}^\top \mat{W}^{\rm gate}_{:, \alpha}\bigr)\bigl(\vecb{x}^\top \mat{W}^{\rm up}_{:, \alpha}\bigr) = \vecb{x}^\top \mat{W}^{\rm gate}_{:, \alpha} \bigl(\mat{W}^{\rm up}_{:, \alpha}\bigr)^\top \vecb{x} = \vecb{x}^\top \mat{B}_{\alpha} \vecb{x}.
\end{equation}
Note that only the symmetric part of $\mat{B}_\alpha$ contributes to the quadratic form, so $\mat{B}_\alpha$ may be replaced by $\tfrac{1}{2}(\mat{B}_\alpha + \mat{B}_\alpha^\top)$ without loss of generality. Propagating through the down-projection and summing over $\alpha$ gives:
\begin{equation}
  \vecb{y}_\beta = \sum_{\alpha=1}^{d_{\rm ff}} \vecb{z}_\alpha \mat{W}^{\rm down}_{\alpha, \beta} = \sum_{\alpha=1}^{d_{\rm ff}} (\vecb{x}^\top \mat{B}_{\alpha} \vecb{x}) \mat{W}^{\rm down}_{\alpha, \beta} = \vecb{x}^\top \left( \sum_{\alpha=1}^{d_{\rm ff}} \mat{W}^{\rm down}_{\alpha, \beta} \mat{B}_{\alpha} \right) \vecb{x} = \vecb{x}^\top \ten{B}_{\beta, :, :} \vecb{x},
\end{equation}
so that the whole layer collapses into a single contraction
\begin{equation}
\label{eq:bilinear-forward-pass}
  \vecb{y}^\top = \ten{B} \times_2 \vecb{x}^\top \times_3 \vecb{x}^\top,
\end{equation}
where $\ten{B} \in \R^{d \times d \times d}$ and each slice $\ten{B}_{\beta, :, :}$ is symmetric. The modes of $\ten{B}$ carry semantic meaning: two input coordinates and one output coordinate, so this is where tensorization is mode-specific in the FFN. Pearce et al. \cite{pearce2025bilinearmlp} use this directly as a mechanistic interpretability tool, discovering circuits from $\ten{B}$.
The practical obstacle is that gated variants remain the default in deployed LLMs \cite{shazeer2020glu}, so obtaining a bilinear FFN requires training from scratch.

With a nonlinear activation, no such representation exists. The three matrices can still be stacked into:
\begin{equation}
\label{eq:SGUD}
  \ten{S} = \bigl[ \mat{W}^{\rm gate}, \mat{W}^{\rm up}, (\mat{W}^{\rm down})^\top \bigr] \in \R^{3 \times d \times d_{\rm ff}},
\end{equation}
where the modes index the projection type, the input dimension, and the hidden dimension; as in Eq. \eqref{eq:QKVO}, these stacks can be further concatenated across layers or across experts in MoE models, yielding higher-order tensors. Unlike $\ten{B}$ in Eq. \eqref{eq:bilinear-forward-pass}, however, $\ten{S}$ carries no multilinear operator meaning: its first mode does not participate in the forward pass.

\subsection{Components Overview}
\label{sec:component-sum}

\Cref{tab:components} provides an overview of the component-wise analysis. The Tensorization Type column follows the mode-specific/imposed classification of \cref{sec:tensor-strategies}, and the Key tradeoff column names the dominant constraint.
\begin{table}[!htbp]
\centering
\caption{Structural overview of tensorized Transformer components. The tensorization strategy follows the classification of \cref{sec:tensor-strategies}.}
\label{tab:components}
\small
\begin{tblr}{
  width = \textwidth,
  colspec = {Q[2.0cm,m,c]Q[4.0cm,m,c]Q[2.4cm,m,c]X[m,c]},
  row{1} = {font=\bfseries},
  cell{2}{1} = {r=3}{},
  cell{5}{1} = {r=2}{},
  cell{7}{1} = {r=3}{},
  rowsep = 3pt,
  hline{1,10} = {1pt},
  hline{2} = {0.6pt},
  hline{5,7} = {0.6pt, gray7},
}
Component & Tensor object & Tensorization strategy & Key tradeoff \\
Embedding layer & $\mat{E}\in\R^{|\set{V}|\times d}$ & imposed & Compression vs.\ lookup latency \\
 & $\mat{E}_{\rm ngram}\in\R^{|\set{V}|^n\times d}$ & mode-specific & Memory tractability vs.\ rank selection difficulty \\
 & $\mat{E}_{\rm morpheme}\in\R^{|\set{M}|\times d}$ & mode-specific & Memory reduction vs.\ expressiveness of representations \\
Attention mechanism & $\ten{Q} \in\R^{B\times H_{\rm Q}\times T\times d_h}; \newline \ten{K},\ten{V}\in\R^{B\times H_{\rm KV}\times T\times d_h}$ & mode-specific & KV-cache compression vs.\ kernel efficiency and quality parity \\
 & $\ten{W}_{QKV}\in\R^{L\times 3\times H\times d\times d_h}$, $\ten{W}_{QKVO}\in\R^{L\times 4\times H\times d\times d_h}$ & mode-specific & Parameter sharing vs.\ reduced independence \\
FFN & $\mat{W}_{\rm up},\mat{W}_{\rm gate}\in\R^{d\times d_{\rm ff}},$ \newline $\mat{W}_{\rm down}\in\R^{d_{\rm ff}\times d}$ & imposed & Compression vs.\ latency (contractions vs.\ GEMM) \\
 & $\ten{B}\in\R^{d\times d\times d}$ (bilinear) & mode-specific & Exact multilinear structure vs.\ training from scratch \\
 & $\ten{S}\in\R^{3\times d\times d_{\rm ff}}$ (stacked) & mode-specific & Joint representation of a layer vs.\ no operator semantics \\
\end{tblr}
\end{table}

\section{Tensorization Across the Lifecycle}
\label{sec:lifecycle-analysis}

This section develops the lifecycle view: it organizes applications of tensor methods for LLMs by stage, following the lifecycle taxonomy introduced in \cref{sec:lifecycle-overview}. Each subsection opens with a formal problem definition, followed by a literature overview and, where coverage is sufficient, a summary table. Stages with sparse literature acknowledge gaps directly.

\subsection{Tokenization}
\label{subsec:tokenization}

The tokenization stage maps raw text to a sequence of subword units drawn from a fixed vocabulary $\set{V}$, typically produced by byte-pair encoding \cite{sennrich2016bpe} or the unigram language model \cite{kudo2018unigramlm}, and commonly implemented with SentencePiece \cite{kudo2018sentencepiece}.
The vocabulary is a set of discrete symbols and the tokenizer is a non-differentiable map, which makes this stage an unusual target for tensor methods, as no continuous multilinear object exists until the tokens are embedded.

To our knowledge, no published work applies tensor methods to the tokenization procedure itself.
The closest tensor work starts one step later, at the representation of tokens: MorphTE \cite{gan2022morphte} injects morphological structure into the embedding table, and TN-gram \cite{zhou2026tensorizingengram} tensorizes $n$-gram embeddings (\cref{subsec:embeddings}). Both take the vocabulary as given. The tokenizer has been made stochastic \cite{kudo2018unigramlm, provilkov2020bpedropout} and even learnable \cite{tay2022charformer}, but not tensorized. We outline what tensorization could offer at this stage in \cref{sec:future-directions}.


\subsection{Embeddings}
\label{subsec:embeddings}

\paragraph{Problem definition.}
Tensorizing the embedding layer can pursue different optimization objectives that mirror the two objectives for pre-training (see \cref{subsec:pre-training}) and compression (see \cref{subsec:compression}): training the embedding matrix from scratch or applying a training-free compression scheme to a pretrained embedding matrix. Both cases affect only the embedding layer while leaving the rest of the model unchanged, unlike the methods discussed in \cref{subsec:pre-training} and \cref{subsec:compression}. This two-way classification is orthogonal to the table-type classification from \cref{sec:component-emb}, since any table type may appear under either objective.

\paragraph{Literature overview.}
Before LLMs, tensor methods in NLP were primarily used for representation learning of linguistic units. Early works constructed representations for subject-verb-object triplets via Tucker-inspired factorization \cite{cruys2013svo} and tensor-algebraic operations on word vectors \cite{etal2014svotensorspaces}. A second line of work emerged from reformulating the embedding-learning problem as matrix factorization of co-occurrence matrices. This motivated tensor extensions that obtain richer representations by aggregating additional information, such as sentiment \cite{rahimi2021tenssent} or frame-semantic role annotations \cite{etal2017frame}, into a higher-order co-occurrence tensor, which was then factorized to produce word representations. However, with the development of LLMs, the objective shifted from tensor-based enrichment to compressing and parameterizing learned embedding tables.

In modern LLMs, the dominant approach for tensorizing the classical embedding layer (see \cref{sec:component-emb}) relies on TT-family decompositions. Hrinchuk et al. \cite{hrinchuk2020tensorized} replace the embedding layer with a TTM representation and train the model from scratch.
This parameterization achieves a substantial compression for Transformer-XL \cite{dai2019transformerxl} on the language modeling task \cite{hrinchuk2020tensorized}. However, this approach requires training from scratch in this parameterization, and, as discussed in \cref{sec:component-emb}, row independence is violated: token ordering may affect representation quality \cite{hrinchuk2020tensorized}. TensorGPT \cite{xu2023tensorgpt} treats each row of the embedding layer independently; specifically, it reshapes and compresses each row via TT-SVD \cite{oseledets2011tensor} applied to an already pretrained model (see \cref{fig:example-embeddings}). This eliminates the need for retraining but may yield less expressive representations and add inference latency.

Beyond the standard embedding table, tensor methods have been applied to two structurally different settings: n-gram embedding table and morpheme-based embedding table (see \cref{sec:component-emb}). In the first case, TN-gram \cite{zhou2026tensorizingengram} parameterizes the n-gram embedding table as a CP representation, reducing the parameter count from exponential $\mathcal{O}(|\set{V}|^nd)$ in $n$ to linear $\mathcal{O}(Rn|\set{V}| + Rd)$. At the same time, each CP factor corresponds to a specific position within the n-gram, making the decomposition interpretable. However, as with TT-Embedding, TN-gram must be trained from scratch since the CP parameterization is intrinsic to the n-gram table. In the second case, MorphTE \cite{gan2022morphte} takes a different direction: instead of compressing a word-level table, it maintains a small morpheme embedding table and constructs word representations as Kronecker products (referred to as tensor products in the original paper) of morpheme vectors. This drastically reduces embedding table size but, again, requires training from scratch to adapt the model to such embeddings.

\paragraph{Comparison.}
\Cref{tab:embeddings-methods} compares the decomposition, embedding table type, and training regime. For standard token embeddings, TT-embeddings learn a factorization of the full table from scratch, while TensorGPT compresses pretrained rows independently. TN-gram and MorphTE learn embeddings from scratch using n-grams and morphemes, respectively.


\begin{table}[!htbp]
\centering
\caption{\footnotesize Structural comparison of tensor methods for embeddings stage.}
\label{tab:embeddings-methods}
\footnotesize
\begin{tblr}{
  width = \textwidth,
  colspec = {Q[3.0cm,m,c]Q[4.5cm,m,c]Q[3.5cm,m,c]X[m,c]},
  row{1} = {font=\bfseries},
  rowsep = 3pt,
  hline{1,6} = {1pt},
  hline{2} = {0.6pt},
  hline{3,4,5} = {0.6pt, gray7},
}
Method & Decomposition & Embedding table & Training \\
TT-embeddings \cite{hrinchuk2020tensorized} & TTM & Classical & from scratch \\
TensorGPT \cite{xu2023tensorgpt} & TT row-wise & Classical & training-free \\
TN-gram \cite{zhou2026tensorizingengram} & CP &  N-gram & from-scratch \\
MorphTE \cite{gan2022morphte} & Kronecker product of \newline morpheme embeddings & Morpheme-based & from scratch \\
\end{tblr}
\end{table}


\subsection{Pre-training}
\label{subsec:pre-training}

\paragraph{Problem definition.}
The optimization goal for tensorized pre-training can be stated as the following constrained optimization problem:
\begin{equation}
\begin{aligned}
& \min_{\ten{W} \in \mathfrak{TN}(\set{R})} \mathcal{L}_{\rm LM} (\ten{W}; \ \mathcal{D}_{\rm LM}), \\
& {\rm s.t.} ~ m(\ten{W}) \leq M,\ c(\ten{W}) \leq C
\end{aligned}
\end{equation}
Here, $\ten{W}$ denotes the tensorized weights of the language model, $\mathfrak{TN}(\set{R})$ is the parametric family of tensor networks given a set of ranks $\set{R}$, and $\mathcal{L}_{\rm LM}$ is the language-modeling loss on a corpus $\mathcal{D}_{\rm LM}$. The budget uses two functions to reflect the different effects of tensorization: $m(\cdot)$ measures memory, typically the stored parameters together with the peak footprint reached during training, whereas $c(\cdot)$ measures computation, either the contraction FLOPs or the latency they induce on the target hardware.
The pre-training objective, unlike the compression objective (see \cref{subsec:compression}), has no reference dense $\mat{W}$, so the tensorized weights $\ten{W}$ are trained from scratch. And, unlike the embedding stage (see \cref{subsec:embeddings}), tensorization here can target any part of the Transformer.

\paragraph{Literature overview.}
Tensorizing attention changes which token-sequence indices interact. For the autoregressive decoders considered here, the choice of indices also determines how causality is enforced. 
Tuformer \cite{liu2022tuformer} changes the coupling between heads. In the MHA diagram in \cref{fig:mha-tensor-diagram}, the central identity tensors tie each attention map to its own head. Tuformer replaces this fixed coupling with a learnable core that can be dense or structured, for example, a Kronecker product. Off-diagonal head couplings allow scores from one head to contribute to another head's weighting of values. MHA is the special case of Tuformer \cite{liu2022tuformer}.
Multi-linear attention \cite{ma2019tensorized} goes further, replacing MHA with a BT tensor network (\cref{eq:btd}). Its $K$ blocks within one layer share the query, key, and value projections as factor matrices, while each term has its own diagonal core $\ten{G}_k \in \R^{R \times R \times R}$ and serves as a head. The diagonal cores keep the number of head-specific parameters small. This construction buys factor sharing and low-rank cores at the price of non-trivial causal masking \cite{ma2019tensorized}.

A second group of methods keeps the attention computation fixed and tensorizes the linear projections \cite{novikov2015tensorizing}. GPT-TTM \cite{chekalina2023ttm} replaces each dense projection with a TTM layer (\cref{sec:component-ffn}). Small TTM ranks reduce the parameter count but can limit the projection's expressivity in practice \cite{li2022hypoformer}. Hypoformer \cite{li2022hypoformer} combines a dense and a TTM branch in a hybrid layer $\mat{W} = \begin{bmatrix} \mat{W}_{\rm dense} \\ \mat{W}_{\rm TTM} \end{bmatrix}$, where $\mat{W}_{\rm dense} \in \R^{\alpha I \times J}$ and $\mat{W}_{\rm TTM} \in \R^{(1-\alpha)I \times J}$. The two branches process complementary fractions $\alpha$ and $1-\alpha$ of the input in parallel, leaving the dense branch free to learn an unrestricted projection of its input.
CoMERA \cite{yang2024comera} uses TT projections to reduce training memory. Its implementation optimizes the contraction order in both the forward and backward passes and uses CUDA Graphs to reduce kernel-launch overhead. CoMERA also learns the TT ranks during training by inserting diagonal matrices between adjacent cores,
\begin{equation*}
  \ten{W} = \ten{G}^{(1)} \leftindex_{-1} \times^1 ~ \mat{D}^{(1)} \leftindex_{-1} \times^1 ~ \ten{G}^{(2)} \leftindex_{-1} \times^1 ~ \dots \leftindex_{-1} \times^1 ~ \mat{D}^{(N - 1)} \leftindex_{-1} \times^1 ~ \ten{G}^{(N)}
\end{equation*}
where $\mat{D}^{(n)} \in \R^{R_n \times R_n}$. The training objective balances task loss and model size. Pruning a diagonal entry zeros out the corresponding slices from both neighboring cores, reducing their internal rank.

Structured matrix parameterizations are closely related to tensor networks. We use the convention $\vecb{y}^{\top}=\vecb{x}^{\top}\mat{W}$ and reshape the input and output into $\ten{X}$ and $\ten{Y}$ in row-major order. Here $\ten{L}$ and $\ten{R}$ denote the factor tensors, and each sum over an input index runs over its full range.
Shapeshifter \cite{pahani2021shapeshifter} reshapes and factorizes each weight matrix into a sum of Kronecker products $\mat{W}=\sum_{r=1}^{R}\mat{L}_r\otimes\mat{R}_r$ (\cref{fig:structured-projections}(a)). Its action on the input is
\[
  \ten{Y}_{j_1,j_2} =
  \sum_{i_1,i_2}\sum_{r=1}^{R}
  \ten{L}_{i_1, j_1, r} \ten{R}_{i_2, j_2, r} \ten{X}_{i_1,i_2},
\]
where $\ten{L}_{i_1, j_1, r} = (\mat{L}_r)_{i_1, j_1}$ and $\ten{R}_{i_2, j_2, r} = (\mat{R}_r)_{i_2, j_2}$. The factors share only the summation index $r$.
The Monarch parameterization \cite{dao2022monarch} writes the column-vector operator as $\mat{W}^{\top}=\mat{P}\mat{L}\mat{P}^{\top}\mat{R}$ (\cref{fig:structured-projections}(b)), where $\mat{L}$ and $\mat{R}$ are block diagonal with $\sqrt{I}$ dense $\sqrt{I}\times \sqrt{I}$ blocks each. The matrix $\mat{P}$ is a special-type permutation matrix. Stacking the blocks of each factor into a third-order tensor, with $\ten{L}_{\alpha, \beta, \gamma} = (\mat{L}_\alpha)_{\beta, \gamma}$ and $\ten{R}_{\alpha, \beta, \gamma} = (\mat{R}_\alpha)_{\beta, \gamma}$, gives
\[
  \ten{Y}_{j_1,j_2} =
  \sum_{i_1,i_2}
  \ten{L}_{j_2,j_1,i_1} \ten{R}_{i_1,j_2,i_2} \ten{X}_{i_1,i_2}.
\]
The factors share the input index $i_1$ and the output index $j_2$. For fixed $(i_1,j_2)$, the corresponding slice of the weight tensor is an outer product over $i_2$ and $j_1$, with entries $\ten{W}_{i_1,i_2,j_1,j_2} = \ten{L}_{j_2,j_1,i_1} \ten{R}_{i_1,j_2,i_2}$.
Qiu et al.\ \cite{qiu2024btt} combine ideas from Kronecker decomposition and Monarch in a two-core block tensor train (BTT) (\cref{fig:structured-projections}(c)). They add a rank index $r$ to the Monarch factors to obtain
\begin{equation}
  \ten{Y}_{j_1,j_2} =
  \sum_{i_1,i_2}\sum_{r=1}^{R}
  \ten{L}_{j_2,j_1,i_1,r} \ten{R}_{i_1,j_2,i_2,r} \ten{X}_{i_1,i_2},
  \label{eq:pretrain-btt}
\end{equation}
where $\ten{L}$ and $\ten{R}$ are fourth-order tensors formed by stacking the corresponding third-order factors from $R$ Monarch matrices.
Each slice at fixed $(i_1,j_2)$ is now a sum of $R$ outer products and has matrix rank at most $R$. With balanced modes, this recovers Monarch at $R=1$ and can represent any dense weight matrix at $R=\sqrt{I}$ \cite{qiu2024btt}.


\begin{figure}[!htbp]
\centering
\begin{tikzpicture}[
  core/.style={circle, draw=black!80, line width=1pt, minimum size=0.76cm,
               inner sep=0pt, font=\small, fill=ColW!40},
  idnode/.style={circle, fill=black!80, minimum size=4pt, inner sep=0pt},
  leg/.style={line width=1pt, black!80},
  slbl/.style={font=\scriptsize, text=black},
]
\definecolor{ColW}{RGB}{252,141,98}
\foreach \p/\x/\panel/\titletext in {K/0/a/Kronecker products,M/3.65/b/Monarch,B/7.30/c/BTT,G/10.95/d/General einsum}{
  \begin{scope}[xshift=\x cm]
    \node[core] (\p L) at (-0.76,0) {$\ten{L}$};
    \node[core] (\p R) at ( 0.76,0) {$\ten{R}$};
    \node[font=\small\bfseries] at (0,1.95) {(\panel) \titletext};
  \end{scope}
}
\foreach \p/\dimlabel in {K/\sqrt{I},M/\sqrt{I},B/\sqrt{I},G/\sqrt[3]{I}}{
  \draw[leg] (\p R.north) -- ++(0,0.72) node[above,slbl] {$\dimlabel$};
  \draw[leg] (\p L.south) -- ++(0,-0.72) node[below,slbl] {$\dimlabel$};
}
\foreach \p/\dimlabel in {K/\sqrt{I},G/\sqrt[3]{I}}{
  \draw[leg] (\p L.north) -- ++(0,0.72) node[above,slbl] {$\dimlabel$};
  \draw[leg] (\p R.south) -- ++(0,-0.72) node[below,slbl] {$\dimlabel$};
}
\foreach \p/\x/\dimlabel in {M/3.65/\sqrt{I},B/7.30/\sqrt{I},G/10.95/\sqrt[3]{I}}{
  \begin{scope}[xshift=\x cm]
    \node[idnode] (\p in)  at (0, 0.66) {};
    \node[idnode] (\p out) at (0,-0.66) {};
    \draw[leg] (\p L) -- (\p in)  -- (\p R);
    \draw[leg] (\p L) -- (\p out) -- (\p R);
    \draw[leg] (\p in)  -- (0, 1.10) node[above,slbl] {$\dimlabel$};
    \draw[leg] (\p out) -- (0,-1.10) node[below,slbl] {$\dimlabel$};
  \end{scope}
}
\foreach \p in {K,B,G}{
  \draw[leg] (\p L) -- (\p R) node[midway,above=1pt,slbl] {$R$};
}
\end{tikzpicture}
\caption{\footnotesize Tensor diagrams for four parameterizations of a square $I\times I$ weight matrix with balanced modes. Input modes are at the top, and output modes are at the bottom.}
\label{fig:structured-projections}
\end{figure}
Potapczynski et al.\ \cite{potapczynski2024generaleinsum} bring all three structures into a general two-factor einsum. Each input or output index can belong to the left factor, the right factor, or both, as shown in \cref{fig:structured-projections}(d). Using indices $(i_1,i_2,i_3)$ for the input and $(j_1,j_2,j_3)$ for the output gives
\[
  \ten{Y}_{j_1,j_2,j_3} =
  \sum_{i_1,i_2,i_3}\sum_{r=1}^{R}
  \ten{L}_{i_1, i_2, j_1, j_2,r}\,\ten{R}_{i_2, i_3, j_2, j_3,r}\,\ten{X}_{i_1,i_2,i_3},
\]
Setting the dimensions of $i_2$ and $j_2$ to one gives a sum of Kronecker products. Setting those of $i_1$ and $j_3$ to one gives BTT, which reduces to Monarch at $R=1$. For square layers, the authors parameterize each index dimension as a power of the layer width. The input exponents sum to one, as do the output exponents. A zero exponent sets the corresponding dimension to one, so the edge can be omitted. They search over these exponents and compare the resulting compute-optimal scaling laws for GPT-2 trained on OpenWebText.
Among the structures tested, full-rank matrices without parameter sharing most closely matched the dense models' loss at a given training compute budget \cite{potapczynski2024generaleinsum}.
Motivated by this result, the authors propose BTT-MoE, which treats each Monarch term in \cref{eq:pretrain-btt} as an expert. For each input token, a learned linear gate selects the two highest-scoring terms and combines their outputs with softmax weights \cite{potapczynski2024generaleinsum}.

\paragraph{Comparison.}
\Cref{tab:pre-training-structure} compares the parameterizations used to train tensorized models from scratch. Attention methods change how queries, keys, and values interact, while the other methods structure the weight matrices of linear projections.


\begin{table}[!htbp]
\centering
\caption{\footnotesize Structural comparison of tensorized pre-training methods.}
\label{tab:pre-training-structure}
\footnotesize
\begin{tblr}{
  width = \textwidth,
  colspec = {X[m,c]X[m,c]},
  row{1} = {font=\bfseries},
  rowsep = 3pt,
  hline{1,14} = {1pt},
  hline{2} = {0.6pt},
  hline{3-13} = {0.6pt, gray7},
}
Method & Decomposition \\
Tuformer \cite{liu2022tuformer} & Learnable head coupling \\
Multi-linear attention \cite{ma2019tensorized} & BT \\
GPT-TTM \cite{chekalina2023ttm} & TTM \\
Hypoformer \cite{li2022hypoformer} & Hybrid dense and TTM\\
CoMERA \cite{yang2024comera} & Rank-adaptive TT \\
Shapeshifter \cite{pahani2021shapeshifter} & Sum of Kronecker products \\
Monarch \cite{dao2022monarch} & Permuted block-diagonal factors \\
BTT \cite{qiu2024btt} & Sum of Monarch matrices \\
General einsum / BTT-MoE \cite{potapczynski2024generaleinsum} & General two-factor einsum / routed BTT \\
\end{tblr}
\end{table}


\subsection{Adaptation}
\label{subsec:peft}

\paragraph{Problem definition.}
Tensorized PEFT generalizes LoRA \cite{hu2022lora} by parametrizing the update as a low-rank tensor network. The general optimization problem for tensorized adaptation is
\begin{equation}
\begin{aligned}
& \min_{\Delta \ten{W} \in \mathfrak{TN}(\set{R})} \mathcal{L}_{\rm task}(\ten{W}_0 + \Delta \ten{W}; \ \mathcal{D}_{\rm task}), \\
& {\rm s.t.} ~ m(\Delta \ten{W}) \leq M,\ c(\Delta \ten{W}) \leq C
\end{aligned}
\end{equation}
where $\ten{W}_0$ is a frozen pretrained backbone, $\Delta \ten{W}$ is constrained to a tensor family $\mathfrak{TN}(\set{R})$ for a given set of ranks $\set{R}$, and $\mathcal{D}_{\rm task}$ is a task-specific dataset. The memory budget $m(\cdot)$ is the same as in \cref{subsec:pre-training}, only applied to the update. The compute budget $c(\cdot)$ depends on how the adapter is deployed. If it is merged into $\ten{W}_0$ after training, inference costs no more than the backbone alone. If it cannot be merged, as in multi-task serving where many adapters share one backbone, its contractions run in every forward pass and have to be counted.

The tensorization of $\Delta \ten{W}$ can take one of two forms. In the \emph{local} approach, each update is reshaped and parameterized as a tensor separately. In the \emph{global} approach, the updates are aggregated across all matrices into a single tensor, similar to the tensors in \eqref{eq:WQKV} and \eqref{eq:SGUD}, and then parameterized jointly.

\paragraph{Literature overview.}
The first line of work follows the local tensorization approach, meaning that each update matrix is tensorized independently. KronA \cite{edalati2022krona} replaces low-rank matrix factorization with a Kronecker rank-1 factorization. As discussed in \cref{subsec:illustrative-examples}, the resulting update matrix can have full rank.
SaMA \cite{nguyen2026sama} generalizes KronA's Kronecker update by using two block-diagonal factors with independent dense blocks, interleaved with fixed permutations in the style of Monarch matrices \cite{dao2022monarch}.
DoTA \cite{hu2024dota} represents each update matrix using an MPO initialized from the MPO decomposition of the pretrained weight $\mat{W}_0 \in \R^{I \times J}$, while keeping the residual matrix $\mat{W}_{\rm res} = \mat{W}_0 - \operatorname{MPO}(\mat{W}_0)$ frozen. Moreover, analogously to QLoRA \cite{dettmers2023qlora}, QDoTA \cite{hu2024dota} was proposed: it trains updates in bFloat16 while casting the frozen parts to NormalFloat4 \cite{dettmers2023qlora}.
LoRETTA \cite{yang2024loretta} provides two variants using TT parameterization of matrices: $\text{LoRETTA}_{\text{adp}}$ incorporates a TT-parameterized Houlsby-inspired adapter \cite{houlsby2019adapters} after each attention and FFN block, whereas $\text{LoRETTA}_{\text{rep}}$ makes a LoRA-like low-rank factorized incremental update $\Delta \mat{W} = \mat{B}\mat{A}$, where the matrices $\mat{B} \in \R^{I \times R}$ and $\mat{A} \in \R^{R \times J}$ are TT-parameterized.
ChainDoRA \cite{yeafi2026chaindora} extends the factor tensorization of $\text{LoRETTA}_{\text{rep}}$ to DoRA's magnitude--direction decomposition \cite{liu2024dora}, using a TT chain to parametrize both factors of the directional update.
Like $\text{LoRETTA}_{\text{rep}}$, TRAC \cite{ye2026trac} parameterizes $\mat{A}$ and $\mat{B}$ as separate tensor trains, with layer-specific trainable cores, frozen cores, and cross-layer shared cores modulated by layer-specific controllers.
TT-LoRA \cite{anjum2024ttlora} parameterizes the update $\Delta \mat{W}$ directly as a single tensor train.
TeRA \cite{gu2026tera} reshapes each weight update into a higher-order tensor and parameterizes it with a Tucker-like network. The randomly initialized core and factor matrices are frozen and shared across adapted layers, and only the diagonal scaling vectors specific to each update are trained.

A small cluster of works is inspired by the quantum tensor network literature. QuanTA \cite{chen2024quanta} reshapes the update matrix and parameterizes it as a quantum-circuit tensor network built from small "gates," i.e., tensors that connect only two axes. The main advantage is that QuanTA's tensor-network representation is not restricted to low rank, since such an update can have high rank \cite{chen2024quanta}. A related but structurally different method, Quantum-PEFT \cite{koikeakino2025quantumpeft}, instead follows the AdaLoRA \cite{zhang2023adalora} SVD-like update parameterization: $\Delta \mat{W} = \mat{U} \mat{\Lambda} \mat{V}^\top$ where $\mat{U} \in \R^{I \times R}$, $\mat{\Lambda} \in \R^{R \times R}$, and $\mat{V} \in \R^{J \times R}$. However, unlike AdaLoRA, which imposes orthogonality inexactly through a regularizer, Quantum-PEFT makes $\mat{U}$ and $\mat{V}$ orthogonal by construction via the Pauli parameterization. Their parameter count then grows logarithmically with the matrix dimension rather than linearly as in LoRA, while the diagonal factor contributes $R$ parameters. This yields comparable downstream performance at a substantially smaller parameter budget.

The second line of work follows the global approach, where all update weights are aggregated into a single tensor and then parameterized using some decomposition.
LoRTA \cite{hounie2024lorta} suggests a CP parameterization for all weights, composed into a 5th-order tensor $\Delta \ten{W} \in \R^{d\times d_h\times H\times L\times M}$, where the modes represent the model and head dimensions, the number of heads, the layer index, and the projection type (query, key, value, output).
LoTR \cite{bershatsky2024lotr} instead proposes composing a third-order tensor $\Delta \ten{W} \in \R^{d \times d \times N}$ by stacking all update matrices along the third mode. Then LoTR applies a Tucker-2 parameterization to this tensor: $\Delta \ten{W} = \ten{G} \times_1 \mat{A} \times_2 \mat{B}$, where the factors $\mat{A}$ and $\mat{B}$ are shared across all stacked projection updates, which helps make fine-tuning more parameter-efficient.
LoRA-CRAFT \cite{dewage2026loracraft} extends LoTR's cross-layer Tucker approach to pretrained attention weights, using HOSVD to obtain a frozen core and factors and training only small square transformations of the factor matrices.
MetaTT \cite{lopezpiqueres2025metatt} constructs a 4-th order update tensor $\Delta \ten{W} \in \R^{d\times L\times M\times d_h}$, where $d$ and $d_h$ are the model and head dimensions, $L$ is the number of layers, and $M$ is the projection type. MetaTT then utilizes a TT parameterization for this tensor. Moreover, this method proposes an adaptive rank optimization strategy \cite{lopezpiqueres2025metatt}.

\paragraph{Comparison.}
\Cref{tab:peft-structure} lists the decomposition and scope of tensorization for each adapter. Local methods parameterize updates separately and may share factors across layers, while global methods factorize several updates jointly.


\begin{table}[!htbp]
\centering
\caption{\footnotesize Structural comparison of tensorized PEFT methods.}
\label{tab:peft-structure}
\footnotesize
\begin{tblr}{
  width = \textwidth,
  colspec = {Q[4.3cm,m,c]X[m,c]Q[4.0cm,m,c]},
  row{1} = {font=\bfseries},
  rowsep = 3pt,
  hline{1,17} = {1pt},
  hline{2} = {0.6pt},
  hline{3,4,5,6,7,8,9,10,11,12,13,14,15,16} = {0.6pt, gray7},
}
Method & Decomposition & Type \\
KronA \cite{edalati2022krona} & Kronecker & Local \\
SaMA \cite{nguyen2026sama} & Permuted block-diagonal factors & Local \\
DoTA / QDoTA \cite{hu2024dota} & MPO & Local \\
$\text{LoRETTA}_{\rm adp}$ \cite{yang2024loretta} & TT adapter & Local \\
$\text{LoRETTA}_{\rm rep}$ \cite{yang2024loretta} & TT-parameterized LoRA factors & Local \\
ChainDoRA \cite{yeafi2026chaindora} & TT-parameterized directional factors & Local \\
TRAC \cite{ye2026trac} & TT with trainable, shared, and frozen cores & Local with cross-layer sharing \\
TT-LoRA \cite{anjum2024ttlora} & TT of the update & Local \\
TeRA \cite{gu2026tera} & Tucker with frozen shared core and factors, trainable diagonal matrices & Local with cross-layer sharing\\
QuanTA \cite{chen2024quanta} & Quantum-circuit tensor network & Local \\
Quantum-PEFT \cite{koikeakino2025quantumpeft} & SVD-like with Pauli factors & Local \\
LoRTA \cite{hounie2024lorta} & CP & Global \\
LoTR \cite{bershatsky2024lotr} & Partial Tucker & Global \\
LoRA-CRAFT \cite{dewage2026loracraft} & Frozen Tucker with trainable small transforms & Global \\
MetaTT \cite{lopezpiqueres2025metatt} & Rank-adaptive TT & Global \\
\end{tblr}
\end{table}

\subsection{Compression}
\label{subsec:compression}

\paragraph{Problem definition.}
Post-training compression is a two-stage process. The first stage minimizes the reconstruction error between the original and the factorized weights
\begin{equation}
\label{eq:compression}
\ten{Z}^* =\arg\min_{\ten{Z}\in\mathfrak{TN}(\set{R})}\|\ten{W}-\ten{Z}\|_F^2,
\end{equation}
where $\ten{W}$ collects the pretrained weights targeted for compression, $\ten{Z}$ is their tensorized approximation, and $\mathfrak{TN}(\set{R})$ is the set of tensors admitting a given decomposition with rank set $\set{R}$; the minimizer $\ten{Z}^*$ is the compressed model.
It is followed by an optional second stage, hereafter called \emph{healing}, that retrains the factorized model on a healing dataset, initialized with the factorized weights $\ten{Z}^*$ obtained from Eq. \eqref{eq:compression}
\begin{equation}
\label{eq:healing}
\min_{\ten{Z} \in \mathfrak{TN}(\set{R})} \mathcal{L}_{\rm heal}(\ten{Z}; \ \mathcal{D}_{\rm heal}), \quad \ten{Z}^{(0)} = \ten{Z}^*.
\end{equation}
Healing can range from short retraining on some dataset \cite{compactifai} to a lightweight knowledge distillation approach \cite{tahaei2021kroneckerbert}, so the healing loss $\mathcal{L}_{\rm heal}$ is not restricted to the pre-training language modeling loss.
The healing stage is not a cosmetic correction, as Zagitov et al. \cite{zagitov2026rethinking} give theoretical and empirical evidence that the reconstruction objective of Eq. \eqref{eq:compression} is misaligned with functional preservation. Consequently, a Frobenius-optimal $\ten{Z}^*$ need not be a good solution in the operator sense. In their experiments, even a lightweight LoRA-style repair recovers only part of the lost quality, which suggests that healing should be a full post-compression training stage.
    
Alternatively, the first stage can minimize the error on activations over a calibration dataset \cite{koikeakino2025latentllm}
\begin{equation}
\label{eq:compression-activation}
\ten{Z}^* = \arg\min_{\ten{Z} \in \mathfrak{TN}(\set{R})}
  \sum_{i \in \set{I}} \lambda_i \,
  \mathbb{E}_{x \sim \mathcal{D}_{\rm cal}}
  \bigl\| (\mat{W}_i - \mat{Z}_i)\, \vecb{x}_i \bigr\|_2^2 .
\end{equation}
Here $\set{I}$ indexes the components being compressed, including individual projections or projection groups that a method factorizes jointly, and $\mat{W}_i$, $\mat{Z}_i$ are the corresponding slices of $\ten{W}$ and $\ten{Z}$. The activation $\vecb{x}_i$ is the input induced at component $i$ by a calibration sample $x \sim \mathcal{D}_{\rm cal}$ in the uncompressed model, and $\mathcal{D}_{\rm cal}$ is the calibration dataset. The weights $\lambda_i > 0$ make the terms proportional, since components can differ in output dimension and activation scale, so an unweighted sum would let the largest term dominate.

\paragraph{Literature overview.} 
One line of work uses Kronecker decomposition, which yields a high compression ratio while keeping the factorized matrices high-rank.
Initially, Tahaei et al. \cite{tahaei2021kroneckerbert} applied a Kronecker rank-1 decomposition of each projection to the BERT model \cite{devlin2019bert}, followed by a KD healing stage.
Continuing this line, KnGPT \cite{edalati2021kroneckergpt} uses the same approach to compress the GPT-2 model \cite{radford2019language}, achieving the same compression ratio as DistilGPT2 \cite{sanh2019distilbert} with better perplexity and substantially lower training time. TQCompressor \cite{abronin2024tqcompressor} then adds column and row permutations to the Kronecker parameterization, which improves the compression ratio while matching quality on benchmarks.

Several works explore Tucker decomposition for model compression. TRAWL \cite{luo2025trawl} was among the first to apply a tensor decomposition to a tensor aggregated from projection matrices, motivated not only by compression but also by noise reduction. TRAWL explores CP and Tucker decomposition applied to a third-order tensor formed by stacking projection matrices one by one.
Following this line, TensorLLM \cite{gu2025tensorllm} proposes operating on a fourth-order tensor that separates the head dimension. Tucker is then applied to this tensor, leaving the head mode uncompressed; it can be viewed as a per-head full Tucker decomposition with shared factors and distinct third-order cores. This yields a high compression ratio with better accuracy than the uncompressed model, for GPT-J \cite{wang2021gpt_j} and LLaMA2 \cite{touvron2023llama2}. However, an accurate decomposition requires a high Tucker rank, so the dense core becomes a new memory bottleneck. To mitigate this issue, LeSTD \cite{li2026lestd} proposes a two-stage scheme: (1) finding the cores and shared factors as in TensorLLM, and (2) sparsifying the fourth-order core by pruning unimportant values. The authors also propose an inference algorithm without reconstruction. It yields better accuracy and matches or exceeds the throughput of TensorLLM and of non-tensor counterparts.

Other decompositions were explored as well. CompactifAI \cite{compactifai} applies MPO decomposition to the pretrained weights of LLaMA2-7B \cite{touvron2023llama2}, then heals on several datasets in less than one epoch. The authors systematically compare this approach against, and in combination with, quantization techniques. Saten \cite{solgi2025saten} also uses sparsity, but in a different way from LeSTD \cite{li2026lestd}. Saten first decomposes the weight matrices with error-based TT-SVD \cite{oseledets2011tensor} and then constructs an additive sparse correction to the residual $\mat{W} - \mat{W}_{\rm TT}$. LatentLLM \cite{koikeakino2025latentllm} follows the activation-aware optimization objective in Eq. \eqref{eq:compression-activation}, building the compression from a joint preconditioned factorization of three projection groups: query with key, value with output, and up with down. Validation across all scales of OPT \cite{zhang2022opt} shows that LatentLLM outperforms activation-aware non-tensor methods.

\paragraph{Comparison.}
\Cref{tab:compression-structure} compares the decompositions used for pretrained weights and the data or training each method requires. The table distinguishes data-free compression, activation calibration, and healing or task-specific fine-tuning.


\begin{table}[!htbp]
\centering
\caption{\footnotesize Structural comparison of tensorized compression methods.}
\label{tab:compression-structure}
\footnotesize
\begin{tblr}{
  width = \textwidth,
  colspec = {Q[4.3cm,m,c]X[m,c]Q[4.3cm,m,c]},
  row{1} = {font=\bfseries},
  rowsep = 3pt,
  hline{1,11} = {1pt},
  hline{2} = {0.6pt},
  hline{3,4,5,6,7,8,9,10} = {0.6pt, gray7},
}
Method & Decomposition & Training regime \\
KnGPT \cite{edalati2021kroneckergpt} & Kronecker & Healing by distillation \\
TQCompressor \cite{abronin2024tqcompressor} & Permuted Kronecker & Healing by distillation \\
TensorLLM \cite{gu2025tensorllm} & Tucker with a full head mode & Data-free \\
LeSTD \cite{li2026lestd} & Tucker with a sparse core & Data-free \\
CompactifAI \cite{compactifai} & MPO & Healing by fine-tuning \\
Saten \cite{solgi2025saten} & TT + sparse residual & Task-specific fine-tuning \\
LatentLLM \cite{koikeakino2025latentllm} & Joint Tucker via alternating SVD & Activation calibration \\
\end{tblr}
\end{table}
\FloatBarrier


\subsection{Inference}
\label{subsec:inference}

\paragraph{Problem definition.} Given the KV-cache tensor produced by MHA for an input $x$: $\ten{C}(x) = \bigl(\ten{K}(x), \ten{V}(x)\bigr) \in\R^{2\times L\times B\times H_{\rm KV}\times T\times d_h}$, the objective for the inference stage can, roughly speaking, be stated as follows:
\begin{equation}
  \begin{aligned}
      & \min_{\Pi} \mathbb{E}_{x \sim \mathcal{D}} \Bigl[ \mu \bigl(\ten{C}(x), \Pi(x)\bigr) \Bigr] \\
      & \rm s.t. \quad \text{memory/latency budget},
  \end{aligned}
\end{equation}
where $\Pi(\cdot)$ is such that, for every $x$ drawn from the input distribution $\mathcal{D}$, $\Pi(x) \in \mathfrak{TN}(\set{R})$ is a tensor-network approximation of the KV-cache, with $\mathfrak{TN}(\set{R})$ denoting the tensor network family given a set of ranks $\set{R}$; $\mu(\cdot, \cdot)$ is a measure of error, such as reconstruction error on the cache tensor or an activation-aware loss, analogous to \cref{subsec:compression}. We do not attempt a fully rigorous formalization here, as no work explicitly poses or solves this joint constrained problem; the common implicit goal across these methods is instead to find an efficient tensor parameterization or decomposition of the KV-cache that substantially reduces memory footprint while preserving quality and maintaining or improving inference throughput.

\paragraph{Literature overview.}
Tensor Product Attention (TPA) \cite{zhang2026tpa} factorizes per-token queries, keys, and values into sums of context-dependent outer products, e.g., $\mat{K}_t=R_{\rm K}^{-1}\sum_{r=1}^{R_{\rm K}}\vecb{a}_r^{\rm K}(x_t)\circ\vecb{b}_r^{\rm K}(x_t)$ with factors in $\R^H$ and $\R^{d_h}$.
Caching the factors takes $(R_{\rm K}+R_{\rm V})(H+d_h)$ scalars per token and layer, compared with $2Hd_h$ for MHA (see \cref{fig:example-inference}).
Kronecker Product Attention (KPA) \cite{nikpour2026kpa} extends this construction to contextual matrix factors, $\mat{K}_t=R_{\rm K}^{-1}\sum_{r=1}^{R_{\rm K}}\mat{A}_r^{\rm K}(x_t)\otimes\mat{B}_r^{\rm K}(x_t)$, with factors in $\R^{H_1\times d_1}$ and $\R^{H_2\times d_2}$, where $H=H_1H_2$ and $d_h=d_1d_2$.
Setting $H_2=d_1=1$ makes each Kronecker term an outer product, recovering the TPA parameterization.
Tucker Attention \cite{klein2026tuckerattention} applies Tucker parameterization to the fused query--key and value--output weight tensors across heads.
With both KV latent ranks equal to $R$, it caches $2R$ scalars per token and layer, or $R$ if the key and value share a down-projection.

For post-hoc compression, Krishnan and Schulz \cite{krishnan2026kvspectra} analyze 1024-token prefill caches from Mistral-7B-v0.3 and LLaMA-2-13B and find that the head and layer modes are nearly full rank, whereas the token and feature modes show more redundancy, especially for keys.
At matched storage, Tucker's mode-specific ranks yield the lowest relative Frobenius error among the tested tensor formats (Tucker, CP, t-SVD, TT) at $2$--$5\times$ compression, and four-mode Tucker gives the lowest value error in the joint KV comparison, while matrix unfoldings give lower key errors \cite{krishnan2026kvspectra}.
JoLT \cite{krishnan2026jolt} builds on this spectral evidence by applying partial Tucker to grouped prefill caches, truncating the token and feature modes while keeping the head and layer modes at full rank.
Additionally, it rotates and quantizes the truncation residual and jointly allocates Tucker ranks and residual bit widths under a cache byte budget.
DecoQuant \cite{liu2024decoquant} also combines decomposition with quantization, applying a two-core MPO decomposition to cached key and value matrices.
It quantizes the larger core, whose values have a narrower range, keeps the smaller core in FP16, and reconstructs the cache in FP16 for attention.
EinSort \cite{koikeakino2026einsort} explores permutations, motivated by the finding that sorting the elements of a matrix yields a substantially better low-rank approximation. Building on this, the framework proposes using an invertible permutation $\pi$ before compression and, after compression, restoring the initial order via $\pi^{-1}$. This poses a new challenge, since the permutation order needs to be stored for each token, which can be costly.

\paragraph{Comparison.}
\Cref{tab:inference} compares methods that reduce KV-cache storage. Architectural methods learn to produce factors from scratch during training, while training-free methods compress the keys and values produced by an existing model.


\begin{table}[!htbp]
\centering
\caption{\footnotesize Structural comparison of tensorized inference methods.}
\label{tab:inference}
\footnotesize
\begin{tblr}{
  width = \textwidth,
  colspec = {Q[5.0cm,m,c]Q[5.0cm,m,c]X[m,c]},
  row{1} = {font=\bfseries},
  rowsep = 3pt,
  hline{1,8} = {1pt},
  hline{2} = {0.6pt},
  hline{3,4,5,6,7} = {0.6pt, gray7},
}
Method & Decomposition & Training \\
TPA \cite{zhang2026tpa} & Sum of outer products & from scratch \\
KPA \cite{nikpour2026kpa} & Sum of Kronecker products & from scratch \\
Tucker Attention \cite{klein2026tuckerattention} & Tucker & from scratch \\
JoLT \cite{krishnan2026jolt} & Partial Tucker + residual quantization & training-free \\
DecoQuant \cite{liu2024decoquant} & MPO + quantization & training-free \\
EinSort \cite{koikeakino2026einsort} & Permutation + tensor decomposition & training-free \\
\end{tblr}
\end{table}

\subsection{Interpretability}
\label{subsec:interpretability}

\paragraph{Problem definition.} 
Interpretability operates under a structurally different goal than the rest of the survey. The preceding stages follow the implicit paradigm of making the model as capable as possible while minimizing memory, training, and inference costs. Interpretability instead aims to understand and explain existing models. Due to the vagueness of this goal, we do not state a formal objective here.

\paragraph{Literature overview.}
After tokenization, interpretability is the least developed stage in this survey, and nearly all of the work in it is recent. The contribution of tensor theory here is not efficiency but description, providing a way to write down what a model computes so that its structure becomes visible. In its weakest form, this contribution is purely notational; Taylor \cite{taylor2024interpretabilitysurvey4} rewrites the mathematical framework of \cite{elhage2021mathematical} in graphical tensor notation, up to the construction of the induction head, so that circuits become diagrams.

The three substantive works on interpretability differ in how they treat multilinearity. Bilinear MLPs \cite{pearce2025bilinearmlp} \emph{impose} it. A Gated Linear Unit without its element-wise nonlinearity is described exactly by a third-order weight tensor, so we can analyze it from the weights alone. We discussed this bilinear construction in \cref{sec:component-ffn} and provide it as the example in \cref{subsec:illustrative-examples}.
PolySAE \cite{koromilas2026polysae} instead \emph{adds} multilinearity where it was missing, since standard sparse autoencoders \cite{cunningham2023sae} reconstruct activations as linear combinations of dictionary atoms and therefore cannot separate a compound concept from the co-occurrence of its parts. PolySAE keeps the linear encoder but extends the decoder with quadratic and cubic terms, made tractable by a low-rank tensor factorization on a shared projection subspace, at roughly $3\%$ extra parameters on GPT-2 small.
TensorLens \cite{atad2026tensorlens} \emph{reformulates} the model as a whole: the entire Transformer stack, including attention, FFNs, normalization, and residual connections, becomes a single input-dependent operator represented as a fourth-order tensor $\ten{T} \in \R^{T \times d \times T \times d}$.
This tensor acts as a generalized attention matrix, replacing both the individual attention maps and their heuristic aggregations across heads and layers, and providing a global description of the model.

\paragraph{Comparison.} 
The interpretability methods described analyze different objects, so we do not provide a comparison table.


\section{Relationship with Neighboring Efficiency Methods}
\label{sec:neighboring-methods}


\begin{table}[!htbp]
\centering
\caption{\footnotesize Comparison of tensor methods with neighboring LLM-efficiency techniques. The table emphasizes interaction and complementarity.}
\label{tab:neighboring}
\footnotesize
\begin{tblr}{
width = \textwidth,
colspec = {Q[2.0cm,m,c]Q[2.5cm,m,c]Q[3.0cm,m,c]Q[3.0cm,m,c]X[m,c]},
row{1} = {font=\bfseries},
rowsep = 3pt,
hline{1,9} = {1pt},
hline{2} = {0.6pt},
hline{3,4,5,6,7,8} = {0.6pt, gray7},
}
Technique & Target & Typical advantages & Typical limitations & How tensors interact \\
Low-rank Adapters \cite{hu2022lora, zhang2023adalora} & Individual matrices or updates & Simple, mature, easy to implement & Limited to two-dimensional structure & Tensorized PEFT generalizes low-rank structure across layers, modules and heads (\cref{subsec:peft})\\
Quantization \cite{frantar2023gptq, dettmers2022llmint8, liu2023llmqat} & Numerical precision & Strong memory reduction, hardware support & Quality degrades at low bit-widths; post-training methods often need calibration data & Tensor factors can be quantized to different precisions \cite{liu2024decoquant} \\
Pruning / sparsity \cite{frantar2023sparsegpt, ashkboos2024slicegpt} & Weights: neurons, blocks, heads & Removes weights and the corresponding compute & Irregular sparsity may be slow & Tensor networks can be combined with structured sparsity \cite{solgi2025saten, li2026lestd} \\
KD \cite{hinton2015distillingknowledge} & Model behavior & Recovers performance in smaller models & Requires data and training & Tensorized student models can be distilled from dense teachers \cite{edalati2021kroneckergpt} \\
Efficient attention \cite{dao2022flashattention, wang2020linformer, deepseekai2025deepseekv32} & Attention compute and memory traffic & Addresses long-context bottleneck & Often does not reduce FFN or embedding parameters & Tensorized attention may complement IO-aware and linear/sparse methods \cite{ma2019tensorized} \\
KV-cache \cite{shazeer2019mqa, ainslie2023gqa, deepseekai2024deepseekv2} & Cached keys and values at inference & Reduces cache memory and bandwidth in long-context decoding & May degrade model quality & Non-tensor methods such as MQA and GQA are special cases of more general tensor parameterizations \cite{zhang2026tpa, klein2026tuckerattention} \\
MoE routing \cite{shazeer2017moe, fedus2022switch} & Parameters activated per token & Scales capacity without activating all parameters & Routing and load-balancing complexity & The expert index becomes an additional mode, so a single factorization can share structure across experts \cite{xu2026tdmoe} \\
\end{tblr}
\end{table}

Tensor methods interact with other LLM-efficiency techniques. \Cref{tab:neighboring} describes, for each neighboring technique, what it targets, its typical advantages and limitations, and exactly how it interacts with tensor methods.
A key observation is that tensor methods do not need to compete with these techniques as mutually exclusive alternatives. In realistic deployments, tensor factors can be quantized, tensorized adapters can be trained on a quantized backbone model, and tensor compression can be followed by knowledge distillation. 

\paragraph{How to read the interactions.}
The last column of \cref{tab:neighboring} contains three conceptually different relationships. First, techniques can be \emph{compositional}: tensorization changes the structural parameter count or operator graph, whereas quantization changes precision, pruning changes support, and knowledge distillation changes the training signal. These methods act on different design variables and can therefore be optimized jointly. Second, a tensor method can \emph{generalize} a neighboring method. Matrix low-rank adaptation is recovered when an update has only two modes, while tensorized PEFT extends sharing across layers, heads, or modules; similarly, MQA and GQA can be viewed as restrictecombination's gain tensor factorizations of attention projections or cached states. Third, methods can \emph{overlap}, i.e., efficient-attention kernels, linear or sparse attention, and tensorized attention may all target the same compute or memory traffic. In this case, the gain of the combination is the marginal gain after the first method has already changed the bottleneck, not the product of two independently reported speedups.

For example, in the case of quantization, a compact memory model is
\begin{equation}
\label{eq:memory-budget}
M_{\rm bits} = \sum_{k=1}^{K} n_k b_k + M_{\rm meta},
\end{equation}
where the tensorized model is stored as $K$ factors/cores, $n_k$ is the number of scalars retained in part $k$, $b_k$ its precision in bits, and $M_{\rm meta}$ collects what the format needs besides the values themselves, such as quantization scales and zero-points.
Written this way, the two techniques act on different arguments of one expression: tensorization determines $n_k$, quantization determines $b_k$, and both are chosen per factor. A sequential pipeline such as decompose then quantize, or decompose an already quantized model, fixes one argument before examining the other and therefore searches only a slice of this space, with no reason for the result to be optimal. Schemes that assign different precisions to different factors are a first step out of that slice \cite{liu2024decoquant}, and the argument is not specific to quantization. 
The last term $M_{\rm meta}$ decides whether structural savings survive into the stored format. For instance, regular structure is almost free to describe, whereas irregular structure, which group carries which scale, which core entries survived pruning, has to be enumerated. Since high compression ratios often come from introducing such irregularity, a ratio computed from the values alone can overstate the memory actually used.

\paragraph{Order of composition and error interaction.}
Even when two techniques target different variables, their order can matter. Decomposition changes the distributions and sensitivities of the resulting factors, which changes how aggressively each factor can be quantized. Pruning before decomposition may destroy a low-rank pattern that the factorization could otherwise exploit, whereas pruning the factors after decomposition introduces a different structured approximation. Distillation can be used either to train a tensorized student directly or as a healing stage after compression. Consequently, the errors of two techniques need not add linearly: they may amplify one another, partially cancel, or act as an implicit regularizer. A combined method should therefore be optimized and evaluated as a joint pipeline rather than assembled from hyperparameters tuned for each technique in isolation.

\paragraph{Bottleneck-aware combinations.}
The appropriate combination depends on which resource dominates the operating regime. If weight bandwidth is the bottleneck, tensorization combined with quantization or structured sparsity can reduce the bytes fetched per token. In long-context decoding, attention and KV-cache methods are more likely to dominate, so evaluate tensor cache factorization alongside MQA/GQA and IO-aware kernels. For adaptation, a tensorized update can reduce trainable state while a quantized backbone controls the frozen-model footprint. Knowledge distillation is different again since it can recover quality or transfer behavior, but its benefit must be weighed against the additional training cost. This stage-specific view prevents crediting a method that improves an inactive resource with an end-to-end gain it cannot deliver.

The limitation can be stated through an Amdahl-style bound. Let $p$ be the fraction of baseline runtime affected by a technique and let $S_{\rm target}$ be the speedup of that fraction. Then the end-to-end speedup satisfies
\begin{equation}
S_{\rm e2e} \leq \frac{1}{(1-p)+p/S_{\rm target}}.
\label{eq:neighboring-amdahl}
\end{equation}
Thus, even an arbitrarily fast tensorized module gives only a modest gain when untensorized FFNs, attention, communication, or framework overhead dominate the remaining runtime. Methods that target disjoint bottlenecks can complement each other, whereas methods that target the same fraction exhibit diminishing returns. This also explains why the same combination may help prefill, decoding, fine-tuning, and batch serving by very different amounts.

\paragraph{Evaluation of combined methods.}
A convincing comparison should include four systems: the dense baseline, the tensor-only method, the neighboring technique alone, and their combination, all at matched quality and under the same hardware and precision. Reporting only the combined result makes it difficult to attribute the gain to tensorization. Moreover, memory accounting should include tensor metadata, quantization scales, sparsity indices, temporary contraction buffers, and adapter state. Latency should be separated into prefill and decoding and reported both per module and end-to-end. Finally, state the order of operations and the tuning budget for every baseline explicitly, and report $\rho_{\rm gap}$ from \cref{eq:rho-gap} for the tensorized part and the full pipeline.


\section{Relationship to Probabilistic Tensor Networks}
\label{sec:prob-tensor-nets}

The connection between tensor decompositions and probabilistic graphical models (PGMs) has been recognized for over a decade, primarily in the context of unsupervised learning and density estimation \cite{hameed2025efficient,han2018unsupervised,miller2021tensor,vieijra2022generative,glasser2019expressive}. This relationship is often invoked to justify tensor-network architectures for sequential data, but the correspondence is subtle and requires careful qualification. In this section, we clarify the precise conditions under which a tensor decomposition corresponds to a PGM, discuss the limitations of these correspondences for real-valued tensor networks, and outline the implications for language modeling.

\subsection{Tensor Decompositions as Parameterizations of Graphical Models}

A joint probability mass function over $N$ discrete random variables $X_1,X_2,\dots, X_N$ can be represented as a tensor $\ten{P} \in \R^{I_1 \times I_2\times \dots \times I_N}$ with non-negative entries summing to one. Many common PGMs correspond to low-rank or structured tensor factorizations of $\ten{P}$, with the decomposition directly encoding the model's conditional independence structure.

\paragraph{CP decomposition and naïve Bayes.}
The CP decomposition  expresses an $N$-th-order tensor as
\begin{equation}
\ten{P}_{i_1,\dots,i_N} = \sum_{r=1}^{R} \lambda_r \mat{A}^{(1)}_{i_1,r} \mat{A}^{(2)}_{i_2,r} \cdots \mat{A}^{(N)}_{i_N,r}.
\end{equation}
If $\ten{P}$ is a joint probability tensor and the entries $\lambda_r, \mat{A}^{(n)}_{i_n,r}$ are non-negative and suitably normalized, then the CP representation corresponds exactly to a naïve Bayes model with a latent class variable $Z \in \{1,2,\dots,R\}$. Specifically, each observed variable $X_n$ is conditionally independent given $Z$, with conditional distributions $\Pr(X_n = i_n \mid Z = r) \propto \mat{A}^{(n)}_{i_n,r}$, and prior $\Pr(Z = r) \propto \lambda_r$ (up to normalization constants). This correspondence is well-known in the tensor decomposition literature and has been exploited for topic modeling \cite{anandkumar2014tensor} and non-negative tensor factorization \cite{cichocki2009nonnegative}.

\paragraph{TT decomposition and hidden Markov models.}
The tensor train decomposition \cite{oseledets2011tensor} represents a tensor as:
\begin{equation}
\ten{P}_{i_1,\dots,i_N} = \mat{G}^{(1)}_{i_1} \mat{G}^{(2)}_{i_2} \cdots \mat{G}^{(N)}_{i_N},
\end{equation}
where each $\mat{G}^{(n)}_{i_n} \in \R^{R_{n-1} \times R_n}$ is a matrix slice of the $n$-th core, and $R_0 = R_N = 1$.

If the entries are non-negative and the cores satisfy certain column-stochastic normalization conditions, the TT format is equivalent to a hidden Markov model (HMM)    \cite{glasser2019expressive}. Specifically, the bond dimension $R_n$ corresponds to the number of hidden states at position $n$, and the core $\mat{G}^{(n)}_{i_n}$ encodes the transition probabilities from hidden state $R_{n-1}$ to $R_n$ together with the emission probability of $X_n = i_n$.
This connection has been used in tensor-network-based density estimation \cite{han2018unsupervised}, generative modeling \cite{glasser2020probabilistic}, and sequence modeling \cite{stoudenmire2016supervised}.

Once again, the correspondence is conditional, since an unconstrained real-valued TT representation of a weight matrix does not correspond to an HMM. The cores are not stochastic matrices, the bond dimensions lack a probabilistic interpretation, and there is no underlying latent process. The TT format is simply a structured, low-rank parameterization that shares the same algebraic form as an HMM's parameterization but without the probabilistic semantics.

\paragraph{Block-term decomposition and more general PGMs.}
The BT decomposition represents a tensor as a sum of Tucker blocks:
\begin{equation}
\ten{P} = \sum_{k=1}^{K} \ten{G}_k \times_1 \mat{A}^{(1)}_k \times_2 \mat{A}^{(2)}_k \cdots \times_N \mat{A}^{(N)}_k.
\end{equation}
When the factors and core are non-negative and normalized, BTD can represent mixtures of latent-variable models, such as mixture of naïve Bayes (where each block is a naïve Bayes component) or more general hierarchical models. Conversely, an unconstrained BTD has no PGM interpretation and should be viewed purely as a flexible tensor parameterization.

\subsection{Tensor Networks as Matrix-Valued Stochastic Processes}

The term ``matrix-valued Markov chain'' is often used informally to describe the TTM or MPO parameterization of a sequence of matrices. This terminology is evocative but not formally precise. An MPO is a tensor network of the form:
\begin{equation}
\ten{W}_{i_1,j_1,\dots,i_N,j_N} = \mat{G}^{(1)}_{i_1,j_1} \mat{G}^{(2)}_{i_2,j_2} \cdots \mat{G}^{(N)}_{i_N,j_N},
\end{equation}
where each $\mat{G}^{(n)}_{i_n,j_n} \in \R^{R_{n-1} \times R_n}$ is a matrix slice indexed by an input-output pair $(i_n, j_n)$.

This representation is \emph{not} a Markov chain in the probabilistic sense. It has no transition probabilities, no state space over which probability mass is conserved, and no stochastic interpretation. The term ``matrix-valued'' means each slice is a matrix, and ``chain'' refers to the sequential contraction of these matrices. It is a useful mnemonic, but it should not be interpreted as implying a probabilistic Markov process unless the cores are explicitly constrained to be column-stochastic and the appropriate normalization conditions are imposed

\subsection{Implications for Language Models}

The correspondence between tensor decompositions and PGMs has several implications for tensor methods in language modeling:

\begin{enumerate}
\item \textbf{Interpretability:} When a tensor decomposition is applied to an embedding table, attention tensor, or activation tensor, the factors do \emph{not} automatically correspond to interpretable latent variables such as topics, syntactic roles, or semantic features. For factors to be interpretable in a probabilistic sense, the decomposition must be non-negative and properly normalized. This is rarely the case in practice; most LLM tensorization uses real-valued decompositions optimized for reconstruction or task loss.

\item \textbf{Generative modeling:} If the goal is to build a generative language model directly from a tensor network, the decomposition must be constrained to represent a valid probability distribution. This requires non-negativity and normalization, which complicate optimization and may reduce expressivity compared to unconstrained real-valued factorizations.

\item \textbf{Sequential structure:} The TT and MPO formats are natural for sequential data, as they impose a chain-like structure that mirrors the temporal or positional ordering of tokens. However, this structural similarity does not imply that the model captures temporal dependencies in a probabilistic sense; it simply means the parameterization aligns with the data's sequential nature.

\item \textbf{Unified view:} Treating tensor decompositions as parameterizations of PGMs provides a unified perspective on tensor methods for language models. It suggests that tensorization can be viewed as imposing a particular conditional independence structure on the model's representations, which may serve as a useful inductive bias. This perspective is particularly relevant for understanding the expressivity and generalization properties of tensorized architectures.
\end{enumerate}


\section{Software overview}
\label{sec:software-overview}

\subsection{Packages and Frameworks.}

\paragraph{Python tensor libraries.}
This paragraph lists the most relevant Python packages compatible with, or specifically designed for, deep learning scenarios. TensorLy \cite{kossaifi2019tensorly} is a general-purpose tensor toolbox that runs on many backends, including NumPy \cite{harris2020numpy}, JAX \cite{bradbury2018jax}, and PyTorch \cite{paszke2019pytorch}, whereas tntorch \cite{usvyatsov2022tntorch} targets PyTorch directly. Packages for tensorized training build upon mature deep learning frameworks: TensorLy-Torch \cite{kossaifi2024tensorlytorch} for PyTorch and T3F \cite{novikov2020t3f} for TensorFlow \cite{abadi2016tensorflow}. Other packages, including TensorKrowch \cite{pareja2024tensorkrowch} and tn4ml \cite{puljak2025tn4ml}, address machine-learning applications more broadly. Finally, quimb \cite{gray2018quimb} supports tensor networks of arbitrary topology across several backends, and cotengra \cite{gray2021hyperoptimized} specializes in optimizing their contraction.

\paragraph{Hardware co-design.} 
A separate line of work co-designs software and hardware for tensorized layers. ETTE \cite{gong2023ette} and Huang et al. \cite{huang2025gvsa} optimize the forward pass for tensorized linear layers to accelerate inference, whereas FETTA \cite{lu2026fetta} and the FPGA design of Tian et al. \cite{tian2025fpga} cover both training and inference.

\paragraph{Others.} 
The remaining tools fall into three groups. Some target languages other than Python, including Julia, C++, and MATLAB; examples include ITensor \cite{fishman2022itensor}, TenDeC++ \cite{huang2019tendec}, and Tensor Toolbox \cite{bader2026tensortoolbox}. Others implement a single decomposition family without specializing in deep learning, as in Scikit-TT \cite{gelss2019scikittt}. A third group has not been updated for five years or more: TTAX \cite{novikov2021ttax}, TorchMPS \cite{miller2019torchmps}, and TedNet \cite{yu2022tednet}.

\subsection{Case studies}

We now examine two recent studies that illustrate how these tools are used in practice at different scales, with one operating at research scale and the other at production scale. Javanmard et al. \cite{javanmard2026mpopicogpt} apply MPO parameterization to every linear layer of PicoGPT \cite{osborne2026picogptjl}, a small, GPT-2-style character-level model. Their work provides an open-source \texttt{MPOLinear} module, whose cores can be initialized from a dense matrix or trained from scratch with standard PyTorch autograd. They validate the tensorized model against a dense baseline on the Tiny Shakespeare dataset. Kozyrev et al. \cite{kozyrev2026minima} propose Minima, a deployment-ready compression pipeline for production-scale LLMs. The pipeline consists of sensitivity analysis, Tucker/TT/Tensor Ring compression, a short healing stage, custom Triton/CUDA kernels, and speculative decoding. The authors validate it on the Qwen3-32B \cite{yang2025qwen3} model.


\section{Discussion}
\label{sec:future-directions}

In this section, we discuss challenges, gaps, and future research directions for tensor methods in LLMs.

\subsection{Compression-Realization Gap}
\label{subsec:rho_gap}

A recurrent problem with tensorized models is that a substantial reduction in parameter
count rarely converts into a proportional speedup, either in training or at inference.
We do not report new measurements here; the aim is to name the phenomenon, to separate
its two sources, and to fix a protocol under which it becomes comparable across methods.
Intuitively, the quantity of interest is
\begin{equation*}
  \rho_{\rm gap} = \frac{\text{theoretical compression}}{\text{practical speedup}},
\end{equation*}
whose numerator is what a paper usually claims and whose denominator is what a user
obtains. Let $B$ denote the total number of bytes, $F$ the FLOPs count, and $T$ the
wall-clock time, with the subscripts ``dense'' and ``TN'' referring to the baseline
dense model and to its tensorized counterpart. We define byte compression, FLOPs
compression, and speedup as
\begin{equation}
  C_B = \frac{B_{\rm dense}}{B_{\rm TN}}, \quad
  C_F = \frac{F_{\rm dense}}{F_{\rm TN}}, \quad
  S_T = \frac{T_{\rm dense}}{T_{\rm TN}}.
\end{equation}
These quantities are comparable only under a common setup, which we require to satisfy:
(i) the same device and numerical precision, so that the peak throughput is shared;
(ii) an optimized dense baseline; (iii) matched quality for the dense and the TN model; (iv) the same scope for all three quantities, either per module or end-to-end; and (v) a declared operating regime, with
prefill and decoding reported separately, as they sit on opposite sides of the
compute-bound/memory-bound boundary.

We define the \emph{compression-realization gap} as
\begin{equation}
  \rho_{\rm gap} = \frac{C_B}{S_T}
  = \underbrace{\frac{C_B}{C_F}}_{\text{algorithmic}} \times
    \underbrace{\frac{C_F}{S_T}}_{\text{realization}},
  \label{eq:rho-gap}
\end{equation}
which splits the gap into two complementary and independently actionable parts. The
algorithmic term depends only on the decomposition and the contraction scheme, not on
the hardware; it exceeds one when FLOPs increase more rapidly than memory as the
decomposition-specific ranks grow, or when a suboptimal contraction order is used. The
realization term admits an exact reading in terms of hardware utilization. Model FLOPs
utilization is defined with respect to the device peak throughput as
$\mathrm{MFU} = F / (T \cdot F_{\rm peak})$, so that $T = F / (\mathrm{MFU} \cdot
F_{\rm peak})$. Substituting this into $S_T$ and using the shared $F_{\rm peak}$
granted by condition (i), we obtain
\begin{equation}
  \frac{C_F}{S_T} = \frac{\mathrm{MFU}_{\rm dense}}{\mathrm{MFU}_{\rm TN}}.
\end{equation}
This identity is informative only in the compute-bound regime, where FLOPs govern the runtime. In the memory-bound regime the
runtime is set by the bytes moved, and the analogous quantity is the achieved
memory-bandwidth utilization.

A value $\rho_{\rm gap} = 1$ means that the system realizes all of the idealized
byte-compression factor as speedup. The typical case is $\rho_{\rm gap} > 1$, and the
two factors attribute it either to a decomposition that buys memory at the price of
arithmetic, or to a model that utilizes the device worse than dense GEMMs. A value
$\rho_{\rm gap} < 1$ is also possible in the memory-bound regime, where shrinking the weights
or the KV-cache reduces exactly the resource that bounds the runtime, so the speedup may
exceed the byte compression. Since the metric only contrasts nominal compression with
realized speedup, it applies equally to quantization \cite{frantar2023gptq,
dettmers2022llmint8, liu2023llmqat}, sparsification \cite{frantar2023sparsegpt,
ashkboos2024slicegpt}, and IO-aware attention kernels \cite{dao2022flashattention},
providing common ground for the comparison tensor and non-tensor methods; we return
to it in \cref{subsec:gaps}.

\subsection{Gaps, Challenges, and Possible Solutions}
\label{subsec:gaps}

This subsection collects the open problems that the reviewed works themselves report.
We took them from the limitations and future-work sections of the surveyed papers and
grouped them by theme. For each gap we describe the challenges that arise there and the
directions that could resolve them. \Cref{tab:gap-map} presents this in compact form,
and the paragraphs that follow discuss each gap in more detail.


\begin{table}[!htbp]
\centering
\caption{\footnotesize Gaps and challenges overview across the tensor methods literature.}
\label{tab:gap-map}
\footnotesize
\begin{tblr}{
  width = \textwidth,
  colspec = {Q[4.0cm,m,c]Q[5.5cm,m,c]X[m,c]},
  row{1} = {font=\bfseries},
  rowsep = 3pt,
  hline{1,10} = {1pt},
  hline{2} = {0.6pt},
  hline{3,4,5,6,7,8,9} = {0.6pt, gray7},
}
Gap & Challenges & Possible solutions \\
Hardware compatibility & Small, sequential, and irregular contractions underutilize GPU parallelism, so fewer parameters need not mean faster computations & Open-source Triton/CUDA libraries with fused and structure-aware kernels, evaluated via measured $\rho_{\rm gap}$ \\
Rank selection & Optimal rank differs across layers and component types; matrix adaptive schemes do not easily extend to tensor ranks & Adaptive rank allocation algorithms, benchmarked against fixed-rank heuristics \\
Tensorization scheme & Schemes with equal parameter count differ in quality and contraction cost; scheme selection is a complex search whose objective is expensive to evaluate & Construct an admissible set of schemes and search it efficiently, scoring candidates by validation loss, memory, and contraction cost \\
New tensor networks & Richer topologies introduce more complex ranks/scheme choices; raise contraction cost and latency  & evaluate underexplored topologies per component/stage, reporting quality, memory, and latency jointly   \\
Training/optimization dynamics & Training can be unstable since the tensor parameterization changes the loss landscape, yet optimizers are tuned for dense weight matrices &  Diagnostics of gradient scale and initialization across cores; optimization routines that account for the tensor structure \\
Scaling & Training tensorized models from scratch at large scales is beyond most academic compute, which affects the pre-training stage most & Measure how quality and compression ratio change with scale, giving an analog of scaling laws for tensorized models \\
Compatibility with neighboring efficiency methods & Orthogonality to other efficiency methods holds in principle, but whether gains add up and how errors compose is unknown & Ablations separating the contribution of tensorization from that of other methods \\
Benchmarking and evaluation & Methods use different baselines, datasets, and per-paper tuning, so reported gains are not directly comparable & A per-stage benchmark on a shared baseline, reporting the compression-realization gap next to quality \\
\end{tblr}
\end{table}

\paragraph{Hardware compatibility.} 
Tensorization of all components and stages shares one common challenge: hardware
compatibility. Modern GPUs are built for dense GEMMs, which are large, regular, and
highly parallelizable, whereas tensor contractions are small, sequential, and
irregular, so a model with fewer parameters need not run faster.
Closing this compression-realization gap is probably a kernel-level problem. Two kinds of GPU kernels could address it. Fused kernels
combine a chain of contractions into a single launch, removing per-launch overhead and
keeping intermediate factors in fast memory instead of writing them back to global
memory. Structure-aware kernels exploit the structure that a decomposition imposes on
the factors, such as sparse cores, instead of treating them as dense arrays. One
practical solution is open-source Triton or CUDA libraries implementing such kernels,
with their benefit reported as measured $\rho_{\rm gap}$ (Eq. \ref{eq:rho-gap}).

\paragraph{Rank selection.}
Ranks are the main control knob of a tensorized model. Uniform rank selection is a
convenient heuristic, but it is not optimal, since the optimal ranks may differ between
layers and component types. Algorithms that allocate ranks adaptively should, in
principle, dominate uniform assignment at an equal parameter budget. At the same time, generalizing
matrix adaptive schemes, such as AdaLoRA \cite{zhang2023adalora}, to tensor ranks is
not straightforward. Several reviewed studies
\cite{yang2024comera, lopezpiqueres2025metatt}, as well as work beyond our scope
\cite{hawkins2019rankselection1, hawkins2021rankselection2, gu2022heat}, propose
adaptive schemes for automatic rank allocation.
However, a universal and generally applicable method for adaptive rank allocation is still missing.

\paragraph{Tensorization scheme.}
A second design choice concerns the tensorization scheme itself, namely how a model parameter
becomes a tensor object, and which of the admissible mappings works best in practice.
For example, in the imposed setting (\cref{sec:tensor-strategies}), turning a weight
matrix $\mat{W}$ into a tensor $\ten{W}$ requires choosing the number of modes and
their sizes, and these choices are not obvious. Different choices may give the same
parameter count but differ in approximation quality and contraction cost. This area
remains under-theorized.

\paragraph{New tensor networks.}
Applying new tensor networks is a natural and promising direction. In the imposed tensorization setting, for instance, nothing in principle restricts the applicable tensor network to TTM; richer topologies are equally valid. Deep tensor networks are potentially more expressive and memory-efficient than shallow, chain-like topologies such as TTM, and they have shown promising results in domains beyond LLMs and NLP \cite{adtn}, suggesting they may transfer to the LLM setting as well. However, the deeper and more complicated the tensor network, the higher the computational latency, since the contraction cost grows. Tensor Ring \cite{zhao2016tensorring}, tree-structured \cite{gracedyck2010h_tucker}, fully-connected \cite{zheng2021fctn}, PEPS \cite{verstraete2008peps}, MERA-inspired \cite{vidal2008mera}, and Kronecker Tensor Decomposition \cite{batselier2017ktd} networks remain underexplored in LLM tensorization.

\paragraph{Training and optimization dynamics.} 
Training a tensorized model with standard techniques and optimizers can be unstable
\cite{barratt2022tnunstabletraining}. One reason is that the tensor parameterization
substantially changes the loss landscape, so optimizers designed and tuned for dense
weight matrices, such as Adam, may not be optimal for tensorized models
\cite{yang2024comera}. What is still missing is an understanding of how tensorized
training behaves in this landscape, including how gradient magnitudes are distributed across
cores, how core initialization affects convergence, and how these quantities evolve
during training. Such diagnostics would also indicate what an optimizer actually has to
correct. Studying optimization routines that account for the tensor structure could
then make training both faster and more stable.
Muon possibly offers a starting point for such structure-aware optimization \cite{li2026tmuon, miasnikov2026monarchmuon}.

\paragraph{Scaling.} 
Training a tensorized language model from scratch is expensive, and training one at several scales, or at a large scale at all, is beyond the compute of most academic groups \cite{chekalina2023ttm}. This affects the pre-training stage most. The open question is how the benefits of tensorization change as the model grows, specifically how compression ratio and quality behave at each scale, thereby providing an analog of scaling laws \cite{kaplan2020scalinglaws} for tensorized language models. Studies of BTT and BTT-MoE have begun to address this question at GPT-2 scale by comparing language-modeling loss as a function of training compute \cite{qiu2024btt,potapczynski2024generaleinsum}. Whether the same trends hold at larger scales remains an open question. Scaling is also a practical
question at the other stages, since how much a model can be compressed depends on how it was trained. Recent models are often distilled and probably harder to compress
\cite{solgi2025saten}. A compression ratio measured on one model therefore need not translate to a newer or larger one.

\paragraph{Compatibility with neighboring efficiency methods.} 
\Cref{sec:neighboring-methods} discusses how tensor methods relate to quantization, pruning, distillation, and other efficiency techniques.
Most of them are orthogonal to tensorization in principle, but their practical interaction, including whether the gains add up and how the errors compose, remains an open question.

\paragraph{Benchmarking and evaluation.}
Numbers reported across papers are often not directly comparable; methods are evaluated on different baselines and datasets, and each is tuned separately, so differences in reported quality may come from the setup, not just the method. Claiming that one method is better than another requires a comparison on a shared baseline, with hyperparameters tuned for every method under the same approach. A benchmark would provide this, and it is probably better to have one per stage, since the stages optimize for different things. Whatever the stage, such a benchmark must report the compression-realization gap (Eq. \ref{eq:rho-gap}) alongside quality. Otherwise, a method that saves parameters on paper but not time on a GPU is indistinguishable from one that saves both.

\subsection{Future Directions}
\label{subsec:future-directions}

The preceding subsection focuses on obstacles that must be removed for existing tensor methods to become reliable and efficient. Here we highlight longer-horizon directions that change where tensorization enters the model lifecycle, which objects it acts on, or what capabilities it is intended to provide. These directions extend the stage-specific literature toward tensor-native language-model design.

\paragraph{Tokenization.} 
No tensor method reaches the tokenizer (\cref{subsec:tokenization}), although relatedness between tokens is lost as soon as the vocabulary is used, as the vocabulary built by BPE and its variants is a set of subword strings, but each string is used only through its index. Consequently, two tokens sharing a morpheme may receive unrelated embeddings. Tensor methods naturally impose structure that reflects this relatedness. MorphTE \cite{gan2022morphte} does this for embeddings, but takes the segmentation from an external analyzer and leaves the vocabulary unchanged. More broadly, the vocabulary and the embedding table are still built sequentially; forming them jointly remains an open problem. The obstacle is that the tokenizer is a discrete, non-differentiable map, so there is no continuous object to factorize.

\paragraph{Interpretability.}
Interpretability is underexplored for the opposite reason. The stage is not empty but new and, in our view, the most promising. Apart from the notational work, every method surveyed in \cref{subsec:interpretability} has appeared within the last year. Together they show that multilinear structure can support understanding in three ways. It can be \emph{added} to the analysis tool, as in PolySAE \cite{koromilas2026polysae}; \emph{reformulated} from a model that is multilinear without having been written that way, as in TensorLens \cite{atad2026tensorlens}; or \emph{imposed} at training time as an inductive bias, as in bilinear layers \cite{pearce2025bilinearmlp}. The last route is the most demanding, since it requires pre-training from scratch, and the most far-reaching, as it turns interpretability into an architectural design constraint.

A separate opportunity is graphical tensor notation \cite{taylor2024interpretabilitysurvey4}. Its immediate benefit is expository: a diagram shows which modes a method contracts, shares, or truncates; accordingly, we encourage authors of tensor interpretability methods to draw their constructions this way. Beyond that, the notation is a reasoning tool in its own right, since asking which contraction orders, factorizations, and shared factors a network admits can reveal new mechanisms.

\paragraph{Lifecycle-level co-design.}
Most existing methods optimize tensorization at one lifecycle stage while treating the surrounding stages as fixed. A tensor-native model could instead co-design the embedding factorization, base-model ranks, adaptation subspaces, post-training compression, and inference contraction schedule. Such coupling matters, since a rank pattern that minimizes pre-training loss may be unsuitable for later PEFT, and a compact algebraic decomposition may be difficult to execute efficiently during decoding. A useful formulation is therefore multi-objective: select tensorization schemes and ranks jointly to optimize quality, training memory, adaptation capacity, serving latency, and energy under explicit resource budgets. Results should be reported as Pareto frontiers.

\paragraph{Tensorized training state and distributed learning.}
The literature focuses primarily on model weights, adapters, and KV caches, but large-scale training also stores activations, gradients, optimizer moments, and communication buffers. These objects have natural modes associated with layers, microbatches, sequence positions, heads or experts, and devices. Exploiting low-rank structure across such modes could reduce both accelerator memory and inter-device communication, extending tensorization from model representation to the full training state. The central difficulty is that these tensors are transient and non-stationary: their useful ranks may change over training, and approximation errors can accumulate through optimization. Future work should therefore evaluate convergence, numerical stability, communicated bytes, synchronization cost, and end-to-end training throughput together.

\paragraph{Conditional and elastic tensorization.}
Nearly all surveyed methods use ranks and contraction graphs that remain fixed for every input. A more flexible model could allocate tensor capacity conditionally by layer, token, context length, task, or confidence, activating additional cores or rank components only when they sufficiently improve the prediction. This would turn rank from a static architecture hyperparameter into a runtime compute budget and could provide controllable quality--latency trade-offs. The systems challenge is substantial: variable contraction shapes interfere with batching, compilation, and fused kernels. Evaluation must consequently include worst-case memory and tail latency, not only average FLOPs or average rank.

\paragraph{Multimodal and modular architectures.}
Multimodal language models and mixture-of-experts architectures expose semantic modes that are absent from a single dense text model, including modality, expert, routing group, spatial position, and temporal position. Tucker, block-term, tree-structured, or shared-core decompositions could separate globally shared linguistic factors from modality- or expert-specific factors. This may offer a principled alternative to duplicating full projection and adapter matrices across modules. Open questions include how to preserve modality-specific capacity, routing balance, and rare-expert behavior while avoiding negative transfer from an overly restrictive shared core. This direction would also test whether semantically meaningful tensor modes are more robust than imposed reshaping at scale.
Existing MoE methods illustrate how tensorization can be used at both the expert and adapter levels, with MPOE \cite{gao2022mpoe} sharing a central MPO core across FFN experts, TT-LoRA MoE \cite{kunwar2025ttloramoe} using top-1 routing over task-specific TT adapters, and TD-MoE \cite{xu2026tdmoe} compressing pretrained expert weights via a joint Tucker decomposition.


\section{Experiments}
\label{sec:experiments}

We focus on pre-training and adaptation, two stages where tensor methods can be compared under a shared training and evaluation protocol.
For stages such as embeddings and inference, the few available works often pursue different objectives, which makes direct comparisons difficult.
For compression, the literature already provides a systematic benchmark \cite{zagitov2026rethinking}.
The code for all experiments is available in our \href{https://github.com/ma-tt-a/awesome-tensor-methods-for-llms}{GitHub} repository.

\subsection{Pre-training}
\label{subsec:exp-pre-training}
The pre-training results reviewed in \cref{subsec:pre-training} span different architectures and datasets.
We compare four factorized parameterizations of the same GPT-2 architecture to answer two questions.
\begin{enumerate}
  \item \textbf{Efficiency.} How much of the parameter savings translates into shorter forward and backward passes, and how does this depend on batch size and compilation?
  \item \textbf{Quality.} How does validation loss change with the parameter budget, and what can the singular value spectra of the learned projections tell us about these differences?
\end{enumerate}

\paragraph{Baseline model.}
We use the dense GPT-2 small model (124M parameters) as our baseline, following the nanoGPT implementation \cite{karpathy2022nanogpt}.

\paragraph{Tensorized models.}
\begin{wraptable}{r}{0.4\linewidth}
\centering
\caption{\footnotesize Ranks for tensorized models.}
\label{tab:tensor_ranks}
\footnotesize
\begin{tabular}{lc}
  \toprule
  Model & Ranks \\
  \midrule
  Low-rank & $224, 288, 432$ \\
  Kronecker & $224, 272, 512$ \\
  TT & $96, 112, 128$ \\
  TTM & $80, 96, 112$ \\
  \bottomrule
\end{tabular}
\end{wraptable}
We replace the joint QKV and output projections in attention and the up- and down-projections in the FFN with our low-rank, Kronecker, TT, or TTM layer implementations. We keep the token and position embeddings dense and tie the language modeling head to the token embedding table in every model. The low-rank model serves as a factorized baseline alongside the dense model.
\Cref{tab:tensor_ranks} gives the rank grid. We choose multiples of 16 around target whole-model compression ratios of $1.25$, $1.5$, and $1.75$, obtaining ratios from $1.18$ to $1.82$ in the training runs.
For TT and TTM, we use the same rank for every core and layer within a run. \Cref{tab:tensor_shapes} gives the fixed mode sizes for each projection type.

\begin{table}[h]
\caption{Tensor shapes by projection type. The first and second tuples give the input and output mode sizes, respectively.}
\label{tab:tensor_shapes}
\footnotesize
\begin{tabular}{lcccc}
  \toprule
  Model & Q, K, V (joint) & Out & FFN Up & FFN Down \\
  \midrule
  Low-rank & $[(768,), (2304,)]$ & $[(768,), (768,)]$ & $[(768,), (3072,)]$ & $[(3072,), (768,)]$ \\
  Kronecker & $[(24, 32), (48, 48)]$ & $[(24, 32), (24, 32)]$ & $[(24, 32), (48, 64)]$ & $[(48, 64), (24, 32)]$ \\
  TT & $[(24, 32), (48, 48)]$ & $[(24, 32), (24, 32)]$ & $[(24, 32), (48, 64)]$ & $[(48, 64), (24, 32)]$ \\
  TTM & $[(8, 8, 12), (12, 12, 16)]$ & $[(8, 8, 12), (8, 8, 12)]$ & $[(8, 8, 12), (12, 16, 16)]$ & $[(12, 16, 16), (8, 8, 12)]$ \\
  \bottomrule
\end{tabular}
\end{table}

\paragraph{Datasets.}
We use the 10B-token subset of FineWeb-Edu \cite{penedo2024fineweb}. We reserve the first 10M tokens for validation and use the following 2.5B tokens for training.

\subsubsection{Experimental setup}
We run the pre-training experiments on a single NVIDIA A100 80\,GB, with 12 CPU cores and 64\,GB of RAM. We implement and train the models in PyTorch. The tensorized layers apply the factors through contractions, as described in \cref{sec:component-ffn}.
For the efficiency benchmark, we refer to uncompiled execution as \texttt{eager} and to \texttt{torch.compile} in \texttt{reduce-overhead} mode as \texttt{compiled}.

We train each configuration from scratch with three seeds and a common recipe. Each run takes 4768 optimizer steps, corresponding to approximately one pass over the training set. We use AdamW \cite{loshchilov2019adamw} with a peak learning rate of $6 \times 10^{-4}$, linear warmup for 235 steps, and cosine decay.
We use micro-batches of 32 sequences of length 1024, accumulate gradients over 16 micro-batches, and train with bfloat16 autocast and TF32 enabled. The training runs use \texttt{torch.compile} with its default mode.

\subsubsection{Results on efficiency}

\begin{figure}[t]
  \centering
  \begin{minipage}{0.49\linewidth}
    \includegraphics[width=\linewidth]{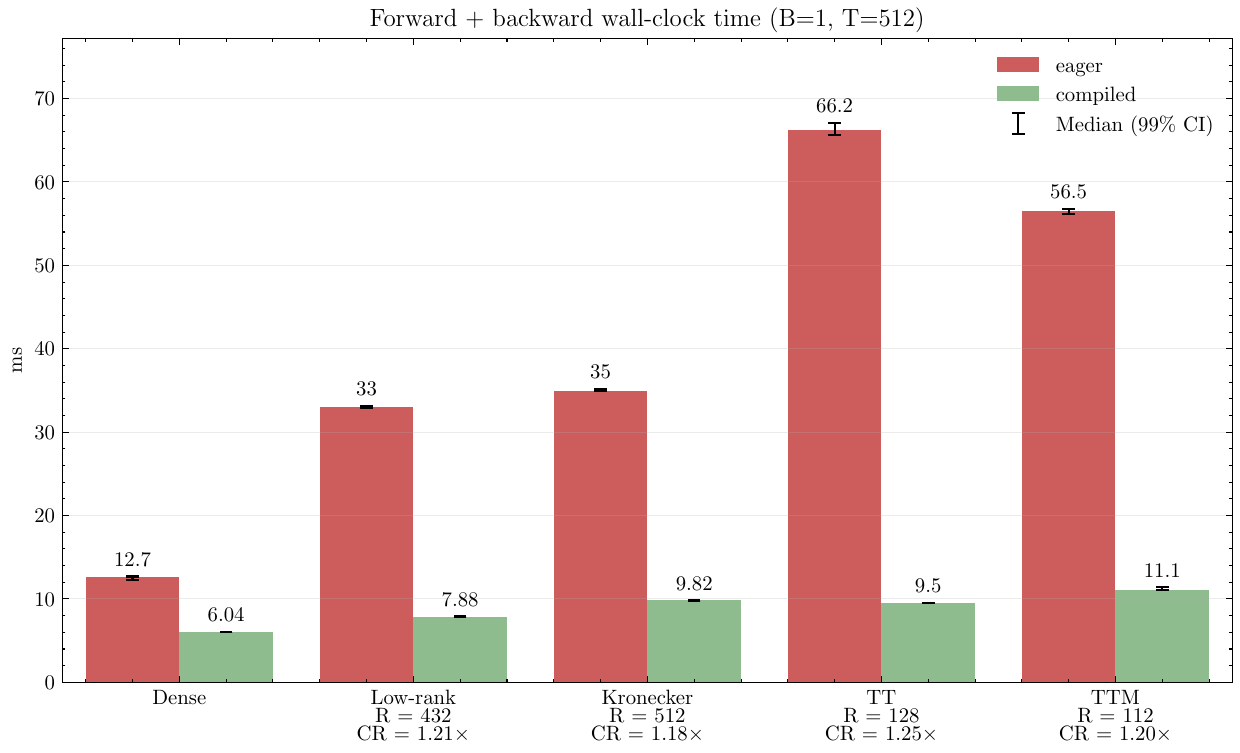}
  \end{minipage}\hfill
  \begin{minipage}{0.49\linewidth}
    \includegraphics[width=\linewidth]{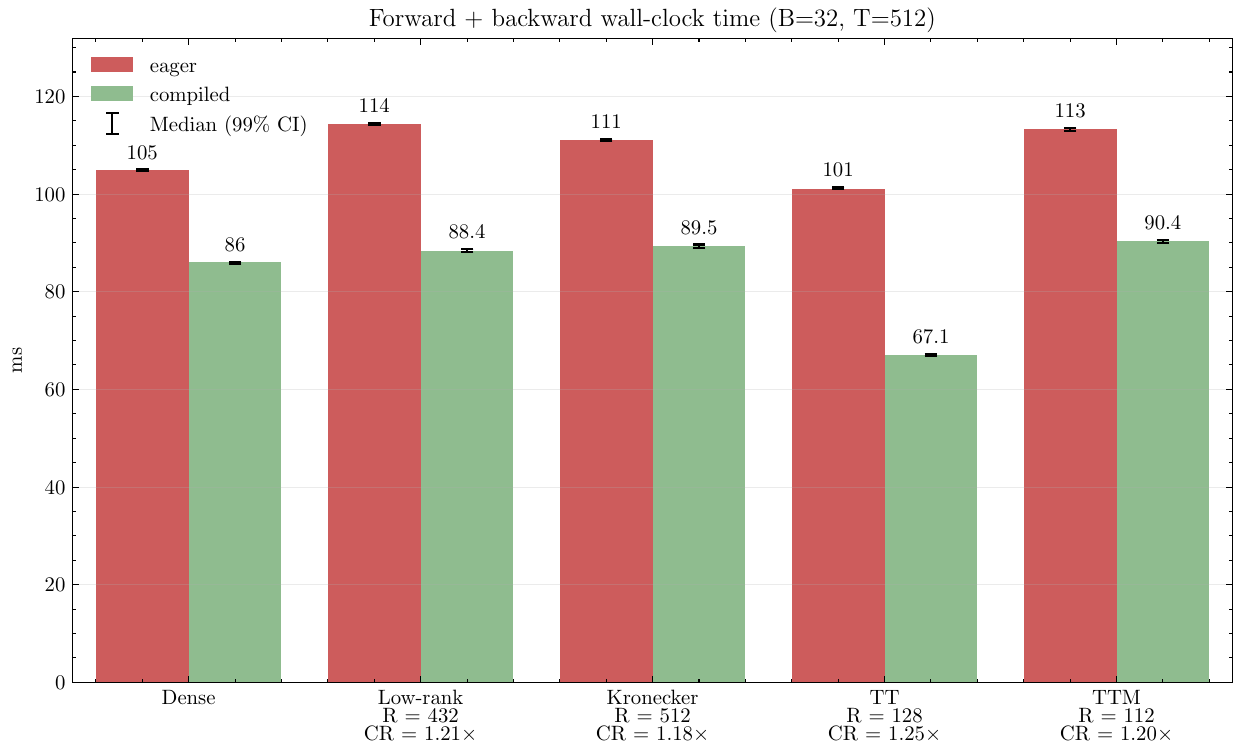}
  \end{minipage}
  \caption{Median time for one forward and backward pass in eager and compiled modes, with 99\% confidence intervals. Sequence length is 512 and batch size is 1 (left) or 32 (right). Each factorized model is shown at its largest tested rank.}
  \label{fig:exp_efficiency_speedup}
\end{figure}

In \cref{sec:component-ffn}, we argued that a sequence of tensor contractions can take longer than a dense matrix multiplication even when it requires fewer FLOPs. To test how much of this overhead we can reduce through compilation, we time a forward and backward pass on synthetic batches after warmup, using bfloat16 autocast in both modes. The measurement excludes the optimizer step.
\Cref{fig:exp_efficiency_speedup} shows the comparison at sequence length 512 for the largest tested rank of each factorized model.

At batch size 1, the factorized models take $2.5$--$5.2\times$ as long as the dense model in eager mode and $1.3$--$1.8\times$ as long in compiled mode, using the dense model in the same mode as the reference. The difference is much smaller at batch size 32. In compiled mode, TT takes 67.1\,ms versus 86.0\,ms for the dense model, and the other three configurations are within 6\% of the dense time.

\begin{figure}[t]
  \centering
  \begin{minipage}{0.49\linewidth}
    \includegraphics[width=\linewidth]{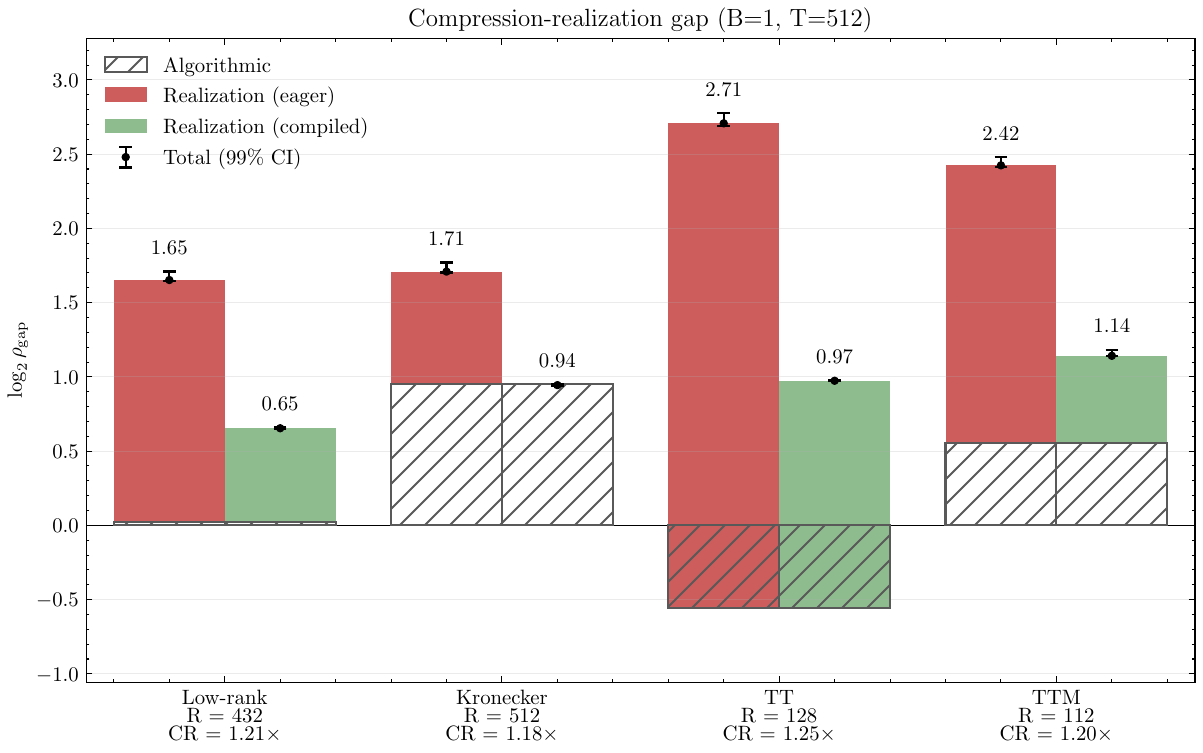}
  \end{minipage}\hfill
  \begin{minipage}{0.49\linewidth}
    \includegraphics[width=\linewidth]{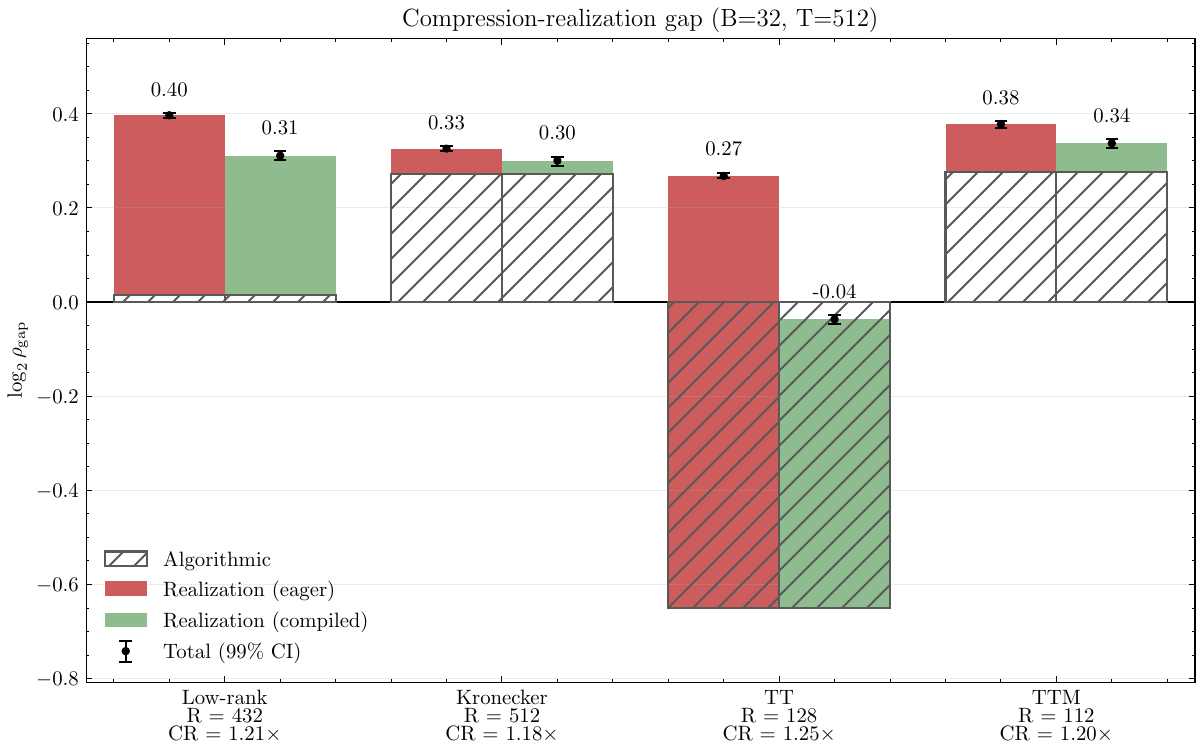}
  \end{minipage}
  \caption{Decomposition of $\log_2 \rho_{\rm gap}$ in eager and compiled modes for the configurations in \cref{fig:exp_efficiency_speedup}. Error bars show 99\% confidence intervals for the total gap. Sequence length is 512 and batch size is 1 (left) or 32 (right).}
  \label{fig:exp_efficiency_gap}
\end{figure}

These timings tell us how much compilation helps. To understand the remaining difference, we use the compression-realization gap $\rho_{\rm gap}$ (\cref{subsec:rho_gap}).
We use parameter-count ratios as a proxy for weight-byte compression at a common storage precision, compare each model with the dense baseline in the same execution mode, and use the FLOPs counted in eager mode for both modes. This gives a diagnostic at fixed configurations, with validation quality reported separately below. In \cref{fig:exp_efficiency_gap}, the algorithmic and realization terms add on the $\log_2$ scale.

The realization term decreases with compilation, especially at batch size 1. It remains the main positive contribution for low-rank and TT at both batch sizes. The algorithmic term is close to zero for low-rank and negative for TT, where FLOPs decrease by a larger factor than the parameter count. Kronecker and TTM save fewer FLOPs than parameters proportionally. After compilation, the algorithmic term dominates for Kronecker at both batch sizes and for TTM at batch size 32. For TTM at batch size 1, the two terms are comparable.

These results motivate two follow-up experiments from the hardware compatibility discussion in \cref{subsec:gaps}. The remaining realization gap for low-rank and TT gives a target for testing fused kernels, especially at small batch sizes. For Kronecker and TTM, the algorithmic term also motivates testing cheaper contraction schedules at a fixed parameter budget.

\subsubsection{Results on language modeling}

\begin{figure}[t]
  \centering
  \begin{minipage}[c]{0.9\linewidth}
    \centering
    \includegraphics[width=\linewidth]{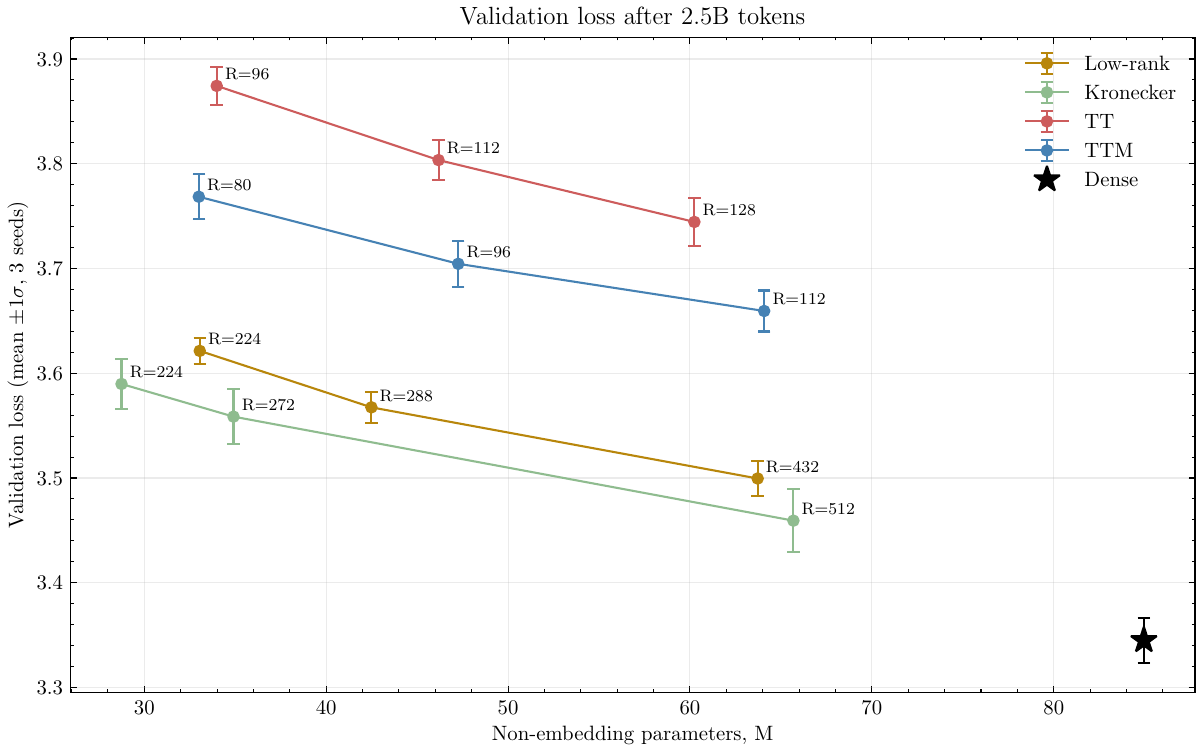}
    \caption{Validation loss after approximately 2.5B training tokens versus the number of non-embedding parameters. Points show the mean over three seeds and error bars show one standard deviation.}
    \label{fig:exp_train}
  \end{minipage}
\end{figure}

We next ask how each parameterization trades off language modeling quality against parameter count. \Cref{fig:exp_train} plots the final validation loss against the number of non-embedding parameters, since the embedding tables have the same size in all models.

Of the factorized models, Kronecker achieves the lowest mean validation loss across the three approximate budget levels, while both TT and TTM trail the low-rank baseline. For example, at roughly 33--35M non-embedding parameters, the mean losses are 3.559 for Kronecker, 3.621 for low-rank, 3.769 for TTM, and 3.874 for TT. The dense model reaches 3.345 with 85.0M non-embedding parameters.

We obtain this ordering with fixed tensor shapes, uniform ranks, and a shared training recipe. A common backbone and data split, as discussed in \cref{subsec:pre-training,subsec:gaps}, provide a starting point for comparing parameterizations. TT and TTM allow different values on each internal rank, so a direct follow-up would compare uniform and adaptive ranks at the same parameter budget. Separate ablations of tensor shapes and optimizers would then show how sensitive the ordering is to those choices.

\subsubsection{Rank analysis}

\begin{figure}[t]
  \centering
  \begin{minipage}{0.49\linewidth}
    \includegraphics[width=\linewidth]{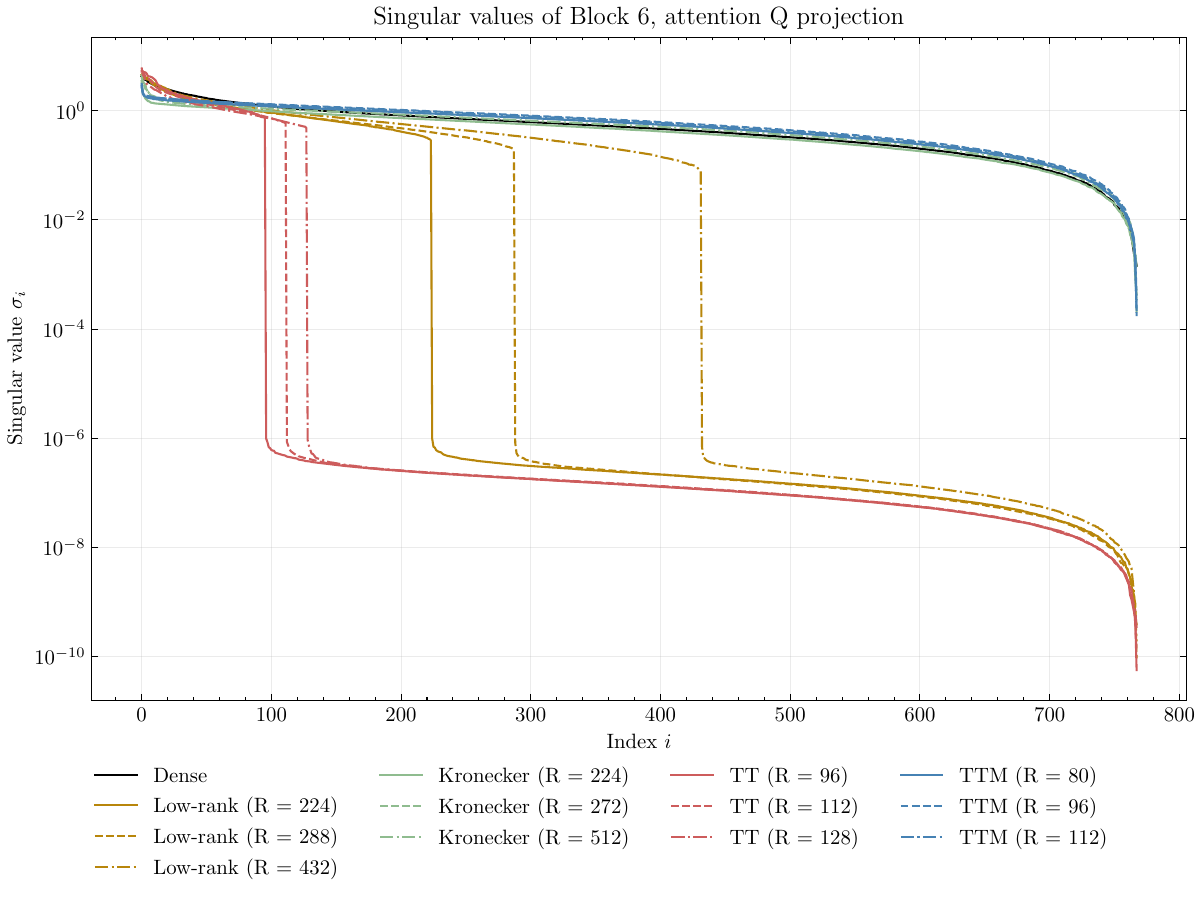}
  \end{minipage}\hfill
  \begin{minipage}{0.49\linewidth}
    \includegraphics[width=\linewidth]{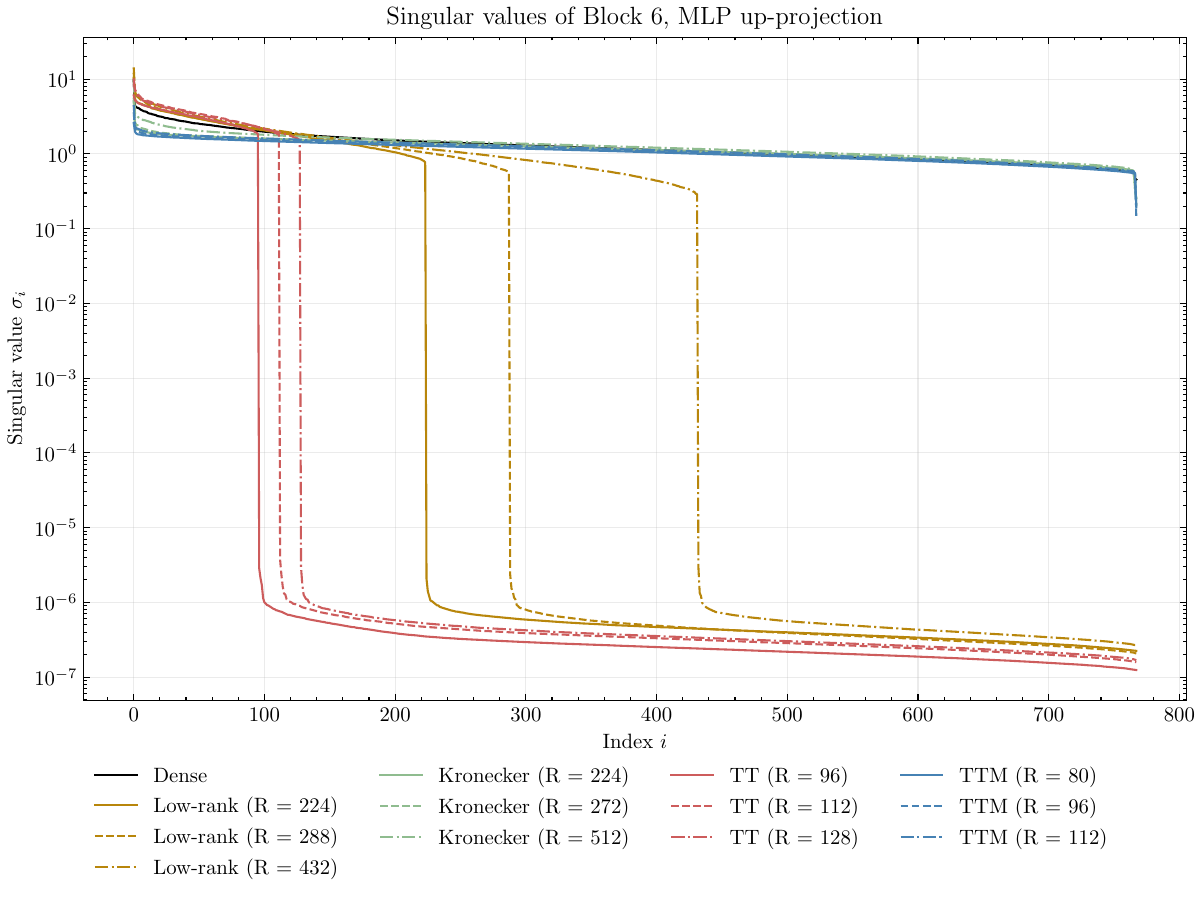}
  \end{minipage}
  \caption{Singular values of the reconstructed attention query projection (left) and FFN up-projection (right) in block 6.}
  \label{fig:exp_train_svd}
\end{figure}

To see what matrix rank explains about differences in validation loss, we examine the singular values of the reconstructed weight matrices. \Cref{fig:exp_train_svd} shows the spectra of the attention query projection and FFN up-projection in block 6.

The spectra of low-rank and TT show a sharp drop after their configured rank. For TT, this follows from the middle connection separating the input and output modes, as described in \cref{sec:component-ffn}. The Kronecker and TTM spectra extend across all 768 singular values without a similar cutoff, consistent with full-rank reconstructions in these examples. This larger matrix rank may help explain why Kronecker and TTM outperform TT at comparable budgets. The low-rank baseline still achieves lower validation loss than TTM, so matrix rank alone cannot explain the ordering.

Within each parameterization, validation loss decreases as the rank and parameter count increase together (\cref{fig:exp_train}). Which layers and cores benefit most from the additional rank remains open. Redistributing rank across layers and cores at a fixed parameter budget would test the adaptive rank allocation proposed in \cref{subsec:gaps} and provide a starting point for the conditional rank schemes in \cref{subsec:future-directions}.

\subsection{PEFT}

To complement the comparison across papers in \cref{subsec:peft}, we evaluate local tensorized adapters on commonsense reasoning using the same frozen Llama-3.2-1B backbone \cite{metaai2024llama321b, unsloth2024llama321b}. We ask whether a tensorized adapter can match LoRA's accuracy with fewer trainable parameters, and whether the matrix ranks of the learned updates help explain accuracy differences.

\paragraph{Baseline adapter.}
We use the \texttt{peft} implementation of LoRA \cite{hu2022lora} with rank 32 and scaling parameter $\alpha=32$ as our baseline.

\paragraph{Tensorized adapters.}
\begin{wraptable}{r}{0.4\linewidth}
    \centering
    \caption{\footnotesize Adapter ranks.}
    \label{tab:peft_tensor_ranks}
    \footnotesize
    \begin{tabular}{lc}
      \toprule
      Model & Ranks \\
      \midrule
    LoRA & $32$ \\
    Kronecker & $16$ \\
    TT & $32$ \\
    TTM & $32$ \\
      \bottomrule
    \end{tabular}
    \end{wraptable}
We implement Kronecker, TT, and TTM adapters through the \texttt{peft} interface. Each adapter learns an additive update $\Delta\mat{W}$ to each frozen projection $\mat{W}$. All adapters target the query, key, and value projections in self-attention and the up- and down-projections in the FFN.

\Cref{tab:peft_tensor_ranks} lists the ranks and \cref{tab:peft_tensor_shapes} gives the mode sizes for each projection type. We compare rank-32 LoRA, rank-16 Kronecker, and the uniform-rank TT and TTM configurations. TT and TTM both use four cores with internal ranks $(32,32,32)$ across all target projections.

\begin{table}[h]
    \centering
    \caption{Tensor shapes by adapter and projection type. The first and second tuples give the input and output mode sizes, respectively.}
    \label{tab:peft_tensor_shapes}
    \footnotesize
    \setlength{\tabcolsep}{3pt}
    \begin{tabular*}{\linewidth}{@{\extracolsep{\fill}}lcccc@{}}
      \toprule
      Adapter & Q & K and V & FFN up & FFN down \\
      \midrule
    LoRA & $[(2048,), (2048,)]$ & $[(2048,), (512,)]$ & $[(2048,), (8192,)]$ & $[(8192,), (2048,)]$ \\
    \addlinespace[2pt]
    Kronecker & $[(128, 16), (32, 64)]$ & $[(128, 16), (8, 64)]$ & $[(128, 16), (32, 256)]$ & $[(32, 256), (128, 16)]$ \\
    \addlinespace[2pt]
    TT & $[(64, 32), (32, 64)]$ & $[(64, 32), (16, 32)]$ & $[(64, 32), (64, 128)]$ & $[(128, 64), (32, 64)]$ \\
    \addlinespace[2pt]
    TTM & $[(8, 8, 4, 8), (8, 8, 4, 8)]$ & $[(8, 8, 4, 8), (4, 4, 4, 8)]$ & $[(8, 8, 4, 8), (16, 8, 4, 16)]$ & $[(16, 8, 4, 16), (8, 8, 4, 8)]$ \\
      \bottomrule
    \end{tabular*}
    \end{table}

\paragraph{Datasets.}
We primarily follow the commonsense reasoning setup of HiRA \cite{huang2025hira}. We fine-tune on the 170,300 training examples in Commonsense170K \cite{hu2023llmadapters} and use its 120-example validation split for model selection.
We evaluate each adapter on the test sets of BoolQ \cite{clark2019boolq}, PIQA \cite{zellers2020piqa}, SIQA \cite{sap2019siqa}, ARC-c and ARC-e \cite{clark2018arc}, OBQA \cite{mihaylov2018obqa}, HellaSwag \cite{zellers2019hellaswag}, and WinoGrande \cite{sakaguchi2021winogrande}.

\subsubsection{Experimental setup}

\begin{wraptable}{r}{0.48\textwidth}
  \centering
  \includegraphics[width=\linewidth]{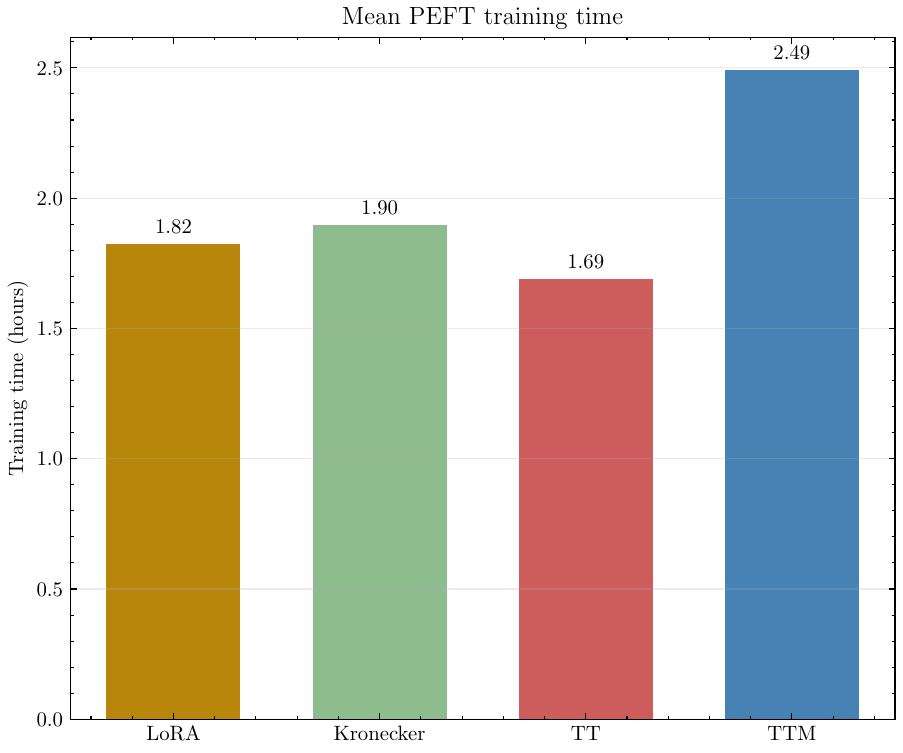}
  \captionof{figure}{Mean training time for three epochs of fine-tuning, in hours.}
  \label{fig:exp_peft_time}
\end{wraptable}

We use the same hardware as for pre-training and train with the \texttt{transformers} and \texttt{peft} libraries \cite{wolf2020transformers, mangrulkar2022peft}. We compile the training model with \texttt{torch.compile}.

We fine-tune for three epochs with AdamW \cite{loshchilov2019adamw}, linear warmup, and cosine decay. We tune the learning rate on the Commonsense170K validation split using LoRA and apply the same rate, $2 \times 10^{-4}$, to all adapters.
Training uses micro-batches of eight and 16 gradient accumulation steps, with bfloat16 mixed precision and TF32 enabled. We mask prompt and padding tokens in the loss. We limit prompts to 256 tokens, or 512 for HellaSwag at test time, and cap responses at 16 tokens.

We evaluate validation loss every 80 optimizer steps and select the checkpoint with the lowest loss. At test time, we generate responses with beam size four and take the first valid answer keyword in the generated response as the prediction. Responses without a valid keyword count as incorrect. We report accuracy on each test set, and the unweighted mean over the eight sets.

\subsubsection{Results on commonsense reasoning}

\Cref{fig:exp_peft_time} shows that TT has the shortest mean training time (1.69\,h) and TTM the longest (2.49\,h).
\Cref{tab:exp_peft} reports test accuracy. Kronecker achieves the highest mean accuracy among the tensorized adapters, reaching $73.15\%$, compared with LoRA's $74.41\%$, with $6.55$M trainable parameters versus $15.20$M for LoRA.

\begin{table}[h]
    \centering
    \caption{Commonsense reasoning accuracy (\%). Parameters are the trainable adapter parameters (in millions). We report the mean $\pm$ sample standard deviation across three seeds for each method. Each entry shows the mean on the first line and the standard deviation on the second.}
    \label{tab:exp_peft}
    \vspace{3pt}
    \footnotesize
    \setlength{\tabcolsep}{2.5pt}
  \begin{tabular*}{\linewidth}{@{\extracolsep{\fill}}lcccccccccc@{}}
    \toprule
    Method & Parameters & BoolQ & PIQA & SIQA & ARC-c & ARC-e & OBQA & HellaS & WinoG & Avg. \\
    \midrule
    LoRA & $15.20$M & \shortstack{$62.95$\\$\pm 2.44$} & \shortstack{$79.33$\\$\pm 0.41$} & \shortstack{$75.42$\\$\pm 0.52$} & \shortstack{$61.77$\\$\pm 0.39$} & \shortstack{$78.97$\\$\pm 0.42$} & \shortstack{$74.00$\\$\pm 0.87$} & \shortstack{$88.65$\\$\pm 0.17$} & \shortstack{$74.19$\\$\pm 0.63$} & \shortstack{$74.41$\\$\pm 0.33$} \\
    \addlinespace[2pt]
    Kronecker & $6.55$M & \shortstack{$65.03$\\$\pm 0.99$} & \shortstack{$78.31$\\$\pm 0.30$} & \shortstack{$73.66$\\$\pm 0.16$} & \shortstack{$59.98$\\$\pm 0.17$} & \shortstack{$76.85$\\$\pm 0.54$} & \shortstack{$72.13$\\$\pm 1.03$} & \shortstack{$86.17$\\$\pm 0.79$} & \shortstack{$73.06$\\$\pm 0.43$} & \shortstack{$73.15$\\$\pm 0.31$} \\
    \addlinespace[2pt]
    TT & $6.13$M & \shortstack{$63.13$\\$\pm 0.73$} & \shortstack{$77.49$\\$\pm 0.27$} & \shortstack{$73.88$\\$\pm 0.64$} & \shortstack{$57.99$\\$\pm 0.18$} & \shortstack{$74.73$\\$\pm 0.82$} & \shortstack{$66.80$\\$\pm 0.92$} & \shortstack{$85.84$\\$\pm 0.65$} & \shortstack{$70.38$\\$\pm 0.39$} & \shortstack{$71.28$\\$\pm 0.44$} \\
    \addlinespace[2pt]
    TTM & $5.93$M & \shortstack{$62.93$\\$\pm 1.99$} & \shortstack{$76.53$\\$\pm 0.21$} & \shortstack{$72.24$\\$\pm 0.58$} & \shortstack{$56.14$\\$\pm 1.33$} & \shortstack{$74.96$\\$\pm 0.39$} & \shortstack{$65.07$\\$\pm 1.21$} & \shortstack{$85.10$\\$\pm 0.11$} & \shortstack{$68.14$\\$\pm 0.37$} & \shortstack{$70.14$\\$\pm 0.04$} \\
    \bottomrule
  \end{tabular*}
  \end{table}

\subsubsection{Rank analysis}

\begin{figure}[!htbp]
  \centering
  \includegraphics[width=\linewidth]{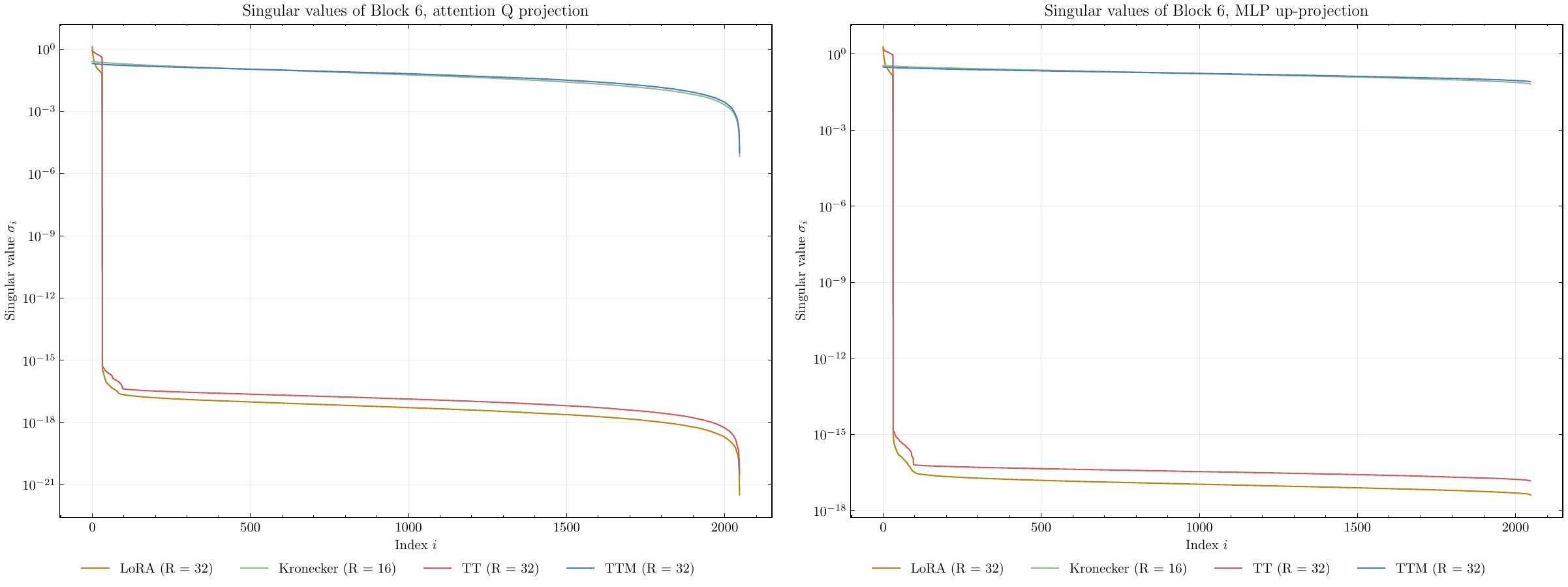}
  \caption{Singular values of learned adapter updates in the attention query and FFN up-projections of block 6. The vertical axis is logarithmic.}
  \label{fig:exp_peft_svd}
\end{figure}

\Cref{fig:exp_peft_svd} shows the singular values of the learned updates $\Delta\mat{W}$ for the attention query and FFN up-projections in block 6. The LoRA and TT spectra drop sharply after the first 32 singular values and continue with a tail at the level of floating-point roundoff. The cutoff follows from the LoRA factorization rank and the central TT internal rank. For Kronecker and TTM, all 2048 singular values in each projection exceed the numerical rank tolerance. Both updates are therefore numerically full rank in these projections, while their recorded mean accuracies differ (\cref{tab:exp_peft}).

\section{Conclusion}
\label{sec:conclusion}

We introduce a taxonomy of tensor methods for LLMs that combines two complementary views of the model lifecycle and Transformer components.
We use this taxonomy to review and compare existing methods by what objectives they serve, what they tensorize, and which decompositions they use.
We also discuss connections to neighboring efficiency methods and probabilistic tensor networks.
We identify open challenges and introduce the compression-realization gap $\rho_{\rm gap}$ to quantify the discrepancy between theoretical memory savings and measured speedup.
Finally, we compare low-rank, Kronecker, tensor train, and tensor train matrix parameterizations in pre-training and PEFT under matched protocols.
We apply $\rho_{\rm gap}$ to the pre-training benchmarks to examine how the relative contributions of contraction cost and hardware utilization vary across decompositions.
We compare final validation loss in pre-training and commonsense reasoning accuracy in PEFT against the respective common non-tensor baselines. Additionally, we examine the singular value spectra of selected weights and adapter updates to assess their numerical rank.


\printbibliography

@inproceedings{zheng2021fctn,
  title={Fully-Connected Tensor Network Decomposition and Its Application to Higher-Order Tensor Completion},
  author={Zheng, Yu-Bang and Huang, Ting-Zhu and Zhao, Xi-Le and Zhao, Qibin and Jiang, Tai-Xiang},
  booktitle={Proceedings of the AAAI Conference on Artificial Intelligence},
  volume={35},
  number={12},
  pages={11071--11078},
  year={2021},
  doi={10.1609/aaai.v35i12.17321}
}

@article{goto2008anatomy,
  title={Anatomy of High-Performance Matrix Multiplication},
  author={Goto, Kazushige and van de Geijn, Robert A.},
  journal={ACM Transactions on Mathematical Software},
  volume={34},
  number={3},
  pages={1--25},
  year={2008},
  publisher={ACM},
  doi={10.1145/1356052.1356053}
}

@article{kolda2009tensor,
  title={Tensor Decompositions and Applications},
  author={Kolda, Tamara G. and Bader, Brett W.},
  journal={SIAM Review},
  volume={51},
  number={3},
  pages={455--500},
  year={2009},
  doi={10.1137/07070111X}
}

@article{cichocki2015tensor,
   title={Tensor Networks for Dimensionality Reduction and Large-scale Optimization: Part 1 Low-Rank Tensor Decompositions},
   volume={9},
   ISSN={1935-8245},
   url={http://dx.doi.org/10.1561/2200000059},
   DOI={10.1561/2200000059},
   number={4-5},
   journal={Foundations and Trends® in Machine Learning},
   publisher={Emerald},
   author={Cichocki, Andrzej and Lee, Namgil and Oseledets, Ivan and Phan, Anh-Huy and Zhao, Qibin and Mandic, Danilo P.},
   year={2016},
   pages={249–429}
}

@article{cichocki2015tensor2,
   title={Tensor Networks for Dimensionality Reduction and Large-Scale Optimizations Part 2 Applications and Future Perspectives},
   volume={9},
   ISSN={1935-8245},
   url={http://dx.doi.org/10.1561/2200000067},
   DOI={10.1561/2200000067},
   number={6},
   journal={Foundations and Trends® in Machine Learning},
   publisher={Emerald},
   author={Cichocki, Andrzej and Phan, Anh-Huy and Zhao, Qibin and Lee, Namgil and Oseledets, Ivan and Sugiyama, Masashi and Mandic, Danilo},
   year={2017},
   month=May, pages={431–673}
}

@article{baggag2025linalg,
  author  = {Baggag, Abdelkader and Saad, Yousef},
  title   = {Deep learning, transformers and graph neural networks: a linear algebra perspective},
  journal = {Numerical Algorithms},
  volume  = {100},
  number  = {4},
  pages   = {2095--2134},
  year    = {2025},
  doi     = {10.1007/s11075-025-02218-2},
  url     = {https://doi.org/10.1007/s11075-025-02218-2}  
}

@article{oseledets2011tensor,
  title={Tensor-Train Decomposition},
  author={Oseledets, Ivan V.},
  journal={SIAM Journal on Scientific Computing},
  volume={33},
  number={5},
  pages={2295--2317},
  year={2011},
  doi={10.1137/090752286}
}

@article{tucker1966some,
  title={Some Mathematical Notes on Three-Mode Factor Analysis},
  author={Tucker, Ledyard R.},
  journal={Psychometrika},
  volume={31},
  number={3},
  pages={279--311},
  year={1966},
  url={https://doi.org/10.1007/BF02289464},
  doi={10.1007/BF02289464}
}

@article{carroll1970analysis,
  title={Analysis of Individual Differences in Multidimensional Scaling via an N-way Generalization of ``Eckart-Young'' Decomposition},
  author={Carroll, J. Douglas and Chang, Jih-Jie},
  journal={Psychometrika},
  volume={35},
  pages={283--319},
  year={1970},
  doi={10.1007/BF02310791}
}

@article{harshman1970foundations,
  title={Foundations of the PARAFAC Procedure: Models and Conditions for an Explanatory Multimodal Factor Analysis},
  author={Harshman, Richard A.},
  journal={UCLA Working Papers in Phonetics},
  volume={16},
  pages={1--84},
  year={1970}
}

@inproceedings{novikov2015tensorizing,
  title={Tensorizing Neural Networks},
  author={Novikov, Alexander and Podoprikhin, Dmitrii and Osokin, Anton and Vetrov, Dmitry},
  booktitle={Advances in Neural Information Processing Systems},
  volume={28},
  year={2015}
}

@article{wang2023tensor,
  title={Tensor Networks Meet Neural Networks: A Survey and Future Perspectives},
  author={Wang, Maolin and Pan, Yu and Xu, Zenglin and Yang, Xiangli and Li, Guangxi and Cichocki, Andrzej},
  journal={arXiv preprint arXiv:2302.09019},
  year={2023}
}

@article{hamreras2025tensorization,
  title={Tensorization is a Powerful but Underexplored Tool for Neural Network Compression},
  author={Hamreras, S. and others},
  journal={arXiv preprint arXiv:2505.20132},
  year={2025}
}

@inproceedings{vaswani2017attention,
 author = {Vaswani, Ashish and Shazeer, Noam and Parmar, Niki and Uszkoreit, Jakob and Jones, Llion and Gomez, Aidan N and Kaiser, \L ukasz and Polosukhin, Illia},
 booktitle = {Advances in Neural Information Processing Systems},
 editor = {I. Guyon and U. Von Luxburg and S. Bengio and H. Wallach and R. Fergus and S. Vishwanathan and R. Garnett},
 pages = {},
 publisher = {Curran Associates, Inc.},
 title = {Attention is All you Need},
 url = {https://proceedings.neurips.cc/paper_files/paper/2017/file/3f5ee243547dee91fbd053c1c4a845aa-Paper.pdf},
 volume = {30},
 year = {2017}
}

@inproceedings{devlin2019bert,
  title={{BERT}: Pre-training of Deep Bidirectional Transformers for Language Understanding},
  author={Devlin, Jacob and Chang, Ming-Wei and Lee, Kenton and Toutanova, Kristina},
  booktitle={Proceedings of NAACL-HLT},
  pages={4171--4186},
  year={2019},
  doi={10.18653/v1/N19-1423}
}

@inproceedings{radford2019language,
  title={Language Models are Unsupervised Multitask Learners},
  author={Radford, Alec and Wu, Jeffrey and Child, Rewon and Luan, David and Amodei, Dario and Sutskever, Ilya},
  booktitle={OpenAI Technical Report},
  year={2019}
}

@inproceedings{brown2020gpt3,
 author = {Brown, Tom and Mann, Benjamin and Ryder, Nick and Subbiah, Melanie and Kaplan, Jared D and Dhariwal, Prafulla and Neelakantan, Arvind and Shyam, Pranav and Sastry, Girish and Askell, Amanda and Agarwal, Sandhini and Herbert-Voss, Ariel and Krueger, Gretchen and Henighan, Tom and Child, Rewon and Ramesh, Aditya and Ziegler, Daniel and Wu, Jeffrey and Winter, Clemens and Hesse, Chris and Chen, Mark and Sigler, Eric and Litwin, Mateusz and Gray, Scott and Chess, Benjamin and Clark, Jack and Berner, Christopher and McCandlish, Sam and Radford, Alec and Sutskever, Ilya and Amodei, Dario},
 booktitle = {Advances in Neural Information Processing Systems},
 editor = {H. Larochelle and M. Ranzato and R. Hadsell and M.F. Balcan and H. Lin},
 pages = {1877--1901},
 publisher = {Curran Associates, Inc.},
 title = {Language Models are Few-Shot Learners},
 url = {https://proceedings.neurips.cc/paper_files/paper/2020/file/1457c0d6bfcb4967418bfb8ac142f64a-Paper.pdf},
 volume = {33},
 year = {2020}
}

@misc{touvron2023llama,
      title={LLaMA: Open and Efficient Foundation Language Models}, 
      author={Hugo Touvron and Thibaut Lavril and Gautier Izacard and Xavier Martinet and Marie-Anne Lachaux and Timothée Lacroix and Baptiste Rozière and Naman Goyal and Eric Hambro and Faisal Azhar and Aurelien Rodriguez and Armand Joulin and Edouard Grave and Guillaume Lample},
      year={2023},
      eprint={2302.13971},
      archivePrefix={arXiv},
      primaryClass={cs.CL},
      url={https://arxiv.org/abs/2302.13971}, 
}

@inproceedings{hrinchuk2020tensorized,
    title = "Tensorized Embedding Layers",
    author = "Hrinchuk, Oleksii  and
      Khrulkov, Valentin  and
      Mirvakhabova, Leyla  and
      Orlova, Elena  and
      Oseledets, Ivan",
    editor = "Cohn, Trevor  and
      He, Yulan  and
      Liu, Yang",
    booktitle = "Findings of the Association for Computational Linguistics: EMNLP 2020",
    month = nov,
    year = "2020",
    address = "Online",
    publisher = "Association for Computational Linguistics",
    url = "https://aclanthology.org/2020.findings-emnlp.436/",
    doi = "10.18653/v1/2020.findings-emnlp.436",
    pages = "4847--4860"
}

@misc{ma2019tensorized,
      title={A Tensorized Transformer for Language Modeling}, 
      author={Xindian Ma and Peng Zhang and Shuai Zhang and Nan Duan and Yuexian Hou and Dawei Song and Ming Zhou},
      year={2019},
      eprint={1906.09777},
      archivePrefix={arXiv},
      primaryClass={cs.CL},
      url={https://arxiv.org/abs/1906.09777}, 
}

@misc{xu2023tensorgpt,
      title={TensorGPT: Efficient Compression of Large Language Models based on Tensor-Train Decomposition}, 
      author={Mingxue Xu and Yao Lei Xu and Danilo P. Mandic},
      year={2024},
      eprint={2307.00526},
      archivePrefix={arXiv},
      primaryClass={cs.CL},
      url={https://arxiv.org/abs/2307.00526}, 
}

@inproceedings{yang2024loretta,
    title = "{L}o{RETTA}: Low-Rank Economic Tensor-Train Adaptation for Ultra-Low-Parameter Fine-Tuning of Large Language Models",
    author = "Yang, Yifan  and
      Zhou, Jiajun  and
      Wong, Ngai  and
      Zhang, Zheng",
    booktitle = "Proceedings of the 2024 Conference of the North American Chapter of the Association for Computational Linguistics: Human Language Technologies (Volume 1: Long Papers)",
    month = jun,
    year = "2024",
    address = "Mexico City, Mexico",
    publisher = "Association for Computational Linguistics",
    url = "https://aclanthology.org/2024.naacl-long.174/",
    doi = "10.18653/v1/2024.naacl-long.174",
    pages = "3161--3176"
}

@inproceedings{anjum2024ttlora,
  author={Anjum, Afia and Eren, Maksim E. and Boureima, Ismael and Alexandrov, Boian and Bhattarai, Manish},
  booktitle={2024 International Conference on Machine Learning and Applications (ICMLA)}, 
  title={Tensor Train Low-rank Approximation (TT-LoRA): Democratizing AI with Accelerated LLMs}, 
  year={2024},
  volume={},
  number={},
  pages={583-590},
  doi={10.1109/ICMLA61862.2024.00085}
  }

@misc{hounie2024lorta,
      title={LoRTA: Low Rank Tensor Adaptation of Large Language Models}, 
      author={Ignacio Hounie and Charilaos Kanatsoulis and Arnuv Tandon and Alejandro Ribeiro},
      year={2025},
      eprint={2410.04060},
      archivePrefix={arXiv},
      primaryClass={cs.CL},
      url={https://arxiv.org/abs/2410.04060}, 
}

@inproceedings{hu2024dota,
      author="Hu, Xiaolin
      and Cheng, Xiang
      and Liu, Peiyu
      and Liu, Wei
      and Luan, Jian
      and Wang, Bin
      and Liu, Yong",
      title={{DoTA}: Weight-Decomposed Tensor Adaptation for Large Language Models},
      booktitle="Advances in Knowledge Discovery and Data Mining ",
      year="2025",
      publisher="Springer Nature Singapore",
      address="Singapore",
      pages="16--27",
      isbn="978-981-96-8186-0"
}

@inproceedings{hu2022lora,
  title={{LoRA}: Low-Rank Adaptation of Large Language Models},
  author={Hu, Edward J. and Shen, Yelong and Wallis, Phillip and Allen-Zhu, Zeyuan and Li, Yuanzhi and Wang, Shean and Wang, Lu and Chen, Weizhu},
  booktitle={International Conference on Learning Representations},
  year={2022}
}

@inproceedings{dettmers2023qlora,
  title={{QLoRA}: Efficient Finetuning of Quantized {LLMs}},
  author={Dettmers, Tim and Pagnoni, Artidoro and Holtzman, Ari and Zettlemoyer, Luke},
  booktitle={Advances in Neural Information Processing Systems},
  year={2023}
}

@inproceedings{sanh2019distilbert,
  title={DistilBERT, a Distilled Version of {BERT}: Smaller, Faster, Cheaper and Lighter},
  author={Sanh, Victor and Debut, Lysandre and Chaumond, Julien and Wolf, Thomas},
  booktitle={NeurIPS EMC2 Workshop},
  year={2019}
}

@misc{frantar2023gptq,
      title={GPTQ: Accurate Post-Training Quantization for Generative Pre-trained Transformers}, 
      author={Elias Frantar and Saleh Ashkboos and Torsten Hoefler and Dan Alistarh},
      year={2023},
      eprint={2210.17323},
      archivePrefix={arXiv},
      primaryClass={cs.LG},
      url={https://arxiv.org/abs/2210.17323}, 
}

@inproceedings{dao2022flashattention,
  title={{FlashAttention}: Fast and Memory-Efficient Exact Attention with {IO}-Awareness},
  author={Dao, Tri and Fu, Daniel Y. and Ermon, Stefano and Rudra, Atri and R\'{e}, Christopher},
  booktitle={Advances in Neural Information Processing Systems},
  volume={35},
  pages={16344--16359},
  year={2022}
}

@inproceedings{wang2020linformer,
  title={{Linformer}: Self-Attention with Linear Complexity},
  author={Wang, Sinong and Li, Belinda Z. and Khabsa, Madian and Fang, Han and Ma, Hao},
  booktitle={arXiv preprint arXiv:2006.04768},
  year={2020}
}

@article{shazeer2019mqa,
      title={Fast Transformer Decoding: One Write-Head is All You Need}, 
      author={Noam Shazeer},
      year={2019},
      eprint={1911.02150},
      archivePrefix={arXiv},
      primaryClass={cs.NE},
      url={https://arxiv.org/abs/1911.02150}, 
}

@article{ainslie2023gqa,
  title={{GQA}: Training Generalized Multi-Query Transformer Models from Multi-Head Checkpoints},
  author={Ainslie, Joshua and Lee-Thorp, James and de Jong, Michiel and Zemlyanskiy, Yury and Lebr\'{o}n, Federico and Sanghai, Sumit},
  journal={arXiv preprint arXiv:2305.13245},
  year={2023}
}

@inproceedings{fedus2022switch,
  title={Switch Transformers: Scaling to Trillion Parameter Models with Simple and Efficient Sparsity},
  author={Fedus, William and Zoph, Barret and Shazeer, Noam},
  booktitle={Journal of Machine Learning Research},
  volume={23},
  number={120},
  pages={1--39},
  year={2022}
}

@article{geva2021ffnmemory,
      title={Transformer Feed-Forward Layers Are Key-Value Memories}, 
      author={Mor Geva and Roei Schuster and Jonathan Berant and Omer Levy},
      year={2021},
      eprint={2012.14913},
      archivePrefix={arXiv},
      primaryClass={cs.CL},
      url={https://arxiv.org/abs/2012.14913}, 
}

@article{elhage2021mathematical,
  title={A Mathematical Framework for Transformer Circuits},
  author={Elhage, Nelson and Nanda, Neel and Olsson, Catherine and Henighan, Tom and Joseph, Nicholas and Mann, Ben and Askell, Amanda and Bai, Yuntao and Chen, Anna and Conerly, Tom and others},
  journal={Transformer Circuits Thread},
  year={2021}
}

@article{orus2014practical,
      title = {A practical introduction to tensor networks: Matrix product states and projected entangled pair states},
      journal = {Annals of Physics},
      volume = {349},
      pages = {117-158},
      year = {2014},
      issn = {0003-4916},
      doi = {https://doi.org/10.1016/j.aop.2014.06.013},
      url = {https://www.sciencedirect.com/science/article/pii/S0003491614001596},
      author = {Román Orús}
}

@inproceedings{stoudenmire2016supervised,
  title={Supervised Learning with Tensor Networks},
  author={Stoudenmire, Edwin M. and Schwab, David J.},
  booktitle={Advances in Neural Information Processing Systems},
  year={2016}
}

@misc{frantar2023sparsegpt,
      title={SparseGPT: Massive Language Models Can Be Accurately Pruned in One-Shot}, 
      author={Elias Frantar and Dan Alistarh},
      year={2023},
      eprint={2301.00774},
      archivePrefix={arXiv},
      primaryClass={cs.LG},
      url={https://arxiv.org/abs/2301.00774}, 
}

@inproceedings{wolf2017weighttying1,
    title = "Using the Output Embedding to Improve Language Models",
    author = "Press, Ofir  and
      Wolf, Lior",
    editor = "Lapata, Mirella  and
      Blunsom, Phil  and
      Koller, Alexander",
    booktitle = "Proceedings of the 15th Conference of the {E}uropean Chapter of the Association for Computational Linguistics: Volume 2, Short Papers",
    month = apr,
    year = "2017",
    address = "Valencia, Spain",
    publisher = "Association for Computational Linguistics",
    url = "https://aclanthology.org/E17-2025/",
    pages = "157--163",
}

@article{inan2017weighttying2,
      title={Tying Word Vectors and Word Classifiers: A Loss Framework for Language Modeling}, 
      author={Hakan Inan and Khashayar Khosravi and Richard Socher},
      year={2017},
      eprint={1611.01462},
      archivePrefix={arXiv},
      primaryClass={cs.LG},
      url={https://arxiv.org/abs/1611.01462}, 
}

@inproceedings{chekalina2023ttm,
    title = "Efficient {GPT} Model Pre-training using Tensor Train Matrix Representation",
    author = "Chekalina, Viktoriia  and
      Novikov, Georgiy  and
      Gusak, Julia  and
      Panchenko, Alexander  and
      Oseledets, Ivan",
    booktitle = "Proceedings of the 37th Pacific Asia Conference on Language, Information and Computation",
    month = dec,
    year = "2023",
    address = "Hong Kong, China",
    publisher = "Association for Computational Linguistics",
    url = "https://aclanthology.org/2023.paclic-1.60/",
    pages = "600--608"
}

@misc{grattafiori2024llama3,
      title={The Llama 3 Herd of Models}, 
      author={{Meta AI}}, 
      year={2024},
      eprint={2407.21783},
      archivePrefix={arXiv},
      primaryClass={cs.AI},
      url={https://arxiv.org/abs/2407.21783}, 
}

@misc{kwon2023paggedattn,
      title={Efficient Memory Management for Large Language Model Serving with PagedAttention}, 
      author={Woosuk Kwon and Zhuohan Li and Siyuan Zhuang and Ying Sheng and Lianmin Zheng and Cody Hao Yu and Joseph E. Gonzalez and Hao Zhang and Ion Stoica},
      year={2023},
      eprint={2309.06180},
      archivePrefix={arXiv},
      primaryClass={cs.LG},
      url={https://arxiv.org/abs/2309.06180}, 
}

@misc{elhage2022superposition,
      title={Toy Models of Superposition}, 
      author={Nelson Elhage and Tristan Hume and Catherine Olsson and Nicholas Schiefer and Tom Henighan and Shauna Kravec and Zac Hatfield-Dodds and Robert Lasenby and Dawn Drain and Carol Chen and Roger Grosse and Sam McCandlish and Jared Kaplan and Dario Amodei and Martin Wattenberg and Christopher Olah},
      year={2022},
      eprint={2209.10652},
      archivePrefix={arXiv},
      primaryClass={cs.LG},
      url={https://arxiv.org/abs/2209.10652}, 
}

@article{edalati2022krona,
      title={KronA: Parameter Efficient Tuning with Kronecker Adapter}, 
      author={Ali Edalati and Marzieh Tahaei and Ivan Kobyzev and Vahid Partovi Nia and James J. Clark and Mehdi Rezagholizadeh},
      year={2022},
      eprint={2212.10650},
      archivePrefix={arXiv},
      primaryClass={cs.CL},
      url={https://arxiv.org/abs/2212.10650}, 
}

@book{golubvanloan,
  author = {Golub, Gene H. and Van Loan, Charles F.},
  edition = {Fourth},
  isbn = {1421407949 9781421407944},
  publisher = {JHU Press},
  refid = {824733531},
  title = {Matrix Computations},
  url = {http://www.cs.cornell.edu/cv/GVL4/golubandvanloan.htm},
  year = 2013
}

@inproceedings{li2026lestd,
title={Le{STD}: {LLM} Compression via Learning-based Sparse Tensor Decomposition},
author={Yi Li and Zhichun Guo and Miao Yin and Bingzhe Li},
booktitle={The Fourteenth International Conference on Learning Representations},
year={2026},
url={https://openreview.net/forum?id=0oHaazjMUX}
}

@inproceedings{zhang2026tpa,
      title={Tensor Product Attention Is All You Need},
      author={Yifan Zhang and Yifeng Liu and Huizhuo Yuan and Zhen Qin and Yang Yuan and Quanquan Gu and Andrew C Yao},
      booktitle={The Thirty-ninth Annual Conference on Neural Information Processing Systems},
      year={2025},
      url={https://openreview.net/forum?id=ECTxVRFhUa}
}

@article{compactifai,
      title={CompactifAI: Extreme Compression of Large Language Models using Quantum-Inspired Tensor Networks}, 
      author={Andrei Tomut and others},
      year={2024},
      eprint={2401.14109},
      archivePrefix={arXiv},
      primaryClass={cs.CL},
      url={https://arxiv.org/abs/2401.14109}, 
}

@inproceedings{li2022hypoformer,
    title = "Hypoformer: Hybrid Decomposition Transformer for Edge-friendly Neural Machine Translation",
    author = "Li, Sunzhu  and
      Zhang, Peng  and
      Gan, Guobing  and
      Lv, Xiuqing  and
      Wang, Benyou  and
      Wei, Junqiu  and
      Jiang, Xin",
    booktitle = "Proceedings of the 2022 Conference on Empirical Methods in Natural Language Processing",
    month = dec,
    year = "2022",
    address = "Abu Dhabi, United Arab Emirates",
    publisher = "Association for Computational Linguistics",
    url = "https://aclanthology.org/2022.emnlp-main.475/",
    doi = "10.18653/v1/2022.emnlp-main.475",
    pages = "7056--7068"
}

@inproceedings{edalati2021kroneckergpt,
    title = "Kronecker Decomposition for {GPT} Compression",
    author = "Edalati, Ali  and
      Tahaei, Marzieh  and
      Rashid, Ahmad  and
      Nia, Vahid  and
      Clark, James  and
      Rezagholizadeh, Mehdi",
    booktitle = "Proceedings of the 60th Annual Meeting of the Association for Computational Linguistics (Volume 2: Short Papers)",
    month = may,
    year = "2022",
    address = "Dublin, Ireland",
    publisher = "Association for Computational Linguistics",
    url = "https://aclanthology.org/2022.acl-short.24/",
    doi = "10.18653/v1/2022.acl-short.24",
    pages = "219--226"
}

@article{tahaei2021kroneckerbert,
      title={KroneckerBERT: Learning Kronecker Decomposition for Pre-trained Language Models via Knowledge Distillation}, 
      author={Marzieh S. Tahaei and Ella Charlaix and Vahid Partovi Nia and Ali Ghodsi and Mehdi Rezagholizadeh},
      year={2021},
      eprint={2109.06243},
      archivePrefix={arXiv},
      primaryClass={cs.CL},
      url={https://arxiv.org/abs/2109.06243}, 
}

@article{bershatsky2024lotr,
      title={LoTR: Low Tensor Rank Weight Adaptation}, 
      author={Daniel Bershatsky and Daria Cherniuk and Talgat Daulbaev and Aleksandr Mikhalev and Ivan Oseledets},
      year={2024},
      eprint={2402.01376},
      archivePrefix={arXiv},
      primaryClass={cs.CL},
      url={https://arxiv.org/abs/2402.01376}, 
}

@inproceedings{gu2025tensorllm,
  author={Gu, Yuxuan and Zhou, Wuyang and Iacovides, Giorgos and Mandic, Danilo},
  booktitle={2025 International Joint Conference on Neural Networks (IJCNN)}, 
  title={TensorLLM: Tensorising Multi-Head Attention for Enhanced Reasoning and Compression in LLMs}, 
  year={2025},
  volume={},
  number={},
  pages={1-8},
  doi={10.1109/IJCNN64981.2025.11228585}
}

@article{lopezpiqueres2025metatt,
      title={Meta{TT}: A Global Tensor-Train Adapter for Parameter-Efficient Fine-Tuning},
      author={Javier Lopez-Piqueres and Pranav Deshpande and Archan Ray and Mattia Jacopo Villani and Marco Pistoia and Niraj Kumar},
      journal={Transactions on Machine Learning Research},
      issn={2835-8856},
      year={2026},
      url={https://openreview.net/forum?id=1HdcPWfA9s},
      note={}
}

@inproceedings{chen2024quanta,
   series={NeurIPS 2024},
   title={{QuanTA}: Efficient High-Rank Fine-Tuning of LLMs with Quantum-Informed Tensor Adaptation},
   url={http://dx.doi.org/10.52202/079017-2928},
   DOI={10.52202/079017-2928},
   booktitle={Advances in Neural Information Processing Systems 37},
   publisher={Neural Information Processing Systems Foundation, Inc. (NeurIPS)},
   author={Chen, Zhuo and Dangovski, Rumen and Dugan, Owen and Loh, Charlotte and Luo, Di and Soljačić, Marin},
   year={2024},
   pages={92210–92245},
   collection={NeurIPS 2024}
}

@inproceedings{koikeakino2025quantumpeft,
      title={Quantum-{PEFT}: Ultra parameter-efficient fine-tuning},
      author={Toshiaki Koike-Akino and Francesco Tonin and Yongtao Wu and Frank Zhengqing Wu and Leyla Naz Candogan and Volkan Cevher},
      booktitle={The Thirteenth International Conference on Learning Representations},
      year={2025},
      url={https://openreview.net/forum?id=dgR6i4TSng}
}

@inproceedings{abronin2024tqcompressor,
  author={Abronin, Vadim and Naumov, Aleksei and Mazur, Denis and Bystrov, Dmitriy and Tsarova, Katerina and Melnikov, Artem and Dolgov, Sergey and Brasher, Reuben and Perelshein, Michael},
  booktitle={2024 IEEE 7th International Conference on Multimedia Information Processing and Retrieval (MIPR)}, 
  title={TQCompressor: Improving Tensor Decomposition Methods in Neural Networks Via Permutations}, 
  year={2024},
  volume={},
  number={},
  pages={503-506},
  doi={10.1109/MIPR62202.2024.00086}}

@inproceedings{kunwar2025ttloramoe,
   series={SC ’25},
   title={TT-LoRA MoE: Using Parameter-Efficient Fine-Tuning and Sparse Mixture-Of-Experts},
   url={http://dx.doi.org/10.1145/3712285.3759888},
   DOI={10.1145/3712285.3759888},
   booktitle={Proceedings of the International Conference for High Performance Computing, Networking, Storage and Analysis},
   publisher={ACM},
   author={Kunwar, Pradip and Vu, Minh N. and Gupta, Maanak and Abdelsalam, Mahmoud and Bhattarai, Manish},
   year={2025},
   month=Nov, pages={1332–1350},
   collection={SC ’25} }

@article{huang2025overtokenized,
      title={Over-Tokenized Transformer: Vocabulary is Generally Worth Scaling}, 
      author={Hongzhi Huang and Defa Zhu and Banggu Wu and Yutao Zeng and Ya Wang and Qiyang Min and Xun Zhou},
      year={2025},
      eprint={2501.16975},
      archivePrefix={arXiv},
      primaryClass={cs.CL},
      url={https://arxiv.org/abs/2501.16975}, 
}

@article{zhou2026tensorizingengram,
      title={Tensorizing Engram: Sharing Latents Across N-Gram Embeddings is Beneficial in LLMs}, 
      author={Wuyang Zhou and Yuxuan Gu and Giorgos Iacovides and Yuning Qiu and Qibin Zhao and Danilo Mandic},
      year={2026},
      eprint={2606.08347},
      archivePrefix={arXiv},
      primaryClass={cs.CL},
      url={https://arxiv.org/abs/2606.08347}, 
}

@article{deepseekai2024deepseekv2,
      title={DeepSeek-V2: A Strong, Economical, and Efficient Mixture-of-Experts Language Model}, 
      author={{DeepSeek-AI}},
      year={2024},
      eprint={2405.04434},
      archivePrefix={arXiv},
      primaryClass={cs.CL},
      url={https://arxiv.org/abs/2405.04434}, 
}

@article{deepseekai2025deepseekv32,
      title={DeepSeek-V3.2: Pushing the Frontier of Open Large Language Models}, 
      author={{DeepSeek-AI}},
      year={2025},
      eprint={2512.02556},
      archivePrefix={arXiv},
      primaryClass={cs.CL},
      url={https://arxiv.org/abs/2512.02556}, 
}

@article{adtn,
  author = {Yong Qing  and Ke Li  and Peng-Fei Zhou  and Shi-Ju Ran },
  title = {Compressing Neural Networks Using Tensor Networks with Exponentially Fewer Variational Parameters},
  journal = {Intelligent Computing},
  volume = {4},
  number = {},
  pages = {0123},
  year = {2025},
  doi = {10.34133/icomputing.0123},
  URL = {https://spj.science.org/doi/abs/10.34133/icomputing.0123},
  eprint = {https://spj.science.org/doi/pdf/10.34133/icomputing.0123},
}

@article{su2023rope,
      title={RoFormer: Enhanced Transformer with Rotary Position Embedding}, 
      author={Jianlin Su and Yu Lu and Shengfeng Pan and Ahmed Murtadha and Bo Wen and Yunfeng Liu},
      year={2023},
      eprint={2104.09864},
      archivePrefix={arXiv},
      primaryClass={cs.CL},
      url={https://arxiv.org/abs/2104.09864}, 
}

@inproceedings{gu2026tera,
    title = "{T}e{RA}: Vector-based Random Tensor Network for High-Rank Adaptation of Large Language Models",
    author = "Gu, Yuxuan  and
      Zhou, Wuyang  and
      Iacovides, Giorgos  and
      Mandic, Danilo",
    booktitle = "Proceedings of the 64th Annual Meeting of the {A}ssociation for {C}omputational {L}inguistics (Volume 1: Long Papers)",
    month = jul,
    year = "2026",
    address = "San Diego, California, United States",
    publisher = "Association for Computational Linguistics",
    url = "https://aclanthology.org/2026.acl-long.106/",
    doi = "10.18653/v1/2026.acl-long.106",
    pages = "2314--2329",
    ISBN = "979-8-89176-390-6"
}

@article{gan2022morphte,
      title={MorphTE: Injecting Morphology in Tensorized Embeddings}, 
      author={Guobing Gan and Peng Zhang and Sunzhu Li and Xiuqing Lu and Benyou Wang},
      year={2022},
      eprint={2210.15379},
      archivePrefix={arXiv},
      primaryClass={cs.CL},
      url={https://arxiv.org/abs/2210.15379}, 
}

@article{vidal2008mera,
   title={Class of Quantum Many-Body States That Can Be Efficiently Simulated},
   volume={101},
   ISSN={1079-7114},
   url={http://dx.doi.org/10.1103/PhysRevLett.101.110501},
   DOI={10.1103/physrevlett.101.110501},
   number={11},
   journal={Physical Review Letters},
   publisher={American Physical Society (APS)},
   author={Vidal, G.},
   year={2008},
   month=Sept 
}

@article{gracedyck2010h_tucker,
  author = {Grasedyck, Lars},
  title = {Hierarchical Singular Value Decomposition of Tensors},
  journal = {SIAM Journal on Matrix Analysis and Applications},
  volume = {31},
  number = {4},
  pages = {2029-2054},
  year = {2010},
  doi = {10.1137/090764189},
  URL = {https://doi.org/10.1137/090764189},
  eprint = {https://doi.org/10.1137/090764189}
}

@inproceedings{cruys2013svo,
    title = "A Tensor-based Factorization Model of Semantic Compositionality",
    author = "Van de Cruys, Tim  and
      Poibeau, Thierry  and
      Korhonen, Anna",
    editor = "Vanderwende, Lucy  and
      Daum{\'e} III, Hal  and
      Kirchhoff, Katrin",
    booktitle = "Proceedings of the 2013 Conference of the North {A}merican Chapter of the Association for Computational Linguistics: Human Language Technologies",
    month = jun,
    year = "2013",
    address = "Atlanta, Georgia",
    publisher = "Association for Computational Linguistics",
    url = "https://aclanthology.org/N13-1134/",
    pages = "1142--1151"
}

@inproceedings{etal2014svotensorspaces,
    title = "Evaluating Neural Word Representations in Tensor-Based Compositional Settings",
    author = "Milajevs, Dmitrijs  and
      Kartsaklis, Dimitri  and
      Sadrzadeh, Mehrnoosh  and
      Purver, Matthew",
    editor = "Moschitti, Alessandro  and
      Pang, Bo  and
      Daelemans, Walter",
    booktitle = "Proceedings of the 2014 Conference on Empirical Methods in Natural Language Processing ({EMNLP})",
    month = oct,
    year = "2014",
    address = "Doha, Qatar",
    publisher = "Association for Computational Linguistics",
    url = "https://aclanthology.org/D14-1079/",
    doi = "10.3115/v1/D14-1079",
    pages = "708--719"
}

@article{rahimi2021tenssent,
  author = {Rahimi, Zahra and Homayounpour, Mohammad Mehdi},
  title = {TensSent: a tensor based sentimental word embedding method},
  year = {2021},
  issue_date = {Aug 2021},
  publisher = {Kluwer Academic Publishers},
  address = {USA},
  volume = {51},
  number = {8},
  issn = {0924-669X},
  url = {https://doi.org/10.1007/s10489-020-02163-8},
  doi = {10.1007/s10489-020-02163-8},
  journal = {Applied Intelligence},
  month = aug,
  pages = {6056–6071},
  numpages = {16}
}

@inproceedings{etal2017frame,
    title = "Frame-Based Continuous Lexical Semantics through Exponential Family Tensor Factorization and Semantic Proto-Roles",
    author = "Ferraro, Francis  and
      Poliak, Adam  and
      Cotterell, Ryan  and
      Van Durme, Benjamin",
    editor = "Ide, Nancy  and
      Herbelot, Aur{\'e}lie  and
      M{\`a}rquez, Llu{\'i}s",
    booktitle = "Proceedings of the 6th Joint Conference on Lexical and Computational Semantics (*{SEM} 2017)",
    month = aug,
    year = "2017",
    address = "Vancouver, Canada",
    publisher = "Association for Computational Linguistics",
    url = "https://aclanthology.org/S17-1011/",
    doi = "10.18653/v1/S17-1011",
    pages = "97--103"
}

@article{dai2019transformerxl,
      title={Transformer-XL: Attentive Language Models Beyond a Fixed-Length Context}, 
      author={Zihang Dai and Zhilin Yang and Yiming Yang and Jaime Carbonell and Quoc V. Le and Ruslan Salakhutdinov},
      year={2019},
      eprint={1901.02860},
      archivePrefix={arXiv},
      primaryClass={cs.LG},
      url={https://arxiv.org/abs/1901.02860}, 
}

@inproceedings{atad2026tensorlens,
    title = "{T}ensor{L}ens: End-to-End Transformer Analysis via High-Order Attention Tensors",
    author = "Atad, Ido Andrew  and
      Zimerman, Itamar  and
      Katz, Shahar  and
      Wolf, Lior",
    editor = "Liakata, Maria  and
      Moreira, Viviane P.  and
      Zhang, Jiajun  and
      Jurgens, David",
    booktitle = "Proceedings of the 64th Annual Meeting of the {A}ssociation for {C}omputational {L}inguistics (Volume 1: Long Papers)",
    month = jul,
    year = "2026",
    address = "San Diego, California, United States",
    publisher = "Association for Computational Linguistics",
    url = "https://aclanthology.org/2026.acl-long.156/",
    doi = "10.18653/v1/2026.acl-long.156",
    pages = "3452--3468",
    ISBN = "979-8-89176-390-6"
}

@misc{yang2024comera,
      title={{CoMERA}: Computing- and Memory-Efficient Training via Rank-Adaptive Tensor Optimization}, 
      author={Zi Yang and Ziyue Liu and Samridhi Choudhary and Xinfeng Xie and Cao Gao and Siegfried Kunzmann and Zheng Zhang},
      year={2024},
      eprint={2405.14377},
      archivePrefix={arXiv},
      primaryClass={cs.LG},
      url={https://arxiv.org/abs/2405.14377}, 
}

@inproceedings{pahani2021shapeshifter,
 author = {Panahi, Aliakbar and Saeedi, Seyran and Arodz, Tom},
 booktitle = {Advances in Neural Information Processing Systems},
 pages = {1337--1350},
 publisher = {Curran Associates, Inc.},
 title = {Shapeshifter: a Parameter-efficient Transformer using Factorized Reshaped Matrices},
 url = {https://proceedings.neurips.cc/paper_files/paper/2021/file/09def3ebbc44ff3426b28fcd88c83554-Paper.pdf},
 volume = {34},
 year = {2021}
}

@misc{javanmard2026mpopicogpt,
      title={Compressing Transformer Language Models via Matrix Product Operator Decomposition: A Case Study on PicoGPT},
      author={Younes Javanmard and Tanmoy Pandit and Masoud Mardani},
      year={2026},
      eprint={2603.28534},
      archivePrefix={arXiv},
      primaryClass={cs.CL},
      url={https://arxiv.org/abs/2603.28534},
}

@misc{kozyrev2026minima,
      title={Minima: A Practical Tensor-Network Compression Pipeline for Production-Scale Large Language Models},
      author={Sergii Kozyrev and Davyd Maiboroda},
      year={2026},
      eprint={2602.01613},
      archivePrefix={arXiv},
      primaryClass={cs.CL},
      url={https://arxiv.org/abs/2602.01613},
}

@article{shazeer2017moe,
      title={Outrageously Large Neural Networks: The Sparsely-Gated Mixture-of-Experts Layer}, 
      author={Noam Shazeer and Azalia Mirhoseini and Krzysztof Maziarz and Andy Davis and Quoc Le and Geoffrey Hinton and Jeff Dean},
      year={2017},
      eprint={1701.06538},
      archivePrefix={arXiv},
      primaryClass={cs.LG},
      url={https://arxiv.org/abs/1701.06538}, 
}

@article{zhang2023adalora,
      title={AdaLoRA: Adaptive Budget Allocation for Parameter-Efficient Fine-Tuning}, 
      author={Qingru Zhang and Minshuo Chen and Alexander Bukharin and Nikos Karampatziakis and Pengcheng He and Yu Cheng and Weizhu Chen and Tuo Zhao},
      year={2023},
      eprint={2303.10512},
      archivePrefix={arXiv},
      primaryClass={cs.CL},
      url={https://arxiv.org/abs/2303.10512}, 
}

@inproceedings{solgi2025saten,
    title = "Saten: Sparse Augmented Tensor Networks for Post-Training Compression of Large Language Models",
    author = "Solgi, Ryan  and
      Zhen, Kai  and
      Swaminathan, Rupak Vignesh  and
      Susanj, Nathan  and
      Mouchtaris, Athanasios  and
      Kunzmann, Siegfried  and
      Zhang, Zheng",
    booktitle = "Findings of the Association for Computational Linguistics: EMNLP 2025",
    month = nov,
    year = "2025",
    address = "Suzhou, China",
    publisher = "Association for Computational Linguistics",
    url = "https://aclanthology.org/2025.findings-emnlp.1287/",
    doi = "10.18653/v1/2025.findings-emnlp.1287",
    pages = "23674--23683",
    ISBN = "979-8-89176-335-7"
}

@misc{kaplan2020scalinglaws,
      title={Scaling Laws for Neural Language Models}, 
      author={Jared Kaplan and Sam McCandlish and Tom Henighan and Tom B. Brown and Benjamin Chess and Rewon Child and Scott Gray and Alec Radford and Jeffrey Wu and Dario Amodei},
      year={2020},
      eprint={2001.08361},
      archivePrefix={arXiv},
      primaryClass={cs.LG},
      url={https://arxiv.org/abs/2001.08361}, 
}

@inproceedings{koikeakino2026einsort,
      title={EinSort: Sorting is All We Need for Tensorizing {LLM}},
      author={Toshiaki Koike-Akino and Jing Liu and Ye Wang},
      booktitle={ICML'26 workshop on CoLoRAI - The 2nd Workshop on Connecting Low-rank Representations in AI},
      year={2026},
      url={https://openreview.net/forum?id=yoIh7UwdAC}
}

@misc{wang2021gpt_j,
  author = {Wang, Ben and Komatsuzaki, Aran},
  title = {{GPT-J-6B: A 6 Billion Parameter Autoregressive Language Model}},
  howpublished = {\url{https://github.com/kingoflolz/mesh-transformer-jax}},
  year = 2021,
  month = May
}

@InProceedings{luo2025trawl,
      author="Luo, Yiran
      and Patel, Het
      and Fu, Yu
      and Ahn, Dawon
      and Chen, Jia
      and Dong, Yue
      and Papalexakis, Evangelos E.",
      title={{TRAWL}: Tensor Reduced and Approximated Weights for Large Language Models},
      booktitle="Data Science: Foundations and Applications",
      year="2025",
      publisher="Springer Nature Singapore",
      address="Singapore",
      pages="402--413",
      isbn="978-981-96-8298-0"
}

@article{touvron2023llama2,
      title={Llama 2: Open Foundation and Fine-Tuned Chat Models}, 
      author={Touvron and others},
      year={2023},
      eprint={2307.09288},
      archivePrefix={arXiv},
      primaryClass={cs.CL},
      url={https://arxiv.org/abs/2307.09288}, 
}

@article{koikeakino2025latentllm,
      title={LatentLLM: Attention-Aware Joint Tensor Compression}, 
      author={Toshiaki Koike-Akino and Xiangyu Chen and Jing Liu and Ye Wang and Pu and Wang and Matthew Brand},
      year={2025},
      eprint={2505.18413},
      archivePrefix={arXiv},
      primaryClass={cs.LG},
      url={https://arxiv.org/abs/2505.18413}, 
}

@article{zhang2022opt,
      title={OPT: Open Pre-trained Transformer Language Models}, 
      author={Susan Zhang and others},
      year={2022},
      eprint={2205.01068},
      archivePrefix={arXiv},
      primaryClass={cs.CL},
      url={https://arxiv.org/abs/2205.01068}, 
}

@misc{metaai2024llama321b,
  author       = {{Meta AI}},
  title        = {Llama 3.2-1B},
  year         = {2024},
  month        = sep,
  howpublished = {\url{https://huggingface.co/meta-llama/Llama-3.2-1B}},
}

@misc{unsloth2024llama321b,
  author       = {{Unsloth AI}},
  title        = {Llama 3.2-1B},
  year         = {2024},
  month        = sep,
  howpublished = {\url{https://huggingface.co/unsloth/Llama-3.2-1B}},
}

@inproceedings{klein2026tuckerattention,
      title={Tucker Attention: A generalization of approximate attention mechanisms},
      author={Timon Klein and Jonas Kusch and Sebastian Sager and Stefan Schnake and Steffen Schotth{\"o}fer},
      booktitle={Forty-third International Conference on Machine Learning},
      year={2026},
      url={https://openreview.net/forum?id=ErcPPRZaiq}
}

@inproceedings{liu2024decoquant,
    title = "Unlocking Data-free Low-bit Quantization with Matrix Decomposition for {KV} Cache Compression",
    author = "Liu, Peiyu  and
      Gao, Ze-Feng  and
      Zhao, Xin  and
      Ma, Yipeng  and
      Wang, Tao  and
      Wen, Ji-Rong",
    booktitle = "Proceedings of the 62nd Annual Meeting of the Association for Computational Linguistics (Volume 1: Long Papers)",
    month = aug,
    year = "2024",
    address = "Bangkok, Thailand",
    publisher = "Association for Computational Linguistics",
    url = "https://aclanthology.org/2024.acl-long.133/",
    doi = "10.18653/v1/2024.acl-long.133",
    pages = "2430--2440"
}

@misc{raffel2023t5,
      title={Exploring the Limits of Transfer Learning with a Unified Text-to-Text Transformer}, 
      author={Colin Raffel and Noam Shazeer and Adam Roberts and Katherine Lee and Sharan Narang and Michael Matena and Yanqi Zhou and Wei Li and Peter J. Liu},
      year={2023},
      eprint={1910.10683},
      archivePrefix={arXiv},
      primaryClass={cs.LG},
      url={https://arxiv.org/abs/1910.10683}, 
}

@misc{dettmers2022llmint8,
      title={LLM.int8(): 8-bit Matrix Multiplication for Transformers at Scale}, 
      author={Tim Dettmers and Mike Lewis and Younes Belkada and Luke Zettlemoyer},
      year={2022},
      eprint={2208.07339},
      archivePrefix={arXiv},
      primaryClass={cs.LG},
      url={https://arxiv.org/abs/2208.07339}, 
}

@misc{liu2023llmqat,
      title={LLM-QAT: Data-Free Quantization Aware Training for Large Language Models}, 
      author={Zechun Liu and Barlas Oguz and Changsheng Zhao and Ernie Chang and Pierre Stock and Yashar Mehdad and Yangyang Shi and Raghuraman Krishnamoorthi and Vikas Chandra},
      year={2023},
      eprint={2305.17888},
      archivePrefix={arXiv},
      primaryClass={cs.CL},
      url={https://arxiv.org/abs/2305.17888}, 
}

@misc{ashkboos2024slicegpt,
      title={SliceGPT: Compress Large Language Models by Deleting Rows and Columns}, 
      author={Saleh Ashkboos and Maximilian L. Croci and Marcelo Gennari do Nascimento and Torsten Hoefler and James Hensman},
      year={2024},
      eprint={2401.15024},
      archivePrefix={arXiv},
      primaryClass={cs.LG},
      url={https://arxiv.org/abs/2401.15024}, 
}

@article{hinton2015distillingknowledge,
      title={Distilling the Knowledge in a Neural Network}, 
      author={Geoffrey Hinton and Oriol Vinyals and Jeff Dean},
      year={2015},
      eprint={1503.02531},
      archivePrefix={arXiv},
      primaryClass={stat.ML},
      url={https://arxiv.org/abs/1503.02531}, 
}

@misc{osborne2026picogptjl,
      title={PicoGPT.jl: From-scratch GPT in pure Julia},
      author={Tobias J. Osborne},
      year={2026},
      howpublished={\url{https://github.com/tobiasosborne/PicoGPT.jl}},
      note={GitHub repository},
}

@article{lathauwer2008btd,
  author = {De Lathauwer, Lieven},
  title = {Decompositions of a Higher-Order Tensor in Block Terms—Part II: Definitions and Uniqueness},
  journal = {SIAM Journal on Matrix Analysis and Applications},
  volume = {30},
  number = {3},
  pages = {1033-1066},
  year = {2008},
  doi = {10.1137/070690729},
  URL = {https://doi.org/10.1137/070690729},
  eprint = {https://doi.org/10.1137/070690729}
}

@article{lathauwer2000hosvd,
      author = {De Lathauwer, Lieven and De Moor, Bart and Vandewalle, Joos},
      title = {A Multilinear Singular Value Decomposition},
      journal = {SIAM Journal on Matrix Analysis and Applications},
      volume = {21},
      number = {4},
      pages = {1253-1278},
      year = {2000},
      doi = {10.1137/S0895479896305696},
      URL = {https://doi.org/10.1137/S0895479896305696},
      eprint = {https://doi.org/10.1137/S0895479896305696}
}

@article{oseledets2010ttm,
  author = {Oseledets, I. V.},
  title = {Approximation of $2^d\times2^d$ Matrices Using Tensor Decomposition},
  journal = {SIAM Journal on Matrix Analysis and Applications},
  volume = {31},
  number = {4},
  pages = {2130-2145},
  year = {2010},
  doi = {10.1137/090757861},
  URL = {https://doi.org/10.1137/090757861},
  eprint = {https://doi.org/10.1137/090757861}
}

@misc{zhao2016tensorring,
      title={Tensor Ring Decomposition}, 
      author={Qibin Zhao and Guoxu Zhou and Shengli Xie and Liqing Zhang and Andrzej Cichocki},
      year={2016},
      eprint={1606.05535},
      archivePrefix={arXiv},
      primaryClass={math.NA},
      url={https://arxiv.org/abs/1606.05535}, 
}

@article{batselier2017ktd,
  author = {Batselier, Kim and Wong, Ngai},
  title = {A constructive arbitrary-degree Kronecker product decomposition of tensors},
  journal = {Numerical Linear Algebra with Applications},
  volume = {24},
  number = {5},
  pages = {e2097},
  doi = {https://doi.org/10.1002/nla.2097},
  url = {https://onlinelibrary.wiley.com/doi/abs/10.1002/nla.2097},
  eprint = {https://onlinelibrary.wiley.com/doi/pdf/10.1002/nla.2097},
  note = {e2097 nla.2097},
  year = {2017}
}

@article{verstraete2008peps,
   title={Matrix product states, projected entangled pair states, and variational renormalization group methods for quantum spin systems},
   volume={57},
   ISSN={1460-6976},
   url={http://dx.doi.org/10.1080/14789940801912366},
   DOI={10.1080/14789940801912366},
   number={2},
   journal={Advances in Physics},
   publisher={Informa UK Limited},
   author={Verstraete, F. and Murg, V. and Cirac, J.I.},
   year={2008},
   month=Mar, pages={143–224}
}

@article{hawkins2019rankselection1,
      title={Bayesian Tensorized Neural Networks with Automatic Rank Selection}, 
      author={Cole Hawkins and Zheng Zhang},
      year={2019},
      eprint={1905.10478},
      archivePrefix={arXiv},
      primaryClass={cs.LG},
      url={https://arxiv.org/abs/1905.10478}, 
}

@article{hawkins2021rankselection2,
      title={Towards Compact Neural Networks via End-to-End Training: A Bayesian Tensor Approach with Automatic Rank Determination}, 
      author={Cole Hawkins and Xing Liu and Zheng Zhang},
      year={2021},
      eprint={2010.08689},
      archivePrefix={arXiv},
      primaryClass={cs.LG},
      url={https://arxiv.org/abs/2010.08689}, 
}

@article{gu2022heat,
      title={HEAT: Hardware-Efficient Automatic Tensor Decomposition for Transformer Compression}, 
      author={Jiaqi Gu and Ben Keller and Jean Kossaifi and Anima Anandkumar and Brucek Khailany and David Z. Pan},
      year={2022},
      eprint={2211.16749},
      archivePrefix={arXiv},
      primaryClass={cs.LG},
      url={https://arxiv.org/abs/2211.16749}, 
}

@article{kruskal1977three,
      title = {Three-way arrays: rank and uniqueness of trilinear decompositions, with application to arithmetic complexity and statistics},
      journal = {Linear Algebra and its Applications},
      volume = {18},
      number = {2},
      pages = {95-138},
      year = {1977},
      issn = {0024-3795},
      doi = {https://doi.org/10.1016/0024-3795(77)90069-6},
      url = {https://www.sciencedirect.com/science/article/pii/0024379577900696},
      author = {Joseph B. Kruskal}
}

@article{sidiropoulos2000cpuniqueness,
      author = {Sidiropoulos, Nicholas D. and Bro, Rasmus},
      title = {On the uniqueness of multilinear decomposition of N-way arrays},
      journal = {Journal of Chemometrics},
      volume = {14},
      number = {3},
      pages = {229-239},
      doi = {https://doi.org/10.1002/1099-128X(200005/06)14:3<229::AID-CEM587>3.0.CO;2-N},
      url = {https://analyticalsciencejournals.onlinelibrary.wiley.com/doi/abs/10.1002/1099-128X%28200005/06%2914%3A3%3C229%3A%3AAID-CEM587%3E3.0.CO%3B2-N},
      eprint = {https://analyticalsciencejournals.onlinelibrary.wiley.com/doi/pdf/10.1002/1099-128X%28200005/06%2914%3A3%3C229%3A%3AAID-CEM587%3E3.0.CO%3B2-N},
      year = {2000}
}

@article{wan2024efficientllmsurvey1,
      title={Efficient Large Language Models: A Survey}, 
      author={Zhongwei Wan and Xin Wang and Che Liu and Samiul Alam and Yu Zheng and Jiachen Liu and Zhongnan Qu and Shen Yan and Yi Zhu and Quanlu Zhang and Mosharaf Chowdhury and Mi Zhang},
      year={2024},
      eprint={2312.03863},
      archivePrefix={arXiv},
      primaryClass={cs.CL},
      url={https://arxiv.org/abs/2312.03863}, 
}

@article{wang2025peftsurvey1,
      title={Parameter-Efficient Fine-Tuning in Large Models: A Survey of Methodologies}, 
      author={Luping Wang and Sheng Chen and Linnan Jiang and Shu Pan and Runze Cai and Sen Yang and Fei Yang},
      year={2025},
      eprint={2410.19878},
      archivePrefix={arXiv},
      primaryClass={cs.CL},
      url={https://arxiv.org/abs/2410.19878}, 
}

@article{kim2025peftcompressionsurvey1,
    author = {Kim, Gun Il and Hwang, Sunga and Jang, Beakcheol},
    title = {Efficient Compressing and Tuning Methods for Large Language Models: A Systematic Literature Review},
    year = {2025},
    issue_date = {October 2025},
    publisher = {Association for Computing Machinery},
    address = {New York, NY, USA},
    volume = {57},
    number = {10},
    issn = {0360-0300},
    url = {https://doi.org/10.1145/3728636},
    doi = {10.1145/3728636},
    journal = {ACM Comput. Surv.},
    month = may,
    articleno = {253},
    numpages = {39}
}

@article{zhu2024compressionsurvey1,
      title={A Survey on Model Compression for Large Language Models}, 
      author={Xunyu Zhu and Jian Li and Yong Liu and Can Ma and Weiping Wang},
      year={2024},
      eprint={2308.07633},
      archivePrefix={arXiv},
      primaryClass={cs.CL},
      url={https://arxiv.org/abs/2308.07633}, 
}

@article{tang2024compressionsurvey2,
      title={A Survey on Transformer Compression}, 
      author={Yehui Tang and Yunhe Wang and Jianyuan Guo and Zhijun Tu and Kai Han and Hailin Hu and Dacheng Tao},
      year={2024},
      eprint={2402.05964},
      archivePrefix={arXiv},
      primaryClass={cs.LG},
      url={https://arxiv.org/abs/2402.05964}, 
}

@article{wang2024compressioninferencesurvey1,
      title={Model Compression and Efficient Inference for Large Language Models: A Survey}, 
      author={Wenxiao Wang and Wei Chen and Yicong Luo and Yongliu Long and Zhengkai Lin and Liye Zhang and Binbin Lin and Deng Cai and Xiaofei He},
      year={2024},
      eprint={2402.09748},
      archivePrefix={arXiv},
      primaryClass={cs.CL},
      url={https://arxiv.org/abs/2402.09748}, 
}

@article{zhou2024efficientinferencesurvey1,
      title={A Survey on Efficient Inference for Large Language Models}, 
      author={Zixuan Zhou and Xuefei Ning and Ke Hong and Tianyu Fu and Jiaming Xu and Shiyao Li and Yuming Lou and Luning Wang and Zhihang Yuan and Xiuhong Li and Shengen Yan and Guohao Dai and Xiao-Ping Zhang and Yuhan Dong and Yu Wang},
      year={2024},
      eprint={2404.14294},
      archivePrefix={arXiv},
      primaryClass={cs.CL},
      url={https://arxiv.org/abs/2404.14294}, 
}

@article{yuan2024efficientinferencesurvey2,
      title={LLM Inference Unveiled: Survey and Roofline Model Insights}, 
      author={Zhihang Yuan and Yuzhang Shang and Yang Zhou and Zhen Dong and Zhe Zhou and Chenhao Xue and Bingzhe Wu and Zhikai Li and Qingyi Gu and Yong Jae Lee and Yan Yan and Beidi Chen and Guangyu Sun and Kurt Keutzer},
      year={2024},
      eprint={2402.16363},
      archivePrefix={arXiv},
      primaryClass={cs.CL},
      url={https://arxiv.org/abs/2402.16363}, 
}

@misc{li2025kvcachesurvey1,
      title={A Survey on Large Language Model Acceleration based on KV Cache Management}, 
      author={Haoyang Li and Yiming Li and Anxin Tian and Tianhao Tang and Zhanchao Xu and Xuejia Chen and Nicole Hu and Wei Dong and Qing Li and Lei Chen},
      year={2025},
      eprint={2412.19442},
      archivePrefix={arXiv},
      primaryClass={cs.AI},
      url={https://arxiv.org/abs/2412.19442}, 
}

@article{miao2025efficientservingsurvey1,
    author = {Miao, Xupeng and Oliaro, Gabriele and Zhang, Zhihao and Cheng, Xinhao and Jin, Hongyi and Chen, Tianqi and Jia, Zhihao},
    title = {Towards Efficient Generative Large Language Model Serving: A Survey from Algorithms to Systems},
    year = {2025},
    issue_date = {January 2026},
    publisher = {Association for Computing Machinery},
    address = {New York, NY, USA},
    volume = {58},
    number = {1},
    issn = {0360-0300},
    url = {https://doi.org/10.1145/3754448},
    doi = {10.1145/3754448},
    journal = {ACM Comput. Surv.},
    month = sep,
    articleno = {15},
    numpages = {37}
}

@article{ji2019tensorsformlsurvey1,
    author={Ji, Yuwang and Wang, Qiang and Li, Xuan and Liu, Jie},
    journal={IEEE Access}, 
    title={A Survey on Tensor Techniques and Applications in Machine Learning}, 
    year={2019},
    volume={7},
    number={},
    pages={162950-162990},
    doi={10.1109/ACCESS.2019.2949814}
}

@article{sidiropoulos2017tensorsforspandml,
    title={Tensor Decomposition for Signal Processing and Machine Learning},
    volume={65},
    ISSN={1941-0476},
    url={http://dx.doi.org/10.1109/TSP.2017.2690524},
    DOI={10.1109/tsp.2017.2690524},
    number={13},
    journal={IEEE Transactions on Signal Processing},
    publisher={Institute of Electrical and Electronics Engineers (IEEE)},
    author={Sidiropoulos, Nicholas D. and De Lathauwer, Lieven and Fu, Xiao and Huang, Kejun and Papalexakis, Evangelos E. and Faloutsos, Christos},
    year={2017},
    month=July, pages={3551–3582}
}

@article{panagakis2021tensorsfornn1,
   title={Tensor Methods in Computer Vision and Deep Learning},
   volume={109},
   ISSN={1558-2256},
   url={http://dx.doi.org/10.1109/JPROC.2021.3074329},
   DOI={10.1109/jproc.2021.3074329},
   number={5},
   journal={Proceedings of the IEEE},
   publisher={Institute of Electrical and Electronics Engineers (IEEE)},
   author={Panagakis, Yannis and Kossaifi, Jean and Chrysos, Grigorios G. and Oldfield, James and Nicolaou, Mihalis A. and Anandkumar, Anima and Zafeiriou, Stefanos},
   year={2021},
   month=May, pages={863–890}
}

@misc{wang2025tnmeetnn,
      title={Tensor Networks Meet Neural Networks: A Survey and Future Perspectives}, 
      author={Maolin Wang and Yu Pan and Zenglin Xu and Guangxi Li and Xiangli Yang and Danilo Mandic and Andrzej Cichocki},
      year={2025},
      eprint={2302.09019},
      archivePrefix={arXiv},
      primaryClass={cs.LG},
      url={https://arxiv.org/abs/2302.09019}, 
}

@article{xinwei2024tensorsfornn2,
  author={Ou, Xinwei and Chen, Zhangxin and Zhu, Ce and Liu, Yipeng},
  journal={Journal of Systems Engineering and Electronics}, 
  title={Low Rank Optimization for Efficient Deep Learning: Making a Balance Between Compact Architecture And Fast Training}, 
  year={2024},
  volume={35},
  number={3},
  pages={509-531},
  doi={10.23919/JSEE.2023.000159}
}

@article{he2026tensorsfornn3,
      title = {A survey of latent factorization of tensor-based model compression: Algorithms, toolboxes and future directions},
      journal = {Neurocomputing},
      volume = {682},
      pages = {133455},
      year = {2026},
      issn = {0925-2312},
      doi = {https://doi.org/10.1016/j.neucom.2026.133455},
      url = {https://www.sciencedirect.com/science/article/pii/S0925231226008520},
      author = {Yaping He and Hao Wu and Weibo Liu and Xin Luo}
}

@article{somvanshi2026interpretabilitysurvey1,
    author = {Somvanshi, Shriyank and Islam, Md Monzurul and Rafe, Amir and Tusti, Anannya Ghosh and Chakraborty, Arka and Baitullah, Anika and Chowdhury, Tausif Islam and Alnawmasi, Nawaf and Dutta, Anandi and Das, Subasish},
    title = {Bridging the Black Box: A Survey on Mechanistic Interpretability in AI},
    year = {2026},
    issue_date = {June 2026},
    publisher = {Association for Computing Machinery},
    address = {New York, NY, USA},
    volume = {58},
    number = {8},
    issn = {0360-0300},
    url = {https://doi.org/10.1145/3787104},
    doi = {10.1145/3787104},
    journal = {ACM Comput. Surv.},
    month = feb,
    articleno = {210},
    numpages = {35}
}

@article{rai2025interpretabilitysurvey2,
      title={A Practical Review of Mechanistic Interpretability for Transformer-Based Language Models}, 
      author={Daking Rai and Yilun Zhou and Shi Feng and Abulhair Saparov and Ziyu Yao},
      year={2025},
      eprint={2407.02646},
      archivePrefix={arXiv},
      primaryClass={cs.AI},
      url={https://arxiv.org/abs/2407.02646}, 
}

@article{bereska2024interpretabilitysurvey3,
      title={Mechanistic Interpretability for AI Safety -- A Review}, 
      author={Leonard Bereska and Efstratios Gavves},
      year={2024},
      eprint={2404.14082},
      archivePrefix={arXiv},
      primaryClass={cs.AI},
      url={https://arxiv.org/abs/2404.14082}, 
}

@article{taylor2024interpretabilitysurvey4,
      title={An introduction to graphical tensor notation for mechanistic interpretability}, 
      author={Jordan K. Taylor},
      year={2024},
      eprint={2402.01790},
      archivePrefix={arXiv},
      primaryClass={cs.LG},
      url={https://arxiv.org/abs/2402.01790}, 
}

@misc{openai2024gpt4,
      title={GPT-4 Technical Report}, 
      author={OpenAI},
      year={2024},
      eprint={2303.08774},
      archivePrefix={arXiv},
      primaryClass={cs.CL},
      url={https://arxiv.org/abs/2303.08774}, 
}

@misc{hoffmann2022scalinglaws2,
      title={Training Compute-Optimal Large Language Models}, 
      author={Jordan Hoffmann and Sebastian Borgeaud and Arthur Mensch and Elena Buchatskaya and Trevor Cai and Eliza Rutherford and Diego de Las Casas and Lisa Anne Hendricks and Johannes Welbl and Aidan Clark and Tom Hennigan and Eric Noland and Katie Millican and George van den Driessche and Bogdan Damoc and Aurelia Guy and Simon Osindero and Karen Simonyan and Erich Elsen and Jack W. Rae and Oriol Vinyals and Laurent Sifre},
      year={2022},
      eprint={2203.15556},
      archivePrefix={arXiv},
      primaryClass={cs.CL},
      url={https://arxiv.org/abs/2203.15556}, 
}

@misc{yang2025qwen3,
      title={Qwen3 Technical Report}, 
      author={{Qwen Team}},
      year={2025},
      eprint={2505.09388},
      archivePrefix={arXiv},
      primaryClass={cs.CL},
      url={https://arxiv.org/abs/2505.09388}, 
}

@misc{deepseekai2026deepseekv4,
      title={DeepSeek-V4: Towards Highly Efficient Million-Token Context Intelligence}, 
      author={{DeepSeek-AI}},
      year={2026},
      eprint={2606.19348},
      archivePrefix={arXiv},
      primaryClass={cs.CL},
      url={https://arxiv.org/abs/2606.19348}, 
}

@misc{kimiteam2026kimik3,
      title={Kimi K3: Open Frontier Intelligence}, 
      author={{Kimi Team}},
      year={2026},
      eprint={2607.24653},
      archivePrefix={arXiv},
      primaryClass={cs.CL},
      url={https://arxiv.org/abs/2607.24653}, 
}

@article{bahdanau2016attention,
      title={Neural Machine Translation by Jointly Learning to Align and Translate}, 
      author={Dzmitry Bahdanau and Kyunghyun Cho and Yoshua Bengio},
      year={2016},
      eprint={1409.0473},
      archivePrefix={arXiv},
      primaryClass={cs.CL},
      url={https://arxiv.org/abs/1409.0473}, 
}

@article{shazeer2020glu,
      title={GLU Variants Improve Transformer}, 
      author={Noam Shazeer},
      year={2020},
      eprint={2002.05202},
      archivePrefix={arXiv},
      primaryClass={cs.LG},
      url={https://arxiv.org/abs/2002.05202}, 
}

@article{hendrycks2023gelu,
      title={Gaussian Error Linear Units (GELUs)}, 
      author={Dan Hendrycks and Kevin Gimpel},
      year={2023},
      eprint={1606.08415},
      archivePrefix={arXiv},
      primaryClass={cs.LG},
      url={https://arxiv.org/abs/1606.08415}, 
}

@article{ramachandran2017swish,
      title={Searching for Activation Functions}, 
      author={Prajit Ramachandran and Barret Zoph and Quoc V. Le},
      year={2017},
      eprint={1710.05941},
      archivePrefix={arXiv},
      primaryClass={cs.NE},
      url={https://arxiv.org/abs/1710.05941}, 
}

@article{meng2023ffnfactuallocation,
      title={Locating and Editing Factual Associations in GPT}, 
      author={Kevin Meng and David Bau and Alex Andonian and Yonatan Belinkov},
      year={2023},
      eprint={2202.05262},
      archivePrefix={arXiv},
      primaryClass={cs.CL},
      url={https://arxiv.org/abs/2202.05262}, 
}

@article{hitchcock1927cp,
      author = {Hitchcock, Frank L.},
      title = {The Expression of a Tensor or a Polyadic as a Sum of Products},
      journal = {Journal of Mathematics and Physics},
      volume = {6},
      number = {1-4},
      pages = {164-189},
      doi = {https://doi.org/10.1002/sapm192761164},
      url = {https://onlinelibrary.wiley.com/doi/abs/10.1002/sapm192761164},
      eprint = {https://onlinelibrary.wiley.com/doi/pdf/10.1002/sapm192761164},
      year = {1927}
}

@incollection{penrose1971tensordiagrams,
  title={Applications of Negative Dimensional Tensors},
  author={Penrose, Roger},
  booktitle={Combinatorial Mathematics and Its Applications},
  publisher={Academic Press},
  pages={221--244},
  year={1971}
}

@article{yokota2024tensorbasics,
      title={Very Basics of Tensors with Graphical Notations: Unfolding, Calculations, and Decompositions}, 
      author={Tatsuya Yokota},
      year={2024},
      eprint={2411.16094},
      archivePrefix={arXiv},
      primaryClass={cs.LG},
      url={https://arxiv.org/abs/2411.16094}, 
}

@article{haastad1990tensor,
      title = {Tensor rank is NP-complete},
      journal = {Journal of Algorithms},
      volume = {11},
      number = {4},
      pages = {644-654},
      year = {1990},
      issn = {0196-6774},
      doi = {https://doi.org/10.1016/0196-6774(90)90014-6},
      url = {https://www.sciencedirect.com/science/article/pii/0196677490900146},
      author = {Johan Håstad},
}

@article{hillar2013most,
      author = {Hillar, Christopher J. and Lim, Lek-Heng},
      title = {Most Tensor Problems Are NP-Hard},
      year = {2013},
      issue_date = {November 2013},
      publisher = {Association for Computing Machinery},
      address = {New York, NY, USA},
      volume = {60},
      number = {6},
      issn = {0004-5411},
      url = {https://doi.org/10.1145/2512329},
      doi = {10.1145/2512329},
      journal = {J. ACM},
      month = nov,
      articleno = {45},
      numpages = {39},
}

@article{de2008tensor,
      author = {de Silva, Vin and Lim, Lek-Heng},
      title = {Tensor Rank and the Ill-Posedness of the Best Low-Rank Approximation Problem},
      journal = {SIAM Journal on Matrix Analysis and Applications},
      volume = {30},
      number = {3},
      pages = {1084-1127},
      year = {2008},
      doi = {10.1137/06066518X},
      URL = {https://doi.org/10.1137/06066518X},
      eprint = {https://doi.org/10.1137/06066518X}
}

@article{schollwock2011dmrg,
   title={The density-matrix renormalization group in the age of matrix product states},
   volume={326},
   ISSN={0003-4916},
   url={http://dx.doi.org/10.1016/j.aop.2010.09.012},
   DOI={10.1016/j.aop.2010.09.012},
   number={1},
   journal={Annals of Physics},
   publisher={Elsevier BV},
   author={Schollwöck, Ulrich},
   year={2011},
   month=Jan, pages={96–192}
}

@article{barratt2022tnunstabletraining,
      title={Improvements to Gradient Descent Methods for Quantum Tensor Network Machine Learning}, 
      author={Fergus Barratt and James Dborin and Lewis Wright},
      year={2022},
      eprint={2203.03366},
      archivePrefix={arXiv},
      primaryClass={cs.LG},
      url={https://arxiv.org/abs/2203.03366}, 
}

@inproceedings{pearce2025bilinearmlp,
      title={Bilinear {MLP}s enable weight-based mechanistic interpretability},
      author={Michael T Pearce and Thomas Dooms and Alice Rigg and Jose Oramas and Lee Sharkey},
      booktitle={The Thirteenth International Conference on Learning Representations},
      year={2025},
      url={https://openreview.net/forum?id=gI0kPklUKS}
}

@article{kossaifi2019tensorly,
  author  = {Jean Kossaifi and Yannis Panagakis and Anima Anandkumar and Maja Pantic},
  title   = {TensorLy: Tensor Learning in Python},
  journal = {Journal of Machine Learning Research},
  year    = {2019},
  volume  = {20},
  number  = {26},
  pages   = {1--6},
  url     = {http://jmlr.org/papers/v20/18-277.html}
}

@article{usvyatsov2022tntorch,
  author  = {Mikhail Usvyatsov and Rafael Ballester-Ripoll and Konrad Schindler},
  title   = {tntorch: Tensor Network Learning with {PyTorch}},
  journal = {Journal of Machine Learning Research},
  year    = {2022},
  volume  = {23},
  number  = {208},
  pages   = {1--6},
  url     = {http://jmlr.org/papers/v23/21-1197.html}
}

@article{novikov2020t3f,
  author  = {Alexander Novikov and Pavel Izmailov and Valentin Khrulkov and Michael Figurnov and Ivan Oseledets},
  title   = {Tensor Train Decomposition on TensorFlow (T3F)},
  journal = {Journal of Machine Learning Research},
  year    = {2020},
  volume  = {21},
  number  = {30},
  pages   = {1-7},
  url     = {http://jmlr.org/papers/v21/18-008.html}
}

@article{pareja2024tensorkrowch,
  title={Tensor{K}rowch: {S}mooth integration of tensor networks in machine learning},
  author={Pareja Monturiol, Jos{\'e} Ram{\'o}n and P{\'e}rez-Garc{\'i}a, David and Pozas-Kerstjens, Alejandro},
  journal={Quantum},
  volume={8},
  pages={1364},
  year={2024},
  publisher={Verein zur F{\"o}rderung des Open Access Publizierens in den Quantenwissenschaften},
  doi={10.22331/q-2024-06-11-1364},
  archivePrefix={arXiv},
  eprint={2306.08595}
}

@article{puljak2025tn4ml,
      title={tn4ml: Tensor Network Training and Customization for Machine Learning},
      author={Ema Puljak and Sergio Sanchez-Ramirez and Sergi Masot-Llima and Jofre Vallès-Muns and Artur Garcia-Saez and Maurizio Pierini},
      year={2025},
      eprint={2502.13090},
      archivePrefix={arXiv},
      primaryClass={cs.LG},
      url={https://arxiv.org/abs/2502.13090},
}

@article{lu2026fetta,
  author={Lu, Jinming and Tian, Jiayi and Li, Hai and Young, Ian. A. and Zhang, Zheng},
  journal={IEEE Transactions on Computer-Aided Design of Integrated Circuits and Systems}, 
  title={FETTA: Flexible and Efficient Hardware Accelerator for Tensorized Neural Network Training}, 
  year={2026},
  volume={45},
  number={9},
  pages={4422-4435},
  doi={10.1109/TCAD.2026.3651426}
}

@article{huang2025gvsa,
      title={A Tensor-Train Decomposition based Compression of LLMs on Group Vector Systolic Accelerator}, 
      author={Sixiao Huang and Tintin Wang and Ang Li and Ao Shen and Kai Li and Keyao Jiang and Mingqiang Huang and Hao Yu},
      year={2025},
      eprint={2501.19135},
      archivePrefix={arXiv},
      primaryClass={cs.AR},
      url={https://arxiv.org/abs/2501.19135}, 
}

@article{tian2025fpga,
  author={Tian, Jiayi and Lu, Jinming and Li, Hai and Wang, Xiangwei and Hao, Cong Callie and Young, Ian and Zhang, Zheng},
  journal={IEEE Transactions on Computer-Aided Design of Integrated Circuits and Systems}, 
  title={Ultra Memory-Efficient On-FPGA Training of Transformers via Tensor-Compressed Optimization}, 
  year={2026},
  volume={45},
  number={3},
  pages={1352-1365},
  doi={10.1109/TCAD.2025.3597244}}

@inproceedings{gong2023ette,
      author = {Gong, Yu and Yin, Miao and Huang, Lingyi and Xiao, Jinqi and Sui, Yang and Deng, Chunhua and Yuan, Bo},
      title = {ETTE: Efficient Tensor-Train-based Computing Engine for Deep Neural Networks},
      year = {2023},
      isbn = {9798400700958},
      publisher = {Association for Computing Machinery},
      address = {New York, NY, USA},
      url = {https://doi.org/10.1145/3579371.3589103},
      doi = {10.1145/3579371.3589103},
      booktitle = {Proceedings of the 50th Annual International Symposium on Computer Architecture},
      articleno = {68},
      numpages = {13},
      location = {Orlando, FL, USA},
      series = {ISCA '23}
}

@article{gray2018quimb,
      doi       = {10.21105/joss.00819},
      url       = {https://doi.org/10.21105/joss.00819},
      year      = {2018},
      publisher = {The Open Journal},
      volume    = {3},
      number    = {29},
      pages     = {819},
      author    = {Gray, Johnnie},
      title     = {quimb: A python package for quantum information and many-body calculations},
      journal   = {Journal of Open Source Software}
}

@article{gray2021hyperoptimized,
      doi = {10.22331/q-2021-03-15-410},
      url = {https://doi.org/10.22331/q-2021-03-15-410},
      title = {Hyper-optimized tensor network contraction},
      author = {Gray, Johnnie and Kourtis, Stefanos},
      journal = {{Quantum}},
      issn = {2521-327X},
      publisher = {{Verein zur F{\"{o}}rderung des Open Access Publizierens in den Quantenwissenschaften}},
      volume = {5},
      pages = {410},
      month = mar,
      year = {2021}
}

@article{fishman2022itensor,
	title={{The ITensor Software Library for Tensor Network Calculations}},
	author={Matthew Fishman and Steven R. White and E. Miles Stoudenmire},
	journal={SciPost Phys. Codebases},
	pages={4},
	year={2022},
	publisher={SciPost},
	doi={10.21468/SciPostPhysCodeb.4},
	url={https://scipost.org/10.21468/SciPostPhysCodeb.4}
}

@misc{novikov2021ttax,
  author       = {Novikov, Alexander and Belousov, Daniil},
  title        = {{TTAX}: Tensor-Train toolbox on {Jax}},
  year         = {2021},
  howpublished = {\url{https://github.com/fasghq/ttax}}
}

@misc{miller2019torchmps,
  author = {Miller, Jacob},
  title = {TorchMPS},
  year = {2019},
  publisher = {GitHub},
  journal = {GitHub repository},
  howpublished = {\url{https://github.com/jemisjoky/torchmps}},
}

@inproceedings{huang2019tendec,
      author={Huang Jiapeng and others},
      booktitle={2019 IEEE 38th International Performance Computing and Communications Conference (IPCCC)}, 
      title={A {C++} Library for Tensor Decomposition}, 
      year={2019},
      pages={1-2},
      doi={10.1109/IPCCC47392.2019.8958752}
}

@article{yu2022tednet,
  author    = {Yu Pan and Maolin Wang and Zenglin Xu},
  title     = {TedNet: {A} Pytorch toolkit for tensor decomposition networks},
  journal   = {Neurocomputing},
  volume    = {469},
  pages     = {234--238},
  year      = {2022}
}

@misc{gelss2019scikittt,
  author       = {Gel{\ss}, Patrick and Klus, Stefan and Scherer, Martin and N{\"u}ske, Feliks and L{\"u}cke, Marvin},
  title        = {{Scikit-TT}: Tensor-Train Computations in {Python}},
  year         = {2019},
  howpublished = {\url{https://github.com/PGelss/scikit_tt}}
}

@misc{bader2026tensortoolbox,

  author       = {Bader, B. W. and Kolda, T. G.},
  title        = {{Tensor Toolbox} for {MATLAB}, Version 3.8},
  year         = {2026},
  howpublished = {\url{https://www.tensortoolbox.org}}
}

@misc{kossaifi2024tensorlytorch,
  author       = {Jean Kossaifi and others},
  title        = {{TensorLy-Torch}: Deep Tensorized Learning},
  year         = {2024},
  howpublished = {\url{https://tensorly.org/torch/dev/}},
}

@article{paszke2019pytorch,
      title={PyTorch: An Imperative Style, High-Performance Deep Learning Library}, 
      author={Adam Paszke and others},
      year={2019},
      eprint={1912.01703},
      archivePrefix={arXiv},
      primaryClass={cs.LG},
      url={https://arxiv.org/abs/1912.01703}, 
}

@article{abadi2016tensorflow,
      title={TensorFlow: A system for large-scale machine learning}, 
      author={Martín Abadi and others},
      year={2016},
      eprint={1605.08695},
      archivePrefix={arXiv},
      primaryClass={cs.DC},
      url={https://arxiv.org/abs/1605.08695}, 
}

@article{harris2020numpy,
      title         = {Array programming with {NumPy}},
      author        = {Charles R. Harris and others},
      year          = {2020},
      month         = sep,
      journal       = {Nature},
      volume        = {585},
      number        = {7825},
      pages         = {357--362},
      doi           = {10.1038/s41586-020-2649-2},
      publisher     = {Springer Science and Business Media {LLC}},
      url           = {https://doi.org/10.1038/s41586-020-2649-2}
}

@software{bradbury2018jax,
  author = {James Bradbury and others},
  title = {{JAX}: composable transformations of {P}ython+{N}um{P}y programs},
  url = {http://github.com/jax-ml/jax},
  version = {0.3.13},
  year = {2018},
}

@misc{zagitov2026rethinking,
      title={Rethinking the Role of Tensor Decompositions in Post-Training LLM Compression}, 
      author={Artur Zagitov and Alexander Miasnikov and Maxim Krutikov and Vladimir Aletov and Gleb Molodtsov and Nail Bashirov and Artem Tsedenov and Aleksandr Beznosikov},
      year={2026},
      eprint={2606.03465},
      archivePrefix={arXiv},
      primaryClass={cs.LG},
      url={https://arxiv.org/abs/2606.03465}, 
}

@misc{sennrich2016bpe,
      title={Neural Machine Translation of Rare Words with Subword Units}, 
      author={Rico Sennrich and Barry Haddow and Alexandra Birch},
      year={2016},
      eprint={1508.07909},
      archivePrefix={arXiv},
      primaryClass={cs.CL},
      url={https://arxiv.org/abs/1508.07909}, 
}

@misc{kudo2018sentencepiece,
      title={SentencePiece: A simple and language independent subword tokenizer and detokenizer for Neural Text Processing}, 
      author={Taku Kudo and John Richardson},
      year={2018},
      eprint={1808.06226},
      archivePrefix={arXiv},
      primaryClass={cs.CL},
      url={https://arxiv.org/abs/1808.06226}, 
}

@inproceedings{kudo2018unigramlm,
    title = "Subword Regularization: Improving Neural Network Translation Models with Multiple Subword Candidates",
    author = "Kudo, Taku",
    editor = "Gurevych, Iryna  and
      Miyao, Yusuke",
    booktitle = "Proceedings of the 56th Annual Meeting of the Association for Computational Linguistics (Volume 1: Long Papers)",
    month = jul,
    year = "2018",
    address = "Melbourne, Australia",
    publisher = "Association for Computational Linguistics",
    url = "https://aclanthology.org/P18-1007/",
    doi = "10.18653/v1/P18-1007",
    pages = "66--75"
}

@inproceedings{provilkov2020bpedropout,
    title = "{BPE}-Dropout: Simple and Effective Subword Regularization",
    author = "Provilkov, Ivan  and
      Emelianenko, Dmitrii  and
      Voita, Elena",
    editor = "Jurafsky, Dan  and
      Chai, Joyce  and
      Schluter, Natalie  and
      Tetreault, Joel",
    booktitle = "Proceedings of the 58th Annual Meeting of the Association for Computational Linguistics",
    month = jul,
    year = "2020",
    address = "Online",
    publisher = "Association for Computational Linguistics",
    url = "https://aclanthology.org/2020.acl-main.170/",
    doi = "10.18653/v1/2020.acl-main.170",
    pages = "1882--1892"
}

@misc{tay2022charformer,
      title={Charformer: Fast Character Transformers via Gradient-based Subword Tokenization}, 
      author={Yi Tay and Vinh Q. Tran and Sebastian Ruder and Jai Gupta and Hyung Won Chung and Dara Bahri and Zhen Qin and Simon Baumgartner and Cong Yu and Donald Metzler},
      year={2022},
      eprint={2106.12672},
      archivePrefix={arXiv},
      primaryClass={cs.CL},
      url={https://arxiv.org/abs/2106.12672}, 
}

@inproceedings{koromilas2026polysae,
      title={Poly{SAE}: Modeling Feature Interactions in Sparse Autoencoders via Polynomial Decoding},
      author={Panagiotis Koromilas and Andreas D. Demou and James Oldfield and Yannis Panagakis and Mihalis Nicolaou},
      booktitle={Forty-third International Conference on Machine Learning},
      year={2026},
      url={https://openreview.net/forum?id=XAhDgsYn3a}
}

@misc{cunningham2023sae,
      title={Sparse Autoencoders Find Highly Interpretable Features in Language Models}, 
      author={Hoagy Cunningham and Aidan Ewart and Logan Riggs and Robert Huben and Lee Sharkey},
      year={2023},
      eprint={2309.08600},
      archivePrefix={arXiv},
      primaryClass={cs.LG},
      url={https://arxiv.org/abs/2309.08600}, 
}

@article{anandkumar2014tensor,
  title={Tensor decompositions for learning latent variable models},
  author={Anandkumar, Animashree and Ge, Rong and Hsu, Daniel and Kakade, Sham M and Telgarsky, Matus},
  journal={The Journal of Machine Learning Research},
  volume={15},
  number={1},
  pages={2773--2832},
  year={2014},
  publisher={JMLR. org}
}

@book{cichocki2009nonnegative,
  title     = {Nonnegative Matrix and Tensor Factorizations: Applications to Exploratory Multi-way Data Analysis and Blind Source Separation},
  author    = {Cichocki, Andrzej and Zdunek, Rafal and Phan, Anh Huy and Amari, Shun-ichi},
  publisher = {John Wiley \& Sons},
  year      = {2009},
  address   = {United Kingdom},
  isbn      = {9780470746660},
  pages     = {xxi, 477}
}

@article{hameed2025efficient,
  title={Efficient Probabilistic Tensor Networks},
  author={Hameed, Marawan Gamal Abdel and Rabusseau, Guillaume},
  journal={arXiv preprint arXiv:2510.00382},
  year={2025}
}

@article{han2018unsupervised,
  title={Unsupervised generative modeling using matrix product states},
  author={Han, Zhao-Yu and Wang, Jun and Fan, Heng and Wang, Lei and Zhang, Pan},
  journal={Physical Review X},
  volume={8},
  number={3},
  pages={031012},
  year={2018},
  publisher={APS}
}

@inproceedings{miller2021tensor,
  title={Tensor networks for probabilistic sequence modeling},
  author={Miller, Jacob and Rabusseau, Guillaume and Terilla, John},
  booktitle={International Conference on Artificial Intelligence and Statistics},
  pages={3079--3087},
  year={2021},
  organization={PMLR}
}

@article{vieijra2022generative,
  title={Generative modeling with projected entangled-pair states},
  author={Vieijra, Tom and Vanderstraeten, Laurens and Verstraete, Frank},
  journal={arXiv preprint arXiv:2202.08177},
  year={2022}
}

@article{glasser2019expressive,
  title={Expressive power of tensor-network factorizations for probabilistic modeling},
  author={Glasser, Ivan and Sweke, Ryan and Pancotti, Nicola and Eisert, Jens and Cirac, Ignacio},
  journal={Advances in neural information processing systems},
  volume={32},
  year={2019}
}

@article{glasser2020probabilistic,
  title={From probabilistic graphical models to generalized tensor networks for supervised learning},
  author={Glasser, Ivan and Pancotti, Nicola and Cirac, J Ignacio},
  journal={IEEE Access},
  volume={8},
  pages={68169--68182},
  year={2020},
  publisher={IEEE}
}

@misc{houlsby2019adapters,
      title={Parameter-Efficient Transfer Learning for NLP}, 
      author={Neil Houlsby and Andrei Giurgiu and Stanislaw Jastrzebski and Bruna Morrone and Quentin de Laroussilhe and Andrea Gesmundo and Mona Attariyan and Sylvain Gelly},
      year={2019},
      eprint={1902.00751},
      archivePrefix={arXiv},
      primaryClass={cs.LG},
      url={https://arxiv.org/abs/1902.00751}, 
}

@inproceedings{xu2026tdmoe,
      title={{TD}-MoE: Tensor Decomposition for MoE Models},
      author={Yuebin Xu and Yanhong Wang and Xuemei Peng and Hui Zang and Chen Minghao and Pengfei Xia and Zeyi Wen},
      booktitle={The Fourteenth International Conference on Learning Representations},
      year={2026},
      url={https://openreview.net/forum?id=D9cnZNZfxX}
}

@article{Laura2026lowranksurvye,
  author={Balzano, Laura and Ding, Tianjiao and Haeffele, Benjamin D. and Kwon, Soo Min and Qu, Qing and Wang, Peng and Wang, Zhangyang and Yaras, Can},
  journal={IEEE Signal Processing Magazine}, 
  title={An Overview of Low-Rank Structures in the Training and Adaptation of Large Models: From Implicit Low-Dimensionality to Efficient Training and Fine-Tuning [Special Issue on the Mathematics of Deep Learning]}, 
  year={2026},
  volume={43},
  number={3},
  pages={52-70},
  doi={10.1109/MSP.2026.3666749}
}

@software{karpathy2022nanogpt,
  author = {Andrej Karpathy},
  title = {NanoGPT},
  url = {https://github.com/karpathy/nanoGPT},
  year = {2022},
}

@inproceedings{penedo2024fineweb,
  title={The FineWeb Datasets: Decanting the Web for the Finest Text Data at Scale},
  author={Guilherme Penedo and Hynek Kydl{\'\i}{\v{c}}ek and Loubna Ben allal and Anton Lozhkov and Margaret Mitchell and Colin Raffel and Leandro Von Werra and Thomas Wolf},
  booktitle={The Thirty-eight Conference on Neural Information Processing Systems Datasets and Benchmarks Track},
  year={2024},
  url={https://openreview.net/forum?id=n6SCkn2QaG}
}

@inproceedings{wolf2020transformers,
    title = "Transformers: State-of-the-Art Natural Language Processing",
    author = "Thomas Wolf and Lysandre Debut and Victor Sanh and Julien Chaumond and Clement Delangue and Anthony Moi and Pierric Cistac and Tim Rault and Rémi Louf and Morgan Funtowicz and Joe Davison and Sam Shleifer and Patrick von Platen and Clara Ma and Yacine Jernite and Julien Plu and Canwen Xu and Teven Le Scao and Sylvain Gugger and Mariama Drame and Quentin Lhoest and Alexander M. Rush",
    booktitle = "Proceedings of the 2020 Conference on Empirical Methods in Natural Language Processing: System Demonstrations",
    month = oct,
    year = "2020",
    address = "Online",
    publisher = "Association for Computational Linguistics",
    url = "https://aclanthology.org/2020.emnlp-demos.6/",
    pages = "38--45"
}

@misc{mangrulkar2022peft,
  title =        {{PEFT}: State-of-the-art Parameter-Efficient Fine-Tuning methods},
  author =       {Sourab Mangrulkar and Sylvain Gugger and Lysandre Debut and Younes Belkada and Sayak Paul and Benjamin Bossan and Marian Tietz},
  howpublished = {\url{https://github.com/huggingface/peft}},
  year =         {2022}
}

@misc{hu2023llmadapters,
      title={LLM-Adapters: An Adapter Family for Parameter-Efficient Fine-Tuning of Large Language Models}, 
      author={Zhiqiang Hu and Lei Wang and Yihuai Lan and Wanyu Xu and Ee-Peng Lim and Lidong Bing and Xing Xu and Soujanya Poria and Roy Ka-Wei Lee},
      year={2023},
      eprint={2304.01933},
      archivePrefix={arXiv},
      primaryClass={cs.CL},
      url={https://arxiv.org/abs/2304.01933}, 
}

@misc{loshchilov2019adamw,
      title={Decoupled Weight Decay Regularization}, 
      author={Ilya Loshchilov and Frank Hutter},
      year={2019},
      eprint={1711.05101},
      archivePrefix={arXiv},
      primaryClass={cs.LG},
      url={https://arxiv.org/abs/1711.05101}, 
}

@inproceedings{
      huang2025hira,
      title={Hi{RA}: Parameter-Efficient Hadamard High-Rank Adaptation for Large Language Models},
      author={Qiushi Huang and Tom Ko and Zhan Zhuang and Lilian Tang and Yu Zhang},
      booktitle={The Thirteenth International Conference on Learning Representations},
      year={2025},
      url={https://openreview.net/forum?id=TwJrTz9cRS}
}

@inproceedings{clark2019boolq,
    title = "{B}ool{Q}: Exploring the Surprising Difficulty of Natural Yes/No Questions",
    author = "Clark, Christopher  and
      Lee, Kenton  and
      Chang, Ming-Wei  and
      Kwiatkowski, Tom  and
      Collins, Michael  and
      Toutanova, Kristina",
    editor = "Burstein, Jill  and
      Doran, Christy  and
      Solorio, Thamar",
    booktitle = "Proceedings of the 2019 Conference of the {N}orth {A}merican Chapter of the {A}ssociation for {C}omputational {L}inguistics: Human Language Technologies, Volume 1 (Long and Short Papers)",
    month = jun,
    year = "2019",
    address = "Minneapolis, Minnesota",
    publisher = "Association for Computational Linguistics",
    url = "https://aclanthology.org/N19-1300/",
    doi = "10.18653/v1/N19-1300",
    pages = "2924--2936"
}

@article{zellers2020piqa,
      title={PIQA: Reasoning about Physical Commonsense in Natural Language},
      volume={34},
      url={https://ojs.aaai.org/index.php/AAAI/article/view/6239},
      DOI={10.1609/aaai.v34i05.6239},
      number={05},
      journal={Proceedings of the AAAI Conference on Artificial Intelligence},
      author={Bisk, Yonatan and Zellers, Rowan and Le bras, Ronan and Gao, Jianfeng and Choi, Yejin},
      year={2020},
      month={Apr.},
      pages={7432–7439}
}

@inproceedings{sap2019siqa,
    title = "Social {IQ}a: Commonsense Reasoning about Social Interactions",
    author = "Sap, Maarten  and
      Rashkin, Hannah  and
      Chen, Derek  and
      Le Bras, Ronan  and
      Choi, Yejin",
    editor = "Inui, Kentaro  and
      Jiang, Jing  and
      Ng, Vincent  and
      Wan, Xiaojun",
    booktitle = "Proceedings of the 2019 Conference on Empirical Methods in Natural Language Processing and the 9th International Joint Conference on Natural Language Processing ({EMNLP}-{IJCNLP})",
    month = nov,
    year = "2019",
    address = "Hong Kong, China",
    publisher = "Association for Computational Linguistics",
    url = "https://aclanthology.org/D19-1454/",
    doi = "10.18653/v1/D19-1454",
    pages = "4463--4473",
}

@misc{clark2018arc,
      title={Think you have Solved Question Answering? Try ARC, the AI2 Reasoning Challenge}, 
      author={Peter Clark and Isaac Cowhey and Oren Etzioni and Tushar Khot and Ashish Sabharwal and Carissa Schoenick and Oyvind Tafjord},
      year={2018},
      eprint={1803.05457},
      archivePrefix={arXiv},
      primaryClass={cs.AI},
      url={https://arxiv.org/abs/1803.05457}, 
}

@inproceedings{mihaylov2018obqa,
    title = "Can a Suit of Armor Conduct Electricity? A New Dataset for Open Book Question Answering",
    author = "Mihaylov, Todor  and
      Clark, Peter  and
      Khot, Tushar  and
      Sabharwal, Ashish",
    editor = "Riloff, Ellen  and
      Chiang, David  and
      Hockenmaier, Julia  and
      Tsujii, Jun{'}ichi",
    booktitle = "Proceedings of the 2018 Conference on Empirical Methods in Natural Language Processing",
    month = oct # "-" # nov,
    year = "2018",
    address = "Brussels, Belgium",
    publisher = "Association for Computational Linguistics",
    url = "https://aclanthology.org/D18-1260/",
    doi = "10.18653/v1/D18-1260",
    pages = "2381--2391",
}

@inproceedings{zellers2019hellaswag,
    title = "{H}ella{S}wag: Can a Machine Really Finish Your Sentence?",
    author = "Zellers, Rowan  and
      Holtzman, Ari  and
      Bisk, Yonatan  and
      Farhadi, Ali  and
      Choi, Yejin",
    editor = "Korhonen, Anna  and
      Traum, David  and
      M{\`a}rquez, Llu{\'i}s",
    booktitle = "Proceedings of the 57th Annual Meeting of the {A}ssociation for {C}omputational {L}inguistics",
    month = jul,
    year = "2019",
    address = "Florence, Italy",
    publisher = "Association for Computational Linguistics",
    url = "https://aclanthology.org/P19-1472/",
    doi = "10.18653/v1/P19-1472",
    pages = "4791--4800",
}

@article{sakaguchi2021winogrande,
      author = {Sakaguchi, Keisuke and Bras, Ronan Le and Bhagavatula, Chandra and Choi, Yejin},
      title = {WinoGrande: an adversarial winograd schema challenge at scale},
      year = {2021},
      issue_date = {September 2021},
      publisher = {Association for Computing Machinery},
      address = {New York, NY, USA},
      volume = {64},
      number = {9},
      issn = {0001-0782},
      url = {https://doi.org/10.1145/3474381},
      doi = {10.1145/3474381},
      journal = {Commun. ACM},
      month = aug,
      pages = {99–106},
      numpages = {8}
}

@inproceedings{liu2022tuformer,
      title={Tuformer: Data-driven Design of Transformers for Improved Generalization or Efficiency},
      author={Xiaoyu Liu and Jiahao Su and Furong Huang},
      booktitle={International Conference on Learning Representations},
      year={2022},
      url={https://openreview.net/forum?id=V0A5g83gdQ_}
}

@misc{nguyen2026sama,
      title={SaMA: Morpho Adaptation via Asymmetric Expansion of Kronecker Product}, 
      author={An Nguyen and Anh Tong},
      year={2026},
      eprint={2611.04462},
      archivePrefix={arXiv},
      primaryClass={cs.LG},
      url={https://arxiv.org/abs/2611.04462}, 
}

@inproceedings{dao2022monarch,
  title = 	 {Monarch: Expressive Structured Matrices for Efficient and Accurate Training},
  author =       {Dao, Tri and Chen, Beidi and Sohoni, Nimit S and Desai, Arjun and Poli, Michael and Grogan, Jessica and Liu, Alexander and Rao, Aniruddh and Rudra, Atri and Re, Christopher},
  booktitle = 	 {Proceedings of the 39th International Conference on Machine Learning},
  pages = 	 {4690--4721},
  year = 	 {2022},
  editor = 	 {Chaudhuri, Kamalika and Jegelka, Stefanie and Song, Le and Szepesvari, Csaba and Niu, Gang and Sabato, Sivan},
  volume = 	 {162},
  series = 	 {Proceedings of Machine Learning Research},
  month = 	 {17--23 Jul},
  publisher =    {PMLR},
  url = 	 {https://proceedings.mlr.press/v162/dao22a.html}
}

@misc{yeafi2026chaindora,
      title={ChainDoRA: Tensor-Train Factorized Weight-Decomposed Low-Rank Adaptation for Parameter-Efficient LLM Fine-Tuning}, 
      author={Ashfak Yeafi and Mehedi Hasan and Md Khairul Islam},
      year={2026},
      eprint={2609.25058},
      archivePrefix={arXiv},
      primaryClass={cs.CL},
      url={https://arxiv.org/abs/2609.25058}, 
}

@inproceedings{liu2024dora,
  title = 	 {{D}o{RA}: Weight-Decomposed Low-Rank Adaptation},
  author =       {Liu, Shih-Yang and Wang, Chien-Yi and Yin, Hongxu and Molchanov, Pavlo and Wang, Yu-Chiang Frank and Cheng, Kwang-Ting and Chen, Min-Hung},
  booktitle = 	 {Proceedings of the 41st International Conference on Machine Learning},
  pages = 	 {32100--32121},
  year = 	 {2024},
  editor = 	 {Salakhutdinov, Ruslan and Kolter, Zico and Heller, Katherine and Weller, Adrian and Oliver, Nuria and Scarlett, Jonathan and Berkenkamp, Felix},
  volume = 	 {235},
  series = 	 {Proceedings of Machine Learning Research},
  publisher =    {PMLR},
  url = 	 {https://proceedings.mlr.press/v235/liu24bn.html}
}

@inproceedings{ye2026trac,
author = {Ye, Bangguo and Zhang, Yuanwei and Zhang, Xiaoqun},
booktitle = {International Conference on Learning Representations},
editor = {C. Vondrick and B. Hariharan and C. Raffel and L. Pinto and D. Yang and A. Faust},
pages = {114288--114321},
title = {TRAC: Tensor-Train based Across-layer Compression for Parameter-Efficient Fine-Tuning},
url = {https://proceedings.iclr.cc/paper_files/paper/2026/file/b9f4c86c2db3aed56dcc811260b944d0-Paper-Conference.pdf},
volume = {2026},
year = {2026}
}

@misc{dewage2026loracraft,
      title={LORA-CRAFT: Cross-layer Rank Adaptation via Frozen Tucker Decomposition of Pre-trained Attention Weights}, 
      author={Kasun Dewage and Marianna Pensky and Shankadeep Mondal and Suranadi De Silva},
      year={2026},
      eprint={2602.17510},
      archivePrefix={arXiv},
      primaryClass={cs.LG},
      url={https://arxiv.org/abs/2602.17510}, 
}

@inproceedings{gao2022mpoe,
    title = "Parameter-Efficient Mixture-of-Experts Architecture for Pre-trained Language Models",
    author = "Gao, Ze-Feng  and
      Liu, Peiyu  and
      Zhao, Wayne Xin  and
      Lu, Zhong-Yi  and
      Wen, Ji-Rong",
    editor = "Calzolari, Nicoletta  and
      Huang, Chu-Ren  and
      Kim, Hansaem  and
      Pustejovsky, James  and
      Wanner, Leo  and
      Choi, Key-Sun  and
      Ryu, Pum-Mo  and
      Chen, Hsin-Hsi  and
      Donatelli, Lucia  and
      Ji, Heng  and
      Kurohashi, Sadao  and
      Paggio, Patrizia  and
      Xue, Nianwen  and
      Kim, Seokhwan  and
      Hahm, Younggyun  and
      He, Zhong  and
      Lee, Tony Kyungil  and
      Santus, Enrico  and
      Bond, Francis  and
      Na, Seung-Hoon",
    booktitle = "Proceedings of the 29th International Conference on Computational Linguistics",
    month = oct,
    year = "2022",
    address = "Gyeongju, Republic of Korea",
    publisher = "International Committee on Computational Linguistics",
    url = "https://aclanthology.org/2022.coling-1.288/",
    pages = "3263--3273"
}

@misc{krishnan2026kvspectra,
      title={Tensor Decomposition of Transformer Key-Value Caches: Spectral Structure and Format Comparison}, 
      author={Rahul Krishnan and Volker Schulz},
      year={2026},
      eprint={2609.28029},
      archivePrefix={arXiv},
      primaryClass={math.NA},
      url={https://arxiv.org/abs/2609.28029}, 
}

@misc{nikpour2026kpa,
      title={Kronecker Product Attention: Memory-Efficient KV Compression via Structured Factorization}, 
      author={Bahareh Nikpour and Hossein Rajabzadeh and Maryam Dialameh and Vahid Partovi Nia and Parsa Omidi},
      year={2026},
      eprint={2610.12657},
      archivePrefix={arXiv},
      primaryClass={cs.LG},
      url={https://arxiv.org/abs/2610.12657}, 
}

@misc{krishnan2026jolt,
      title={A JoLT for the KV cache: Near-Lossless KV Cache Compression via Joint Rank-bit Allocation}, 
      author={Rahul Krishnan and Volker Schulz},
      year={2026},
      eprint={2607.12550},
      archivePrefix={arXiv},
      primaryClass={cs.LG},
      url={https://arxiv.org/abs/2607.12550}, 
}

@inproceedings{qiu2024btt,
      title={Compute Better Spent: Replacing Dense Layers with Structured Matrices}, 
      author={Shikai Qiu and Andres Potapczynski and Marc Finzi and Micah Goldblum and Andrew Gordon Wilson},
      booktitle={International Conference on Machine Learning},
      year={2024},
      eprint={2406.06248},
      archivePrefix={arXiv},
      primaryClass={cs.LG},
      url={https://arxiv.org/abs/2406.06248}, 
}

@inproceedings{potapczynski2024generaleinsum,
      author = {Potapczynski, Andres and Qiu, Shikai and Finzi, Marc and Ferri, Christopher and Chen, Zixi and Goldblum, Micah and Bruss, C. Bayan and De, Christopher and Wilson, Andrew},
      booktitle = {Advances in Neural Information Processing Systems},
      doi = {10.52202/079017-0126},
      editor = {A. Globerson and L. Mackey and D. Belgrave and A. Fan and U. Paquet and J. Tomczak and C. Zhang},
      pages = {3857--3881},
      publisher = {Curran Associates, Inc.},
      title = {Searching for Efficient Linear Layers over a Continuous Space of Structured Matrices},
      url = {https://proceedings.neurips.cc/paper_files/paper/2024/file/0729c38b421cd66f2313ab397d099e37-Paper-Conference.pdf},
      volume = {37},
      year = {2024}
      }

@misc{li2026tmuon,
      title={T-MUON: Tensorizing Muon via t-SVD for Large Language Model Pre-training}, 
      author={Dongyang Li and Mingsong Yan and Zheng Zhang and Sui Tang},
      year={2026},
      eprint={2610.24938},
      archivePrefix={arXiv},
      primaryClass={cs.LG},
      url={https://arxiv.org/abs/2610.24938}, 
}

@misc{miasnikov2026monarchmuon,
      title={Monarch-Muon: Structure-Aware Muon for Fast and Memory-Efficient Pretraining}, 
      author={Alexander Miasnikov and Ruslan Kabirov and Gleb Molodtsov and Dmitrii Vlashchuk and Vladimir Aletov and Nail Bashirov and Daniil Tikhonov and Maxim Krutikov and Aleksandr Shestakov and Aleksandr Beznosikov},
      year={2026},
      eprint={2610.17429},
      archivePrefix={arXiv},
      primaryClass={cs.LG},
      url={https://arxiv.org/abs/2610.17429}, 
}

\newpage


\begin{appendices}

\section{Pre-training}
\label{appendix:pre-training}

\localtableofcontents

\subsection{Backbone and projection parameterizations}
\label{appendix:projection-parameterizations}

All models use the GPT-2 small architecture, following nanoGPT \cite{karpathy2022nanogpt}. \Cref{tab:gpt2_config} gives the shared configuration. In each of the 12 blocks, we replace four weight matrices: the joint QKV projection, the attention output projection, and the FFN up- and down-projections. Each projection has its own factors. We use the same rank and mode sizes for a given projection type across blocks (\cref{tab:tensor_ranks,tab:tensor_shapes}). The token and position embeddings remain dense. The language modeling head shares the token embedding weights.

\begin{table}[ht]
  \centering
  \caption{Shared GPT-2 configuration and total parameter count of the dense baseline.}
  \label{tab:gpt2_config}
  \begin{tabular}{lc}
    \toprule
    Setting & Value \\
    \midrule
    Transformer blocks & 12 \\
    Hidden size & 768 \\
    FFN intermediate size & 3072 \\
    Attention heads / head dimension & 12 / 64 \\
    Context length in training & 1024 \\
    Model vocabulary size & 50,304 \\
    Activation & GELU \\
    Dropout & 0 \\
    Linear and layer normalization biases & Disabled \\
    Dense parameters & 124,373,760 \\
    \bottomrule
  \end{tabular}
\end{table}

For an input row vector $\vecb{x} \in \R^{1 \times I}$, a projection returns $\vecb{x}\mat{W}$ with $\mat{W} \in \R^{I \times J}$. We write $I=\prod_{n=1}^{N} I_n$ and $J=\prod_{n=1}^{N} J_n$ when the input and output axes are tensorized. The mode order in \cref{tab:tensor_shapes} determines how we flatten them into matrix indices.
\begin{itemize}[leftmargin=*]
  \item \textbf{Dense.} We train the $IJ$ entries of $\mat{W}$ directly.
  \item \textbf{Low-rank.} We set $\mat{W}=\mat{A}\mat{B}^{\top}$ with $\mat{A}\in\R^{I\times R}$ and $\mat{B}\in\R^{J\times R}$. This uses $R(I+J)$ parameters and applies two matrix multiplications.
  \item \textbf{Kronecker.} We set $\mat{W}=\sum_{r=1}^{R}\mat{A}_r\otimes\mat{B}_r$ with $\mat{A}_r\in\R^{I_1\times J_1}$ and $\mat{B}_r\in\R^{I_2\times J_2}$. The parameter count is $R(I_1 J_1+I_2 J_2)$, where $R$ counts Kronecker terms.
  \item \textbf{TT.} We use a chain with mode sizes $(d_1,\ldots,d_{2N})=(I_1,\ldots,I_N,J_1,\ldots,J_N)$ and cores $\ten{G}^{(k)}\in\R^{R_{k-1}\times d_k\times R_k}$. Define $(s_1,\ldots,s_{2N})=(i_1,\ldots,i_N,j_1,\ldots,j_N)$. The entry at multi-indices $\boldsymbol{i}$ and $\boldsymbol{j}$ is the product of the core slices, which yields a scalar,
  \[
    W_{\boldsymbol{i},\boldsymbol{j}}
    = \ten{G}^{(1)}_{:,s_1,:}\cdots\ten{G}^{(2N)}_{:,s_{2N},:}.
  \]
  With $R_0=R_{2N}=1$, the cores contain $\sum_{k=1}^{2N} R_{k-1}d_k R_k$ parameters. We use $N=2$, so each projection has four cores.
  \item \textbf{TTM.} Each core carries one input mode and one output mode, $\ten{G}^{(n)}\in\R^{R_{n-1}\times I_n\times J_n\times R_n}$. The matrix entry is
  \[
    W_{\boldsymbol{i},\boldsymbol{j}}
    = \ten{G}^{(1)}_{:,i_1,j_1,:}\cdots\ten{G}^{(N)}_{:,i_N,j_N,:}.
  \]
  With $R_0=R_N=1$, the cores contain $\sum_{n=1}^{N} R_{n-1}I_n J_n R_n$ parameters. We use $N=3$, so each projection has three cores.
\end{itemize}

Every internal TT or TTM rank equals the configured $R$. During training and timing, we contract the input directly with the factors (cores). We reconstruct the matrix for the singular value analysis.

\paragraph{Initialization.}
We draw dense projection weights from a zero-mean normal distribution with standard deviation $\sigma=0.02$. The attention output and FFN down-projections use $\sigma=0.02/\sqrt{2L}$ with $L=12$. For a factorized projection, we draw entries of all $m$ factors independently with standard deviation $(\sigma/\sqrt{P})^{1/m}$. Here $P$ is the number of summands in one reconstructed entry, equal to $R$ for low-rank and Kronecker and to the product of internal ranks for TT and TTM. This gives each reconstructed entry variance $\sigma^2$ at initialization.

\subsection{Data and training protocol}

We use the FineWeb-Edu token splits described in \cref{subsec:exp-pre-training} \cite{penedo2024fineweb}. The training loop reads the 2.5B-token training stream in order. Each example uses a 1024-token window as input and the same window shifted by one token as the next-token targets. Each optimizer step covers $32 \times 16 \times 1024 = 524{,}288$ input tokens. The 4768 steps consume 2,499,805,184 tokens per run.

We train all 13 configurations from scratch with seeds 0, 1, and 2 on one NVIDIA A100 80\,GB, using PyTorch 2.6.0 with CUDA 12.4. \Cref{tab:training_hyperparams} gives the shared recipe. We accumulate gradients from the mean of the 16 microbatch losses. We clip the accumulated gradient by its $\ell_2$ norm. AdamW applies weight decay to parameter tensors with at least two dimensions, including the projection factors and embedding matrices.

\begin{table}[ht]
  \centering
  \caption{Shared hyperparameters for pre-training from scratch.}
  \label{tab:training_hyperparams}
  \begin{tabular}{lc}
    \toprule
    Setting & Value \\
    \midrule
    Optimizer & Fused AdamW \cite{loshchilov2019adamw} \\
    AdamW $(\beta_1,\beta_2)$ and $\epsilon$ & $(0.9,0.95)$ and $10^{-8}$ \\
    Weight decay / gradient clip & $0.1$ / $1.0$ \\
    Peak / minimum learning rate & $6\times10^{-4}$ / $6\times10^{-5}$ \\
    Learning rate schedule & Linear warmup (235 steps) and cosine decay \\
    Optimizer steps & 4768 \\
    Microbatch size / accumulation steps & 32 / 16 \\
    Sequences per optimizer step & 1024 \\
    Training objective & Next-token cross-entropy \\
    Precision & bfloat16 autocast with TF32 enabled \\
    Compilation & \texttt{torch.compile}, default mode \\
    \bottomrule
  \end{tabular}
\end{table}

\subsection{Measurement protocol}

\paragraph{Validation loss.}
We evaluate at initialization, every 100 optimizer steps, and at the final step. Each evaluation averages next-token cross-entropy over ten batches of 32 windows drawn from random start positions in the 10M-token validation stream. The reported loss is the estimate at step 4768. For the quality comparison, we subtract both embedding tables from the total parameter count, a fixed $768(50{,}304+1{,}024)=39{,}419{,}904$ parameters in every run. The tied token embedding is counted once in the total.

\paragraph{Forward and backward time.}
The efficiency benchmark uses the same GPU and software versions as pre-training. For each configuration, we build the model and use synthetic input and target tensors sampled uniformly over the model vocabulary. We test sequence length 512 at batch sizes 1, 4, 8, 16, 32, 48, 64, and 96. We use bfloat16 autocast and enable TF32 in both eager and compiled modes. Compilation uses \texttt{reduce-overhead} with a full graph and static shapes.

We use a fresh process for each combination of model, batch size, and execution mode. After five warmup passes, we collect 11 groups of 25 timings in that process. Each timing uses synchronized CUDA events around a pass that clears gradients, computes the language modeling loss, and backpropagates. We place the optimizer update and compilation outside the timed region. We take the median within each group and report the median of the 11 group medians with a nonparametric 99\% confidence interval.

\paragraph{FLOPs and the compression-realization gap.}
We use PyTorch's \texttt{FlopCounterMode} to count the forward and backward FLOPs for supported operations in a separate eager pass at each batch size. Both modes use these same counts in \cref{eq:rho-gap}. We estimate $C_B$ from the ratio of total parameter counts, including embeddings, at a common weight precision. We compute $C_F$ from the counted FLOPs and $S_T$ from the medians for each model and the dense baseline in the same execution mode. Each comparison fixes the batch size and sequence length. The resulting gap describes the benchmark configurations, whose validation losses are measured separately.

\subsection{Efficiency across ranks and batch sizes}
\label{appendix:efficiency}

\Cref{fig:appendix_efficiency_speedup} shows the full timing grid. 
\Cref{fig:appendix_efficiency_gap} shows how the algorithmic and realization terms vary across the same grid.

\begin{figure}[p]
  \centering
  \includegraphics[width=0.98\linewidth,height=0.88\textheight,keepaspectratio]{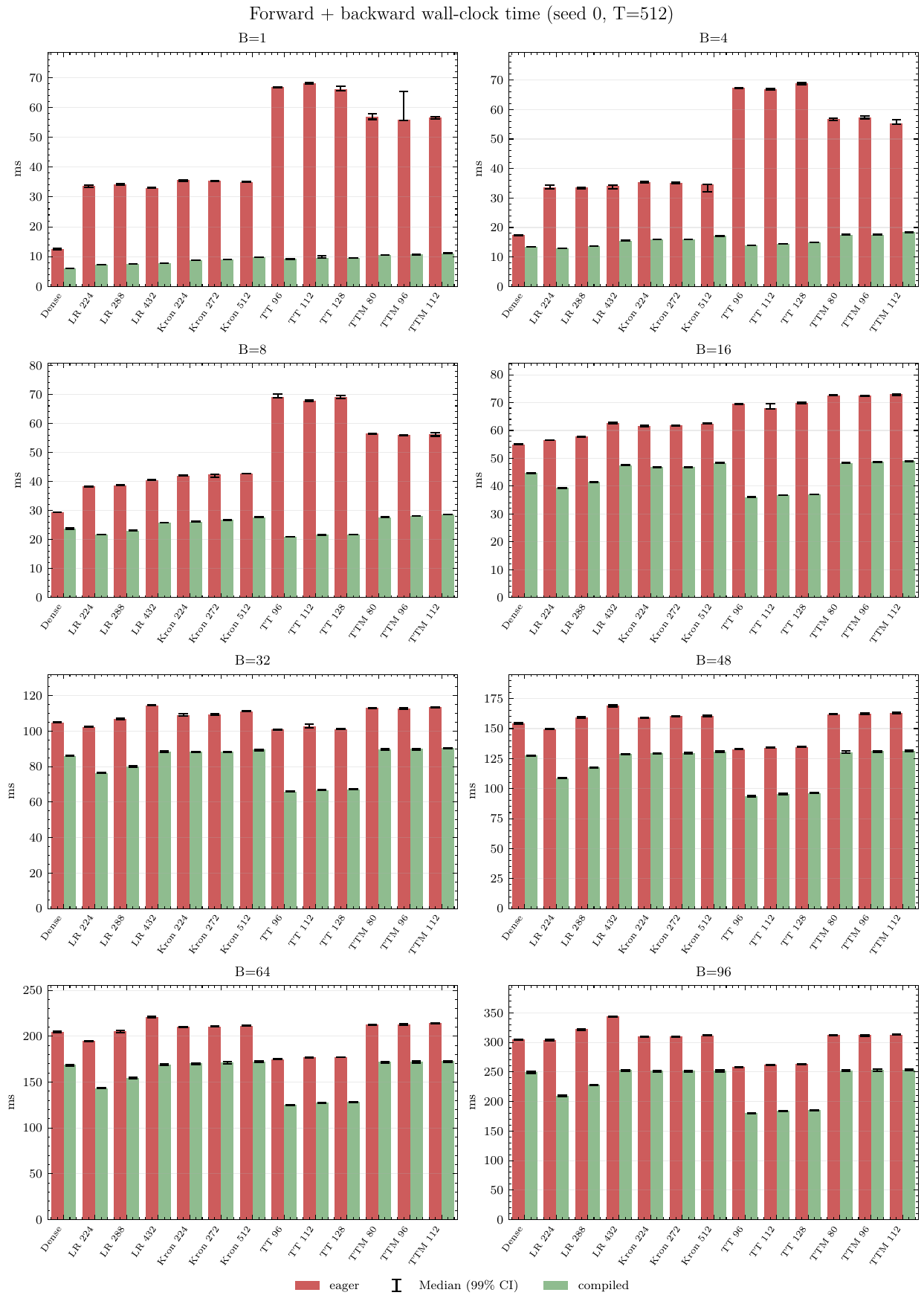}
  \caption{Median forward and backward time in milliseconds for the dense baseline and all tested ranks in eager and compiled modes. Each panel gives a different batch size $B$. Sequence length is 512. Error bars show 99\% confidence intervals for the median.}
  \label{fig:appendix_efficiency_speedup}
\end{figure}

\begin{figure}[p]
  \centering
  \includegraphics[width=0.98\linewidth,height=0.88\textheight,keepaspectratio]{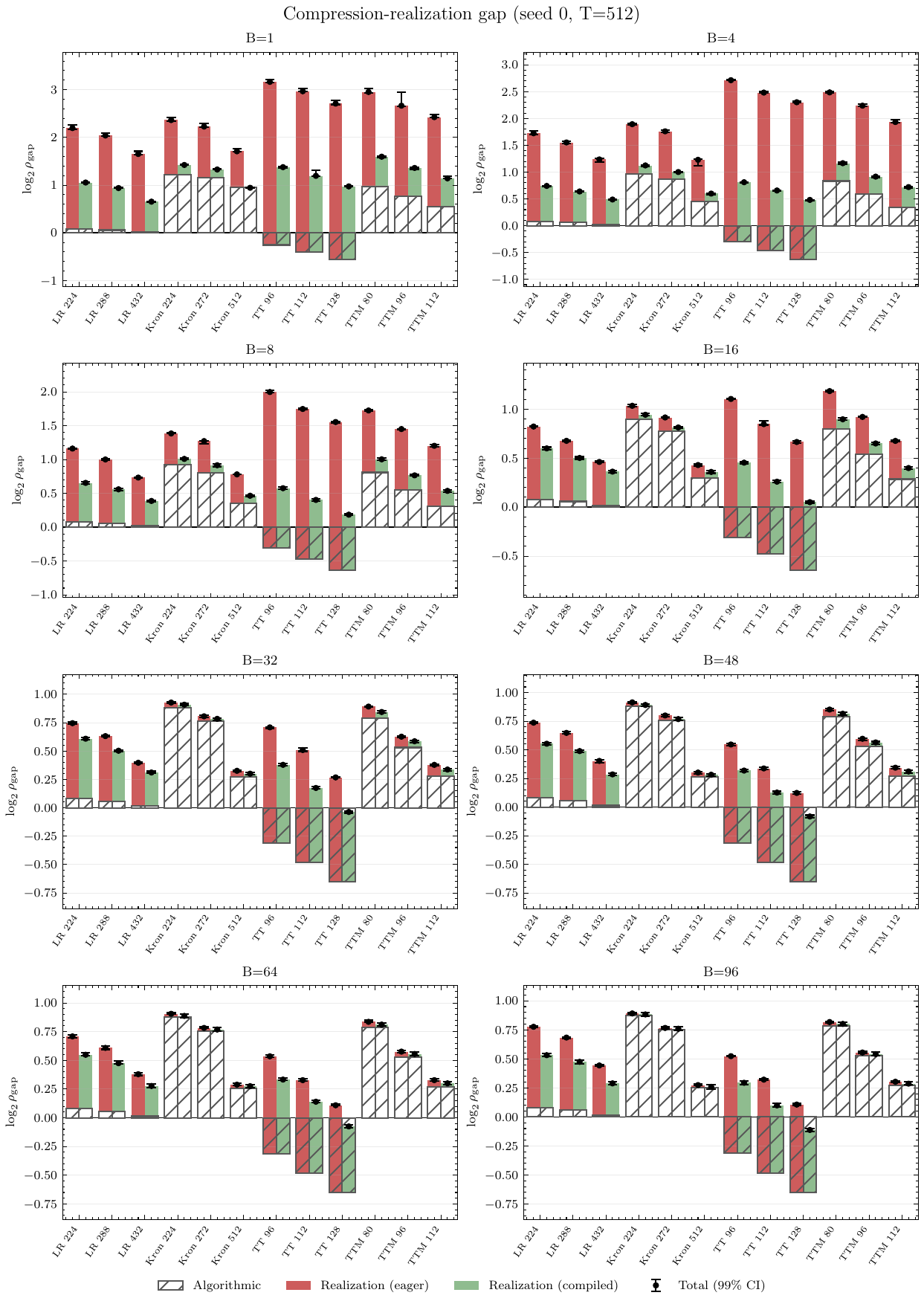}
  \caption{Compression-realization gap for all factorized configurations at sequence length 512. The hatched bars show $\log_2(C_B/C_F)$, which is the same for both execution modes. The colored bars show $\log_2(C_F/S_T)$ for eager and compiled execution. Their signed sum gives $\log_2\rho_{\rm gap}$. Error bars show 99\% confidence intervals for the total.}
  \label{fig:appendix_efficiency_gap}
\end{figure}

\subsection{Validation loss at each parameter budget}

\Cref{tab:e2_full_training} gives the per-seed values behind \cref{fig:exp_train}. Within each family, the final validation loss decreases with rank for all three seeds. The dense model has the lowest loss in each seed. Kronecker with $R=224$ uses 28.75M non-embedding parameters and reaches a mean loss of 3.590, below the 3.621 of low-rank with $R=224$ at 33.05M. At the largest tested budgets, Kronecker with $R=512$ reaches 3.459 at 65.69M non-embedding parameters, and low-rank with $R=432$ reaches 3.500 at 63.72M. The budgets are approximate, and the ranks have different meanings across parameterizations.

\begin{table}[ht]
  \centering
  \small
  \caption{Final FineWeb-Edu validation cross-entropy at step 4768 for each configuration and seed. Parameter counts are in millions. The last column gives the mean and sample standard deviation over the three seeds, computed before rounding.}
  \label{tab:e2_full_training}
  \begin{tabular}{lrrrrrrr}
    \toprule
     & & \multicolumn{2}{c}{Parameters (M)} & \multicolumn{4}{c}{Validation loss} \\
    \cmidrule(lr){3-4}\cmidrule(lr){5-8}
    Model & $R$ & Total & Non-emb. & Seed 0 & Seed 1 & Seed 2 & Mean $\pm$ SD \\
    \midrule
    Dense     & --  & 124.37 & 84.95 & 3.370 & 3.331 & 3.335 & $3.345 \pm 0.021$ \\
    \midrule
    Low-rank  & 224 &  72.47 & 33.05 & 3.635 & 3.611 & 3.619 & $3.621 \pm 0.012$ \\
              & 288 &  81.91 & 42.49 & 3.584 & 3.557 & 3.561 & $3.567 \pm 0.015$ \\
              & 432 & 103.14 & 63.72 & 3.518 & 3.484 & 3.496 & $3.500 \pm 0.017$ \\
    \midrule
    Kronecker & 224 &  68.17 & 28.75 & 3.617 & 3.579 & 3.573 & $3.590 \pm 0.024$ \\
              & 272 &  74.32 & 34.90 & 3.589 & 3.541 & 3.545 & $3.559 \pm 0.027$ \\
              & 512 & 105.11 & 65.69 & 3.493 & 3.435 & 3.451 & $3.459 \pm 0.030$ \\
    \midrule
    TT        &  96 &  73.40 & 33.98 & 3.894 & 3.859 & 3.870 & $3.874 \pm 0.018$ \\
              & 112 &  85.60 & 46.18 & 3.818 & 3.810 & 3.782 & $3.804 \pm 0.019$ \\
              & 128 &  99.66 & 60.24 & 3.767 & 3.721 & 3.746 & $3.744 \pm 0.023$ \\
    \midrule
    TTM       &  80 &  72.42 & 33.00 & 3.793 & 3.754 & 3.759 & $3.769 \pm 0.021$ \\
              &  96 &  86.68 & 47.26 & 3.730 & 3.692 & 3.692 & $3.704 \pm 0.022$ \\
              & 112 & 103.50 & 64.08 & 3.668 & 3.637 & 3.673 & $3.659 \pm 0.020$ \\
    \bottomrule
  \end{tabular}
\end{table}

\subsection{Singular value spectra of the learned weights}

We reconstruct the final projection matrices and plot their singular values in descending order on a logarithmic scale. We examine the joint QKV projection and its three column blocks separately, along with the attention output and both FFN projections. Each matrix has a smaller dimension of 768. \Cref{fig:appendix_rank_analysis_block_0,fig:appendix_rank_analysis_block_6,fig:appendix_rank_analysis_block_11} show the first, seventh, and last Transformer blocks, using zero-based indices.

The spectra show sharp cutoffs after the configured ranks of the low-rank and TT models across the plotted projections. For low-rank, the product $\mat{A}\mat{B}^{\top}$ has rank at most $R$. For TT, grouping the input and output cores on either side of the middle rank factors the matrix through a dimension of size $R_N=R$. Selecting the columns for Q, K, or V preserves this upper bound. Singular values past the cutoff reflect finite-precision reconstruction and decomposition. Kronecker and TTM have no analogous cutoff at the configured $R$. Their spectra extend across all 768 indices in these blocks, with a steeper tail in the individual attention projections than in the FFN projections. The spectra describe matrix rank in one seed. At similar budgets, low-rank still reaches lower validation loss than TTM (\cref{tab:e2_full_training}), so these cutoffs alone cannot explain the loss ordering.

\begin{figure}[p]
  \centering
  \includegraphics[width=0.98\linewidth,height=0.88\textheight,keepaspectratio]{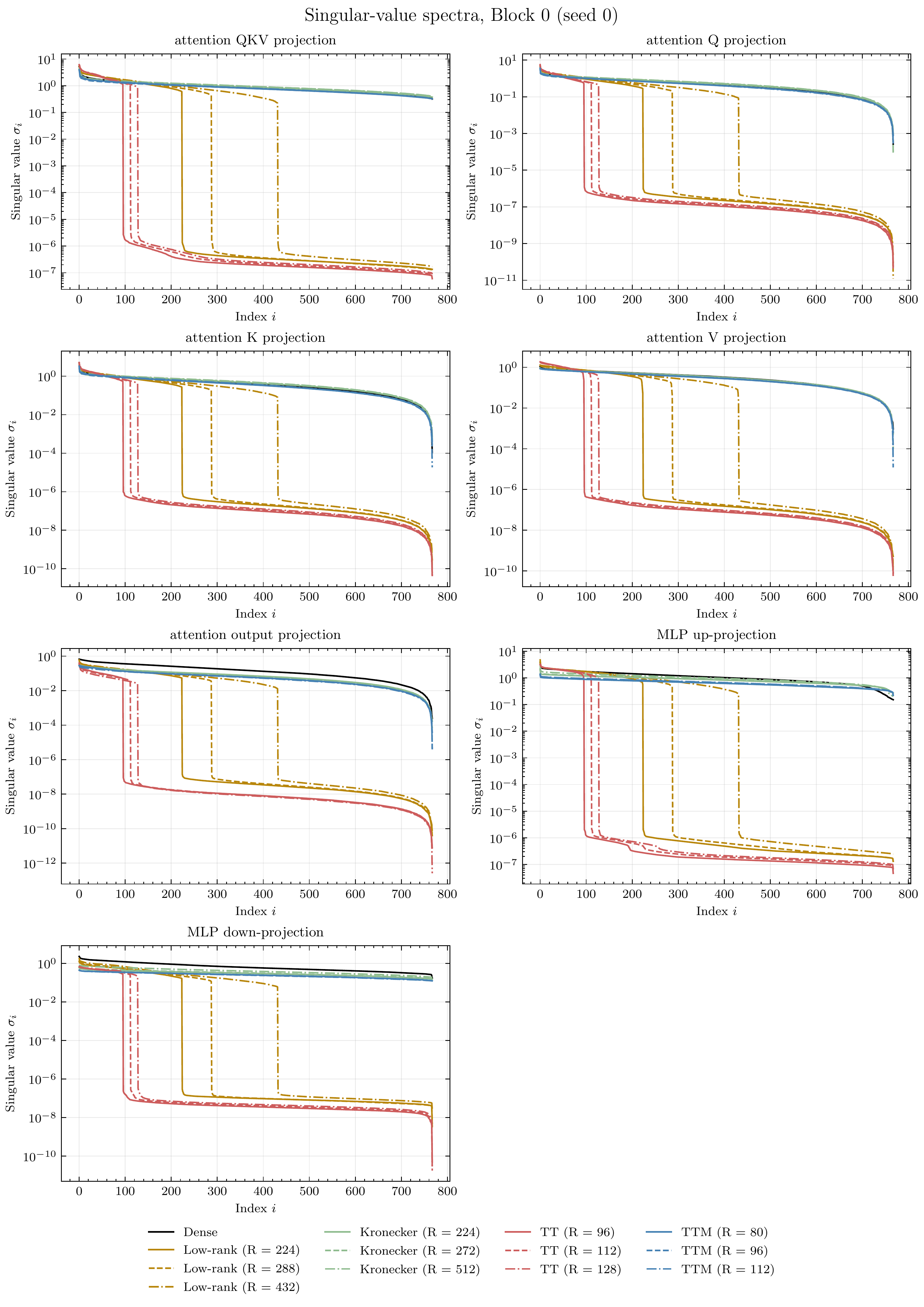}
  \caption{Singular value spectra of the final attention and FFN projection matrices in block 0. The panels include the joint QKV matrix, its individual Q, K, and V blocks, the attention output matrix, and both FFN matrices.}
  \label{fig:appendix_rank_analysis_block_0}
\end{figure}

\begin{figure}[p]
  \centering
  \includegraphics[width=0.98\linewidth,height=0.88\textheight,keepaspectratio]{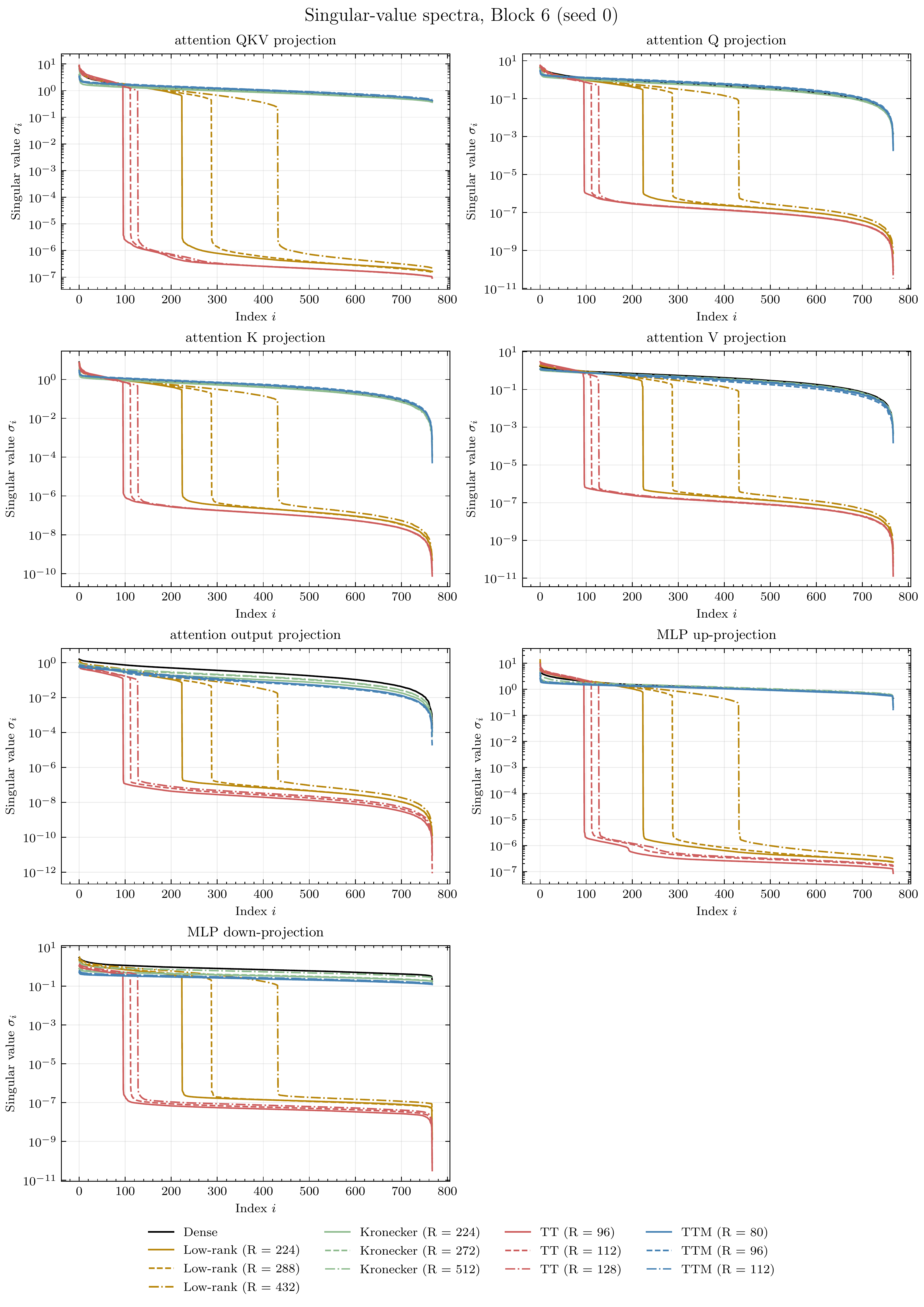}
  \caption{Singular value spectra of the final projection matrices in block 6. Panels and model configurations match \cref{fig:appendix_rank_analysis_block_0}.}
  \label{fig:appendix_rank_analysis_block_6}
\end{figure}

\begin{figure}[p]
  \centering
  \includegraphics[width=0.98\linewidth,height=0.88\textheight,keepaspectratio]{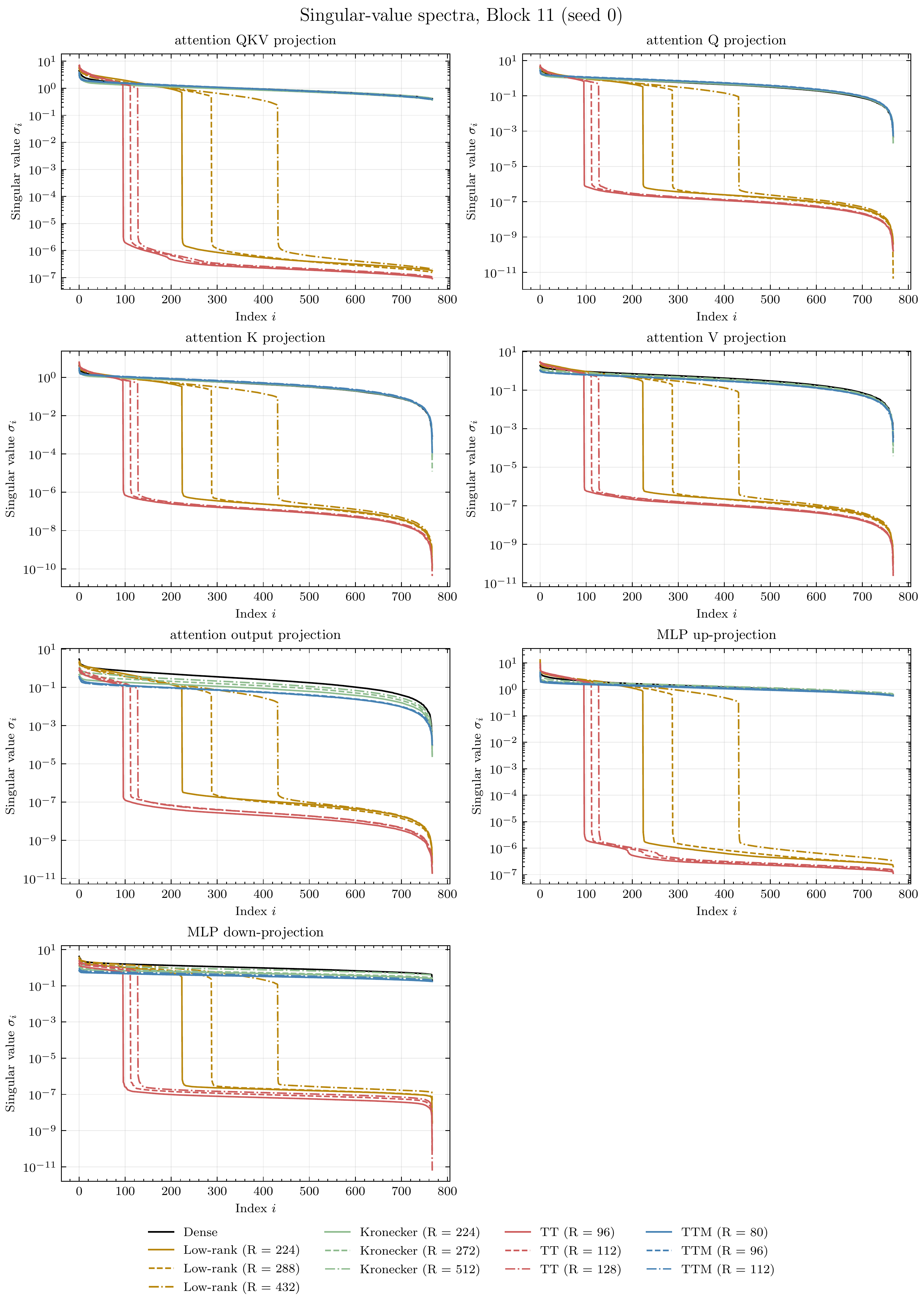}
  \caption{Singular value spectra of the final projection matrices in block 11 for. Panels and model configurations match \cref{fig:appendix_rank_analysis_block_0}.}
  \label{fig:appendix_rank_analysis_block_11}
\end{figure}

\FloatBarrier
\section{Parameter-efficient fine-tuning}
\label{appendix:peft}

\localtableofcontents

\subsection{Backbone and adapter configurations}

We use the \texttt{unsloth/Llama-3.2-1B} checkpoint \cite{metaai2024llama321b,unsloth2024llama321b}. The model has 16 Transformer blocks, a hidden size of 2048, and an FFN intermediate size of 8192. It uses 32 query heads and eight key/value heads of dimension 64. We freeze the backbone and train separate additive updates to the Q, K, V, FFN up-, and FFN down-projections in each block. The number of adapted projections is $16 \times 5 = 80$.

LoRA uses rank 32 and $\alpha=32$, giving the update a scale of $\alpha/R=1$ \cite{hu2022lora}. The tensorized adapters use the Kronecker, TT, and TTM parameterizations in \cref{appendix:projection-parameterizations}, with the mode sizes in \cref{tab:peft_tensor_shapes}. The adapter branch applies its factors directly to the layer input and adds the result to the frozen projection output. We initialize the tensorized update to zero by zeroing its final factor or core. Thus, the initial projection equals the pretrained projection.

\Cref{tab:appendix_peft_params} gives the trainable parameter counts. These counts exclude the frozen backbone.
\begin{table}[ht]
  \centering
  \caption{Configured ranks and trainable parameter counts for the main PEFT comparison.}
  \label{tab:appendix_peft_params}
  \begin{tabular}{lcc}
    \toprule
    Adapter & Ranks & Trainable parameters \\
    \midrule
    LoRA & $32$ & 15,204,352 \\
    Kronecker & $16$ & 6,553,600 \\
    TT & $32$ & 6,127,616 \\
    TTM & $32$ & 5,931,008 \\
    \bottomrule
  \end{tabular}
\end{table}

\subsection{Training and checkpoint selection}

We use the 170,300-example Commonsense170K training split and the associated 120-example validation split \cite{hu2023llmadapters,huang2025hira}. Each example uses an instruction template with separate \texttt{Instruction} and \texttt{Response} fields. The prompt starts with a beginning-of-sequence token and the target ends with an end-of-sequence token. We tokenize the prompt and target separately. We left-pad prompts to 256 tokens and right-pad targets to 16 tokens, truncating to the same limits. We concatenate the two sequences and compute next-token cross-entropy on the target tokens. Prompt and padding labels are masked.

We fine-tune for three epochs with AdamW using the \texttt{transformers} Trainer and \texttt{peft} \cite{wolf2020transformers,mangrulkar2022peft}. We select a learning rate of $2\times10^{-4}$ on the LoRA validation run and use it for all evaluated adapters. The schedule warms up linearly for 5\% of optimizer steps and then decays with a cosine schedule. We use weight decay of 0.1 and a gradient clipping threshold of 1.0. We use a single GPU, bfloat16 mixed precision with TF32 enabled, and \texttt{torch.compile} in default mode. Training uses microbatches of eight examples and accumulates gradients for 16 steps, giving an effective batch size of 128. Validation uses batches of eight. We evaluate and save a checkpoint every 80 optimizer steps and use the checkpoint with the lowest validation loss for test generation.

\subsection{Test protocol}

We evaluate on the eight benchmark test sets in \cref{tab:appendix_peft_eval}.
At test time, we use the same instruction template with an empty response field. We left-pad prompts within each batch of eight and truncate them to 256 tokens, or 512 for HellaSwag. We generate up to 16 new tokens using four beams with sampling disabled. We decode the generated continuation and take the first whole-word match to a valid answer keyword for the dataset, ignoring case. If the continuation has no match, we count the example as incorrect. The score is the fraction of correct answers for each dataset, and the aggregate score is the unweighted mean over the eight datasets.

\begin{table}[ht]
  \centering
  \caption{Test sets and accepted answer keywords for the commonsense reasoning evaluation.}
  \label{tab:appendix_peft_eval}
  \begin{tabular}{lrl}
    \toprule
    Dataset & Examples & Answer keywords \\
    \midrule
    BoolQ      & 3,270  & \texttt{true}, \texttt{false} \\
    PIQA       & 1,838  & \texttt{solution1}--\texttt{solution2} \\
    SIQA       & 1,954  & \texttt{answer1}--\texttt{answer3} \\
    ARC-c      & 1,172  & \texttt{answer1}--\texttt{answer5} \\
    ARC-e      & 2,376  & \texttt{answer1}--\texttt{answer5} \\
    OBQA       & 500    & \texttt{answer1}--\texttt{answer4} \\
    HellaSwag  & 10,042 & \texttt{ending1}--\texttt{ending4} \\
    WinoGrande & 1,267  & \texttt{option1}--\texttt{option2} \\
    \bottomrule
  \end{tabular}
\end{table}

\end{appendices}

\end{document}